\documentclass[twoside,11pt]{article}
\usepackage[hyphens]{url}

\usepackage{jmlr2e}
\usepackage{microtype}      % microtypography
\usepackage{bm}
\usepackage{setspace}
\usepackage[ruled,vlined]{algorithm2e} 
\usepackage{xcolor}
\usepackage{mathtools}
\usepackage{float}
\usepackage{tikz}
\usepackage{longtable}
\usepackage{multirow}
\usepackage{booktabs} % To thicken table lines
\usepackage[export]{adjustbox}
\usetikzlibrary{bayesnet}
\usepackage{subcaption}
\usepackage{caption} 
\usepackage{mathrsfs}
\newcommand{\bigCI}{\mathrel{\text{\scalebox{1.07}{$\perp\mkern-10mu\perp$}}}}

\DeclareMathOperator{\Tr}{tr}
\DeclareMathOperator{\supp}{supp}
\DeclareMathOperator{\diag}{diag}
\DeclareMathOperator*{\argmax}{arg\,max}
\DeclareMathOperator*{\argmin}{arg\,min}
\newcommand{\diagentry}[1]{\mathmakebox[1.8em]{#1}}
\newcommand{\xddots}{%
  \raise 4pt \hbox {.}
  \mkern 6mu
  \raise 1pt \hbox {.}
  \mkern 6mu
  \raise -2pt \hbox {.}
}
\newcommand{\vertiii}[1]{{\left\vert\kern-0.25ex\left\vert\kern-0.25ex\left\vert #1 
    \right\vert\kern-0.25ex\right\vert\kern-0.25ex\right\vert}}

\newcommand{\abs}[1]{\bigg\lvert#1\bigg\rvert}

\newcommand\numberthis{\addtocounter{equation}{1}\tag{\theequation}}
\newtheorem{case}{Case}

\numberwithin{subcase}{case}
\usepackage{mathtools}% Loads amsmath

\newtheorem{assumption}{Assumption}

\ShortHeadings{Learning Gaussian DAGs from Network Data}{Li, Madrid-Padilla, and Zhou}
\firstpageno{1}

\begin{document}

\title{Learning Gaussian DAGs from Network Data}

\author{\name Hangjian\ Li \email lihangjian123@ucla.edu \\
    %   \name Michael I.\ Jordan \email jordan@cs.berkeley.edu \\
    %   \addr Division of Computer Science and Department of Statistics\\
    %   University of California\\
    %   Berkeley, CA 94720-1776, USA
        \name Oscar Hernan Madrid Padilla \email oscar.madrid@stat.ucla.edu\\
       \name Qing Zhou \email zhou@stat.ucla.edu \\
       \addr Department of Statistics\\
       University of California, Los Angeles\\
       Los Angeles, CA 90095, USA}

\editor{}

\maketitle

\begin{abstract}%   <- trailing '%' for backward compatibility of .sty file
Structural learning of directed acyclic graphs (DAGs) or Bayesian networks has been studied extensively under the assumption that data are independent. We propose a new Gaussian DAG model for dependent data which assumes the observations are correlated according to an undirected network. Under this model, we develop a method to estimate the DAG structure given a topological ordering of the nodes. The proposed method jointly estimates the Bayesian network and the correlations among observations by optimizing a scoring function based on penalized likelihood. We show that under some mild conditions, the proposed method produces consistent estimators after one iteration. Extensive numerical experiments also demonstrate that by jointly estimating the DAG structure and the sample correlation, our method achieves much higher accuracy in structure learning. When the node ordering is unknown, through experiments on synthetic and real data, we show that our algorithm can be used to estimate the correlations between samples, with which we can de-correlate the dependent data to significantly improve the performance of classical DAG learning methods.
\end{abstract}


\section{Introduction}
\label{sec:intro}
Bayesian networks (BNs) with structure given by a directed acyclic graph (DAG) are a popular class of graphical models in statistical learning and causal inference. Extensive research has been done to develop new methods and theories to estimate DAG structures and its parameters from data. In this study we focus on the Gaussian DAG model defined as follows. Let $\mathcal{G}^* = (V, E)$ be a DAG that represents the structure of a BN for $p$ random variables $X_1,\ldots, X_p$. The vertex set $V = \{1,\ldots, p\}$ represents the set of random variables and the edges set $E = \{(j,i)\in V\times V: j\to i\}$ represents the directed edges in $\mathcal{G}^*$.  Let $\Pi_i = \{j\in V: (j,i)\in E\}$ denote the parent set of vertex $i$. A data matrix $X \in\mathbb{R}^{n\times p}$ is generated by the following Gaussian linear structural equations induced by $\mathcal{G}^*$:
\begin{equation}\label{eq:1.1}
X_{j} = \sum_{k \in \Pi_j}\beta_{kj}^*X_{k} + \varepsilon_{j},\quad \varepsilon_{j} =(\varepsilon_{1j},\ldots,\varepsilon_{nj}) \sim\mathcal{N}_{n}\left(0,\omega_{j}^{*2}I_n\right),
\end{equation} 
for $j=1,\ldots,p$, where $X_j$ is the $j$th column in $X$, $\omega_j^{*2}$ the error variance, and $B^* = (\beta_{kj}^*)_{p\times p}$ is the weighted adjacency matrix (WAM) of $\mathcal{G}^*$ such that $\beta_{kj}^* \neq 0$ if and only if $(k, j) \in E$, and $\beta_{jj}^* = 0$. The errors $\{\varepsilon_j\}$ are independent, and $\varepsilon_j$ is independent of $X_k$ for $k \in\Pi_j$. The goal is to estimate the structure of $\mathcal{G}^*$ from $X$, which is equivalent to estimating the support of $B^*$. 

A key assumption under \eqref{eq:1.1} is that the rows of $X$ are jointly independent as the covariance matrix of each $\varepsilon_j$ is diagonal. Under such i.i.d. assumption, many structure learning algorithms for DAGs have been developed, which can be largely categorized into three groups: score-based, constraint-based, and hybrid of the two. Score-based methods search for the optimal DAG by maximizing a scoring function such as minimum description length \citep{Roos2017}, BIC \citep{BIC}, and Bayesian scores \citep{Heckerman1995,Cooper1992} with various search strategies, such as order-based search \citep{NIPS2016_6232,Schmidt:2007:LGM:1619797.1619850,qiaoling}, greedy search \citep{Ramsey2017,Chickering:2003:OSI:944919.944933}, and coordinate descent \citep{fu2013learning,JMLR:v16:aragam15a,gu2019penalized}. Constraint-based methods, such as the PC algorithm in \cite{spirtes2000causation}, perform conditional independence tests among variables to construct a skeleton and then proceed to orient some of the edges. There are also hybrid methods such as in \cite{mmhc} and \cite{hybrid} that combine the above two approaches.

%A key assumption in all the aforementioned methods is that the rows of the data matrix $X$ are independent. 
In real applications, however, it is common for observations to be dependent as in network data, which violates the i.i.d assumption for the aforementioned methods. For example, when modeling the characteristics of an individual in a social network, the observed characteristics from different individuals can be dependent because they belong to the same social group such as friends, family and colleagues who often share similar features. Another example appears when modeling a gene regulatory network from individuals that are potentially linked genetically. When estimating brain functional networks, we often have a matrix of fMRI measurements for each individual, $X\in\mathbb{R}^{T\times \nu}$, across $T$ time points and $\nu$ brain regions of interests. The existence of correlations across both time points and brain regions often renders the estimates from standard graphical modeling methods inaccurate \citep{brain}. 
Motivated by these applications, we are interested in developing a Gaussian DAG model that can take into account the dependence between observations. Based on this model, we will develop a learning algorithm that can simultaneously infer the DAG structure and the sample dependencies. Moreover, since many real-world networks are sparse, we also want our method to be able to learn a sparse DAG and scale to large number of vertices. A sparsity constraint on the estimated DAG can also effectively prevent over-fitting and greatly improve the computational efficiency. Lastly, we would like to have theoretical guarantees on the consistency and finite-sample accuracy of the estimators. With these requirements in mind, we seek to
\begin{enumerate}
    \item Develop a novel Bayesian network model for network data;
    \item Develop a method that can jointly estimate a sparse DAG and the sample dependencies under the model;
    \item Establish finite-sample error bound and consistency of our estimators;
    \item Achieve good empirical performance on both synthetic and real data sets. 
\end{enumerate}

 Because $X$ is defined by the Gaussian noise vectors $\varepsilon_j$ according to the structural equations in \eqref{eq:1.1}, dependence among the rows of $X$ may be introduced by modeling the covariance structure among the variables $\varepsilon_{1j},\ldots, \varepsilon_{nj}$ in $\varepsilon_j$. Based on this observation, we will use an undirected graph $G^*$ to define the sparsity pattern in the precision matrix of $\varepsilon_j$. When $G^*$ is an empty graph, the variables in $\varepsilon_j$ are independent as in the classical Gaussian DAG model. However, when $G^*$ is not empty, $X$ follows a more complex matrix normal distribution, and the variance is defined by the product of two covariance matrices, one for the DAG $\mathcal{G}^*$ and the other for the undirected graph $G^*$. As a result, estimating the structure of the DAG $\mathcal{G}^*$ as well as other model parameters under the sparsity constraints in both graphs is a challenging task. We will start off by assuming that a topological ordering $\pi^*$ of $\mathcal{G}^*$ is given so that the search space for DAGs can be largely reduced. However, due to the presence of the second graph for network data, the usual likelihood-based objective function used in traditional score-based methods is non-convex. The constraint-based methods do not naturally extend to network data either due to the dependence among the individuals in $X$, which complicates the conditional independence tests. In order to find a suitable objective function and develop an optimization algorithm, we exploit the biconvex nature of a regularized likelihood score function and develop an effective blockwise coordinate descent algorithm with a nice convergence property. If the topological ordering of the DAG is unknown, it is impossible to identify a unique DAG from data due to the Markov equivalence of DAGs \citep{Chickering:2003:OSI:944919.944933}. Moreover, due to the lack of independence, it is very difficult to estimate the equivalence class defined by $\mathcal{G}^*$. In this case, we take advantage of an invariance property of the matrix normal distribution. Under some sparsity constraint on $\mathcal{G}^*$, we show that even with a random ordering, we can still get a good estimate of the covariance of $\varepsilon_j$, which can be used to decorrelate $X$ so that existing DAG learning algorithms can be applied to estimate an equivalence class of $\mathcal{G}^*$.

The remainder of the paper is structured as follows. In Section \ref{sec:model} we introduce a novel Gaussian DAG model for network data and discuss its connections with some existing models. We propose a structural learning algorithm for the model and go through its details in Section \ref{sec:method}. Section \ref{sec:theory} is devoted to our main theoretical results, as well as their implications under various high-dimensional asymptotic settings. Section \ref{sec:exp} reports numerical results of our method with detailed comparisons with some competing methods on simulated data. Section \ref{sec:gene} presents an application of our method on a real single-cell RNA sequencing data set. All proofs are deferred to the Appendix. 

\noindent
{\bf Notations} For the convenience of the reader, we now summarize some notations to be used throughout the paper. We write $\mathcal{G}^*$ and $G^*$ for the true DAG and the true undirected graph, respectively.  Let 
$\Omega^* := \diag(\omega_j^{*2})$ be a $p\times p$ diagonal matrix of error variances, $B^*$ denote the true WAM of $\mathcal{G}^*$, and $s := \sup_j\|\beta_j^*\|_0$ denote the maximum number of parents of any node in $\mathcal{G}^*$. Furthermore,  $X_j$  denotes the $j$th column of $X$ for  $j =1 ,\ldots,p$, and $x_i$ denotes  the $i$th row of $X$ for  $i = 1,\ldots,n$. Given two sequences $f_n$ and $g_n$, we write $f_n \lesssim g_n$ if $f_n = O(g_n)$, and $f_n \asymp g_n$ if $f_n \lesssim g_n$ and $g_n \lesssim f_n$. Denote by $[p]$ the index set $\{1,\ldots,p\}$. For $x \in \mathbb{R}^n$, we denote by $\|x\|_q$ its $\ell_q$ norm for $q\in[0,\infty]$. For $A \in \mathbb{R}^{n\times m}$,  $\|A\|_2 = \sup_v\{\|Av\|_2: \|v\|_2 \leq 1, v\in\mathbb{R}^m\}$ is the operator norm of $A$, $\|A\|_f$ is the Frobenius norm of $A$, $\|A\|_\infty = \max_{i,j}|a_{ij}|$ is the element-wise maximum norm of $A$, and $\vertiii{A}_\infty  = \max_{i\in[n]}\sum_{j=1}^m|a_{ij}|$ is the maximum row-wise $\ell_1$ norm of $A$.  Denote by $\sigma_{\min}(A)$ and $\sigma_{\max}(A)$, respectively, the smallest and the largest singular values of a matrix $A$. Let $|S|$ be the size of a set $S$.

\section{A Novel DAG Model for Dependent Data}
\label{sec:model}
% \subsection{Review of Bayesian network}
% Not sure if this section is necessary.
We model sample dependency through an undirected graph $G^*$ on $n$ vertices, with each vertex representing an observation $x_i,i\in[n]$, and the edges representing the conditional dependence relations among them. More explicitly, let $A(G^*)$ be the adjacency matrix of $G^*$ so that
$$A(G^*)_{ij} = 0 \Rightarrow x_i \bigCI x_j | x_{\setminus\{i,j\}},\quad \forall\ i\neq j.$$
Suppose we observe not only the dependent samples $\{x_i\}_{i=1}^n$  but also the graph (network) $G^*$. We generalize the structural equation model (SEM) in (\ref{eq:1.1}) to 
\begin{equation}\label{eq:model}
    X_{j} = \sum_{k \in \Pi_j}\beta_{kj}^*X_{k} + \varepsilon_{j},\quad \varepsilon_{j} =(\varepsilon_{1j},\ldots,\varepsilon_{nj}) \sim\mathcal{N}_{n}\left(0,\omega_{j}^{*2}\Sigma^*\right),
\end{equation}
where $\Sigma^*\in\mathbb{R}^{n\times n}$ is positive definite. The support of the precision matrix $\Theta^* = \left(\Sigma^{*}\right)^{-1}$ is restricted by $\text{supp}(\Theta^*)\subseteq A(G^*)$. We fix $\omega_{1}^* = 1$ so that the model is identifiable. Note that when $\Sigma^* = I_n$, the SEM (\ref{eq:model}) reduces to (\ref{eq:1.1}). Hence, the classical Gaussian DAG model in (\ref{eq:1.1}) is a special case of our proposed model (\ref{eq:model}). Under the more general model \eqref{eq:model}, we are facing a more challenging structural learning problem: Given dependent data $X$ generated from a DAG $\mathcal{G}^*$ and the undirected graph $G^*$ that encodes the sample dependencies, we want to estimate the DAG coefficients $B^*$, the noise variance $\Omega^* = \diag(\omega_j^{*2})$, and the precision matrix $\Theta^*$ of the samples. Before introducing our method, let us look at some useful properties of model $\eqref{eq:model}$ first. 

\subsection{Semi-Markovian Model}
\begin{figure}
\centering
\begin{subfigure}[b]{0.4\linewidth}
\begin{tikzpicture}[scale=.9, transform shape, minimum size=2em]
 \tikzstyle{every node} = [draw, shape=circle]
 \node[fill=gray!30] (a) at (0, 0) {$x_{11}$};
 \node[fill=gray!30] (b) at (2, 0) {$x_{12}$};
 \node[fill=gray!30] (c) at (4, 0) {$x_{13}$};
 \node[fill=gray!30] (d) at (0, -3) {$x_{21}$};
 \node[fill=gray!30] (e) at (2, -3) {$x_{22}$};
 \node[fill=gray!30] (f) at (4, -3) {$x_{23}$};
 \node (eps1) at (-1, -1) {$\varepsilon_{11}$};
 \node (eps2) at (1, -1) {$\varepsilon_{12}$};
 \node (eps3) at (3, -1) {$\varepsilon_{13}$};
 \node (eps4) at (-1, -2) {$\varepsilon_{21}$};
 \node (eps5) at (1, -2) {$\varepsilon_{22}$};
 \node (eps6) at (3, -2) {$\varepsilon_{23}$};
 \foreach \from/\to in {a/b, b/c, eps1/a, eps2/b, eps3/c, d/e,e/f,eps4/d,eps5/e,eps6/f}
 \draw [->] (\from) -- (\to);
 \draw [->] (a) to [out=35, in=145] (c);
 \draw [->] (d) to [out=325, in=215] (f);
 \end{tikzpicture}
 \caption{Markovian DAG model}
 \label{fig:1}
 \end{subfigure}
\begin{subfigure}[b]{0.4\linewidth}
\begin{tikzpicture}[scale=.9, transform shape, minimum size=2em]
 \tikzstyle{every node} = [draw, shape=circle]
 \node[fill=gray!30] (a) at (0, 0) {$x_{11}$};
 \node[fill=gray!30] (b) at (2, 0) {$x_{12}$};
 \node[fill=gray!30] (c) at (4, 0) {$x_{13}$};
 \node[fill=gray!30] (d) at (0, -3) {$x_{21}$};
 \node[fill=gray!30] (e) at (2, -3) {$x_{22}$};
 \node[fill=gray!30] (f) at (4, -3) {$x_{23}$};
 \node (eps1) at (-1, -1) {$\varepsilon_{11}$};
 \node (eps2) at (1, -1) {$\varepsilon_{12}$};
 \node (eps3) at (3, -1) {$\varepsilon_{13}$};
 \node (eps4) at (-1, -2) {$\varepsilon_{21}$};
 \node (eps5) at (1, -2) {$\varepsilon_{22}$};
 \node (eps6) at (3, -2) {$\varepsilon_{23}$};
 \foreach \from/\to in {a/b, b/c, eps1/a, eps2/b, eps3/c, d/e,e/f,eps4/d,eps5/e,eps6/f}
 \draw [->] (\from) -- (\to);
 \draw [->] (a) to [out=35, in=145] (c);
 \draw [->] (d) to [out=325, in=215] (f);
 \draw [thick,dotted] (eps1) to [out=180,in=180] (eps4);
 \draw [thick,dotted] (eps2) to [out=180,in=180] (eps5);
 \draw [thick,dotted] (eps3) to [out=180,in=180] (eps6);
\end{tikzpicture}
\caption{Semi-Markovian DAG model}
\label{fig:2}
\end{subfigure}
\caption{Graphical representations of the models in (\ref{eq:1.1}) and (\ref{eq:model}).}
\label{fig:12}
\end{figure}% 

The distinction between (\ref{eq:1.1}) and (\ref{eq:model}) becomes more clear when we regard (\ref{eq:model}) as a semi-Markovian causal model \citep{tian2006characterization}. 
Following its causal reading \citep{pearlbiometrika}, we can represent each variable $z_i$ in a DAG $\mathcal{G}$ on vertices $\{z_1,\ldots,z_p\}$ using a deterministic function:
\begin{equation}\label{eq:3}
    z_i = f_i(\Pi_i,u_i),\quad i\in[p],
\end{equation}
where $\Pi_i$ is the set of  parents of node $z_i$ in $G$ and $u_i$ are noises, sometimes also referred to as background variables. The model (\ref{eq:3}) is \textit{Markovian} if the noise variables $u_i$ are jointly independent, and it is \textit{semi-Markovian} if they are dependent. Now for a data matrix $X$ with $n=2$ and $p=3$, consider the DAG models defined, respectively, by \eqref{eq:1.1} and \eqref{eq:model} over all six random variables $x_{11},x_{12},x_{13},x_{21},x_{22},x_{23}$. 
Under SEM (\ref{eq:1.1}) we model $x_1=(x_{11},x_{12},x_{13})$ and $x_2=(x_{21},x_{22},x_{23})$ using the same SEM and assume they are independent, as shown in Figure~\ref{fig:1}.\footnote{Independent background variables are often omitted in the graph, but we include them here to better illustrate the differences between the two models.}
In contrast, the model proposed in (\ref{eq:model}) allows observations to be dependent by relaxing the independence assumption between $\varepsilon_{1k}$ and $\varepsilon_{2k}, k=1,2,3$. If we use dashed edges to link correlated background variables, then we arrive at a semi-Markovian DAG model as shown in Figure~\ref{fig:2}. 
In general, the variables $x_{i1}, \ldots, x_{ip}$ in each individual under the semi-Markovian model satisfy the same conditional independence constraints defined by a DAG, while the background variables $\varepsilon_{1j}, \ldots, \varepsilon_{nj}$ across the $n$ individuals are dependent. When estimating the DAG structure with such data, the correlations among individuals will reduce the effective sample size. Therefore, we need to take the distribution of the correlated $\varepsilon_i$ into account. 

%\subsection{Matrix normal graphical model}
\subsection{Matrix Normal Distribution}

Our model \eqref{eq:model} defines a matrix normal distribution for $X$. To see this, note that $\varepsilon=(\varepsilon_{ij})_{n\times p}$ in (\ref{eq:model}) follows a matrix normal distribution:
$$\varepsilon \sim \mathcal{N}_{n,p}\left(0, \Sigma^*, \Omega^*\right) \Leftrightarrow \text{vec}(\varepsilon) \sim \mathcal{N}_{np}(0, \Omega^* \otimes \Sigma^*),$$
where $\text{vec}(\cdot)$ is the vectorization operator and $\otimes$ is the Kronecker product.
Then, the random matrix $X$ satisfies
\begin{equation}\label{eq:matrixnormal}
X\sim\mathcal{N}_{n,p}(0, \Sigma^*, \Psi^*),
\end{equation}
where $\Psi^* = \left(I - B^*\right)^{-\top}\Omega^*\left(I - B^*\right)^{-1}$. 
From the properties of a matrix normal distribution, we can easily prove the following lemma which will come in handy when estimating the row covariance matrix $\Sigma^*$ from different orderings of nodes. Given a permutation $\pi$ of the set $[p]$, define $P_\pi$ as the permutation matrix such that $hP_\pi = (h_{\pi^{-1}(1)},\ldots, h_{\pi^{-1}(p)})$ for any row vector $h = (h_1,\ldots, h_p)$.  
\begin{lemma}\label{lem:cor}
If $X$ follows the model \eqref{eq:model}, then for any permutation $\pi$ of $[p]$ we have
$$XP_{\pi}\sim\mathcal{N}_{n,p}(0, \Sigma^*, P_{\pi}^\top\Psi^* P_{\pi}).$$
\end{lemma}
Although matrix normal distributions have been studied extensively in the past, the structural learning problem we consider here is quite unique. First of all, previous studies on matrix normal model usually assume we observe $m$ copies of $X$ and the MLE exists when $m \geq \max\{{p}/{n},{n}/{p}\} + 1$ \citep{dutill}. In our case, we only observe one copy of $X$ and thus the MLE does not exist without additional sparsity constraints. \cite{2009arXiv0906.3465A} proposed to use $\ell_1$ regularization to estimate the covariance matrices when $m=1$, but the estimation relies on the assumption that the model is transposable, meaning that the two components $(\Sigma, \Psi)$ of the covariance are symmetric and can be estimated in a symmetric fashion. In model \eqref{eq:model}, however, the two covariance components have different structural constraints and cannot be estimated in the same way. Lastly, practitioners are often interested in estimating large Bayesian networks with hundreds or more nodes under certain sparsity assumptions on the WAM $B$. For example, for methods that minimize a score function to estimate the covariances, adding a sparsity regularization term on $\Psi = \left(I - B\right)^{-\top}\Omega\left(I - B\right)^{-1}$ to the score function does not necessarily lead to a sparse estimate of $B$. In this paper, we propose a new DAG estimation method under the assumption that both the underlying undirected network among individuals and the Bayesian network are sparse. We are not interested in estimating $\Psi$ but a sparse factorization of $\Psi$ represented by the WAM $B$. This would require imposing sparsity constraints on $B$ itself instead of on $\Psi$. This is different from the recent work by \cite{tsiligkaridis2013convergence}, \cite{2009arXiv0906.3465A}, and \cite{zhou2014gemini} on the Kronecker graphical lasso.

\subsection{Score-equivalence}
The likelihood function of the proposed model \eqref{eq:model} also satisfies the desired score-equivalence property. To see this,  let $\beta_j = (\beta_{1j},\ldots,\beta_{pj})^\top$ be the $j$th column of the WAM $B$. 
Define an $n\times n$ sample covariance matrix of $\varepsilon_1/\omega_1, \dots,\varepsilon_p/\omega_p$ from $X$ as
\begin{align}\label{eq:S}
S(\Omega,B) = \frac{1}{p}\sum_{j=1}^p \frac{1}{\omega_j^2}\left(X_j - X\beta_j\right)\left(X_j - X\beta_j\right)^\top. 
\end{align}
Then the negative log-likelihood $L(B, \Omega, \Theta\mid X)$ from \eqref{eq:model} is given by
\begin{align}\label{eq:lik}
2L(B,\Omega,\Theta\mid X)= n\log\det\Omega - p\log\det\Theta + p\Tr(\Theta S(\Omega, B)).
\end{align} 
Due to the dependence among observations, it is unclear whether the well-known score-equivalence property for Gaussian DAGs \citep{Chickering:2003:OSI:944919.944933} still holds for our model. Let $(\widehat B(\mathcal{G}), \widehat\Omega(\mathcal{G}), \widehat\Theta(\mathcal{G}))$ denote the MLE of $\left(B, \Omega, \Theta\right)$ given a DAG $\mathcal{G}$ and the support restriction on $\Theta$. Then, the following theorem confirms the score-equivalence property for our DAG model. 
\begin{theorem}{(Score equivalence)}\label{thm:score_equivalence}
Suppose $\mathcal{G}_1$ and $\mathcal{G}_2$ are two Markov equivalent DAGs on the same set of $p$ nodes. If the MLEs $(\widehat B(\mathcal{G}_m), \widehat\Omega(\mathcal{G}_m), \widehat\Theta(\mathcal{G}_m))$, $m=1,2$, exist for the matrix $X=(x_{ij})_{n\times p}$, then 
\begin{equation*}
L(\widehat B(\mathcal{G}_1), \widehat\Omega(\mathcal{G}_1), \widehat\Theta(\mathcal{G}_1)\mid X) = L(\widehat B(\mathcal{G}_2), \widehat\Omega(\mathcal{G}_2), \widehat\Theta(\mathcal{G}_2)\mid X).
\end{equation*}
\end{theorem}
This property justifies the evaluation of estimated DAGs using common model selection criterion such as AIC and BIC. For examples, we show in Section \ref{sec:exp} that one can use BIC scores to select the optimal penalty level for our proposed DAG estimation algorithm.

% \section{Review of Bayesian networks}
% \label{sec:review}
% \input{review}

\section{Methods}
\label{sec:method}
We have discussed the properties of our novel DAG model for dependent data and the unique challenges faced by the structural learning task. In this section, we develop a new method to estimate the parameters in model $\eqref{eq:model}$. Our estimator is defined by the minimizer of a score function that derives from a penalized log-likelihood. In order to explain our method, let us start from the penalized negative log-likelihood function:
\begin{align}\label{generic loss}
     f(B,\Omega,\Theta):= 2L(B,\Omega,\Theta\mid X) + \rho_1(B) + \rho_2(\Theta), \quad B \in \mathcal{D},
\end{align}
where $\mathcal{D}$ is the space of WAMs for DAGs and $\rho_1$ and $\rho_2$ are some penalty functions. This loss function is difficult to minimize due to the non-convexity of $L$ and the exponentially large search space of DAGs. One way to reduce the search space is to assume a given topological ordering. A \textit{topological ordering} of a DAG $\mathcal{G}$ with $p$ vertices is a permutation $\pi$ of indices $(1,\ldots, p)$ such that for every edge $(i, j) \in E(\mathcal{G})$, $\pi^{-1}(i) < \pi^{-1}(j)$. Recall that a WAM $B$ is defined as $(\beta_{kj})_{p\times p}$ such that $\beta_{kj} \neq 0$ if and only if $(k, j) \in E(\mathcal{G})$; therefore, given a topological ordering $\pi$, we can define a set $\mathcal{D}(\pi)$ of WAMs compatible to $\pi$ such that all $B \in \mathcal{D}(\pi)$ are strictly upper triangular after permuting its rows and columns according to $\pi$. %The space of upper-triangular matrices is much smaller than the space of DAGs. 
Given a topological ordering $\pi$, the loss function \eqref{generic loss} becomes
\begin{align}\label{biconvex}
     f(B,\Omega,\Theta)=- p\log\det\Theta + \sum_{j=1}^p\frac{1}{\omega_j^2}\left\|LX_j - LX\beta_j\right\|_2^2 + \rho_1(B) + \rho_2(\Theta), \quad B\in\mathcal{D}(\pi),
\end{align}
where $L$ is the Cholesky factor of $\Theta$ (i.e. $\Theta = L^\top L$). If $\rho_1(\cdot)$ and $\rho_2(\cdot)$ are convex loss functions and the noise covariance matrix $\Omega = \diag(\omega_j^2)$ is known, \eqref{biconvex} will be a bi-convex function in $(B, \Theta)$, which can be minimized using iterative methods such as coordinate descent. \cite{Tseng2001} showed that the coordinate descent algorithm in bi-convex problems converges to a stationary point. Inspired by this observation, we propose the following two-step algorithm:
\begin{description}
    \item Step 1: Pre-estimate $\Omega^*$ to get $\widehat\Omega = \diag(\hat\omega_{j}^2)$.
    
    \item Step 2: Estimate $\widehat B$ and $\widehat\Theta$ by minimizing a biconvex score function derived from the penalized negative log-likelihood conditioning on $\hat\omega_j$.
\end{description}
Many existing noise estimation methods for high-dimensional linear models can be used to estimate $\widehat{\Omega}$ in Step 1 such as scaled lasso/MCP \citep{scaledlasso}, natural lasso \citep{jacob}, and refitted cross-validation \citep{fan}. We will present the natural estimator of $\Omega$ and discuss a few other alternatives in Section \ref{sec:natural}. Importantly, the statistical properties of the chosen estimator $\widehat\Omega$ in Step 1 will affect the properties of the $\widehat\Theta$ and $\widehat B$ we get in Step 2, and thus we must choose the estimator carefully. We leave the detailed discussion of the theoretical properties of $\widehat\Omega$ and their implications to Section \ref{sec:theory}.
Suppose $\widehat\Omega$ is given, we propose the following estimator for Step 2:
\begin{align}\label{obj}
\begin{split}
    \left(\widehat\Theta, \widehat B(\pi)\right) = \argmin_{\Theta \succ 0, B\in\mathcal{D}(\pi)} &\left\{- p\log\det\Theta + \sum_{j=1}^p\frac{1}{\hat\omega_j^2}\left\|LX_j - LX\beta_j\right\|_2^2 \right. \\
    &\quad \left. + \frac{\lambda_1}{\hat\omega_j^2}\|\beta_j\|_1 + \lambda_{2}\|\Theta\|_1 \right\}.
\end{split}
\end{align}
The $\ell_1$ regularization on $\beta_j/\hat\omega_j^2$ not only helps promote sparsity in the estimated DAG but also prevents the model from over-fitting variables that have small variances. The $\ell_1$ regularization on $\Theta$ ensures that the estimator is unique and can improve the accuracy of $\widehat\Theta$ by controlling the error carried from the previous step. We will discuss how to control the estimation errors in more detail in Section~\ref{sec:theory}. 

In Section \ref{sec:bcd}, we assume a topological ordering $\pi^*$ of the true DAG $\mathcal{G}^*$ is known. In this case, we will order the columns of $X$ according to $\pi^*$ %: $X_{j} = X_{\pi^{*}(j)}$ 
so that for each $j$, only the first $j-1$ entries in $\beta_j$ can be nonzero. When minimizing \eqref{obj}, we fix $\beta_{jk} = 0$ for $k \geq j$ and the resulting $\widehat B$ is guaranteed to be upper-triangular. If $\pi^*$ is unknown,  we show in Section \ref{sec:unknownorder} how the score function in \eqref{obj} is still useful for estimating $\Theta^*$ and describe a method of de-correlation so that standard DAG learning methods can be applied on the de-correlated data.

\subsection{Block Coordinate Descent}\label{sec:bcd}
We denote an estimate of the true precision matrix $\Theta^*$ at iteration $t$ by $\widehat{\Theta}^{(t)}$. We also write $\widehat{L}^{(t)}$ and $L^*$ for the Cholesky factors of the $\widehat\Theta^{(t)}$ and $\Theta^*$, respectively. %Thus, $L^{*\top} L^* = \Theta^*$. 
Since \eqref{obj} is biconvex, it can be solved by iteratively minimizing over $\Theta$ and $B$, i.e., using block coordinate descent. Consider the $t$th iteration of block coordinate descent. Fixing $\widehat{\Theta}^{(t)}$, the optimization problem in \eqref{obj} becomes the standard Lasso problem \citep{tibshirani1996regression} for each $j$: 
\begin{align}\label{eq:1step}
    \hat\beta_j^{(t+1)} = \argmin_{\beta_j} \frac{1}{2n} \|\widehat{L}^{(t)}X_j - \widehat{L}^{(t)}X\beta_j\|_2^2 + \lambda_n\|\beta_j\|_1,\quad \lambda_n = \lambda_1/(2n),
\end{align}
where $\widehat\Theta^{(t)} = \widehat L^{(t)\top}\widehat L^{(t)}$ is the Cholesky decomposition. Since the columns of $X$ are ordered according to $\pi$, we can set $\hat\beta_{ij}^{(t+1)} = 0$ for $i = j, j+1, \ldots, p$ and reduce the dimension of feasible $\beta_j$ in \eqref{eq:1step} to $j-1$. In particular, $\hat\beta_{1}^{(t+1)}$ is always a zero vector. Fixing $\widehat B^{(t+1)}$, solving for $\widehat{\Theta}^{(t+1)}$ is equivalent to a graphical Lasso problem with fixed support \citep{logdet}
\begin{align}\label{glasso}
    \widehat{\Theta}^{(t+1)} = 
    \argmin_{\Theta\succ 0,\ \supp(\Theta)\subseteq A(G^*)}\
    -\log\det\Theta + \Tr(\widehat{S}^{(t+1)}\Theta) + \lambda_p\|\Theta\|_1, 
    % L_p(\Theta\mid\widehat{B},\widehat\Omega, X) =
\end{align}
where $\widehat{S}^{(t+1)} = S(\widehat\Omega, \widehat B^{(t+1)})$ and $\lambda_p = \lambda_2/p$. The details of the method are given in Algorithm~\ref{alg1}.

\begin{algorithm}
\DontPrintSemicolon
\SetAlgoLined
% \KwResult{Write here the result}
\SetKwInOut{Input}{Input}
\SetKwInOut{Output}{Output}
\Input{$X$, $\Theta^{(0)}$, $\widehat\Omega$, $\rho, A(G^*), T$}
\BlankLine
\While{$ \max\left\{\|\widehat\Theta^{(t+1)} - \widehat\Theta^{(t)}\|_f,\|\widehat B^{(t+1)} - \widehat B^{(t)}\|_f\right\} > \rho$ and $t < T$}{
   \For{$j = 1,\dots,p$}{
        $\hat\beta_j^{(t+1)} \xleftarrow{}$ Lasso regression \ (\ref{eq:1step})\;
    }
    $\widehat\Theta^{(t+1)} \xleftarrow[]{} $ graphical Lasso with support restriction \eqref{glasso}\;   
    $t \xleftarrow[]{} t + 1$
}
\Output{$\widehat B\leftarrow\widehat{B}^{(t)}, \widehat\Theta\leftarrow\widehat\Theta^{(t)}$}
\caption{Block coordinate descent (BCD) algorithm}
\label{alg1}
\end{algorithm}

As shown in Proposition \ref{prop:convergence}, Algorithm \ref{alg1} will converge to a stationary point of the objective function \eqref{obj}. The stationary point here is defined as a point where all directional directives are nonnegative \citep{Tseng2001}. 

\begin{proposition}\label{prop:convergence}
Let $\{(\widehat B^{(t)},\widehat\Theta^{(t)}):t=1,2,\ldots\}$ be a sequence generated by the block coordinate descent Algorithm \ref{alg1} for any $\lambda_1,\lambda_2>0$. Then for almost all $X\in\mathbb{R}^{n\times p}$, every cluster point of $\{(\widehat B^{(t)},\widehat\Theta^{(t)})\}$ is a stationary point of the objective function in \eqref{obj}.
\end{proposition}

\subsection{A Natural Estimator of \texorpdfstring{$\Omega$}{Omega}}
\label{sec:natural}
We restrict our attention mostly to sparse undirected graphs $G$ consisting of $N$ connected components,  which implies that the row precision matrix $\Theta$ is block-diagonal: %in \eqref{block_diag}.
% Under these assumptions, the sparsity of the the network $\mathcal{G}$ depends primarily on the number of diagonal blocks in $\Theta$.
\begin{align}\label{block_diag}
    \Theta = 
\begin{pmatrix}
    \diagentry{\Theta_1}\\
    &\diagentry{\Theta_2}\\
    &&\diagentry{\xddots}\\
    &&&\diagentry{\Theta_N}\\
\end{pmatrix}.
\end{align}
The support of $\Theta$ inside each diagonal block $\Theta_i$ could be dense. %is assumed to be dense or close to dense and $0$ elsewhere:
This type of network is often seen in applications where individuals in the network form clusters: nodes in the same cluster are densely connected and those from different clusters tend to be more independent from each other. The underlying network $G$ will be sparse if the individuals are from a large number of small clusters. In other words, the sparsity of $G$ depends primarily on the number of diagonal blocks in $\Theta$. More general network structures also are considered in the numerical experiments in Section~\ref{sec:exp}. 

%Suppose the total number of diagonal blocks in $\Theta^*$ is $N$. 
Given the block-diagonal structure of $\Theta$ in \eqref{block_diag}, there are a few ways to estimate $\Omega$. We use the \textit{natural estimator} introduced by \cite{jacob}. We estimate $\hat\omega_j^2$ using independent samples in $X$ according to the block structure of $\Theta^*$. Let $B \subseteq [n]$ be a row index set and $A^B$ denote the submatrix formed by selecting rows from a matrix $A_{n\times m}$ with row index $i \in B$.
We draw one sample from each block and form a smaller $N\times p$ design matrix $X^B$. It is not difficult to see that $X_j^B = X^B\beta_j^* + \varepsilon_j^B$. Next define the natural estimator of $\omega_j^{*2}$ for $j\in[p]$ as in \cite{jacob}:
\begin{align}\label{eq:variance}
    \hat\omega_j^2 = \min_{\beta_j} \left\{\frac{1}{N} \|X_j^B - X^B\beta_j\|_2^2 + 2\lambda_N\|\beta_j\|_1\right\},
\end{align}
where $\lambda_N >0$ is a tuning parameter. 
In Section \ref{sec:theory}, we discuss %show how the rate of $\lambda_N$ affects the rate of 
the estimation error rate of $\widehat\Omega$. Alternative methods, such as scaled lasso \citep{scaledlasso} and the Stein's estimator \citep{noiselevel}, can also be used to estimate $\omega_j^2$.

\subsection{Estimating DAGs with Unknown Ordering}\label{sec:unknownorder}
%\revision{Recall that given any permutation $\pi$ of the nodes in the true DAG $\mathcal{G}^*$, there exists another DAG $\mathcal{G}_\pi = (V, E_\pi)$ associated with the same distribution $\mathbb{P}$ \citep{raskutti2018learning}:
%\begin{align}
%\left(\pi(j), \pi(k)\right) \in E_\pi \Longleftrightarrow X_{\pi(j)} \not\!\perp\!\!\!\perp X_{\pi(k)} \mid X_{\{\pi(1), \pi(2),\ldots, \pi(k-1)\} \setminus \{\pi(j)\}} \ \text{in}\ \mathbb{P} \ \text{and}\ j < k.
%\end{align}} 
Given any permutation $\pi$ of $[p]$, there exists a DAG $\mathcal{G}_\pi$ such that (i) $\pi$ is a topological sort of $\mathcal{G}_\pi$ and (ii) the joint distribution $\mathbb{P}$ of the $p$ random variables factorizes according to $\mathcal{G}_{\pi}$.
Under the assumption that the true DAG $\mathcal{G}^*$ is sparse, i.e., the number of nonzero entries in $\beta_j^*$ is at most $s$ for all $j$, for any random ordering $\pi^\prime$ we choose, the corresponding DAG $\mathcal{G}_{\pi^\prime}$ is also likely to be sparse where the number of parents for each node is less than some positive constant $s^\prime$. Following this intuition, we can randomly pick a permutation $\pi^\prime$ for the nodes and apply Algorithm~\ref{alg1} on $X_{\pi^\prime}:=XP_{\pi^\prime}$, where $(XP_{\pi^\prime})_{ij} = X_{i\pi^\prime(j)}$. If the sparsity $s^\prime$ is small compared to the sample size $n$, the estimate $\hat\beta_{ij}^\prime$ we get from solving the Lasso problem $\eqref{eq:1step}$ will be consistent as well (we discuss the error bound on $\hat\beta_{ij}$ in details in Section \ref{sec:theory}). Moreover, since the covariance $\Theta^*$ is invariant to permutations by Lemma \ref{lem:cor}, the resulting estimate $\widehat{\Theta}$ under the
random ordering $\pi^\prime$ will still be a good estimate of $\Theta^*$.
With the Cholesky factor $\widehat L$ of $\widehat \Theta$, we de-correlate the rows of $X$ and treat 
\begin{align}\label{eq:dcor}
\widehat X=\widehat L X,
\end{align}
as the new data. Because the row correlations in $\widehat X$ vanish, we can apply existing structure learning methods which require independent observations to learn the underlying DAG. We find that this de-correlation step is able to substantially improve the accuracy of structure learning by well-known state-of-the-art methods, such as the greedy equivalence search (GES) \citep{Chickering:2003:OSI:944919.944933} and the PC algorithm \citep{spirtes2000causation}. See Section \ref{sec:exp} for more details.

\section{Main Theoretical Results}
\label{sec:theory}
In this section, we present our main theoretical results for Algorithm~\ref{alg1} assuming a true ordering is given. Section \ref{sec:omega_consistency} is devoted to the error bounds of $\widehat{\Omega}$ using the natural estimator. We state our main theorem, Theorem \ref{thm_main}, in Section \ref{sec:b_theta}, along with some important corollaries. Finally, in Section \ref{sec:compare}, we compare the error rates of our estimators with those in related problems. Before we start, let us introduce some additional notations used in this section.\\

\noindent {\bf Notations} Let the errors of $\widehat{L}$ and $\widehat{\Theta}$ be defined as $\widehat{\Delta}_{chol} := \widehat{L} - L^*$ and $\widehat{\Delta}_{prec} := \widehat{\Theta} - \Theta^*$. Let $\widehat\Delta_j := \hat\beta_j - \beta_j^* \in\mathbb{R}^{p}$ denote the estimation error of the $j$th column of $B^*$. Let 
\begin{align*}
\bar\beta = \sup_{_{1\leq i,j\leq p}}|\beta_{ij}^*|, \quad\bar\omega = \sup_{1\leq j\leq p} \omega_j^*,  \quad\bar\psi^2 = \sup_{1\leq j\leq p}\Psi_{jj}^*,    
\end{align*}
where $\Psi^* = (I-B^*)^{-\top}\Omega^*(I-B^*)^{-1}$.
In the proofs, we also use $X_{i\cdot}$ and $X_{\cdot j}$ to denote the $i$th row and $j$th column of $X$, respectively. Let $\widetilde{X} = L^*X
\sim\mathcal{N}(0, \Psi^*\otimes I_n)$ and $\Tilde{\varepsilon} = L^*\varepsilon$. Then the rows of $\widetilde{X}$, i.e. $\Tilde{x}_i,i\in[n]$, are i.i.d from $\mathcal{N}(0, \Psi^*)$.
Let $m$ denote the maximum degree of the undirected graph $G^*$, which is allowed to grow with $n$. Following the setup in \cite{logdet}, the set of non-zero entries in the precision matrix is denoted as $\supp(\Theta^*) := \{(i,j)\mid \Theta_{ij}^*\neq 0\}$.
Let us use the shorthand $S$ and $S^c$ to denote the support and its complement in the set $[n]\times [n]$, respectively. Define the following constants:
\begin{align}\label{kappa_const}
\begin{split}
    \kappa_{\Sigma^*} &:= \vertiii{\Sigma^*}_\infty = \max_{1\leq i\leq n}\sum_{j=1}^n|\Sigma_{ij}^*|,\\
    \Gamma^*_{SS} &:= \left[\Theta^{*-1} \otimes \Theta^{*-1}\right]_{SS} \in\mathbb{R}^{(|S|+n)\times(|S|+n)}, \\ 
    \kappa_{\Gamma^*} &:= \vertiii{(\Gamma^*_{SS})^{-1}}_\infty.
\end{split}
\end{align}

\subsection{Consistency of \texorpdfstring{$\widehat\Omega$}{hatOmega} under Block-diagonal \texorpdfstring{$\Theta^*$}{theta}}\label{sec:omega_consistency}
Recall from Section \ref{sec:method} that $\widehat\Omega$ is pre-estimated at Step 1 in our two-step learning procedure. As we will discuss in more detail in Section \ref{sec:b_theta}, the accuracy of $\widehat\Theta$ obtained in Step 2 using Algorithm \ref{alg1} depends on the accuracy of $\widehat\Omega$. %Therefore, in order to get consistent estimators of $B^*$ and $\Theta^*$, we need to get a consistent estimator of $\Omega^*$ first. 
Existing methods for estimating the error variance in linear models, such as the scaled Lasso \citep{scaledlasso}, square-root lasso \citep{sqrtlasso}, and natural lasso \citep{jacob}, often assume independence among samples, which is not necessarily true under our network setting. However, if we assume the network of the samples is block diagonal and the samples form many small clusters, we would be able to collect independent samples from different clusters. This intuition suggests that existing methods are readily applicable in our setting to get consistent estimates of $\Omega^*$, as long as there are enough independent clusters in the undirected network $G^*$.

Formally, suppose $\Theta^*$ has a block-diagonal structure defined in \eqref{block_diag}. Let $N$ be the number of blocks. If we use the \textit{natural estimator} from \cite{jacob} (described in Section \ref{sec:method}) to get $\hat\omega_j$, then we have the following error bound: %the $\hat\omega_j$ are consistent.

\begin{lemma}\label{lem:omega}
Let $X$ be generated from \eqref{eq:model} and assume $\Theta^*$ is block-diagonal with $N$ blocks. Recall that $s = \sup_{j}\|\beta_j^*\|_0$. Let $\widehat\Omega$ be the natural estimator defined in $\eqref{eq:variance}$ with
 $$\lambda_N = 12\bar\psi\bar\omega\left(\sqrt{\frac{2\log p}{N}} + \sqrt{\frac{2\log 2 + 4\log p}{N}}\right),$$ then with probability at least $1 + 1/p^2 - 3/p$,
\begin{align*}
    % \sup_{1\leq j\leq p}\abs{\hat\omega_j^2 - \frac{\|\varepsilon_j^{B}\|_2^2}{N}} \leq \lambda_N\sup_{1\leq j\leq p}\|\beta_j^*\|_1 \leq \lambda_N s\bar\beta, \\
    \sup_{1\leq j\leq p}\abs{\hat\omega_j^2 - \omega_j^{*2}} \leq \lambda_Ns\bar\beta + 4\bar\omega\sqrt{\frac{\log 2 + \log p}{N}}. 
\end{align*}
\end{lemma}
Lemma \ref{lem:omega} shows that the maximum error of $\hat\omega_j$ is upper bounded by $C\cdot s\sqrt{{\log p}/{N}}$ where $C > 0$ is constant, and thus is consistent as long as $s\sqrt{{\log p}/{N}}\to 0$. When $p$ is fixed and $N$ increases, there are more diagonal blocks in $\Theta^*$, indicating more independent samples, and thus, $\hat\omega_j^2$ will be more accurate. When $N = n$, $\Theta^*$ becomes a diagonal matrix and the rows of $X$ become i.i.d. Another useful quantity for our subsequent discussion is the estimation error of $1/\hat\omega_j^2$. Let $r(\widehat\Omega)$ be defined as: 
\begin{align}\label{def: rnp}
    r(\widehat\Omega) := \sup_{1\leq j\leq p}\abs{\frac{1}{\hat\omega_j^2} - \frac{1}{\omega_j^{*2}}}.
\end{align}
We can easily show that the following lemma holds:
\begin{lemma}\label{lem:rho_consistency}
Suppose $\frac{1}{2}\inf_{1\leq j\leq p}\omega_j^{*2} - \sup_{1\leq j\leq p}\abs{\hat\omega_j^2 - \omega_j^{*2}} > b > 0$. Then 
\begin{align*}
    r(\widehat\Omega) \leq \frac{1}{b^4}\sup_{1\leq j\leq p}\abs{\hat\omega_j^2 - \omega_j^{*2}}.
\end{align*}
\end{lemma}
When $s(\log p / N)^{1/2}$ is small enough, Lemma \ref{lem:omega} implies that $b$ can be taken as a positive constant  with high probability.

\subsection{Error Bounds and Consistency of \texorpdfstring{$\widehat{B}^{(1)}$}{} and \texorpdfstring{$\widehat{\Theta}^{(1)}$}{}}\label{sec:b_theta}

%In the previous section we described how to obtain a consistent $\widehat\Omega$. 
Applying Algorithm \ref{alg1} with a pre-estimated $\widehat\Omega$ as input gives us $\widehat{B}^{(t)}$ and $\widehat\Theta^{(t)}$. In this section, we study the error bounds and consistency of $\widehat{B}^{(t)}$ and $\widehat\Theta^{(t)}$. Although Algorithm \ref{alg1} is computational efficient and has 
a desirable  convergence behavior as described in Proposition \ref{prop:convergence}, there are technical difficulties in establishing the consistency of $\widehat{B}^{(\infty)}$ and $\widehat\Theta^{(\infty)}$ after convergence, due to the dependence between $(\widehat{B}^{(t)}, \widehat\Theta^{(t)})$ across iterations. However, we show that as long as we have a suitable initial estimate satisfying Assumption \ref{assump: theta}, Algorithm \ref{alg1} can produce consistent estimators after one iteration, i.e., $(\widehat{B}^{(1)}, \widehat\Theta^{(1)})$ is consistent.

\begin{assumption}\label{assump: theta}
There exists a constant $0 < M \leq \sigma_{\min}^2(L^*)$, such that the initial estimate $\widehat\Theta^{(0)}$ satisfies
\begin{align*}
 \|\widehat\Theta^{(0)} - \Theta^*\|_2 \leq M. 
\end{align*}
\end{assumption}
Assumption \ref{assump: theta} states that the initial estimate $\widehat\Theta^{(0)}$ is inside an operator norm ball centered at the true parameter $\Theta^*$ with a radius smaller than a constant $M$. The constant $M$ is less than or equal to the smallest eigenvalue of $\Theta^*$. Assumption \ref{assump: theta} may not be easy to verify in practice, but since it only requires $\widehat\Theta^{(0)}$ to be within an $\ell_2$-ball of constant radius around $\Theta^*$, it is not difficult for Assumption \ref{assump: theta} to be met if $\Theta^*$ is sparse and normalized. Under Assumption \ref{assump: theta} we can establish finite-sample error bounds for $(\widehat{B}^{(1)}, \widehat\Theta^{(1)})$. Recall that $m$ denotes the maximum degree of the undirected graph $G^*$ and $s = \sup_{1\leq j\leq p}\|\beta_j^*\|_0$. Define 
\begin{align}\label{Rbar}
    \bar{R}(s, p, n) := \max\left\{6\bar\omega r(\widehat\Omega), \frac{72\bar\omega\bar\psi s}{b}\sqrt{\frac{\log p\log(\max\{n,p\})^2}{n}} \right\},
\end{align}
which depends on $r(\widehat\Omega)$, the error of $\widehat{\Omega}$ defined in \eqref{def: rnp}.
%With this notation we now present a consistency result.

\begin{theorem}\label{thm_main}
Consider a sample matrix $X$ from model \eqref{eq:model}. Let $\widehat{\Theta}^{(1)}, \widehat{B}^{(1)}$ be the estimates after one iteration of the Algorithm \ref{alg1}, given initial estimator $\widehat\Theta^{(0)}$ satisfying Assumption~\ref{assump: theta}. %Let $r(\widehat\Omega) = \sup_{1\leq j\leq p}|1/\hat\omega_j - 1/\omega_j^{*2}|$ where $\hat\omega_j$ are pre-estimated 
Suppose $b > 0$ as defined in Lemma \ref{lem:rho_consistency}. Pick the regularization parameters in \eqref{eq:1step} and \eqref{glasso} such that
\begin{align*}
    \lambda_n &\geq 12\bar\psi\bar\omega\left(\sqrt{\frac{2\log p}{n}} + \sqrt{\frac{2\log 2 + 4\log p}{n}}\right),\\
    \lambda_p &\geq 40\sqrt{2}\sqrt{\frac{\tau\log n + \log 4}{p}} + \bar{R}(s, p, n),
\end{align*}
where $\tau > 2$ and $r(\widehat\Omega)$ is defined in \eqref{def: rnp}. %\eqref{Rbar}. 
Let $\bar\kappa=\sigma_{\min}(\Psi^*)$. Then for some positive constant $c_1$, we have
\begin{align*}
        \sup_j\|\hat\beta_j^{(1)} - \beta_j^*\|_2 \leq \frac{\sqrt{s}}{c_1\bar\kappa}\lambda_n,
\end{align*}
with probability at least $(1-2/p)^2 - \left\{1/(\exp\{n/32\} - 1)  + 1/n^{\tau-2} + 5n^2 / \max\{n,p\}^4 \right\}$.
If in addition $n, p$ satisfy
\begin{align}
\begin{split}\label{main:sample_size}
    3200\log(4n^\tau)\max\left\{160, 24mC\right\} &\leq p,\\
    \max\left\{r(\widehat\Omega), \frac{4s\bar\omega^3 }{b}\sqrt{\frac{\log p\log^2\max\{n,p\}}{n}}\right\} &\leq 1/(24C),
\end{split}
\end{align}
where $C=\max\left\{\kappa_{\Sigma^*}\kappa_{\Gamma^*},\kappa_{\Sigma^*}^3\kappa_{\Gamma^*}^2 \right\}$, we also have
\begin{gather*}
    \|\widehat\Theta^{(1)} - \Theta^* \|_2 \leq  4\kappa_{\Gamma^*}m\lambda_p,
\end{gather*}
with the same probability.
\end{theorem}

We leave the detailed proof for Theorem \ref{thm_main} to the Appendix. The quantities $\kappa_{\Gamma^*}$ and $\kappa_{\Sigma^*}$ defined in \eqref{kappa_const} measure, respectively, the scale of the entries in $\Sigma^*$ and the inverse Hessian matrix $\Gamma^{*-1}_{SS}$ of the graphical Lasso log-likelihood function \eqref{glasso}, and they may scale with $n$ and $p$ in Theorem \ref{thm_main}. To simplify the following asymptotic results, we assume they are bounded by a constant as $n,p\to\infty$; see \cite{logdet} for a related discussion. %When $\kappa_{\Gamma^*}$ and $\kappa_{\Sigma^*}$ are constant w.r.t. $n$ and $p$ \citep{logdet}, the convergence rate of $\|\widehat\Theta^{(1)} - \Theta^*\|_\infty$ can be upper bounded by the sum of the bias function $\bar{R}$ defined in \eqref{Rbar} and the inverse tail function $\bar\delta_f(p;n^\tau)$ defined in \eqref{eq:tail}}: 
In addition, assume $\bar\kappa, \bar\psi, \bar\omega$ stay bounded as well. Then, under the conditions in Theorem \ref{thm_main}, we have for fixed positive constants $c_2, c_3, c_4$  that
\begin{gather}
    \sup_j\|\hat\beta_j^{(1)} - \beta_j^*\|_2^2 \leq c_2 s\frac{\log p}{n},\nonumber\\
    \|\widehat\Theta^{(1)} - \Theta^* \|_2 \leq c_3 m  \left(\sqrt{\frac{\tau\log n}{p}} + \max\left\{r(\widehat\Omega), c_4s\sqrt{\frac{\log p\log^2\max\{n,p\}}{n}}\right\}\right),\label{thetaerr}
\end{gather}
with high probability. For simplicity, we assume that $\Theta^*$ consists of $N$ blocks as in \eqref{block_diag} hereafter.
 If $\widehat\Omega$ satisfies the convergence rate specified in Lemmas \ref{lem:omega} and  \ref{lem:rho_consistency}, i.e., $r(\widehat\Omega) \lesssim  s\sqrt{\log p/N}$, then the sample constraints in \eqref{main:sample_size} are satisfied as long as
 %when $n$ is large, 
\begin{align}\label{simple_samplesize}
      m\log n \lesssim p,\quad s^2\log p \lesssim N,\quad s^2\log^3\max\{n,p\} \lesssim  n.
\end{align}
%for $n \geq 4$. 
As a result, we have the following two asymptotic results. The first one considers the scaling $p\gg n$ under which DAG estimation is high-dimensional. The second one considers the case $n\gg p$ so that the estimation of $\Theta^*$ is a high-dimensional problem.

\begin{cor}\label{cor:p>n}
Suppose the sample size and the number of blocks satisfy 
$$p\gg N\log^2p \gtrsim n\gtrsim N \gg \log p\to\infty.$$
Assume $\bar\beta, \bar\omega, \bar\psi, \kappa_{\Gamma^*},\kappa_{\Sigma^*} < \infty$ as $n, p\to\infty$ and $r(\widehat\Omega) \lesssim  s\sqrt{\log p/N}$. Then under the same assumptions as Theorem \ref{thm_main}, we have
\begin{align*}
 \sup_j\|\hat\beta_j^{(1)} - \beta_j^*\|_2^2 &= O_p\left(s\frac{\log p}{n}\right),\\
    \|\widehat\Theta^{(1)} - \Theta^* \|_2 &= O_p\left( ms\sqrt{\frac{\log^3p}{n}}\right).
\end{align*}
\end{cor}

\begin{cor}\label{cor:p<n}
Suppose the sample size and block numbers satisfy
$$n \gg s^2p\log p\log n \gtrsim N\gtrsim s^2p \to\infty.$$
Assume $\bar\beta, \bar\omega, \bar\psi,\kappa_{\Gamma^*},\kappa_{\Sigma^*} < \infty$ as $n, p \to\infty$ and $r(\widehat\Omega) \lesssim  s\sqrt{\log p/N}$. 
Then under the same assumptions as Theorem \ref{thm_main}, we have 
\begin{align*}
 \sup_j\|\hat\beta_j^{(1)} - \beta_j^*\|_2^2 &= O_p\left(s\frac{\log p}{n}\right),\\
    \|\widehat\Theta^{(1)} - \Theta^* \|_2 &= O_p\left(m\sqrt{\frac{\log n}{p}}\right).
\end{align*}
\end{cor}

\begin{remark}
Although we derived the consistency of $\widehat{\Theta}^{(1)}$ and $\widehat B^{(1)}$ in the above two corollaries under the setting where $\Theta^*$ is block-diagonal, these consistency properties still hold even when $\Theta^*$ is not block-diagonal. The main purpose of the block-diagonal setting is to provide an example where we can conveniently control the error of $\widehat\Omega$. But in practice $\Theta^*$ does not have to be block-diagonal. In particular, we did not assume any block-diagonal structure of $\Theta^*$ for the error bounds in Theorem~\ref{thm_main}.  It can be seen that the error bound of $\widehat B^{(1)}$ does not depend on the error of $\widehat{\Omega}$ at all.
% \textcolor{red}{  It can be seen that the error bound of $\widehat B^{(1)}$ does not dependent on the error of $\widehat{\Omega}$ to the extend that (\ref{main:sample_size}) must hold. }
Hence, the accuracy of $\widehat\Omega$ has no impact on the accuracy of $\widehat B^{(1)}$. The error bound of $\widehat\Theta^{(1)}$ in \eqref{thetaerr} is determined by the trade-off among three terms, one of which is the error rate $r(\widehat\Omega)$ as in \eqref{def: rnp}. This is supported by our numerical results as well. In Section~\ref{sec:exp}, we demonstrate, with both simulated and real networks where $\Theta^*$ is not block-diagonal, that our proposed BCD method can still accurately estimate $\Theta^*$ and $B^*$ whenever a relatively accurate $\widehat\Omega$ is provided.
\end{remark}

\subsection{Comparison to Other Results}\label{sec:compare}
If the data matrix $X$ consists of i.i.d. samples generated from the Gaussian linear SEM \eqref{eq:1.1}, assuming the topological sort of the vertices is known, the DAG estimation problem in \eqref{obj} is reduced to solving the standard Lasso regression in \eqref{eq:1step} with $\widehat{L}^{(t)} = I_n$, and thus independent of the initial $\widehat\Theta^{(0)}$ estimator. Under the \textit{restricted eigenvalue condition} %\eqref{eq:condition2} 
and letting $\lambda_n \asymp \sqrt{\log p/n}$, it is known the Lasso estimator has the following optimal rate for $\ell_2$ error \citep{lassooptimal}:
\begin{align*}
    \sup_{j}\|\hat\beta_j - \beta_j^*\|_2^2 = O_p\left(s\frac{\log p}{n}\right).
\end{align*}
When the data are dependent, Theorem \ref{thm_main} shows that the estimator from Algorithm \ref{alg1} can achieve the same optimal rate if we make the extra assumptions above. In particular, what we need is a reasonably good initial $\widehat\Theta^{(0)}$ estimate such that $\|\widehat\Theta^{(0)} - \Theta^*\|_2 \leq M$ for a small positive constant $M$.

On the other hand, if the underlying DAG is an empty graph and $\Omega^* = I_p$, the problem of estimating $\Theta^*$ can be solved using graphical Lasso in \eqref{glasso} because the data (columns in $X$) are i.i.d. The sample variance $\widehat{S}$ would also be an unbiased estimator of $\Sigma^*$. In this case, \cite{logdet} showed that 
\begin{align*}
    \|\widehat\Theta - \Theta^*\|_2 = O_p\left(m\sqrt{\frac{\log n}{p}}\right).
\end{align*}
This results does not require knowing $\supp(\Theta^*)$ but assumes a \textit{mutual incoherence} condition on the Hessian of the log likelihood function. In our case, $\widehat{S}^{(1)}$ is biased due to the accumulated errors from the previous Lasso estimation as well as $\widehat\Omega$. As a result, there is an extra bias term $\bar{R}(s,n,p)$ in $\|\widehat{S}^{(1)} - \Sigma^* \|_\infty$ (see Lemma \ref{lem:R} in Appendix \ref{app:theta}) compared to the i.i.d. setting:
\begin{align*}
    \|\widehat{S}^{(1)} - \Sigma^* \|_\infty = \bar\delta_f(p, n^\tau) + \bar{R}(s,n,p),
\end{align*}
where $\bar\delta_f(p, n^\tau) \asymp \sqrt{{\log n}/{p}}$ is the classical graphical Lasso error rate, and 
$$\bar{R}(n,p,s) \asymp \max\left\{r(\widehat\Omega), s\sqrt{\log p\log\max\{n,p\}/n}\right\}$$ 
depends on the estimation errors of $\widehat{B}^{(1)}$ and $\widehat\Omega$. When $n \gg p$ and $r(\widehat\Omega)$ is dominated by $\sqrt{\log(n)/p}$, we get the same rate for the $\ell_2$ consistency of $\widehat\Theta$ (Corollary~\ref{cor:p<n}) under slightly more strict constraint on the sample size \eqref{simple_samplesize}. %with slightly smaller probability. 
If $n \ll p$, then the $\ell_2$ error rate is determined by $\max\{r(\widehat\Omega), s(\log^3p/n)^{1/2}\}$. Suppose $\Theta^*$ is block-diagonal. If the number of blocks $N$ is much smaller than $n$, then the $\ell_2$ rate will be dominated by $r(\widehat\Omega) \asymp s\sqrt{\log p/ N}$, which could be slower than the optimal graphical Lasso rate. But that is expected due to the error introduced in the DAG estimates $\widehat{B}^{(1)}$ and $\widehat{\Omega}$.

% \section{Proof outline}
% \label{sec:theory2}
% \input{theory2}

\section{Numerical Experiments}
\label{sec:exp}
Under the assumption that observations generated from a DAG model are dependent, we will evaluate the performance of the block coordinate descent (BCD) algorithm, i.e., Algorithm \ref{alg1}, in recovering the DAG compared to traditional methods that treat data as independent. We expect that the BCD method would give more accurate structural estimation than the baselines by taking the dependence information into account. When a topological ordering of the true DAG is known, we can identify a DAG from data using BCD. When the ordering is unknown, the BCD algorithm may still give an accurate estimate of the row correlations that are invariant to node-wise permutations according to Lemma \ref{lem:cor}. The estimated row correlation matrix can then be used to de-correlate the data so that traditional DAG learning algorithms would be applicable. We will demonstrate this idea of de-correlation with numerical results as well.

\subsection{Simulated Networks}

We first perform experiments on simulated networks for both ordered and unordered cases. To apply the BCD algorithm, we need to set values for $\lambda_1$ and $\lambda_2$ in \eqref{obj}. Since the support of $\Theta^*$ is restricted to $G^*$, we simply fixed $\lambda_2$ to a small value ($\lambda_2=0.01$) in all the experiments. For each data set, we computed a solution path from the largest $\lambda_{1\max}$, for which we get an empty DAG, to $\lambda_{1\min} = \lambda_{1\max}/100$. The optimal $\lambda_1$ was then chosen by minimizing the BIC score over the DAGs on the solution path. 

We generated random DAGs with $p$ nodes and fixed the total number of edges $s_0$ in each DAG to $2p$. The entries in the weighted adjacency matrix $B^*$ of each DAG were drawn uniformly from  $[-1,-0.1]\cup [0.1,1]$, and $\omega_j^*$'s were sampled uniformly from $[0.1,2]$. In our simulations of $\Theta^*$, we first considered networks with a clustering structure, i.e., $\Theta^*$ was block-diagonal as in \eqref{block_diag}. We fixed the size of the clusters to $20$ or $30$, and within each cluster, the individuals were correlated according to the following four covariance structures.
\begin{itemize}
    \item Toeplitz: $\Sigma_{ij}^* = 0.3^{|i-j|/5}$.
    % \item Exponential decay: $\Theta_{ij}^* = 0.3^{|i-j|/5}$.
    \item Equal correlation: $\Sigma_{ij}^* = 0.7$ if $i\neq j$, and $\Sigma_{ii}^* = 1$.
    \item Star-shaped: $\Theta_{1j}^* = \Theta_{i1}^* = a, i,j\geq 2, a \in (0,1)$, and $\Theta_{ii}^* = 1$. 
    \item Autoregressive (AR): $\Theta^*_{ij} = 0.7^{|i-j|} \ \text{if} \ |i-j| \leq \lceil b/ 4\rceil$; $\Theta^*_{ij} = 0 \ \text{otherwise}$, where $b$ is the cluster size. 
\end{itemize}

Toeplitz covariance structure implies that the observations are correlated as in a Markov chain. Equal correlation structure represents the cases when all observations are fully connected in a cluster. Star-shaped and AR structures capture intermediate dependence levels. Besides these block-diagonal covariances, we also considered a more general covariance structure defined through \textit{stochastic block models} (SBM), in which $G^*$ consists of several clusters and nodes within a cluster have a higher probability to be connected than those from different clusters. More explicitly, we generated $\Theta^*$ as follows:
\begin{enumerate}
    \item Let $\mathcal{B}_1,\ldots, \mathcal{B}_L$ be $L$ clusters with varying sizes that form a partition of $\{1,\ldots, n\}$, where the number of clusters $L$ ranges from 5 to 10 in our experiments.  
    Define a probability matrix $P\in \mathbb{R}^{n\times n}$ where 
    $P_{ij} = 0.5$ if $i,j \in \mathcal{B}_l, l\in\{1,\ldots, L\}$; otherwise, $P_{ij} = 0.1$. 
    \item Construct the adjacency matrix $A$ of $G^*$:
    $$A_{ij} \sim \text{Bern}(P_{ij}).$$
    \item Sample $\Theta_{ij}^\prime \sim \text{Unif}[-5,5]$ if $A_{ij} = 1$. Otherwise, $\Theta_{ij}^\prime=0$. To ensure a positive-definite $\Theta^*$, we then perform the following transformations to get $\Theta^*$:
    \begin{align}\label{pdtheta}
    \begin{split}
        \widetilde{\Theta} &= (\Theta^\prime + \Theta^{\prime^\top} )/2\\
        \Theta^* &= \widetilde{\Theta} - \left(\sigma_{\min}(\widetilde{\Theta}) - 0.01\right) \cdot I_n
    \end{split}
    \end{align}
\end{enumerate}
Under the stochastic block model, two nodes from different clusters in $G^*$ are connected with probability 0.1, so $\Theta^*$ is not block-diagonal in general. As explained in Section \ref{sec:b_theta}, our proposed BCD algorithm does not require $\Theta^*$ to be block-diagonal in practice to produce accurate estimates of $B^*$ and $\Theta^*$. Our numerical experiments will confirm this theory and demonstrate the robustness of the BCD method.

We compared the BCD algorithm with its competitors under both high-dimensional ($p > n$) and low-dimensional ($p < n$) settings with respect to DAG learning.
% : we set $(n,p)$ to be either $(200, 400)$ or $(150, 300)$ for $p > n$ and $(n, p) = (200, 100)$ for $p < n$. 
For each $(n,p)$ and each type of covariances, we simulated 10 random DAGs and then generated one data set following equation (\ref{eq:model}) for each DAG. Thus, we had 10 results for each of the $2\times 5 = 10$ simulation settings. In the end, we averaged the results over the 10 simulations under each setting for comparison. 

\subsubsection{Learning with given ordering}
This subsection provides additional results for the simulation studies in Section 5.2.1 in the paper. 
Assuming the nodes in the DAG are sorted according to a given topological ordering, we compared our BCD algorithm against a baseline setting which fixes $\Theta^* = I_n$. In other words, the baseline algorithm ignores the dependencies among observations when estimating the DAG with BCD. The block sizes in $\Theta^*$ were set to $20$ in all cases except SBM whose block sizes ranged from 5 to 25. Among other estimates, both algorithms return an estimated weighted adjacency matrix $\widehat B$ for the optimal $\lambda_1$ selected by BIC. For the BCD algorithm, we use $\widehat B$ and $\widehat\Theta$ for $\widehat B^{(\infty)}$ and $\widehat\Theta^{(\infty)}$ after convergence (see Algorithm~\ref{alg1}). Note that, since $\widehat\Theta^{(0)}$ is initialized to $I_n$ by default in the BCD algorithm, the estimated $\widehat{B}$ from the baseline algorithm is the same as the estimate $\widehat{B}^{(1)}$ from BCD after one iteration. 

We also included the Kronecker graphical Lasso (KGLasso) algorithm \citep{2009arXiv0906.3465A, tsiligkaridis2013convergence} mentioned in Section \ref{sec:model} in our comparison, which estimates both $\widehat\Psi$ and $\widehat\Theta$ via graphical Lasso in an alternating fashion. When estimating $\Theta^*$, KGLasso also makes use of its block-diagonal structure. After KGLasso converges, we perform Cholesky factorization on $\widehat\Psi = (I - \widehat{B})^{-\top}\widehat\Omega(I - \widehat{B})^{-1}$  according to the given ordering to obtain $\widehat B$ and $\widehat\Omega$. A distinction between BCD and KGLasso is that KGLasso imposes a sparsity regularization on $\Psi$ instead of $B$, so the comparison between these two will highlight the importance of imposing sparsity directly on the Cholesky factor. 

Given the estimate $\widehat B$ from a method,  we hard-thresholded the entries in $\widehat B$ at a threshold value $\bar{\tau}$ to obtain an estimated DAG.
To compare the three methods, we chose $\bar{\tau}$ such that they predicted roughly the same number of edges (E). Then we calculated the number of true positives (TP), false positives (FP) and false negatives (FN, missing edges), and two overall accuracy metrics: Jaccard index (TP / (FP + $s_0$)) and structural Hamming distances (SHD = FP+FN). Note that, there were no reserved edges (i.e., estimated edges whose orientation is incorrect) because the ordering of the nodes was given. Detailed comparisons are summarized in Table \ref{tab:ordered1(n<p)} and Table \ref{tab:ordered2(n>p)}. In general, the BCD algorithm outperformed the competitors by having more true positives and less false positives in every case. Because the KGLasso method does not impose sparsity directly on the DAG structure, it suffered from having too many false negatives after thresholding when $p > n$. When $p < n$, the correlations between observations had a more significant impact on the estimation accuracy for DAGs. As a result, BCD and KGLasso which take this correlation into account performed better than the baseline. In particular, BCD substantially reduced the number of missing edges (FNs) and FDR, compared to the baseline. Both BCD and KGLasso yielded accurate estimates of $\widehat{\Theta}$ when $n < p$. When $n > p$, as the sample size $p$ for estimating $\Theta^*\in\mathbb{R}^{n\times n}$ decreased relative to the dimension $n$, $\widehat\Theta$ became less accurate. The difference in the accuracy of $\widehat\Theta=\widehat\Theta^{(\infty)}$ and $\widehat\Theta^{(1)}$ was not significant.

\begin{table}
\centering
\resizebox{1\columnwidth}{!}{
\begin{tabular}{clc|crrccrc}
  \toprule
  $\Theta$-Network  &Method& ($n,p,s_0$) & E & FN & TP & FDR & JI & SHD & err($\widehat\Theta$) (err($\widehat\Theta^{(1)}$))\\
  \midrule
  & BCD & (150, 300, 600) & 686.2 & 214.0 &  386.0 & 0.355& 0.443& 514.2& 0.00034 (0.00032)\\
  equi-cor& Baseline & (150, 300, 600) & 642.4 & 240.3& 359.7 & 0.383&0.410& 523.0 &  ---\\
  & KGLasso &(150, 300, 600) &756.2& 504.9& 95.1&  0.822&    0.080 & 1166.0 &  0.00019\\ 
  \\
  & BCD& (200, 400, 800) & 535.0 & 306.4& 493.6& 0.077 & 0.586 & 347.8 & 0.00143 (0.00833)\\
  toeplitz &Baseline&(200, 400, 800) &549.5&425.2& 374.8&0.317  & 0.384 &  599.9 & ---\\
  &KGLasso & (200, 400, 800) &550.6&  634.3& 165.7&    0.698&    0.139 & 1019.2 &  0.01617\\
  \\ 
   &BCD  &(200, 400, 800) & 543.1 &301.1&498.9& 0.081&0.591&345.3 & 0.00006 (0.00051)\\
   star& Baseline  &(200, 400, 800) & 515.7& 338.3&461.7 & 0.103 &  0.541 & 392.3 & ---\\
   & KGLasso  &(200, 400, 800) &495.8& 504.3& 295.7&0.403& 0.295& 704.4 & 0.00233\\
   \\
%   &BCD & (150, 300, 600)& 562.2 &203.0&397.0& 0.283&            0.521&368.2 & 0.28466 (0.40801)\\
%   exp\_decay& Baseline &(150, 300, 600)& 463.9& 264.0& 336.0& 0.271&    0.462&  391.9 & ---\\
%   &KGLasso  & (150, 300, 600)&484.9&344.7& 255.3& 0.445& 0.311 &574.3 & 1.41436\\
  &BCD& (100, 200, 400)& 253.1&193.8&206.2& 0.184& 0.461& 240.7 & 0.00247 (0.00219)\\
  AR(5)&Baseline& (100, 200, 400)& 245.8&  208.9& 191.1&0.217& 0.420& 263.6  & ---\\
  &KGLasso& (100, 200, 400)& 270.4& 271.1& 128.9& 0.521&   0.237& 412.6 & 0.02587\\
  \\
  &BCD & (100, 300, 600)&  343.3 & 313.6 & 286.4 & 0.159 & 0.435 & 370.5 & 0.51266 (0.52297)\\
  SBM& Baseline &(100, 300, 600)&344.6 & 338.4 & 261.6 & 0.233 & 0.383 & 421.4 &--- \\
  &KGLasso  & (100, 300, 600)&301.3 & 510.7 & 89.3 & 0.696 & 0.110 & 722.7 & 0.46201 \\
  \bottomrule
\end{tabular}}
\caption{Results for ordered DAGs on simulated data when $n < p$. The last column shows the $\ell_2$-estimation errors of $\widehat{\Theta}$ and $\widehat\Theta^{(1)}$ normalized by the true support size. The numbers in the brackets are errors after one iteration of BCD. Each number corresponds to the average over 10 simulations.}
\label{tab:ordered1(n<p)}
\end{table}

\begin{table}
\centering
\resizebox{1\columnwidth}{!}{
\begin{tabular}{clc|crrccrc}
  \toprule
  $\Theta$-Network& Method& ($n,p,s_0$) & E & FN & TP & FDR & JI & SHD & err($\widehat\Theta$) (err($\widehat\Theta^{(1)}$))\\
  \midrule
  & BCD& (200, 100, 200) & 143.0 & 75.1 & 124.9 & 0.119 & 0.570 & 93.2 & 0.00211 (0.00199)\\
   equi-cor&Baseline & (200, 100, 200) & 140.0 & 103.8 & 96.2 & 0.305 & 0.394 & 147.6 & ---\\
   &KGLasso & (200, 100, 200) & 149.0 & 129.5&  70.5& 0.501&   0.255& 208.0 &0.00207\\ 
  \\
  &BCD & (200, 100, 200) & 167.1 & 56.7 & 143.3 & 0.137 & 0.639 & 80.5 & 0.51735 (0.68703)\\
  toeplitz& Baseline& (200, 100, 200)& 166.7 & 104.8 & 95.2 & 0.425 & 0.351 & 176.3 & ---\\
   &KGLasso &(200, 100, 200) & 158.6&  70.6& 129.4& 0.176& 0.564&  99.8 & 0.85023\\
  \\
  & BCD& (200, 100, 200) & 186.9 & 50.6 & 149.4 & 0.199 & 0.629 & 88.1 & 0.54769 (0.39316)\\
 star &Baseline & (200, 100, 200) & 171.7 & 69.2 & 130.8 & 0.236 & 0.543 & 110.1 & ---\\
   & KGLasso    &(200, 100, 200) &183.0& 66.9& 133.1&  0.271&0.532&116.8 & 0.18472\\
  \\
%  &BCD& (200, 100, 200) & 152.2 & 69.7 & 130.3 & 0.143 & 0.587 & 91.6 & 1.60270 (1.48587)\\
%  exp\_decay &Baseline& (200, 100, 200) & 154.7 & 73.8 & 126.2 & 0.179 & 0.552 & 102.3 & ---\\
% &KGLasso&(200, 100, 200)&148.9&72.1& 127.9& 0.139& 0.578& 93.1 & 0.04273\\
  &BCD & (200, 100, 200) & 185.9 & 52.2 & 147.8 & 0.200 & 0.622 & 90.3  & 0.01644 (0.01184)\\
  AR(5)&Baseline & (200, 100, 200) & 180.1 & 66.2 & 133.8 & 0.253 & 0.545 & 112.5& ---\\
  &KGLasso& (200, 100, 200)& 177.0 & 62.7 & 137.3&   0.215&0.574& 102.4 & 0.01038\\
  \\
  &BCD& (300, 100, 200) &  140.1 & 72.2 & 127.8 & 0.086 & 0.602 & 84.5 & 0.31189 (0.31448)\\
 SBM &Baseline& (300, 100, 200) &  139.9 & 85.7 & 114.3 & 0.181 & 0.507 & 111.3 & ---\\
&KGLasso&(300, 100, 200) &128.8 & 161.1 & 38.9 & 0.689 & 0.134 & 251.0& 0.32263\\
  \bottomrule
\end{tabular}}
\caption{Results for ordered DAGs on simulated data when $n > p$.}
\label{tab:ordered2(n>p)}
\end{table}

\begin{figure}
    \centering
    \begin{adjustbox}{minipage=\linewidth,scale=0.9}
        \begin{subfigure}[t]{0.2\textwidth}
        \centering
        \includegraphics[width=\textwidth]{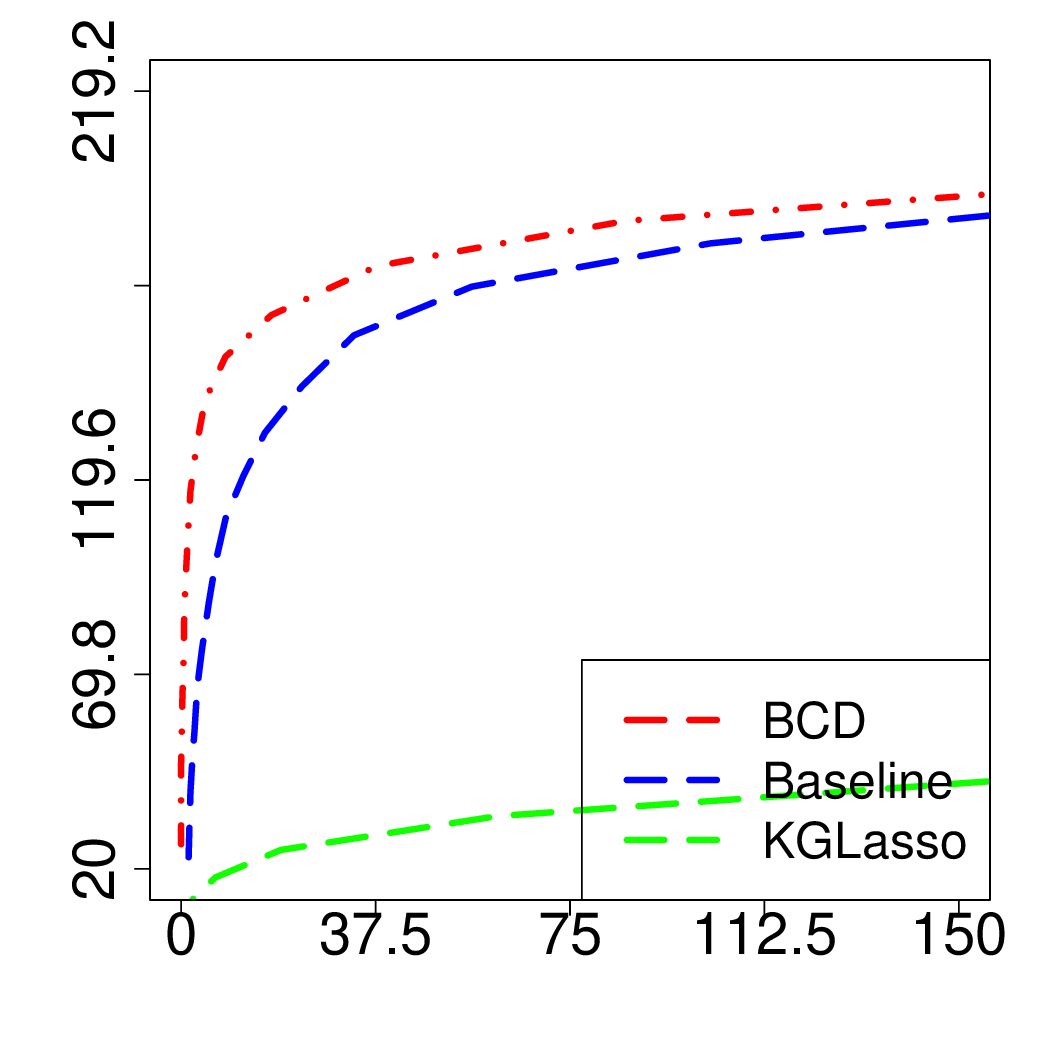}
        \caption{Equal Corr}
    \end{subfigure}%
    \begin{subfigure}[t]{0.2\textwidth}
        \centering
        \includegraphics[width=\textwidth]{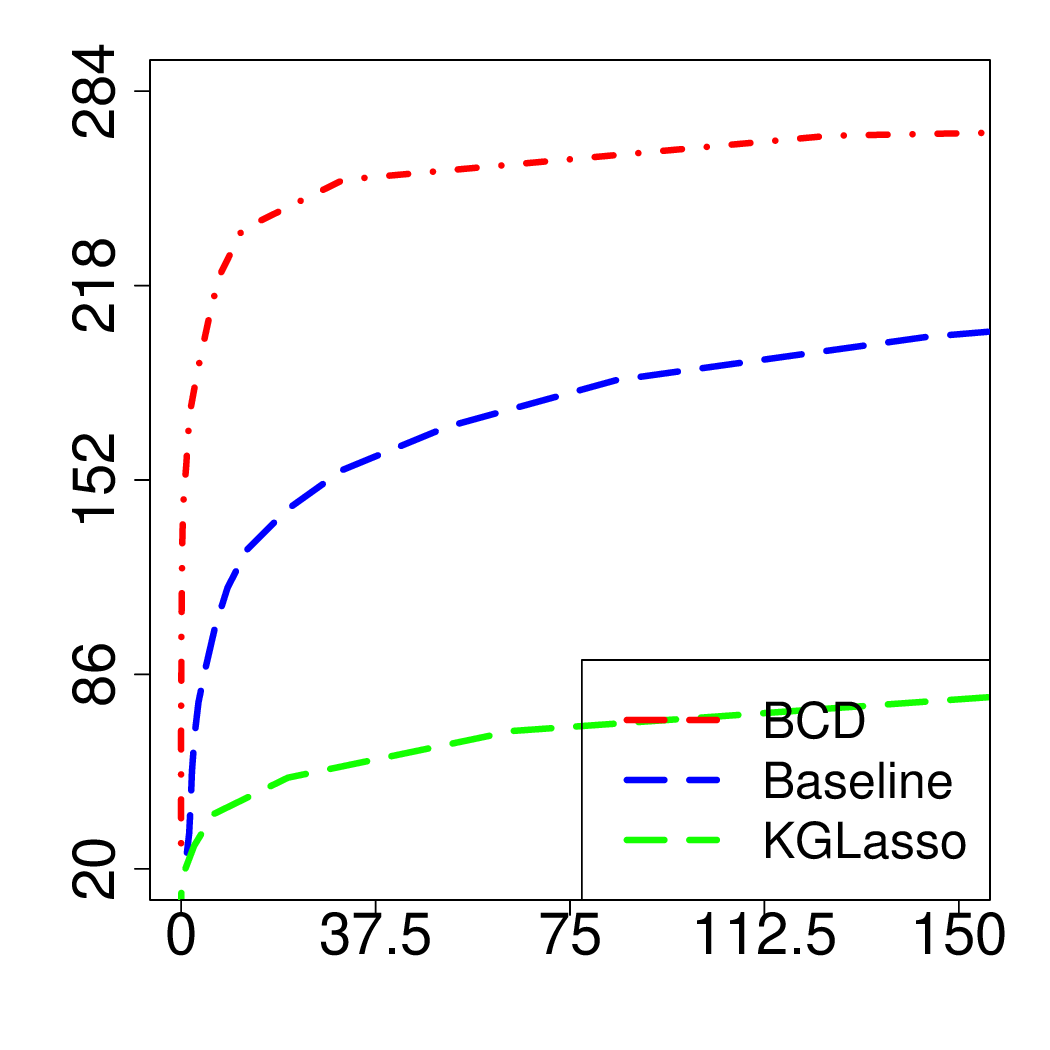}
        \caption{Toeplitz}
    \end{subfigure}%
    \begin{subfigure}[t]{0.2\textwidth}
        \centering
        \includegraphics[width=\textwidth]{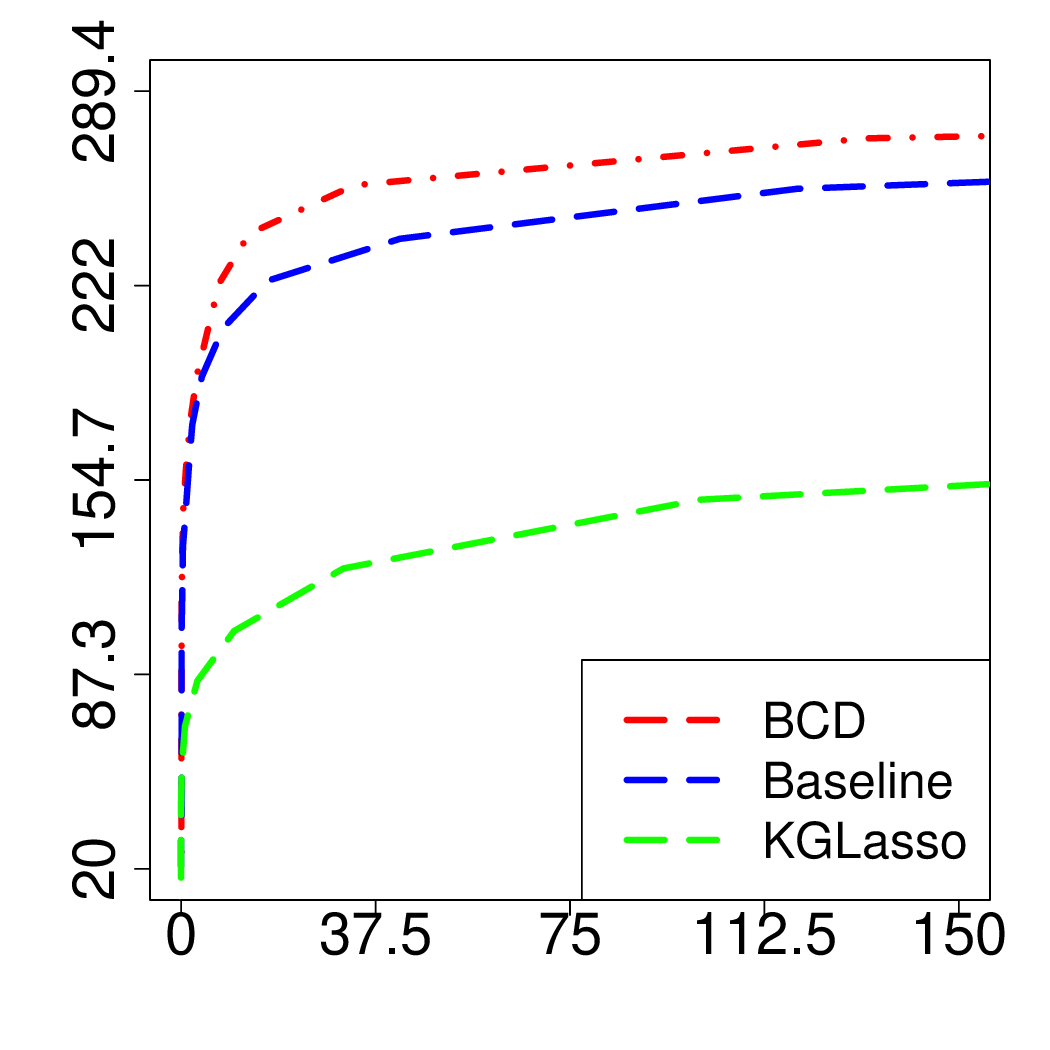}
        \caption{Star}
    \end{subfigure}%
    % \begin{subfigure}[t]{0.2\textwidth}
    %     \centering
    %     \includegraphics[width=\textwidth]{figures/ordered/007.eps}
    %     \caption{Exp. Decay}
    % \end{subfigure}%
     \begin{subfigure}[t]{0.2\textwidth}
        \centering
        \includegraphics[width=\textwidth]{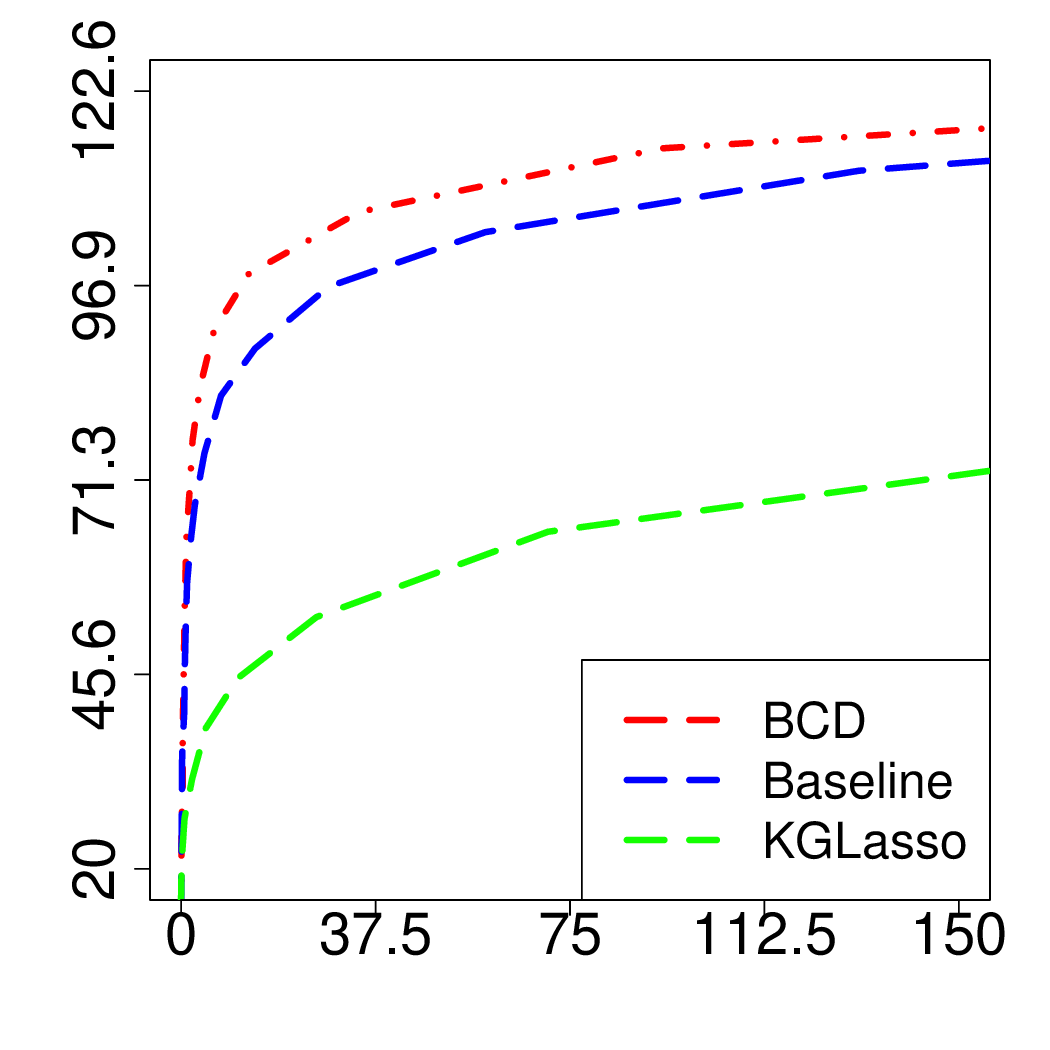}
        \caption{AR}
    \end{subfigure}%
   \begin{subfigure}[t]{0.2\textwidth}
        \centering
        \includegraphics[width=\textwidth]{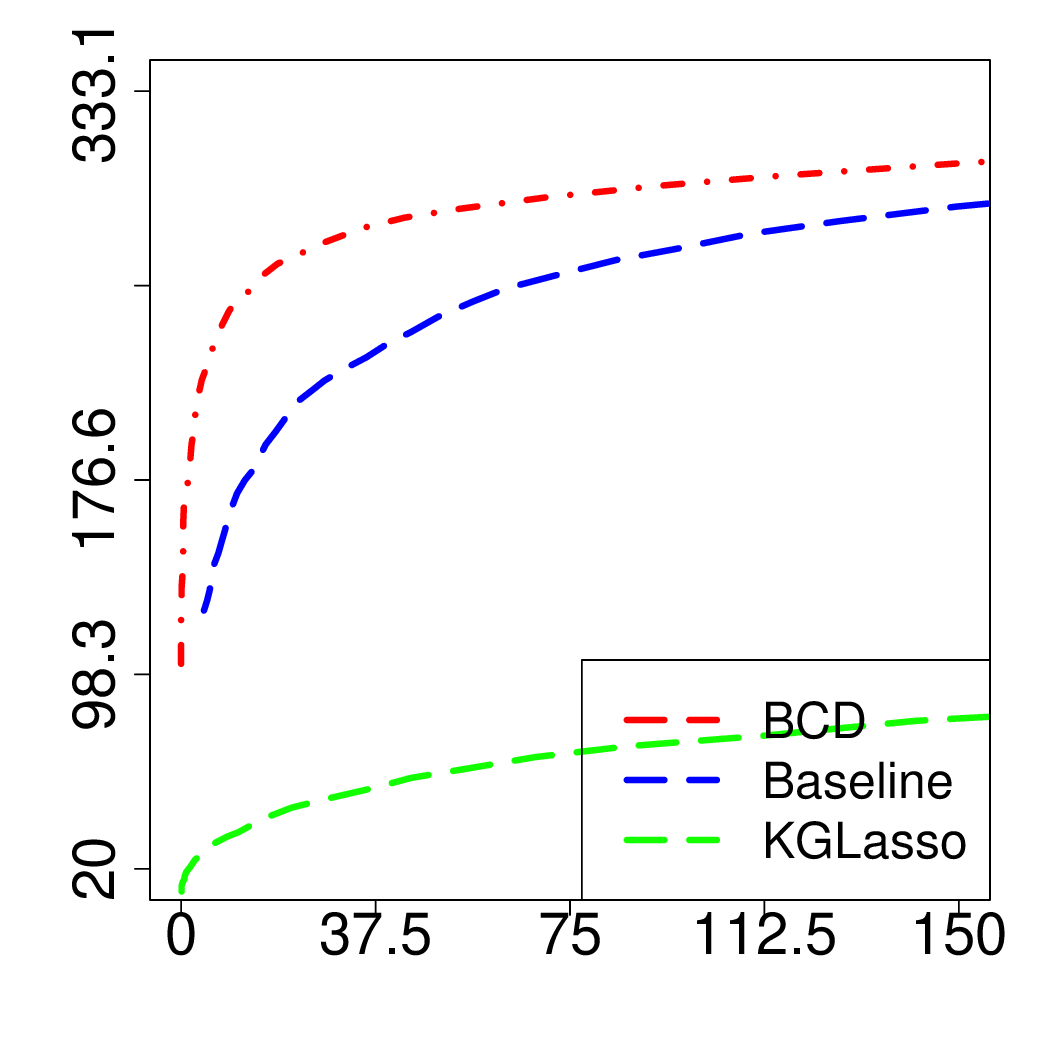}
        \caption{SBM}
    \end{subfigure}%
    
    \begin{subfigure}[t]{0.2\textwidth}
        \centering
        \includegraphics[width=\textwidth]{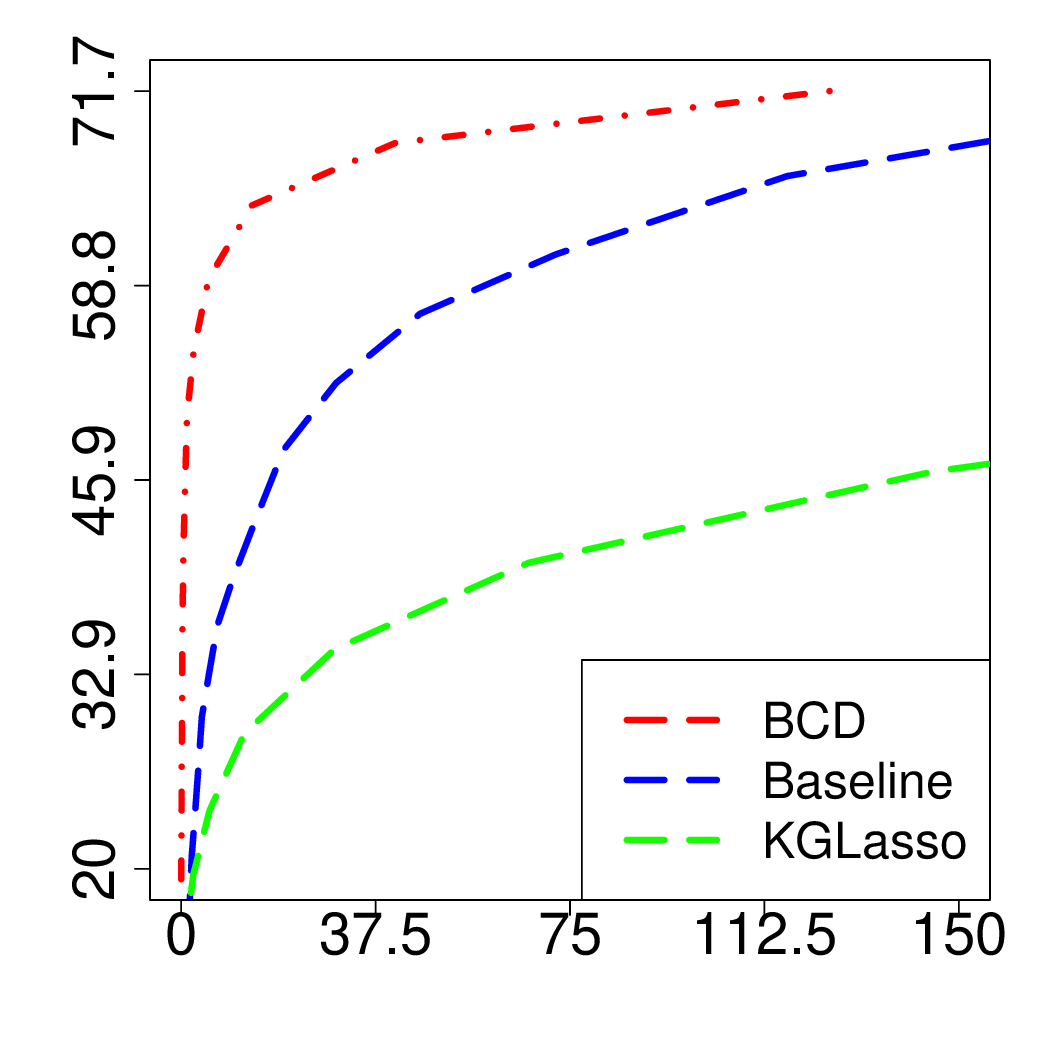}
        \caption{Equal Corr}
    \end{subfigure}%
    \begin{subfigure}[t]{0.2\textwidth}
        \centering
        \includegraphics[width=\textwidth]{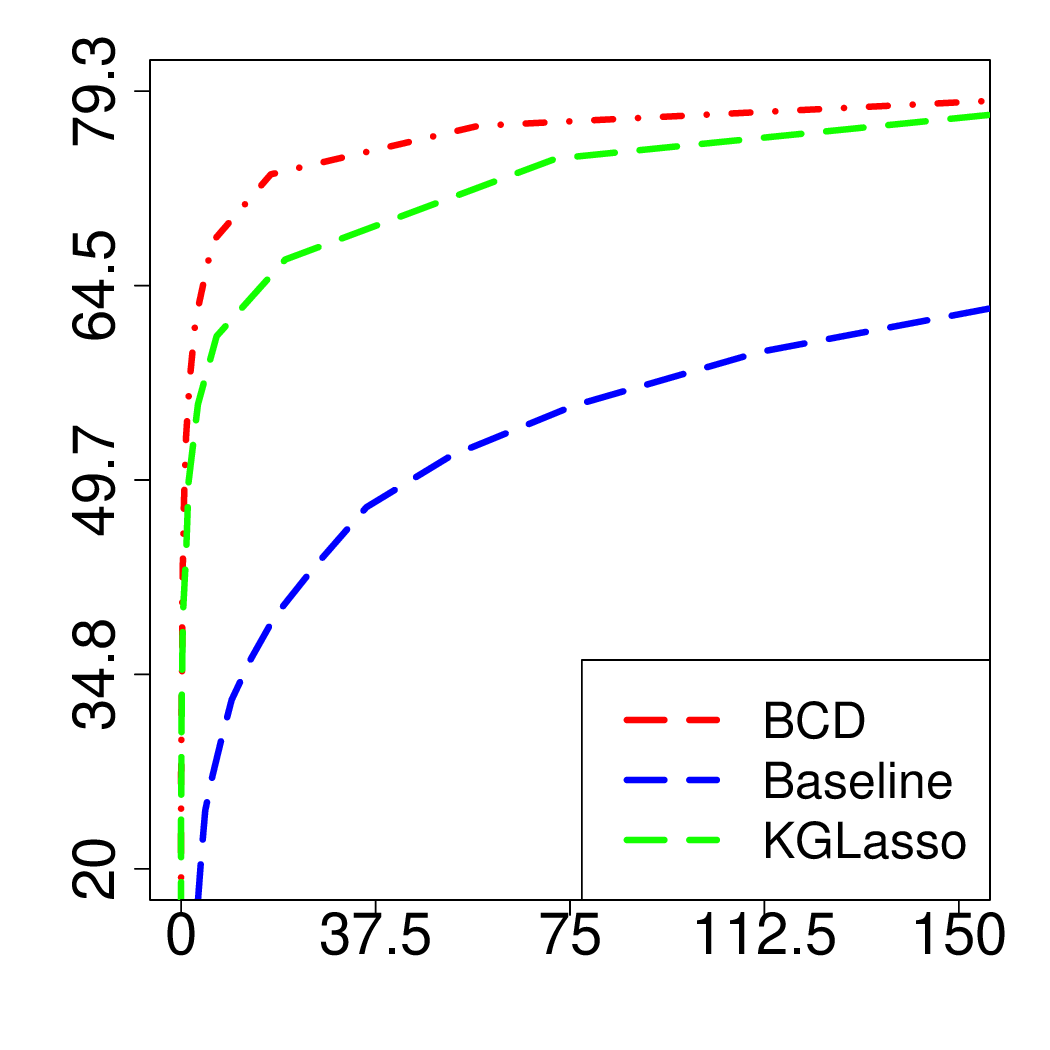}
        \caption{Toepltiz}
    \end{subfigure}%
    \begin{subfigure}[t]{0.2\textwidth}
        \centering
        \includegraphics[width=\textwidth]{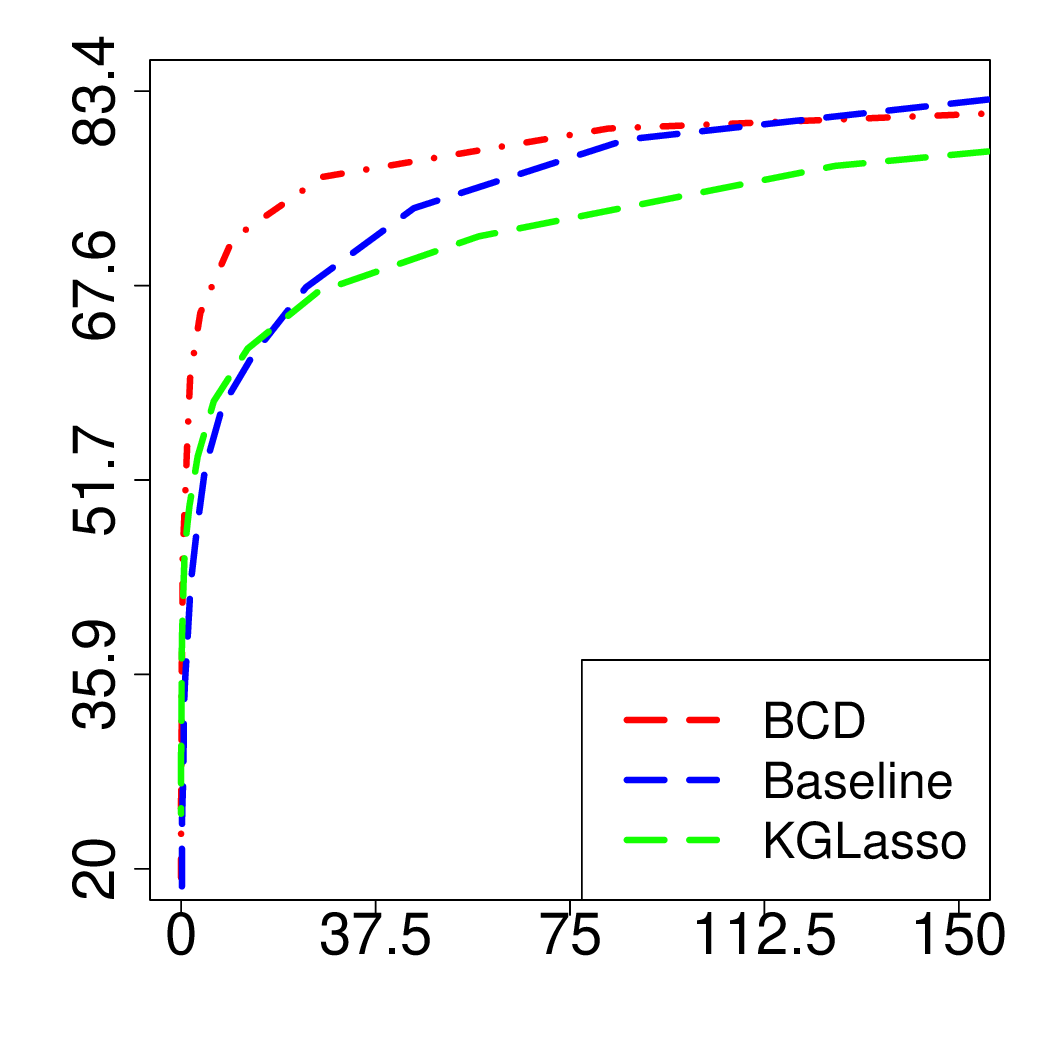}
        \caption{Star}
    \end{subfigure}%
    % \begin{subfigure}[t]{0.2\textwidth}
    %     \centering
    %     \includegraphics[width=\textwidth]{figures/ordered/121005.eps}
    %     \caption{Exp. Decay}
    % \end{subfigure}%
    \begin{subfigure}[t]{0.2\textwidth}
        \centering
        \includegraphics[width=\textwidth]{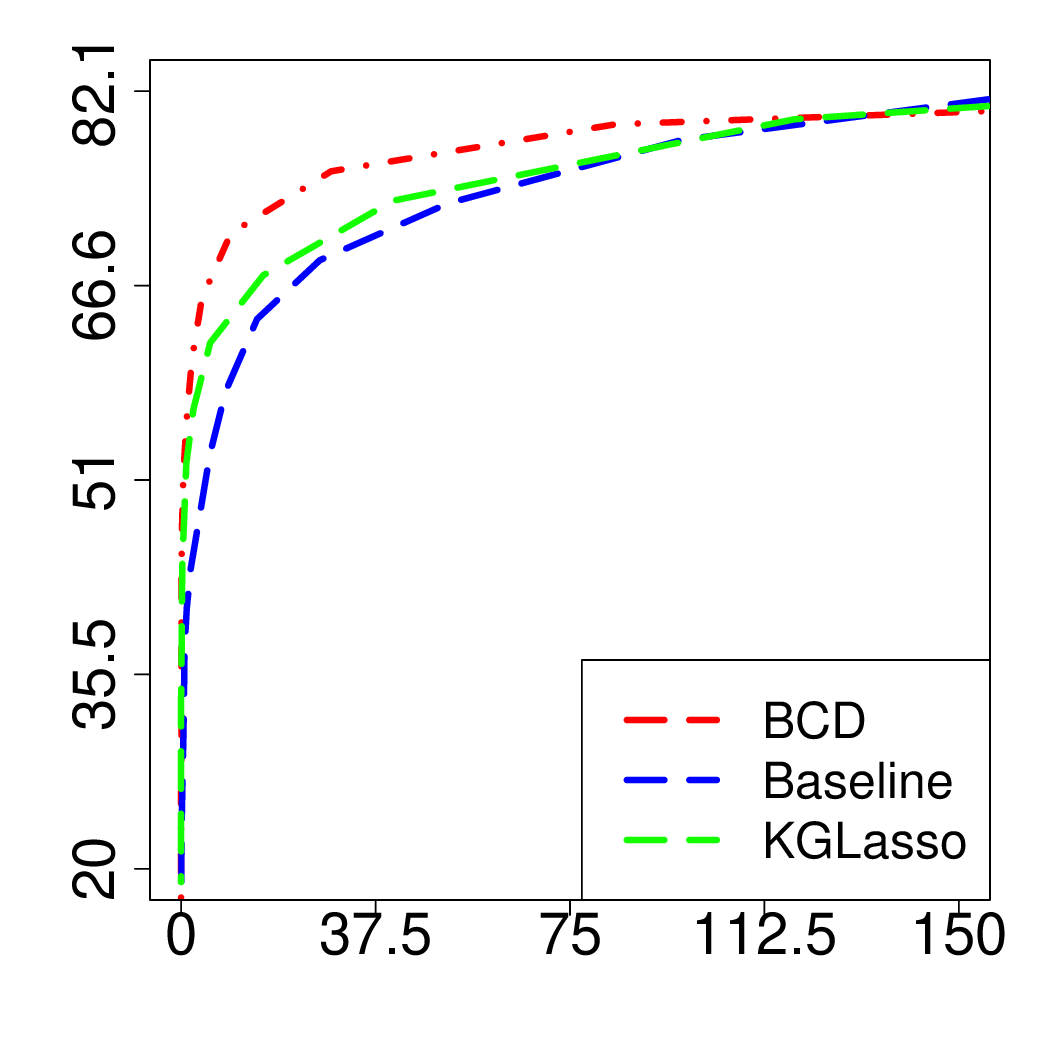}
        \caption{AR}
    \end{subfigure}%
    \begin{subfigure}[t]{0.2\textwidth}
        \centering
        \includegraphics[width=\textwidth]{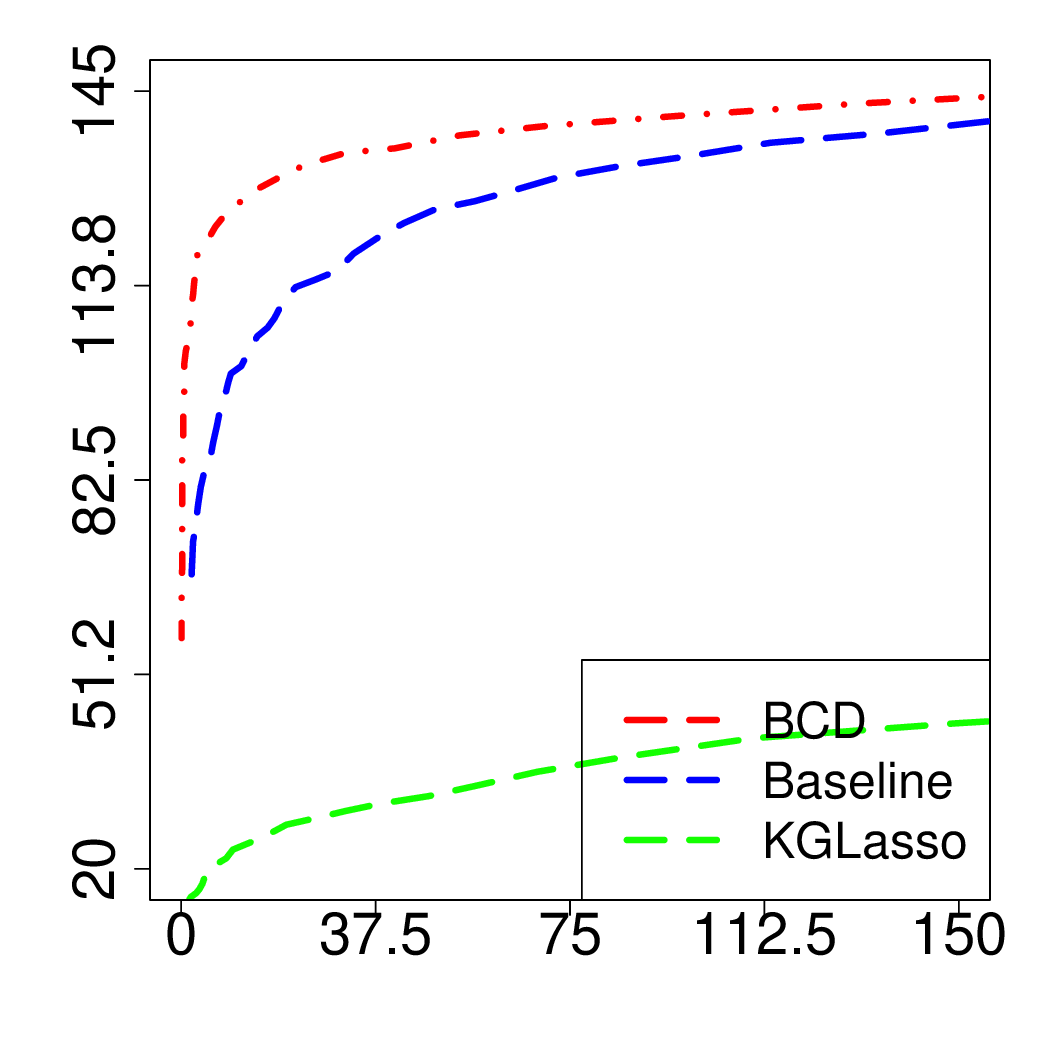}
        \caption{SBM}
    \end{subfigure}%
    \end{adjustbox}
    \caption{ROC curves of BCD, baseline, and KGLasso on simulated and sorted DAGs: x-axis reports the number of false positive edges and y-axis true positive edges. Top row: $n < p$. Bottom row: $n > p$. Each data point in the ROC curves corresponds to the average over 10 simulations.}
    \label{fig:roc}
\end{figure}

Figure \ref{fig:roc} shows the ROC curves of all three methods over a sequence of $\bar{\tau}$ under the 10 settings. The $\bar{\tau}$ sequence contains $30$ equally spaced values in $[0,0.5]$. The BCD algorithm uniformly outperformed the others in terms of the area under the curve (AUC) with substantial margins when $n < p$. When $n > p$, the BCD still did better than the other two most of the time but its lead over KGLasso and baseline was not as significant in some cases. This was largely due to insufficient regularization on $\widehat\Theta$. Fixing $\lambda_2 = 0.01$ in this case implies $\lambda_p = 0.01/p = 0.0001$ in the graphical Lasso step \eqref{glasso} of BCD, resulting in severe overestimates of the magnitude of the entries in $\Theta^*$. After we increased $\lambda_p$ to $0.1$ which is still quite small, the BCD indeed outperformed the other two methods by much larger margins. KGLasso also performed much better when $n >p$ as shown in Table \ref{tab:ordered2(n>p)} and Figure \ref{fig:roc}. This is expected because when $n$ is large compared to $p$, the dependence among individuals will have a larger impact on the accuracy of the estimation of DAGs. Since KGLasso is designed to iteratively estimate $\Theta^*$ and $\Psi^*$, the more accurate estimates of $\Theta^*$ as reported in Table \ref{tab:ordered2(n>p)} compensated for the relatively inaccurate $\widehat B$. 

{}

\begin{figure}
    \centering
    \begin{adjustbox}{minipage=\linewidth,scale=0.9}
     
    \begin{subfigure}[t]{0.2\textwidth}
        \centering
        \includegraphics[width=\textwidth]{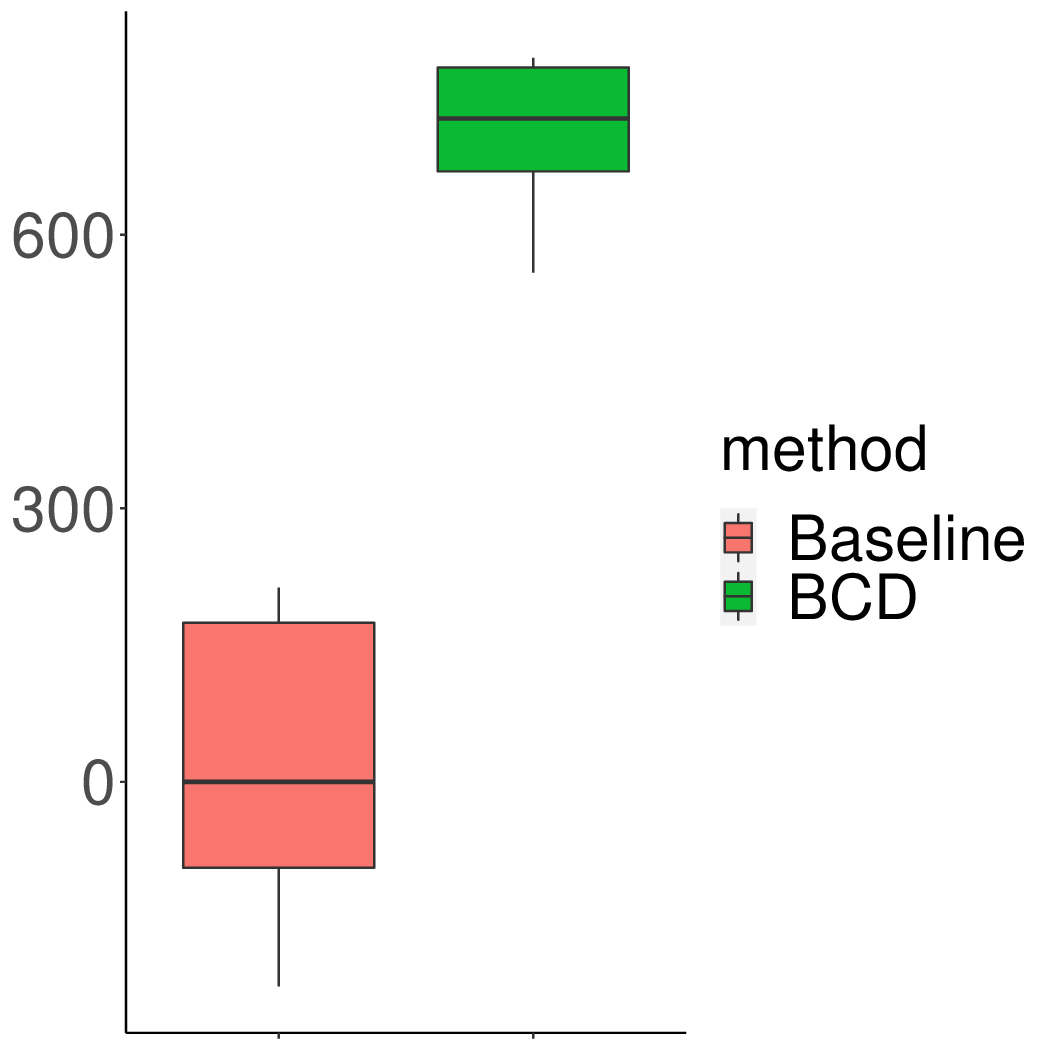}
        \caption{Equal Corr}
    \end{subfigure}%
    \begin{subfigure}[t]{0.2\textwidth}
        \centering
        \includegraphics[width=\textwidth]{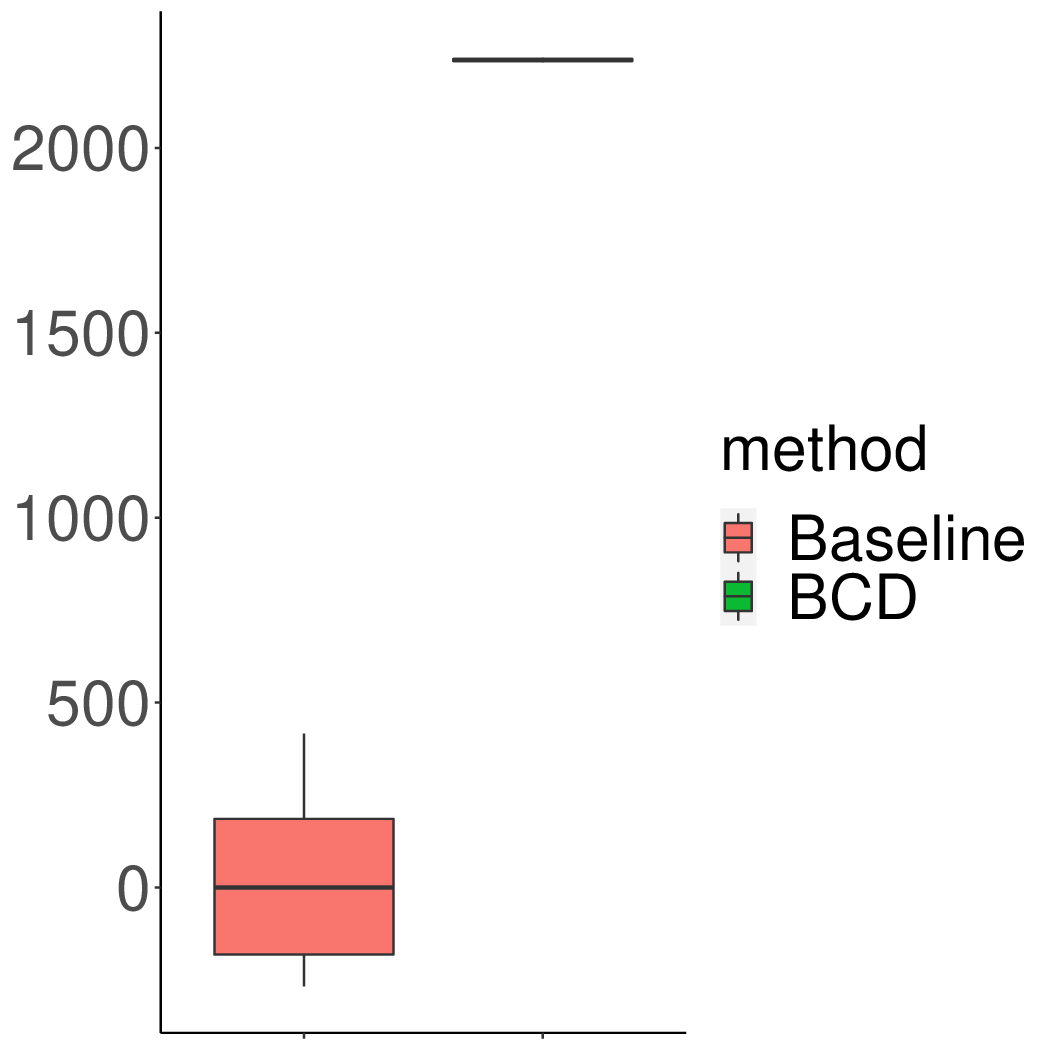}
        \caption{Toeplitz}
    \end{subfigure}%
    \begin{subfigure}[t]{0.2\textwidth}
        \centering
        \includegraphics[width=\textwidth]{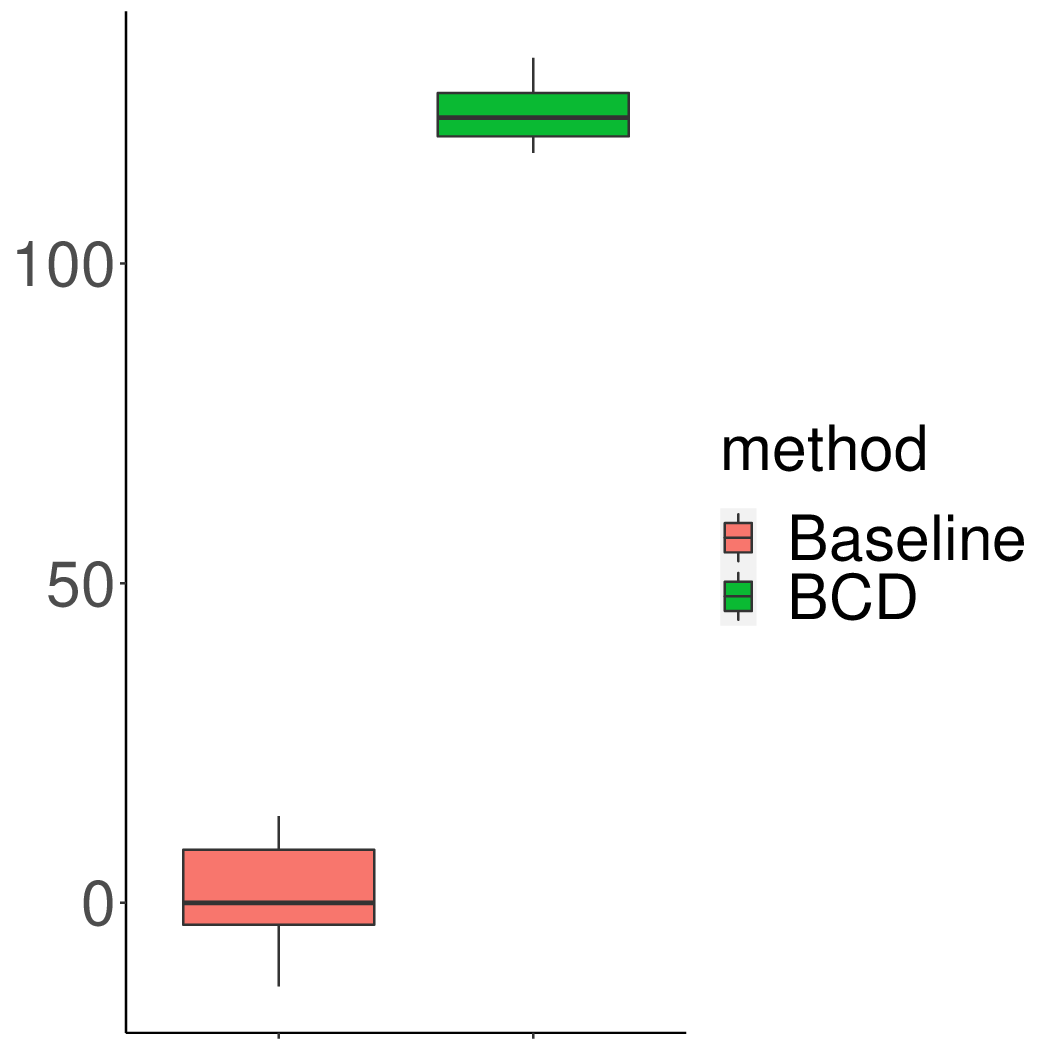}
        \caption{Star}
    \end{subfigure}%
    % \begin{subfigure}[t]{0.2\textwidth}
    %     \centering
    %     \includegraphics[width=\textwidth]{figures/ordered/testll/007-2testll.eps}
    %     \caption{Exp Decay}
    % \end{subfigure}%
    \begin{subfigure}[t]{0.2\textwidth}
        \centering
        \includegraphics[width=\textwidth]{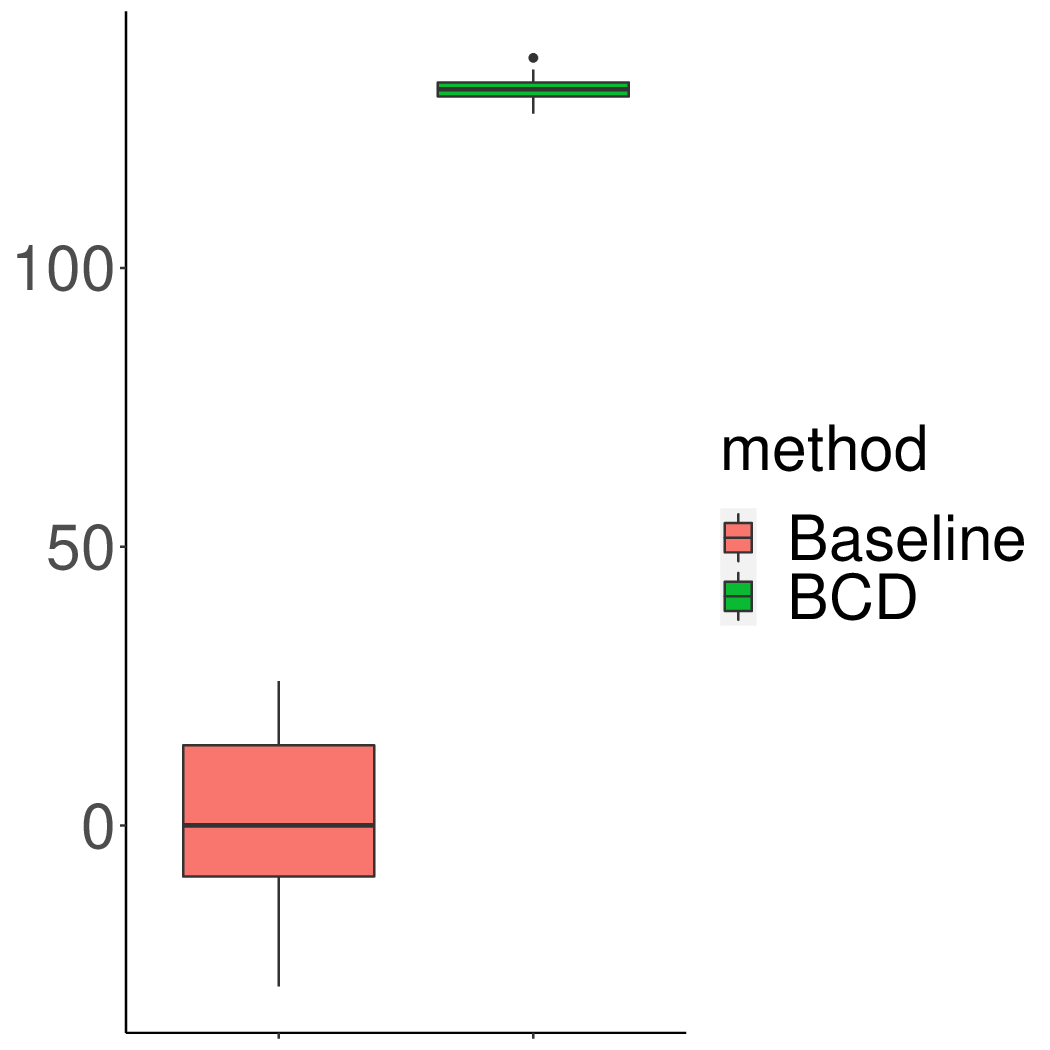}
        \caption{AR}
    \end{subfigure}%
     \begin{subfigure}[t]{0.2\textwidth}
        \centering
        \includegraphics[width=\textwidth]{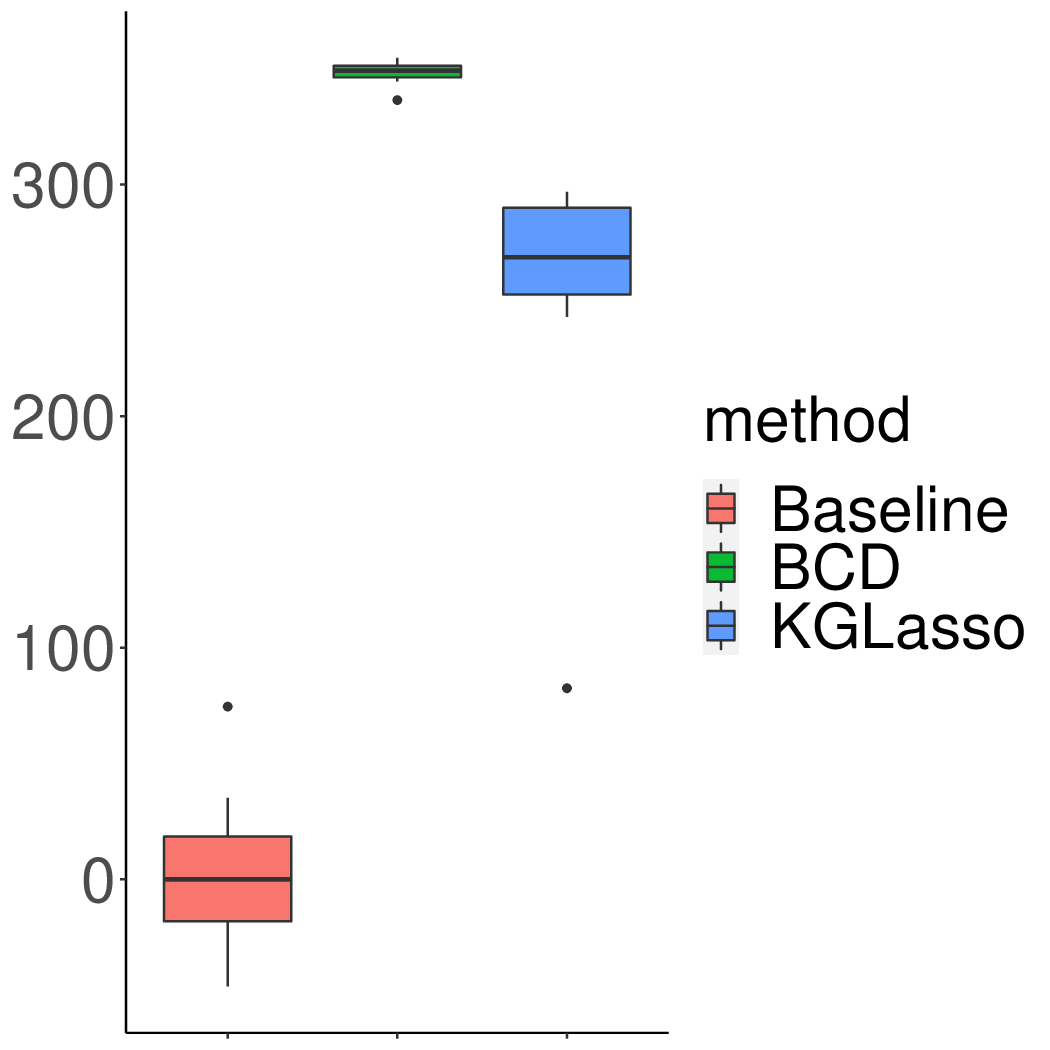}
        \caption{SBM}
    \end{subfigure}%
    
    \begin{subfigure}[t]{0.2\textwidth}
        \centering
        \includegraphics[width=\textwidth]{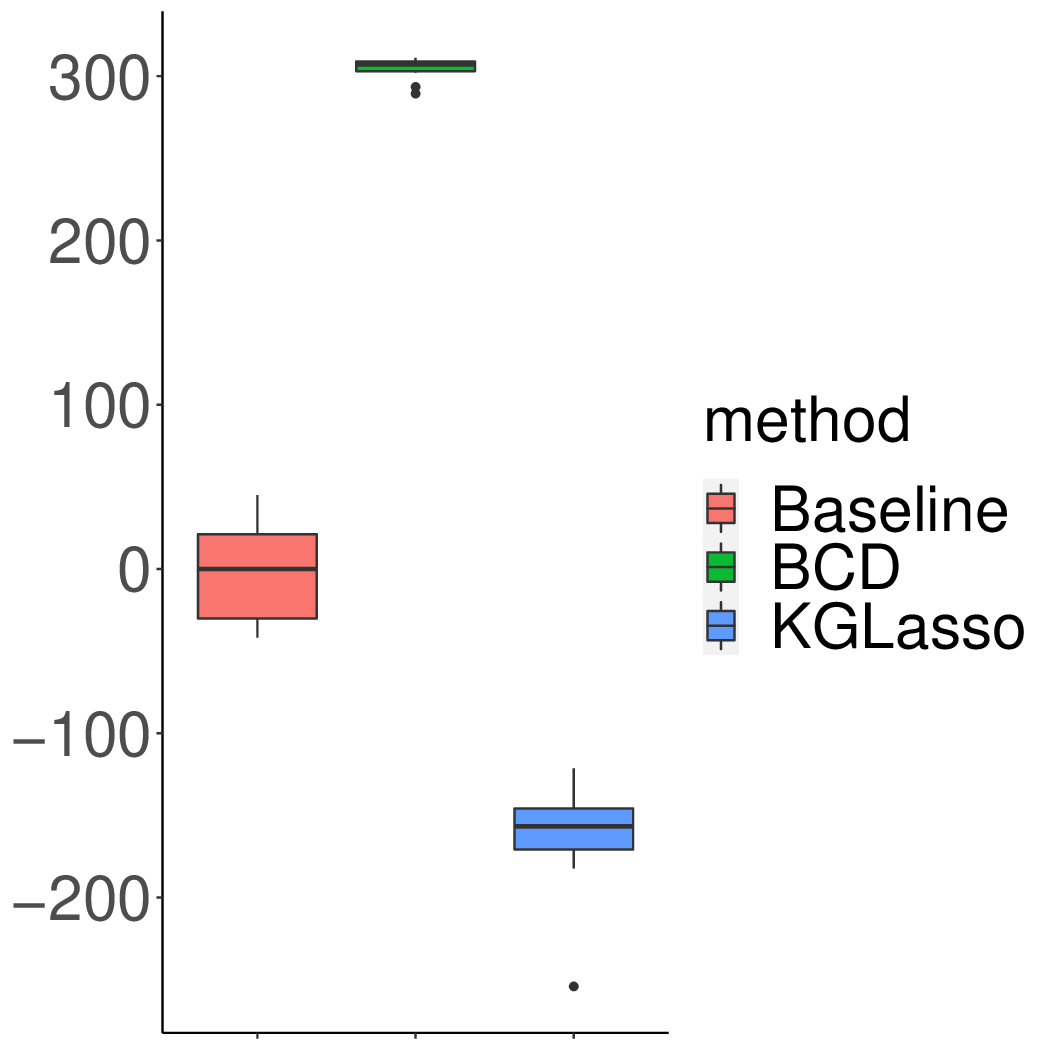}
        \caption{Equal Corr}
    \end{subfigure}%
    \begin{subfigure}[t]{0.2\textwidth}
        \centering
        \includegraphics[width=\textwidth]{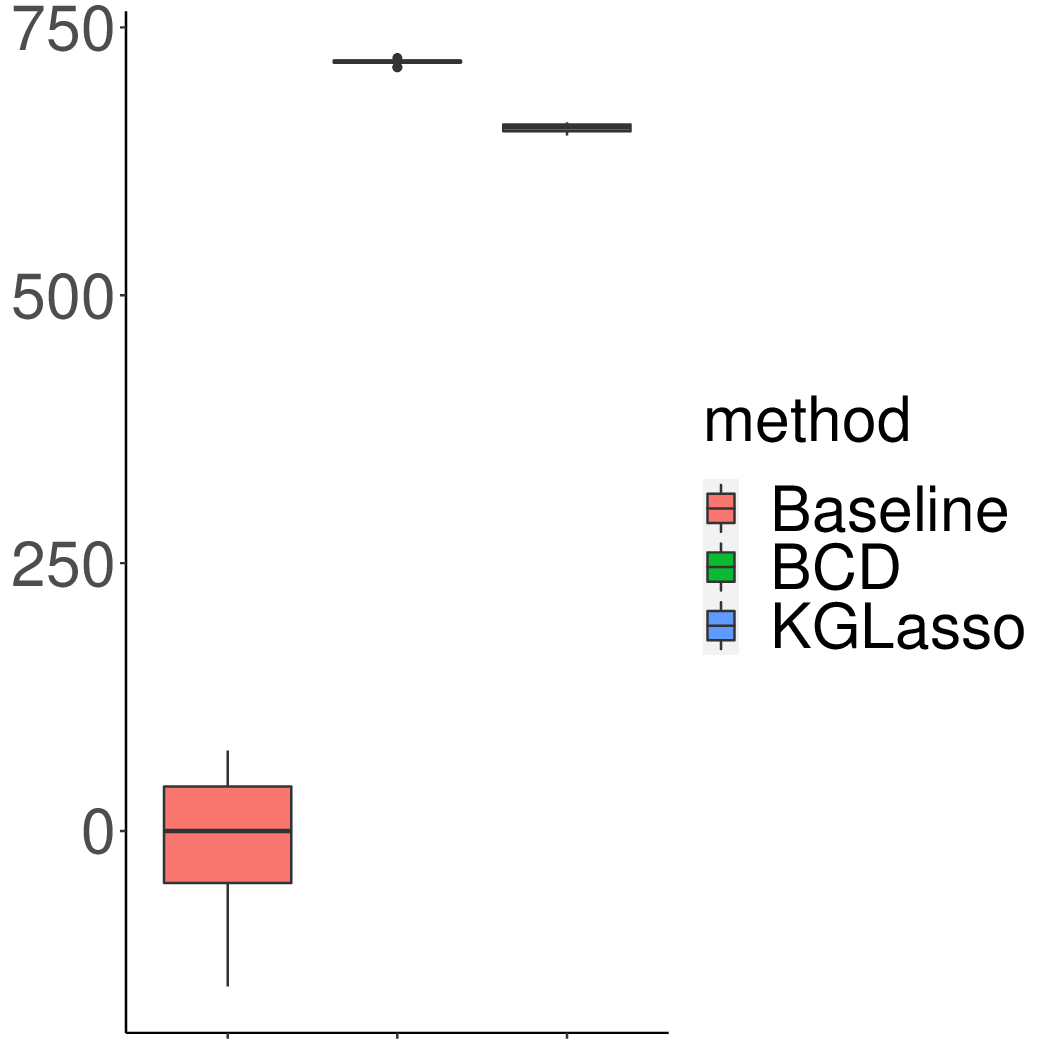}
        \caption{Toeplitz}
    \end{subfigure}%
    \begin{subfigure}[t]{0.2\textwidth}
        \centering
        \includegraphics[width=\textwidth]{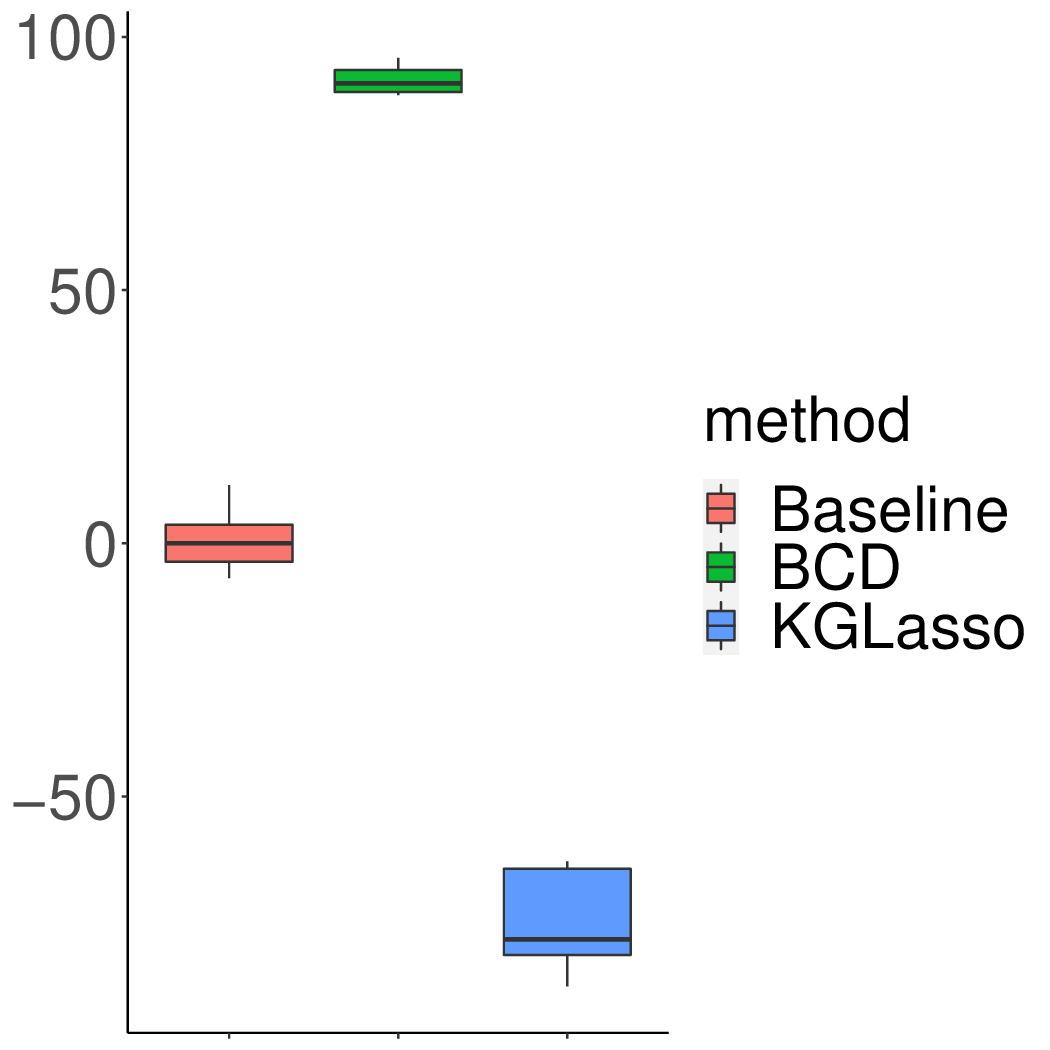}
        \caption{Star}
    \end{subfigure}%
    % \begin{subfigure}[t]{0.2\textwidth}
    %     \centering
    %     \includegraphics[width=\textwidth]{figures/ordered/testll/121005-2testll.eps}
    %     \caption{Exp Decay}
    % \end{subfigure}%
    \begin{subfigure}[t]{0.2\textwidth}
        \centering
        \includegraphics[width=\textwidth]{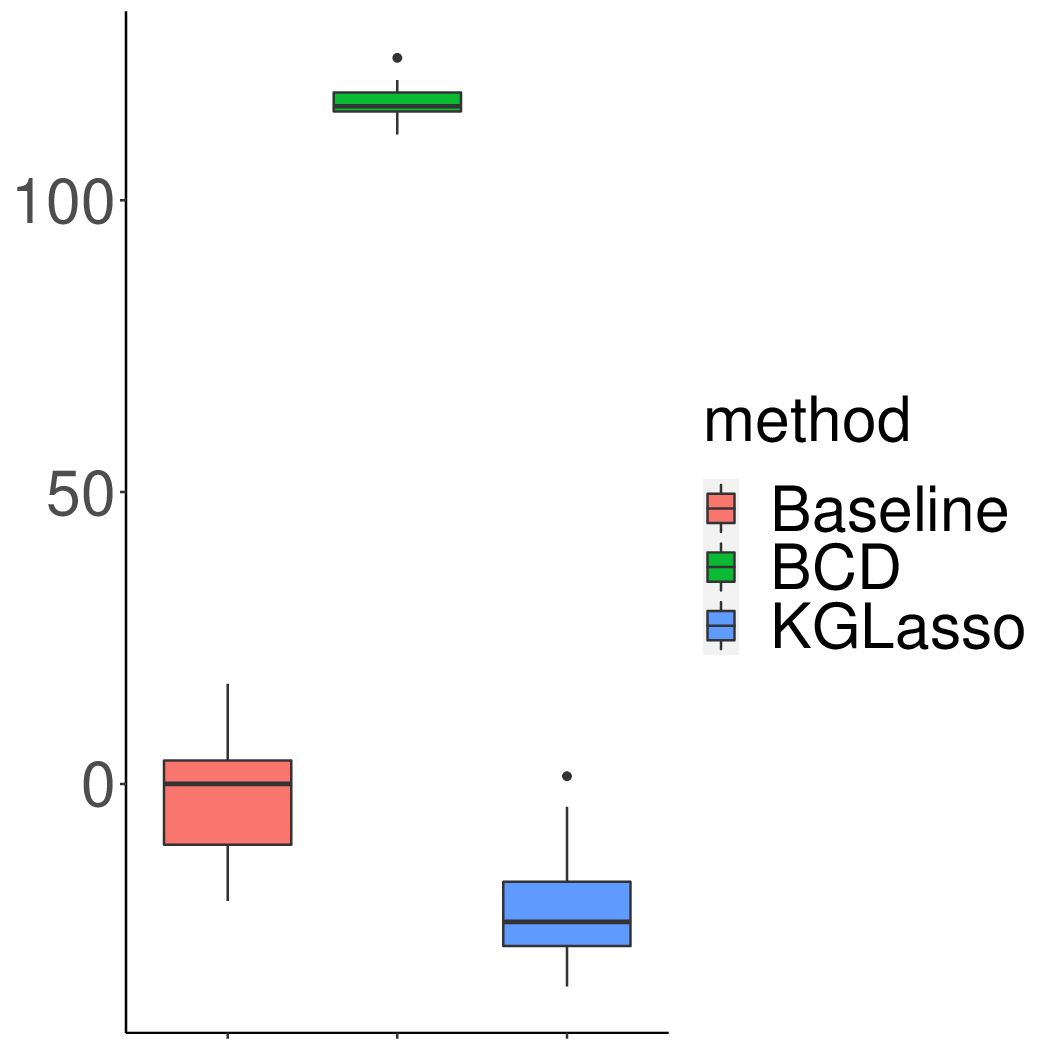}
        \caption{AR}
    \end{subfigure}%
    \begin{subfigure}[t]{0.2\textwidth}
        \centering
        \includegraphics[width=\textwidth]{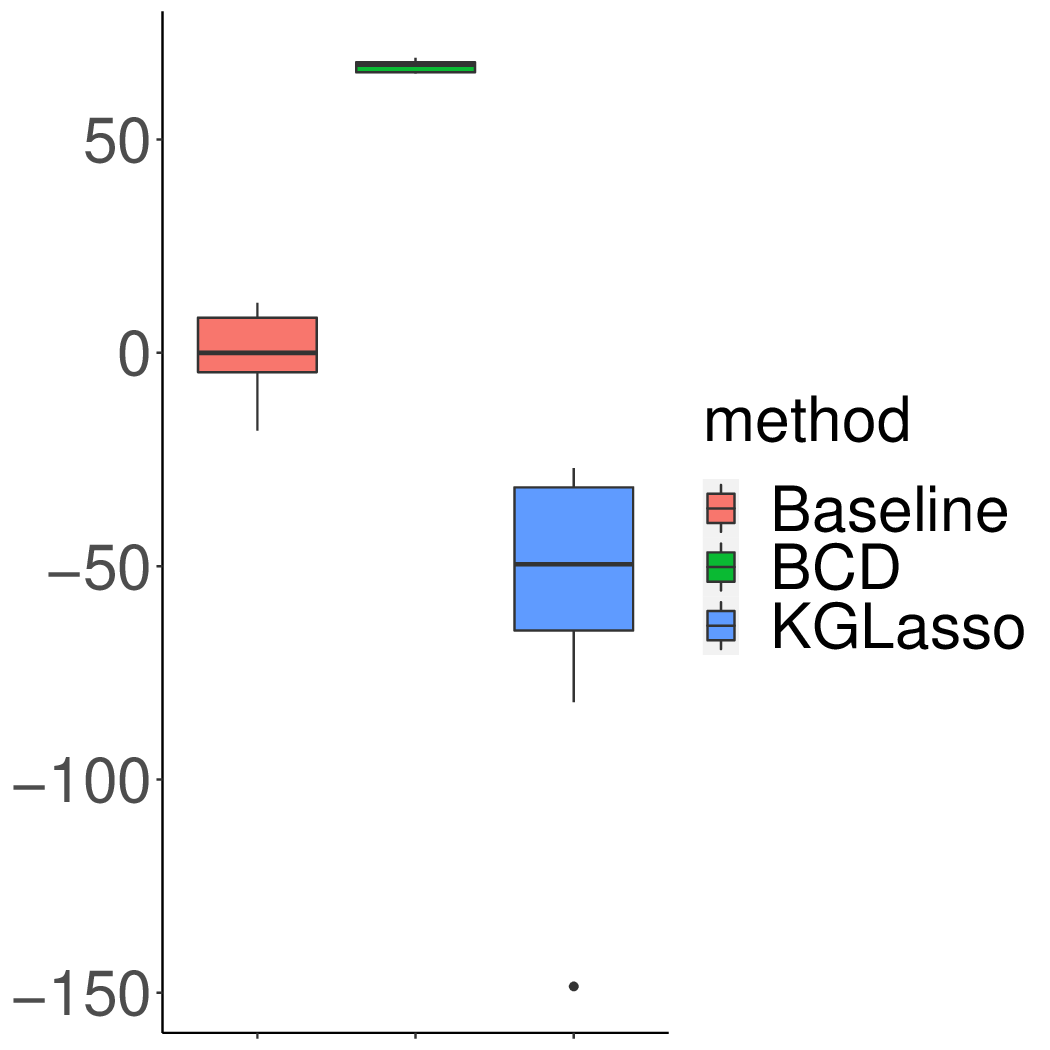}
        \caption{SBM}
    \end{subfigure}%
    \end{adjustbox}
    \caption{Normalized test data log-likelihood of BCD and baseline methods on simulated sorted DAGs. Top row: $n < p$. Bottom row: $n > p$. Each boxplot contains $10$ data points from the 10 repeated experiments.}
    \label{fig:testll2}
\end{figure}

We also compared test data log-likelihood among the three methods. Specifically, under each setting, we generated a test sample matrix $X_{\text{test}}$ from the true distribution for each of the 10 repeated simulations and computed $-L(\widehat B, \widehat\Theta, \widehat\Omega \mid X_{\text{test}})$ using the estimates from the three methods following equation \eqref{eq:lik}. Figure~\ref{fig:testll2} shows the boxplots of the test data log-likelihood, normalized by $\sqrt{np}$ after subtracting the median of the baseline method: $\mathcal{L}_{\text{plot}} = \left(\mathcal{L}_0 - \text{median}(\mathcal{L}_{0}^{\text{baseline}})\right) / \sqrt{np}$, where $\mathcal{L}_0$ is the original test data log-likelihood. The top row shows the test log-likelihood when $n < p$, where we did not include the data for KGLasso in four cases because its test data log-likelihood values were too small to fit in the same plot with the other two methods. The bottom row shows the results for $n > p$. For both cases, we see that the test data log-likelihood of the BCD method (in green) is consistently higher than that of the other methods.

\subsubsection{Learning with de-correlation}
\label{sec:decor}

When the natural ordering is unknown, we focus on estimating the row-wise covariance $\Sigma^*$. Given $\widehat\Sigma$ we can de-correlate the data by Equation \eqref{eq:dcor} and apply existing structural learning methods. In this study, we compared the performances of three structure learning methods before and after de-correlation: GES  \citep{Chickering:2003:OSI:944919.944933} and sparsebn \citep{sparsebn} which are score-based methods implemented respectively in the R packages \texttt{rcausal} \citep{Ramsey2017} and \texttt{sparsebn}, and PC \citep{spirtes2000causation} which is a constraint-based method implemented in \texttt{pcalg} \citep{pcalg}. All three methods rely on the independent data assumption, so we expect the de-correlation step to improve their performances significantly. Different from the previous comparison, the ordering of the nodes is unknown so GES and PC return an estimated CPDAG (completed acyclic partially directed graph) instead of a DAG. Thus, in the following comparisons, we converted both the estimated DAG from sparsebn and the true DAG to CPDAGs, so that all the reported metrics are computed with respect to CPDAGs. 

\begin{figure}
    \centering
    \begin{adjustbox}{minipage=\linewidth,scale=0.9}
    \begin{subfigure}[t]{0.2\textwidth}
        \centering
        \includegraphics[width=\textwidth]{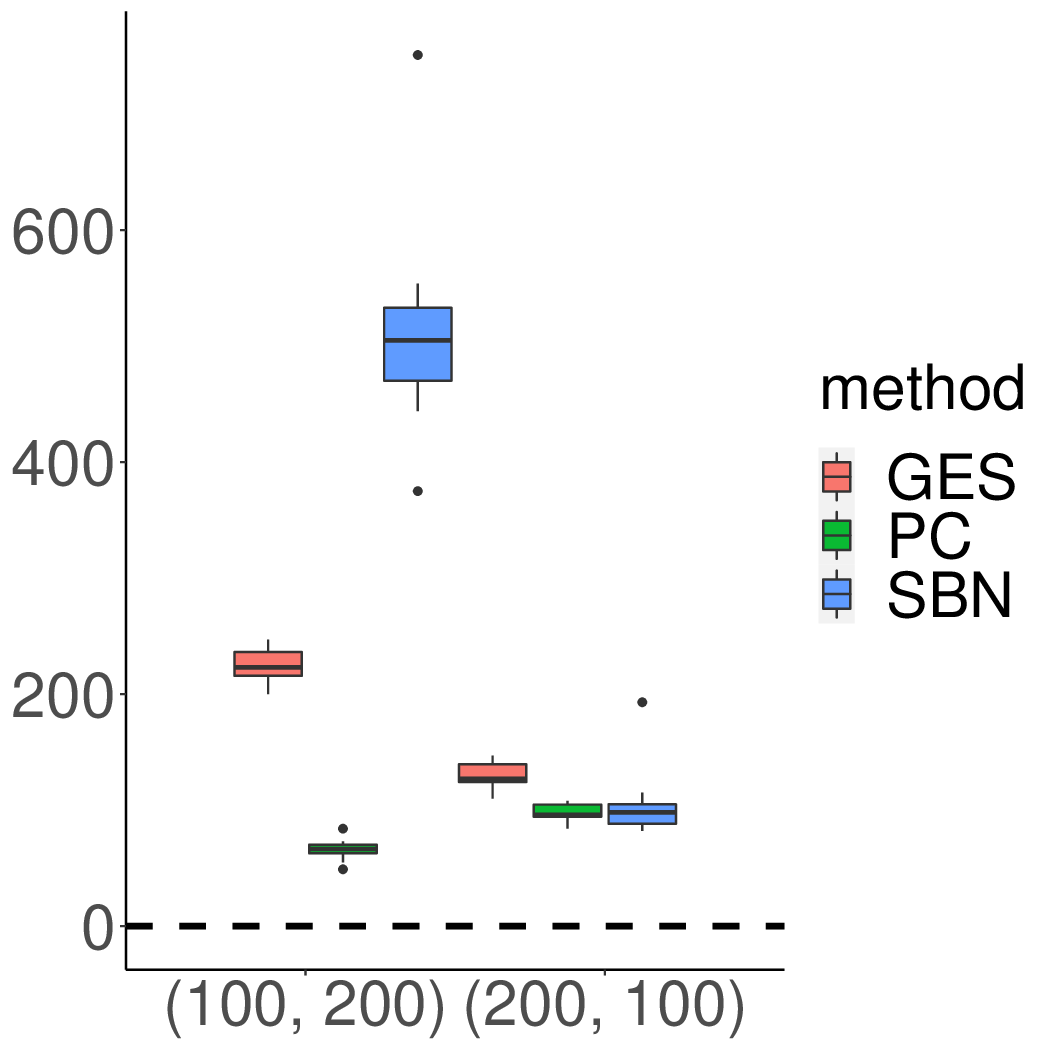}
        \caption{Equal Corr}
    \end{subfigure}%
    \begin{subfigure}[t]{0.2\textwidth}
        \centering
        \includegraphics[width=\textwidth]{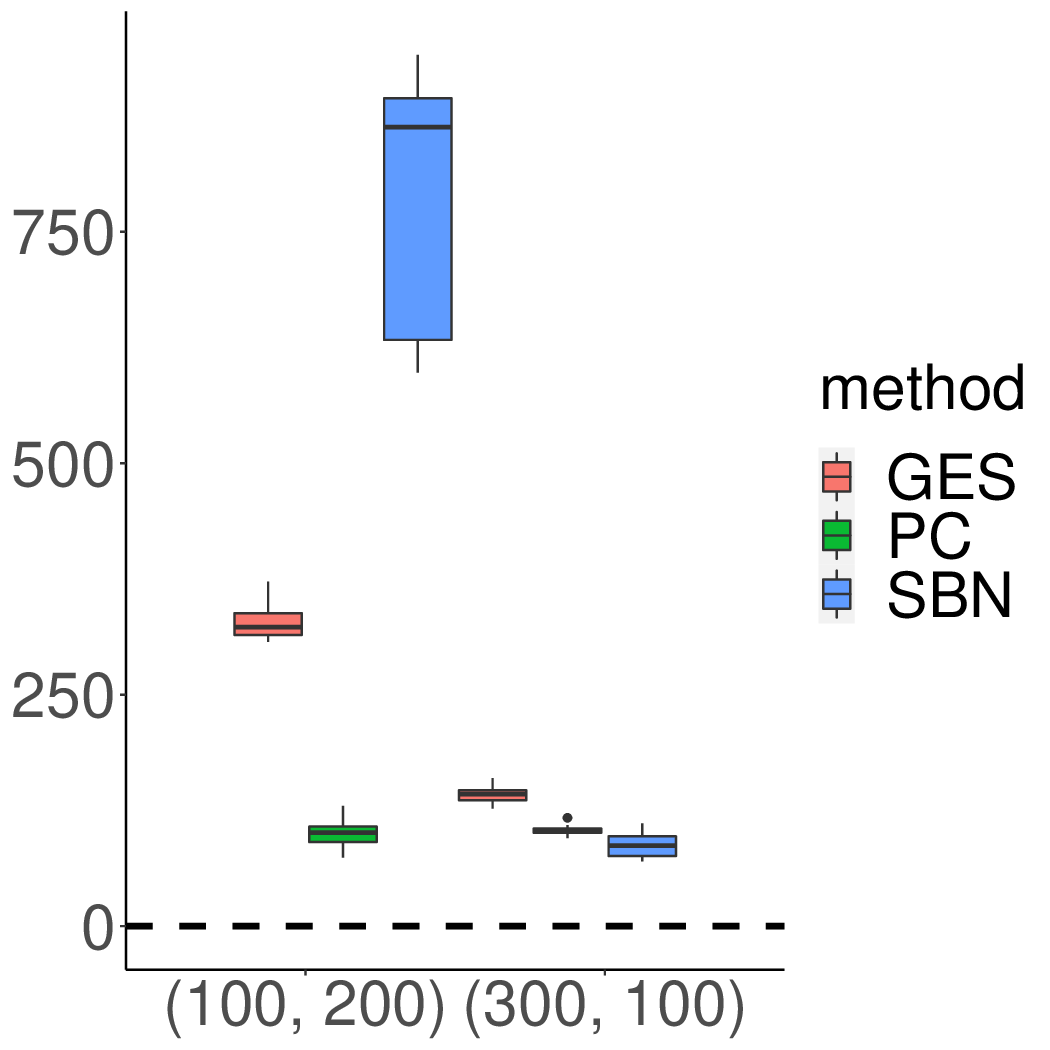}
        \caption{Toeplitz}
    \end{subfigure}%
    \begin{subfigure}[t]{0.2\textwidth}
        \centering
        \includegraphics[width=\textwidth]{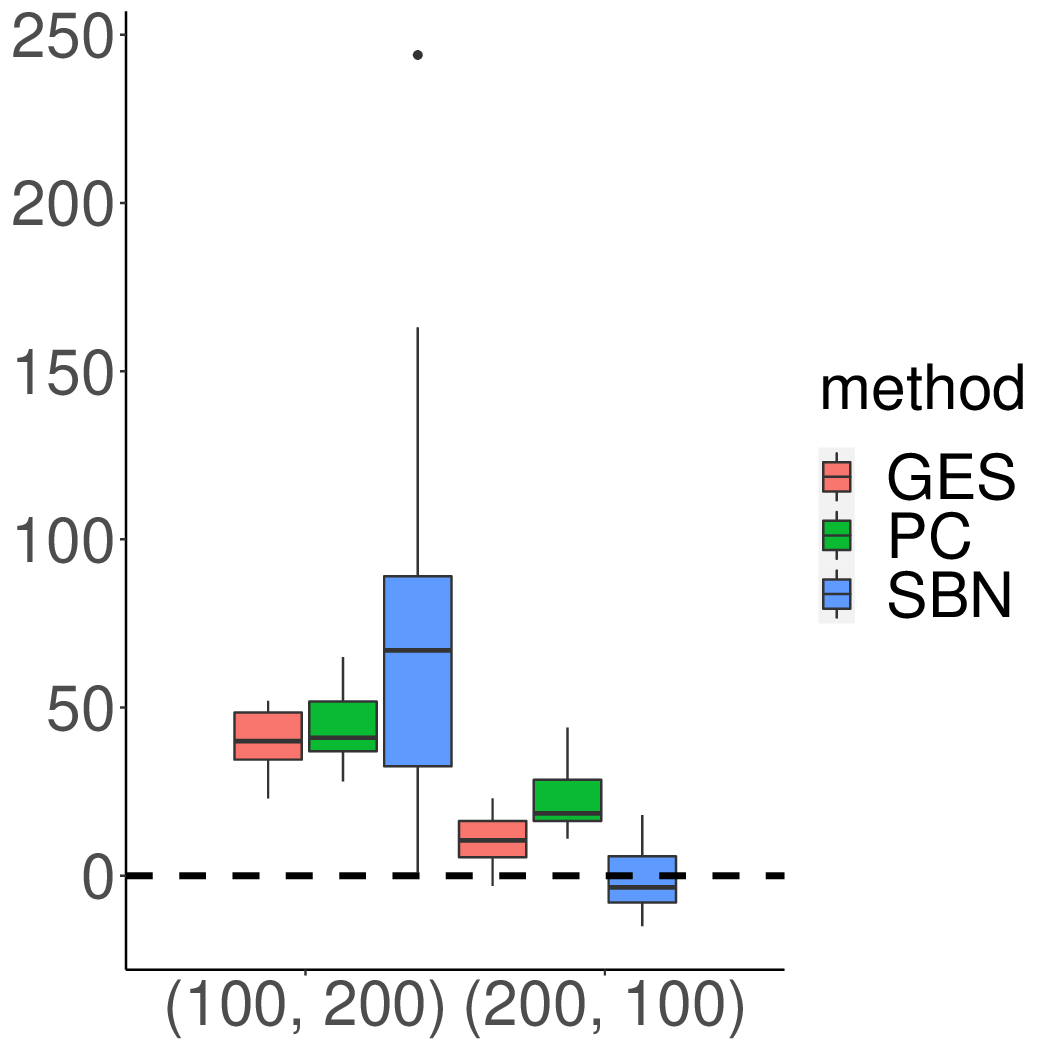}
        \caption{Star}
    \end{subfigure}%
    % \begin{subfigure}[t]{0.2\textwidth}
    %     \centering
    %     \includegraphics[width=\textwidth]{figures/unordered/ExpDecay-shd.eps}
    %     \caption{Exp Decay}
    % \end{subfigure}%
    \begin{subfigure}[t]{0.2\textwidth}
        \centering
        \includegraphics[width=\textwidth]{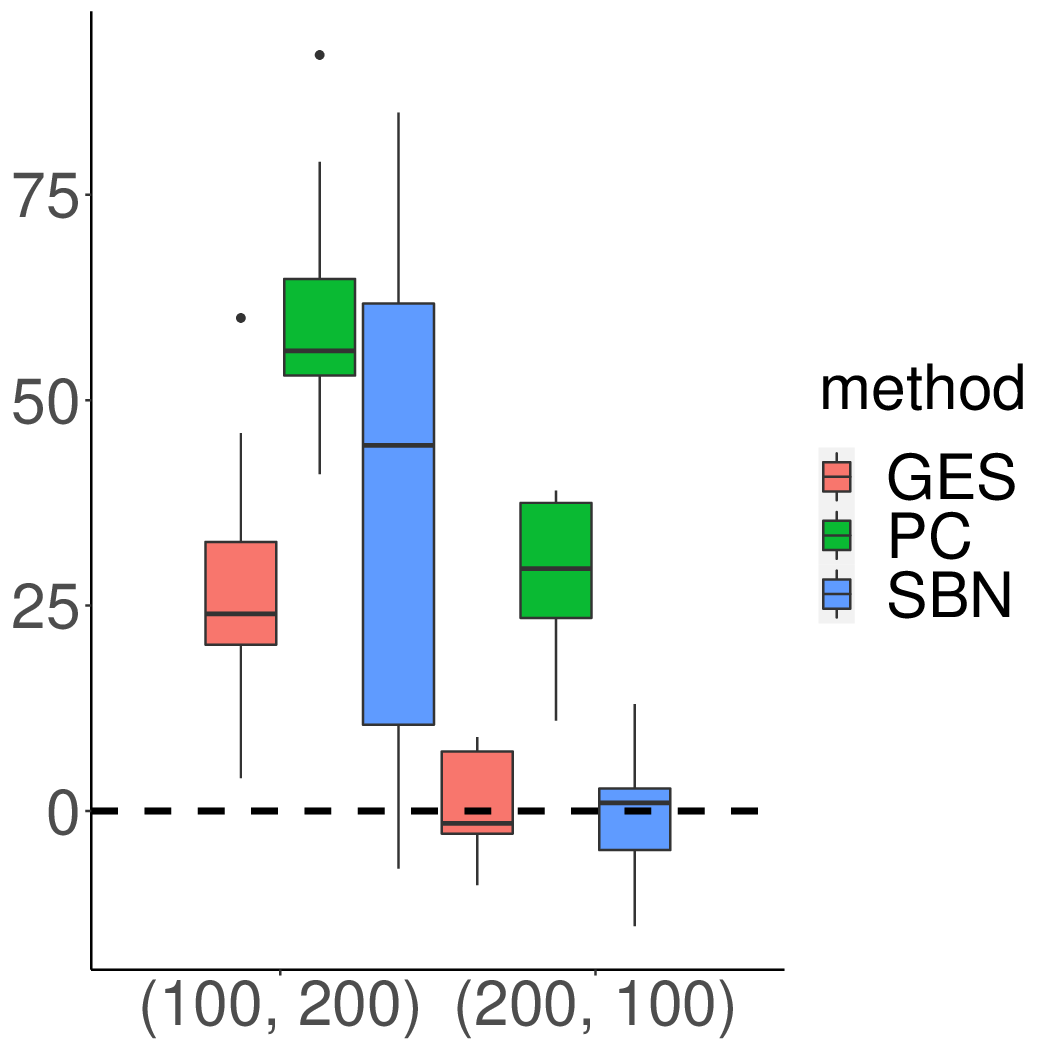}
        \caption{AR}
    \end{subfigure}%
    \begin{subfigure}[t]{0.2\textwidth}
        \centering
        \includegraphics[width=\textwidth]{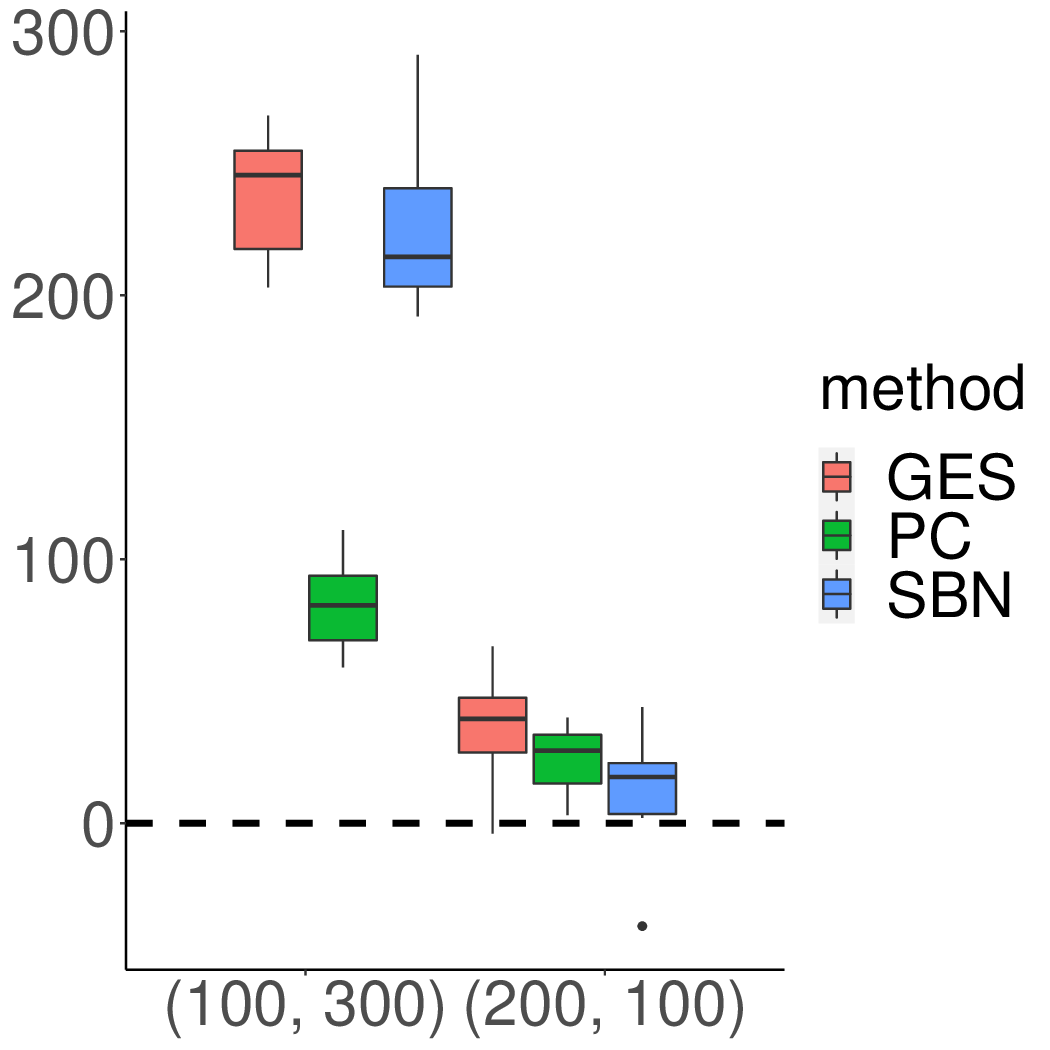}
        \caption{SBM}
    \end{subfigure}%
    
    \begin{subfigure}[t]{0.2\textwidth}
        \centering
        \includegraphics[width=\textwidth]{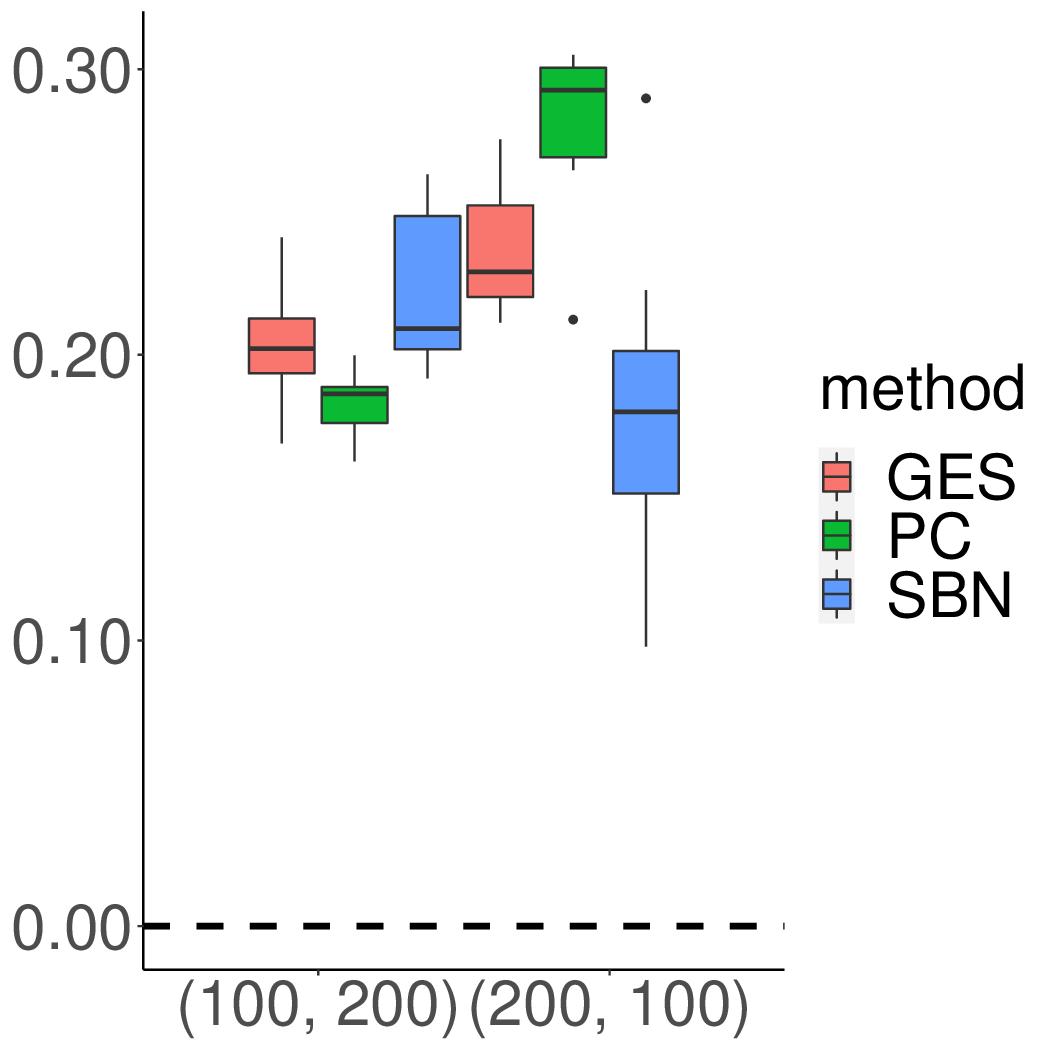}
        \caption{Equal Corr}
    \end{subfigure}%
    \begin{subfigure}[t]{0.2\textwidth}
        \centering
        \includegraphics[width=\textwidth]{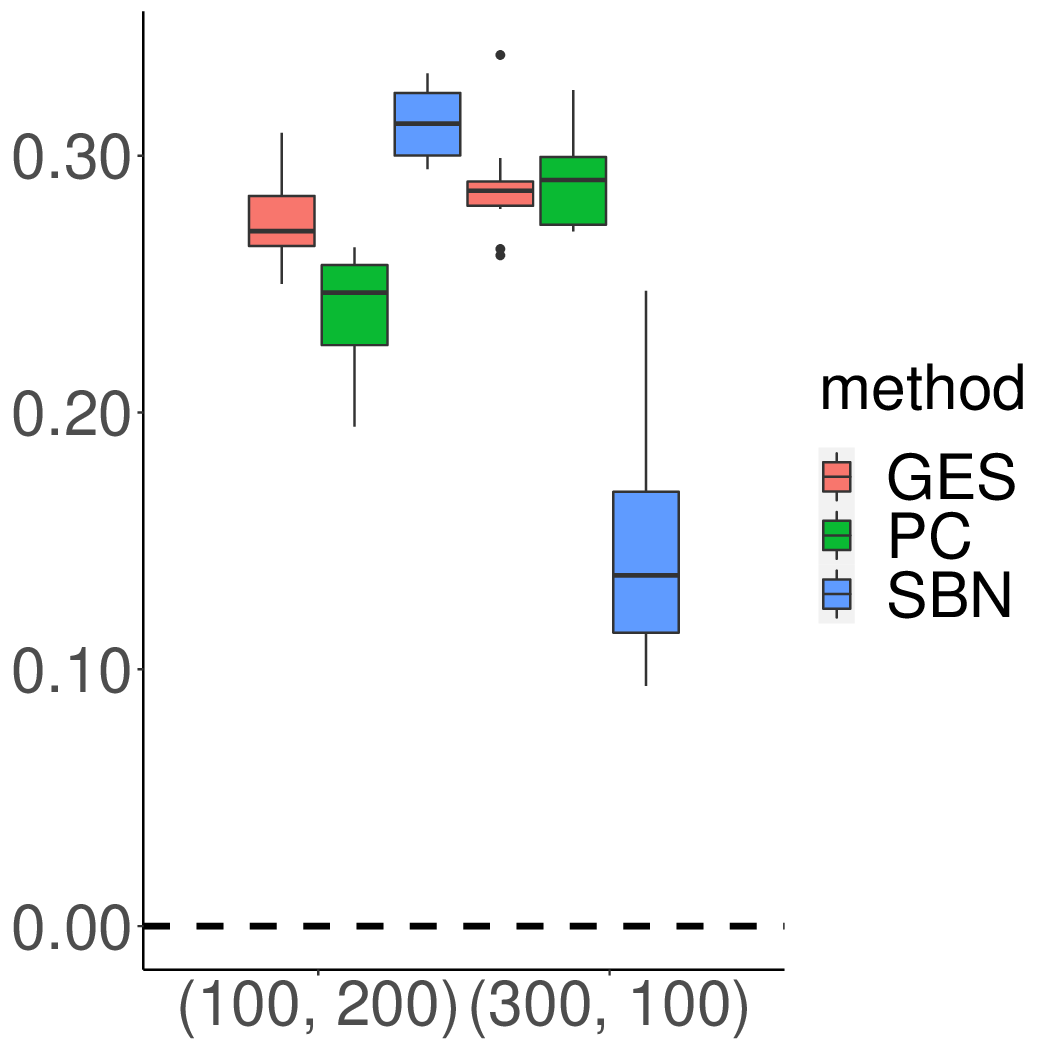}
        \caption{Toeplitz}
    \end{subfigure}%
    \begin{subfigure}[t]{0.2\textwidth}
        \centering
        \includegraphics[width=\textwidth]{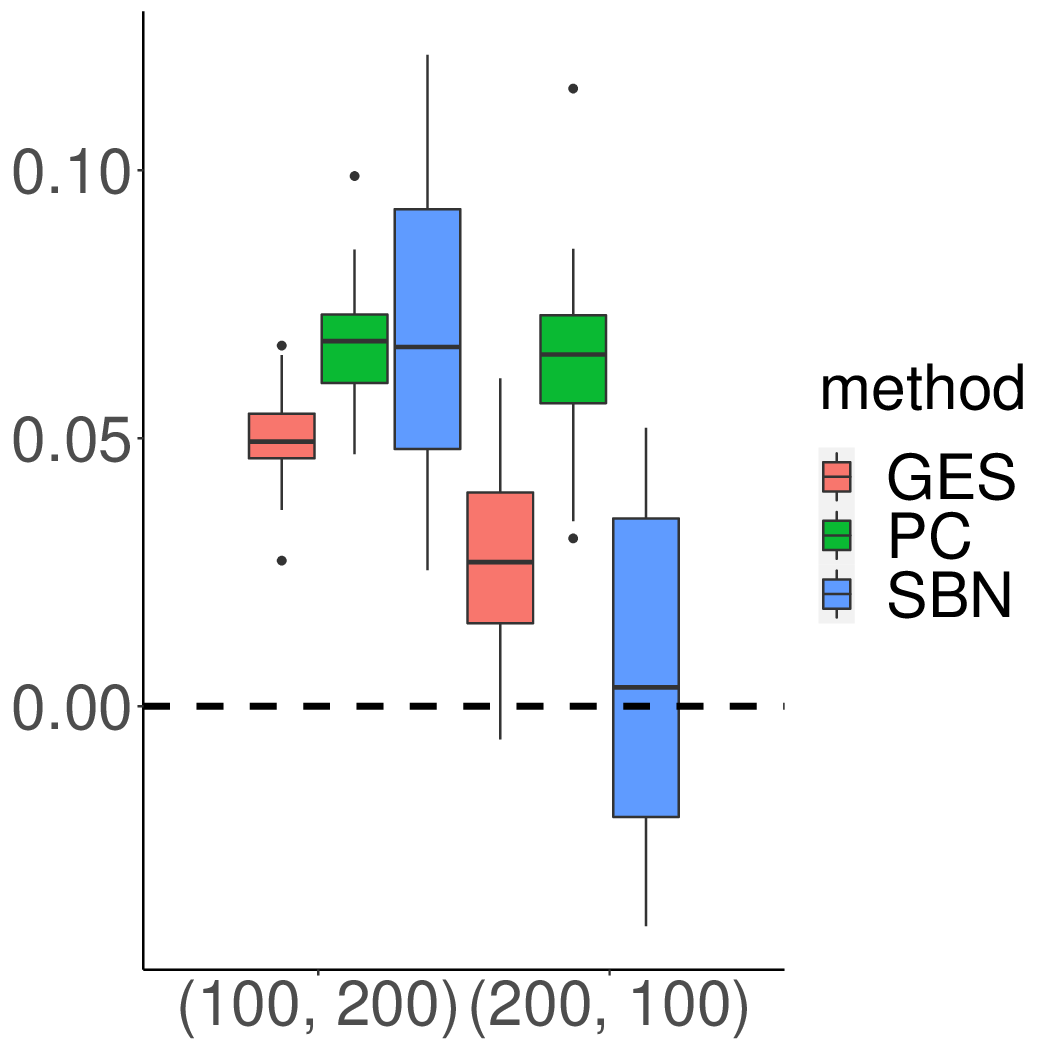}
        \caption{Star}
    \end{subfigure}%
    % \begin{subfigure}[t]{0.2\textwidth}
    %     \centering
    %     \includegraphics[width=\textwidth]{figures/unordered/ExpDecay-ja.eps}
    %     \caption{Exp Decay}
    % \end{subfigure}%
    \begin{subfigure}[t]{0.2\textwidth}
        \centering
        \includegraphics[width=\textwidth]{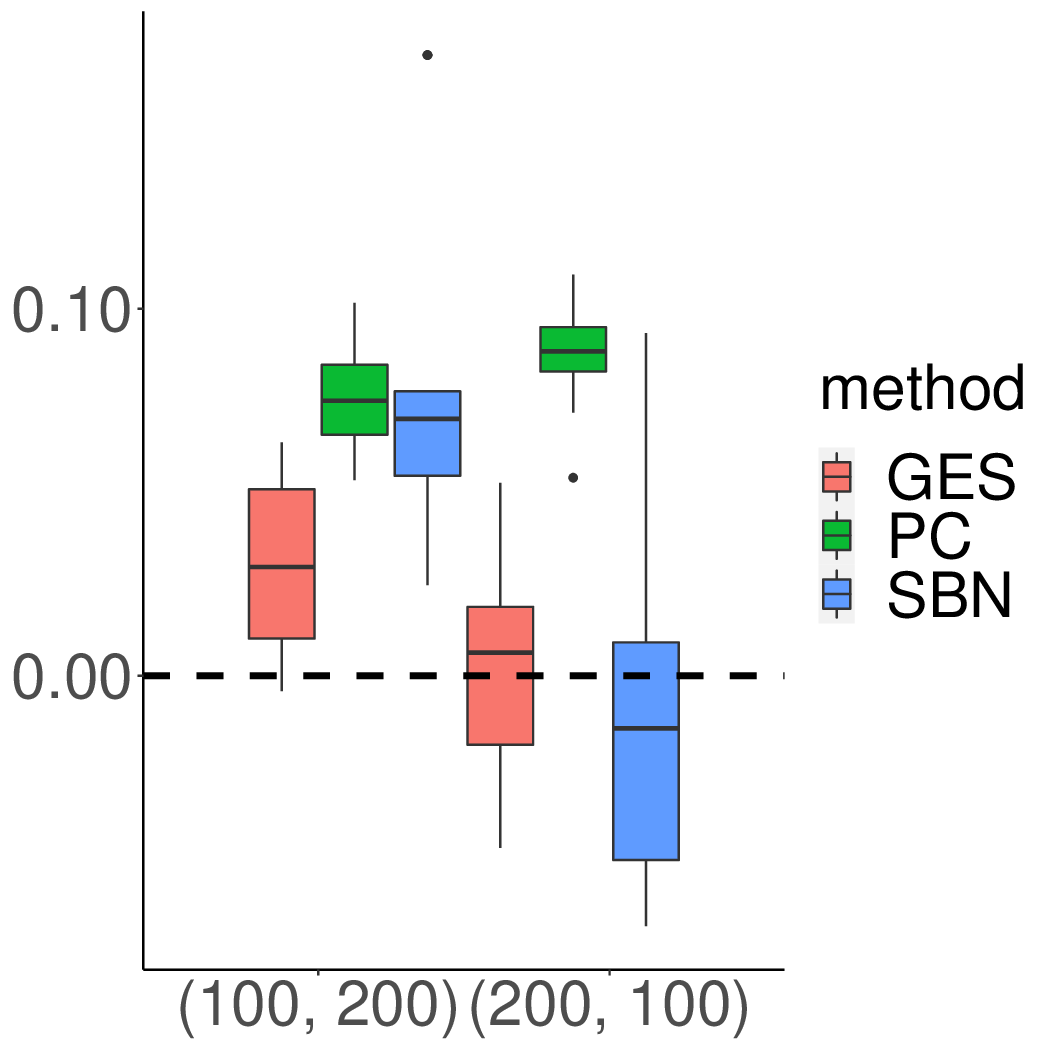}
        \caption{AR}
    \end{subfigure}%
      \begin{subfigure}[t]{0.2\textwidth}
        \centering
        \includegraphics[width=\textwidth]{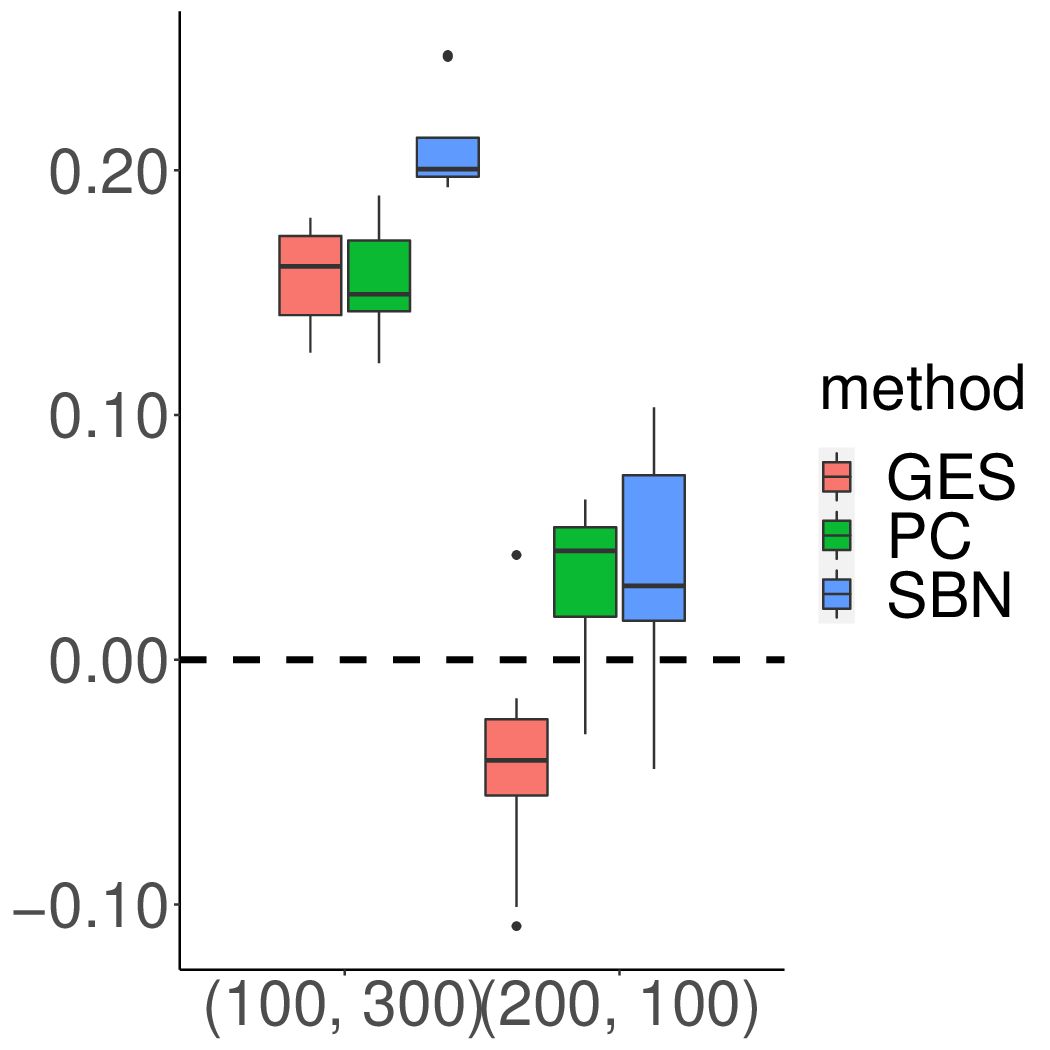}
        \caption{SBM}
    \end{subfigure}%
    \end{adjustbox}
    \caption{Decrease in SHD (top row) and increase in Jaccard index (bottom row) via de-correlation on simulated unsorted DAGs, % y-axis represents the difference between the SHD (JI) before and after decorrelation. 
    with x-axis reporting the value of $(n,p)$. In each panle, the three boxplots on the left and the three on the right correspond to the cases of $n < p$ and $n > p$, respectively. Each boxplot contains 10 data points from 10 simulations.}% Block size $= 30$.}
    \label{fig:SHD}
\end{figure}

\begin{figure}
    \centering
    \begin{adjustbox}{minipage=\linewidth,scale=0.9}
    \begin{subfigure}[t]{0.2\textwidth}
        \centering
        \includegraphics[width=\textwidth]{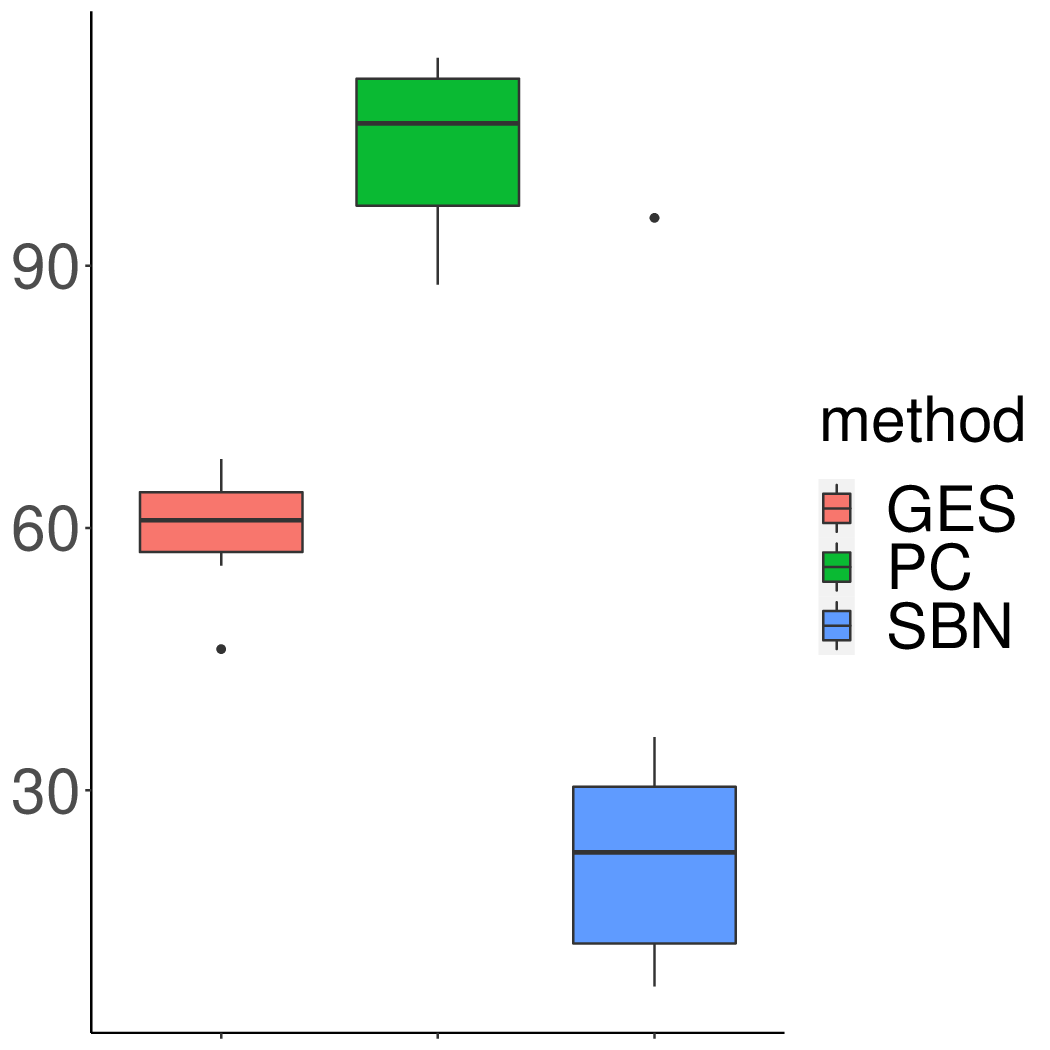}
        \caption{Equal Corr}
    \end{subfigure}%
    \begin{subfigure}[t]{0.2\textwidth}
        \centering
        \includegraphics[width=\textwidth]{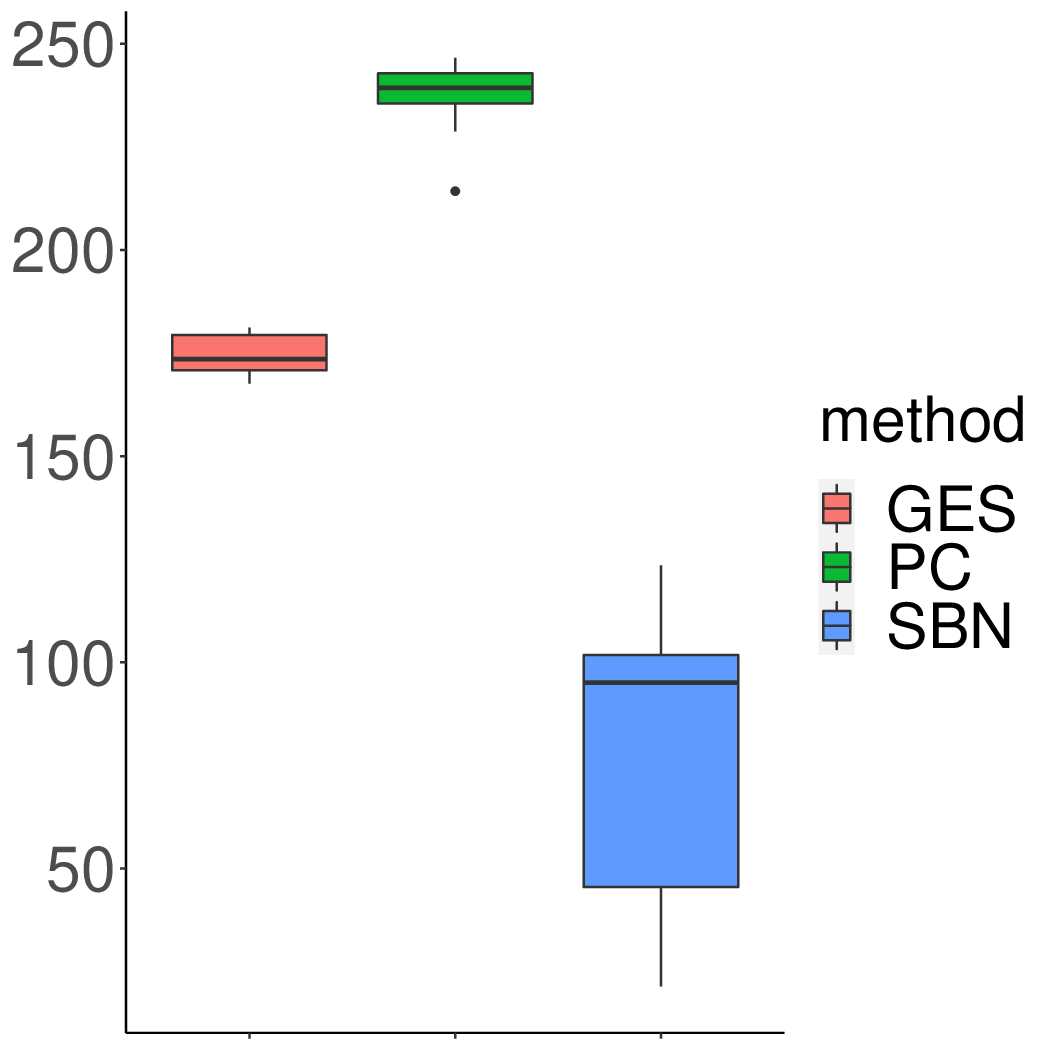}
        \caption{Toeplitz}
    \end{subfigure}%
    \begin{subfigure}[t]{0.2\textwidth}
        \centering
        \includegraphics[width=\textwidth]{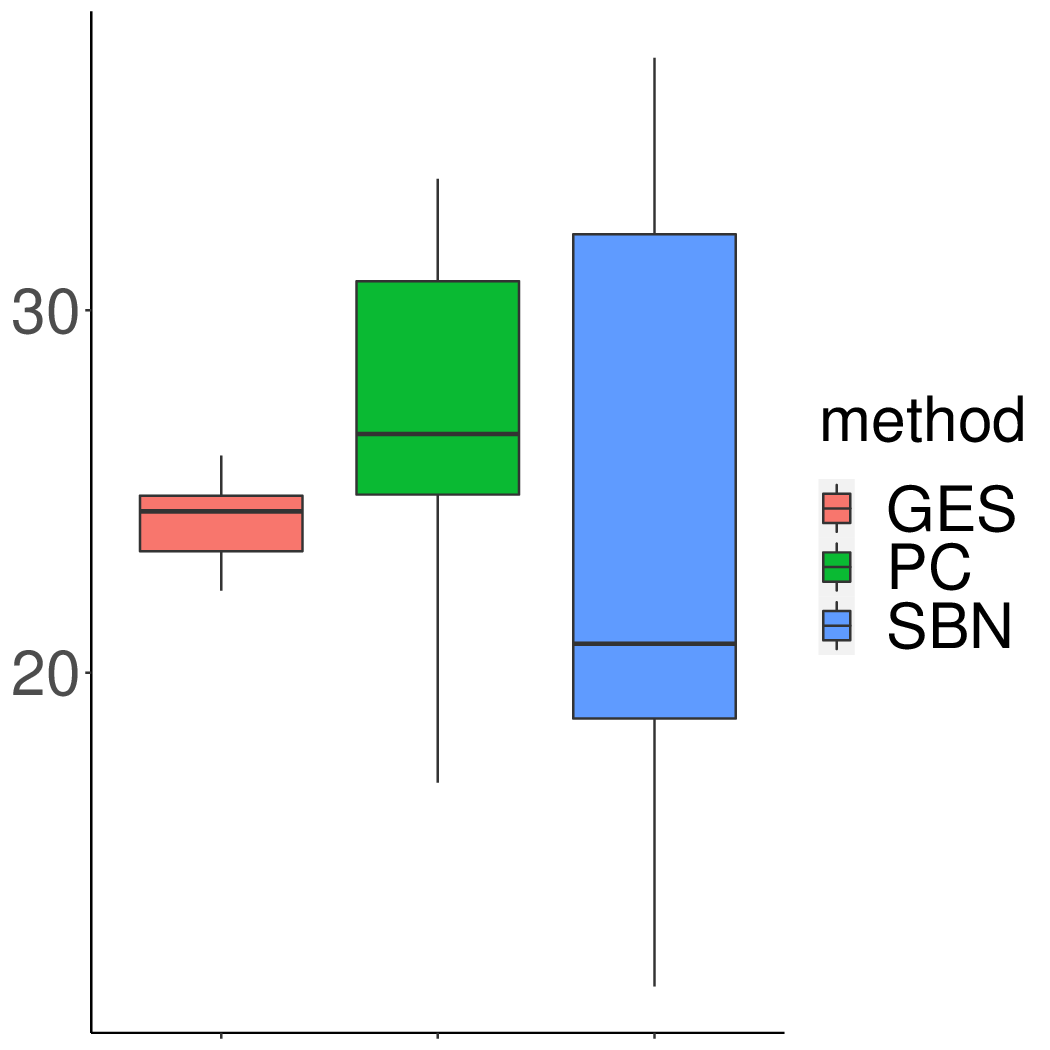}
        \caption{Star}
    \end{subfigure}%
    % \begin{subfigure}[t]{0.2\textwidth}
    %     \centering
    %     \includegraphics[width=\textwidth]{figures/unordered/testll/00055-2testll.eps}
    %     \caption{Exp Decay}
    % \end{subfigure}%
    \begin{subfigure}[t]{0.2\textwidth}
        \centering
        \includegraphics[width=\textwidth]{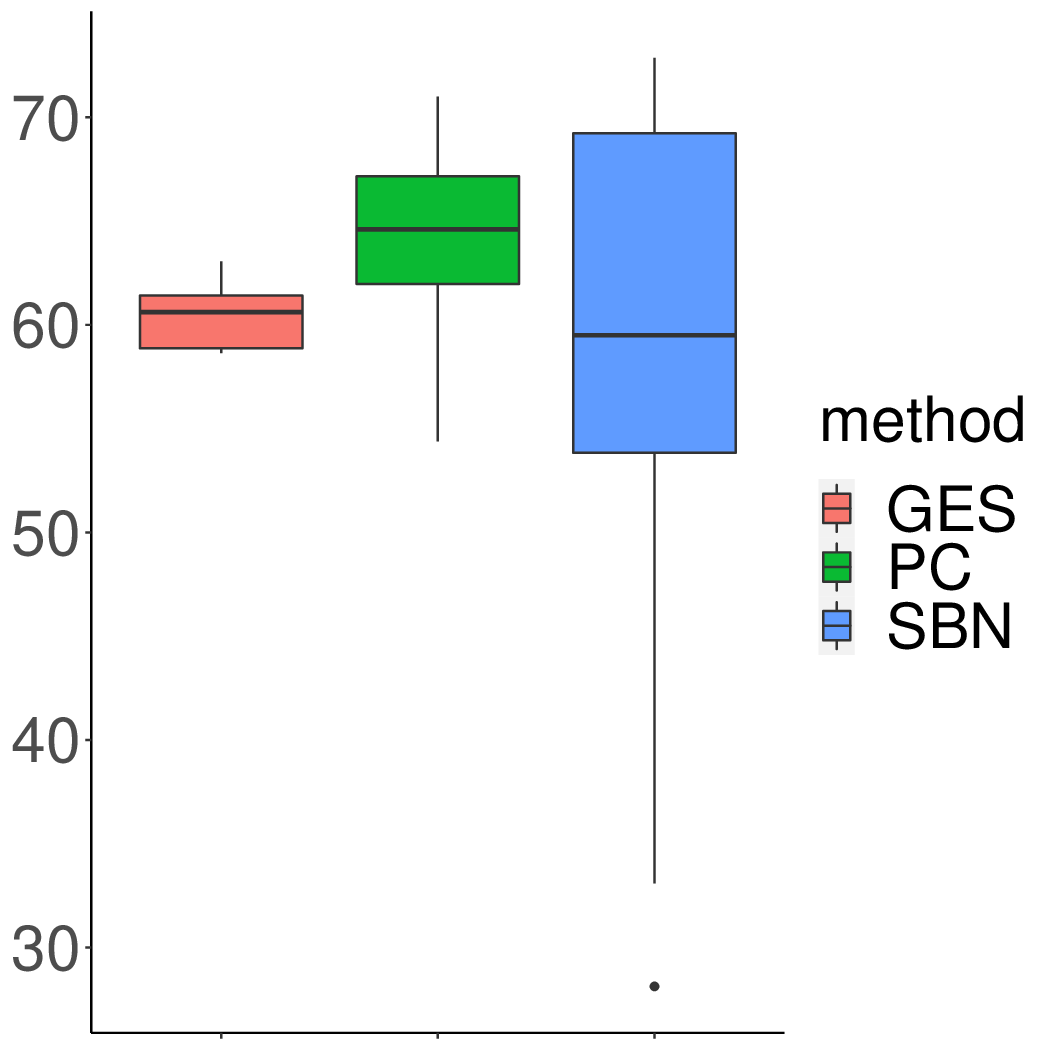}
        \caption{AR}
    \end{subfigure}%
     \begin{subfigure}[t]{0.2\textwidth}
        \centering
        \includegraphics[width=\textwidth]{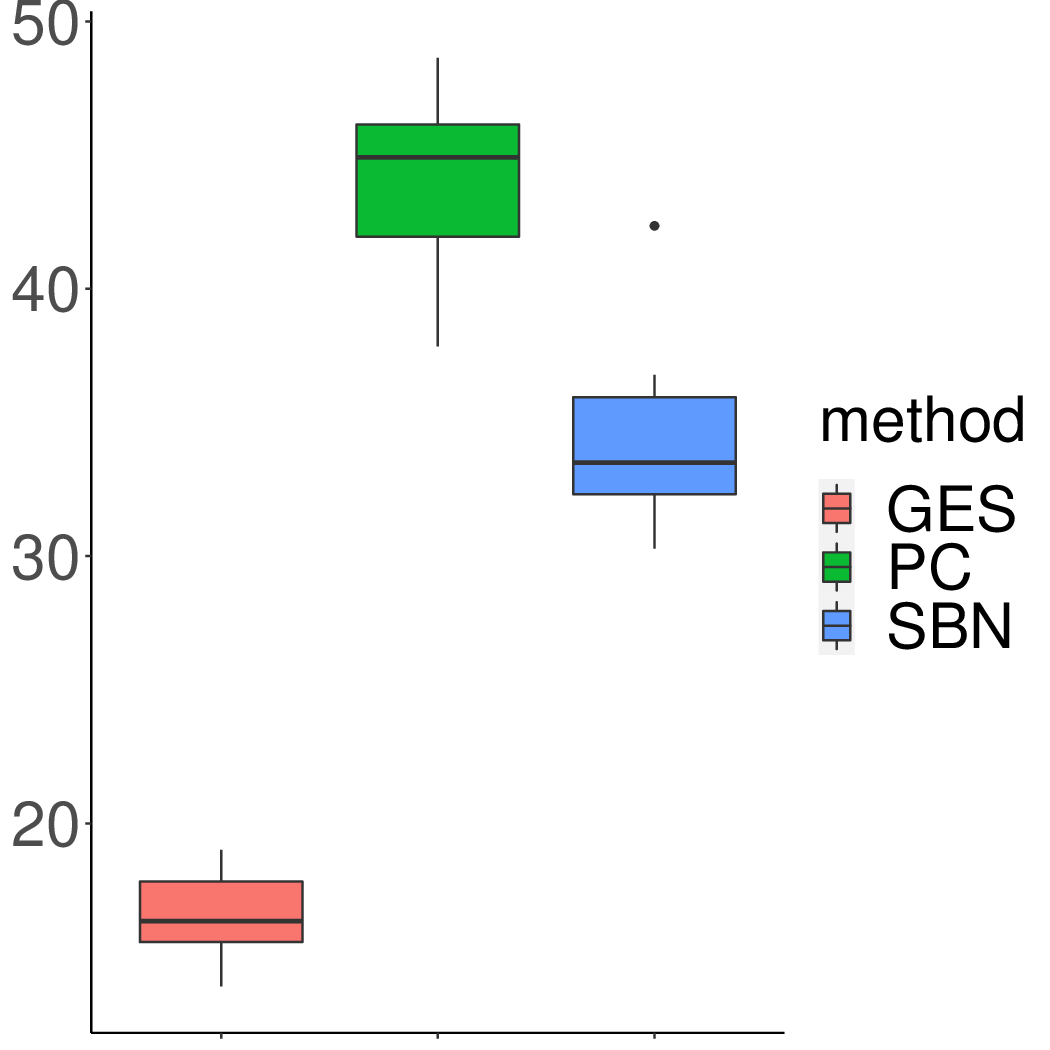}
        \caption{SBM}
    \end{subfigure}%
    
    \begin{subfigure}[t]{0.2\textwidth}
        \centering
        \includegraphics[width=\textwidth]{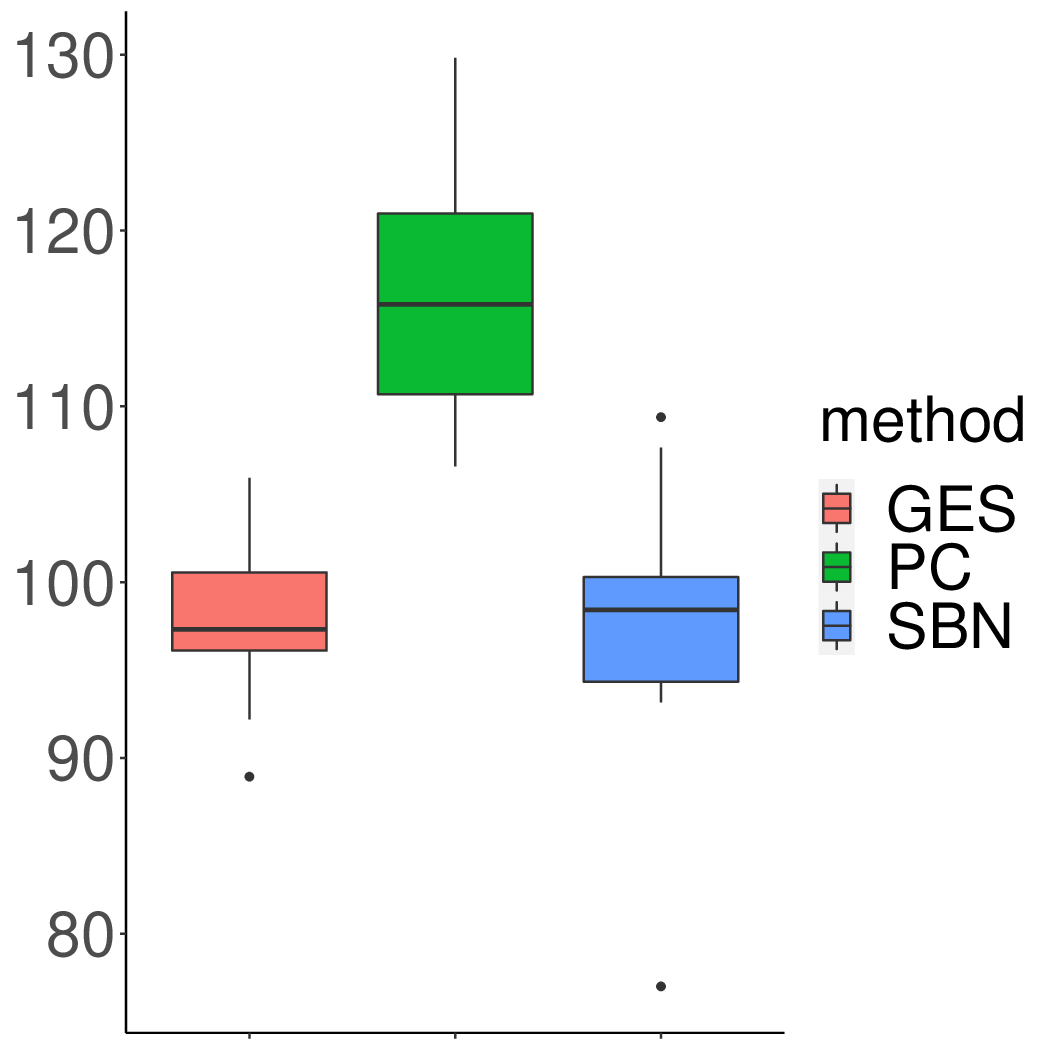}
        \caption{Equal Corr}
    \end{subfigure}%
    \begin{subfigure}[t]{0.2\textwidth}
        \centering
        \includegraphics[width=\textwidth]{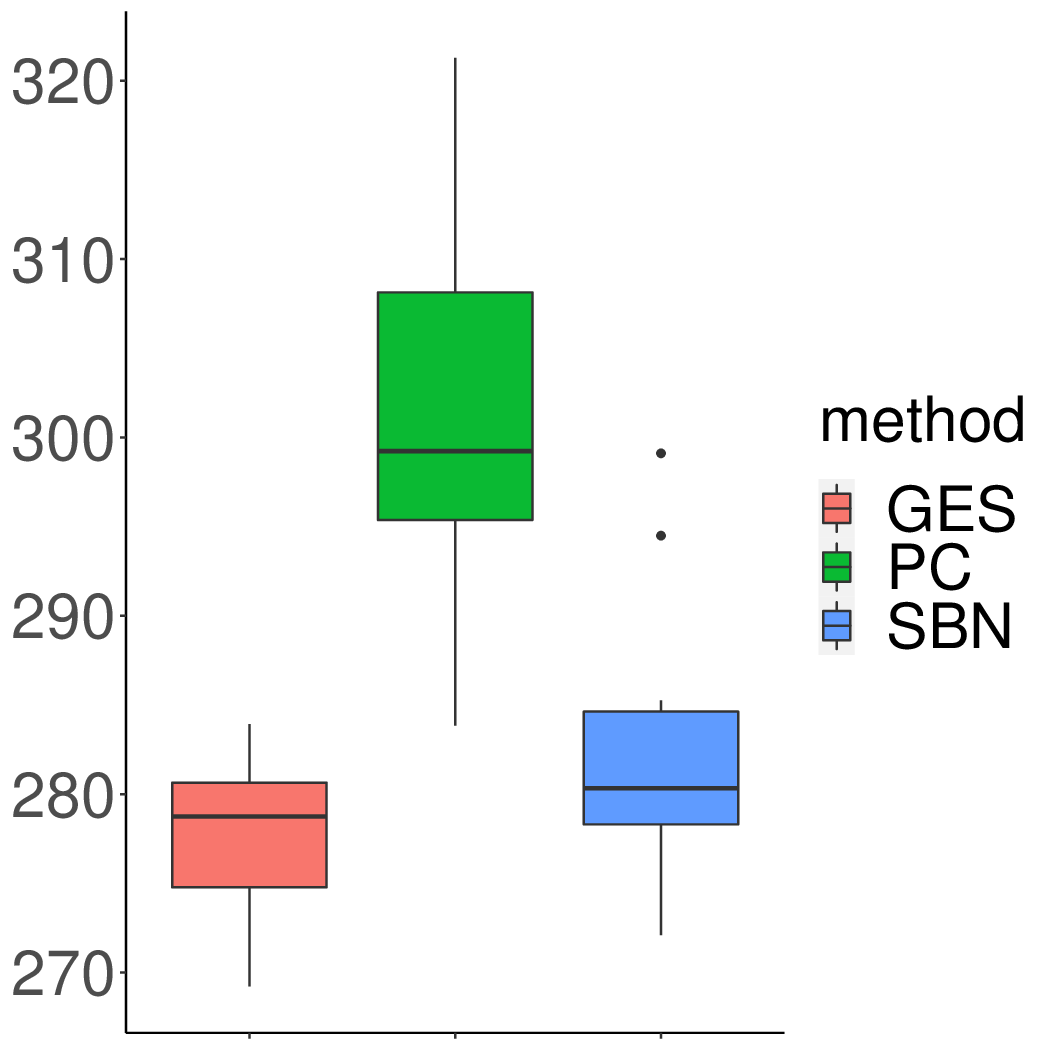}
        \caption{Toeplitz}
    \end{subfigure}%
    \begin{subfigure}[t]{0.2\textwidth}
        \centering
        \includegraphics[width=\textwidth]{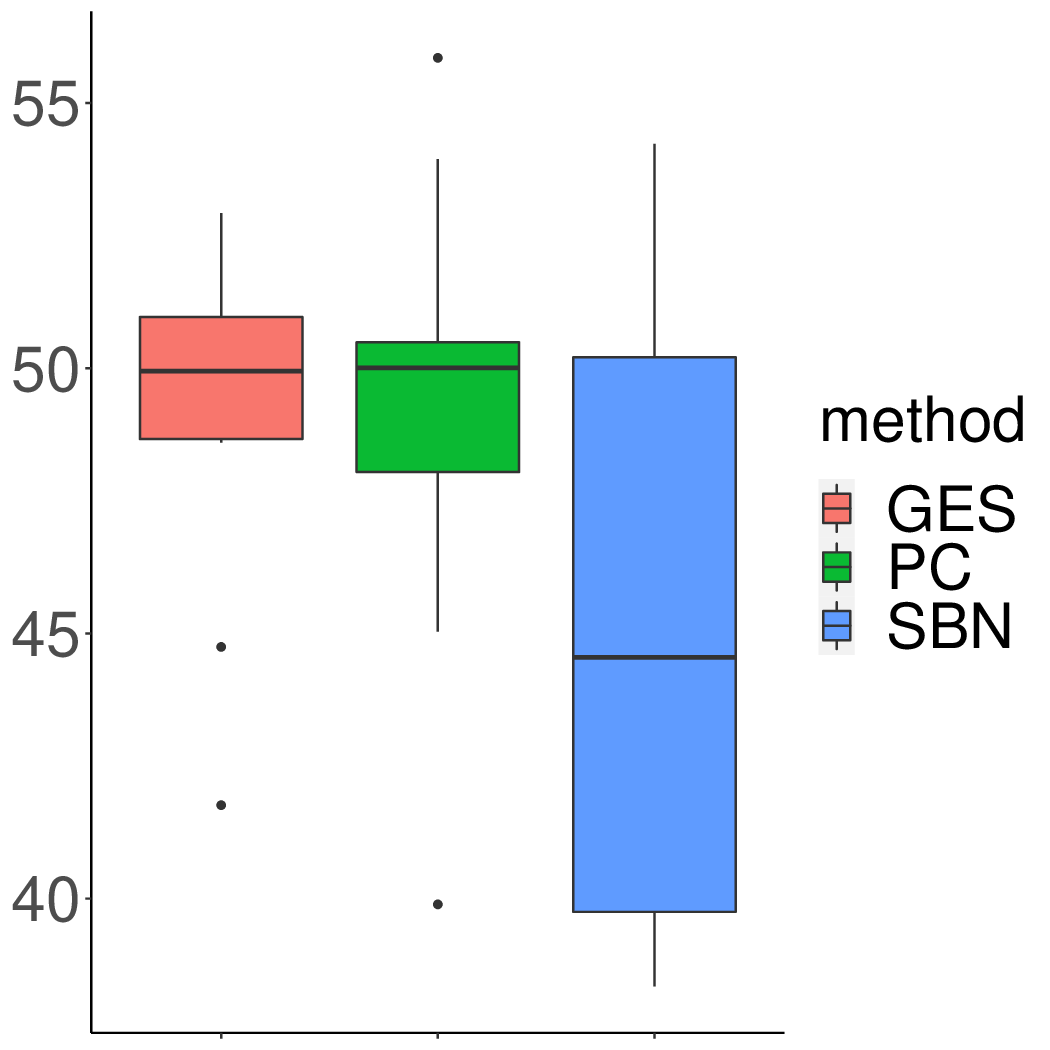}
        \caption{Star}
    \end{subfigure}%
    % \begin{subfigure}[t]{0.2\textwidth}
    %     \centering
    %     \includegraphics[width=\textwidth]{figures/unordered/testll/321004-2testll.eps}
    %     \caption{Exp Decay}
    % \end{subfigure}%
    \begin{subfigure}[t]{0.2\textwidth}
        \centering
        \includegraphics[width=\textwidth]{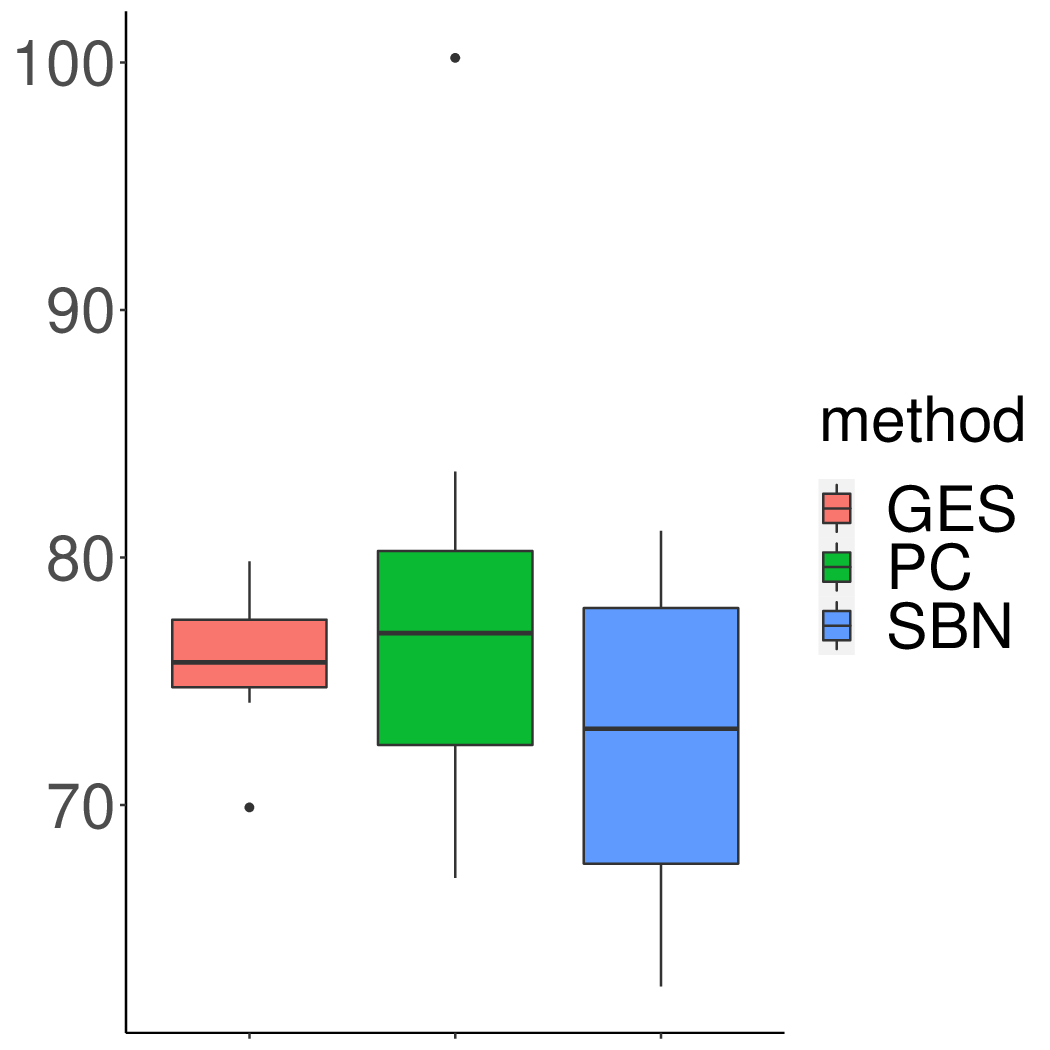}
        \caption{AR}
    \end{subfigure}%
     \begin{subfigure}[t]{0.2\textwidth}
        \centering
        \includegraphics[width=\textwidth]{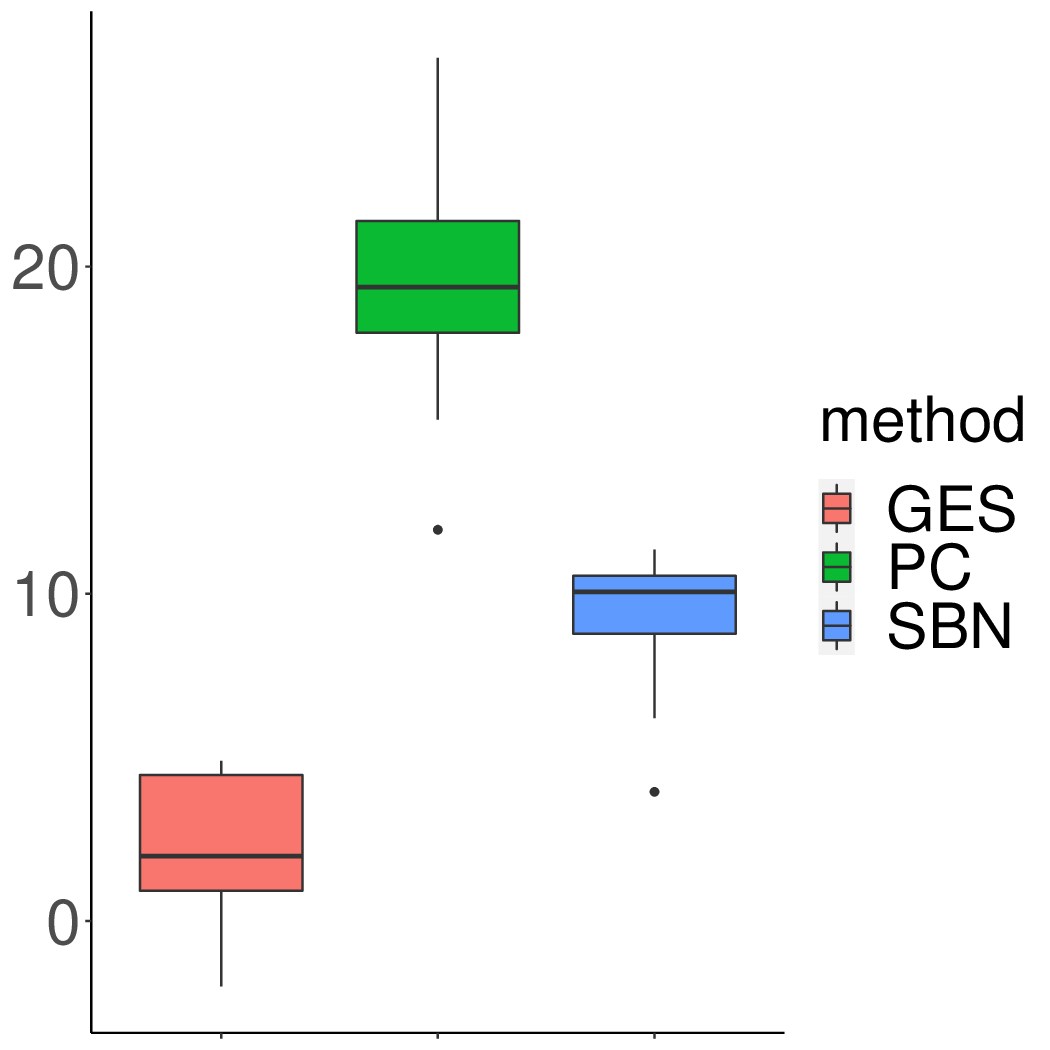}
        \caption{SBM}
    \end{subfigure}%
    \end{adjustbox}
    \caption{Increase in the normalized test data log-likelihood after decorrelation on simulated unsorted DAGs. Top row: $n < p$. Bottom row: $n > p$. }
    \label{fig:testll3}
\end{figure}

As before, we divided the cases into $n < p$ and $n > p$. The block size for the four block-diagonal $\Theta$ was fixed to 30. The estimated Cholesky factor $\widehat L$ of $\widehat\Theta$ used for de-correlating $X$ in \eqref{eq:dcor} was calculated by our BCD algorithm with tuning parameter $\lambda_1$ selected by BIC. %We set  $\alpha = 0.1$ according to the ROC. 
Figure~\ref{fig:SHD} shows the decrease in SHD and increase in Jaccard index via de-correlation of GES, PC and sparsebn on 10 random DAGs, generated under each row-covariance structure and each sample size. For almost all types of covariances and $(n, p)$ settings we considered, there is significant improvement of all three methods in estimating the CPDAG structures after de-correlation. Additional tables with detailed results can be found in the Supplementary Material. Before decorrelation, GES and sparsebn, both score-based methods, tend to significantly overestimate the number of edges, resulting in high false positives, so does PC in some of the cases. After decorrelation, both GES and sparsebn had significant improvements and outperformed PC, as long as $\widehat\Theta$ was accurately estimated. The test data log-likelihood (normalized by $\sqrt{np}$) of all three algorithms also increased significantly after decorrelation as shown in Figure~\ref{fig:testll3}.

\subsection{Experiments on Real Network Structures}

In this section, we look at the performance of the BCD algorithm on real network structures. We took four real DAGs from the bnlearn repository \citep{bnlearn}: \texttt{Andes}, \texttt{Hailfinder}, \texttt{Barley}, \texttt{Hepar2}, and two real undirected networks from \texttt{tnet} \citep{tnet}: \texttt{facebook} \citep{facebook}  and \texttt{celegans\_n306} \citep{celegan}. Only the structures (supports) of these real networks were used and the parameters of the edges were simulated as follows. Given a DAG structure, we sampled the coefficients $\beta_j^*$ uniformly from $[-1,-0.1]\cup [0.1,1]$. Given the support of $\Theta^*$, we generated $\Theta^{\prime}_{ij}$  uniformly from $[-5, 5]$. Then, we applied the transformations in \eqref{pdtheta} to get $\Theta^*$. %the entries $\widetilde\Theta_{ij}^*$ were sampled uniformly from $[-5, 5]$. \revision{Then, we applied the transformations in \eqref{pdtheta} to get $\Theta^*$.}
%took the following steps to make sure $\Theta^*$ is symmetric and positive definite: (1) $\Theta_{\text{temp}}^* := (\Tilde{\Theta}^* + \Tilde{\Theta}^{*\top}) / 2$. (2) $\Theta^* := \Theta_{\text{temp}}^* - \gamma_{\min}\left(\Theta_{\text{temp}}^*\right)I_n$ \qzcmt{change $\gamma_{\min}$ to $\sigma_{\min}$?}.
In order to increase the size of the underlying DAG and show the scalability of the algorithm, we duplicated the DAGs above to form larger networks. In Section \ref{sec:real_ordered} and \ref{sec:real_random}, we again consider undirected networks consisting of several disconnected subgraphs, corresponding to a block-diagonal structure in $\Theta^*$. Each of the subgraphs was sub-sampled from the original real network. In Section \ref{sec:real_general} we present experiments on more general $\Theta^*$ without a block-diagonal structure. The $\omega_j^*$ were uniformly sampled from $[0.1, 2]$ as before. With these parameters, we generated observational samples $X$ following the structural equation \eqref{eq:model}.

\subsubsection{Learning with given ordering}\label{sec:real_ordered}
\begin{figure}[ht]
    \centering
    \begin{adjustbox}{minipage=\linewidth,scale=.8}
    \begin{subfigure}[t]{0.25\textwidth}
        \centering
        \includegraphics[width=\textwidth]{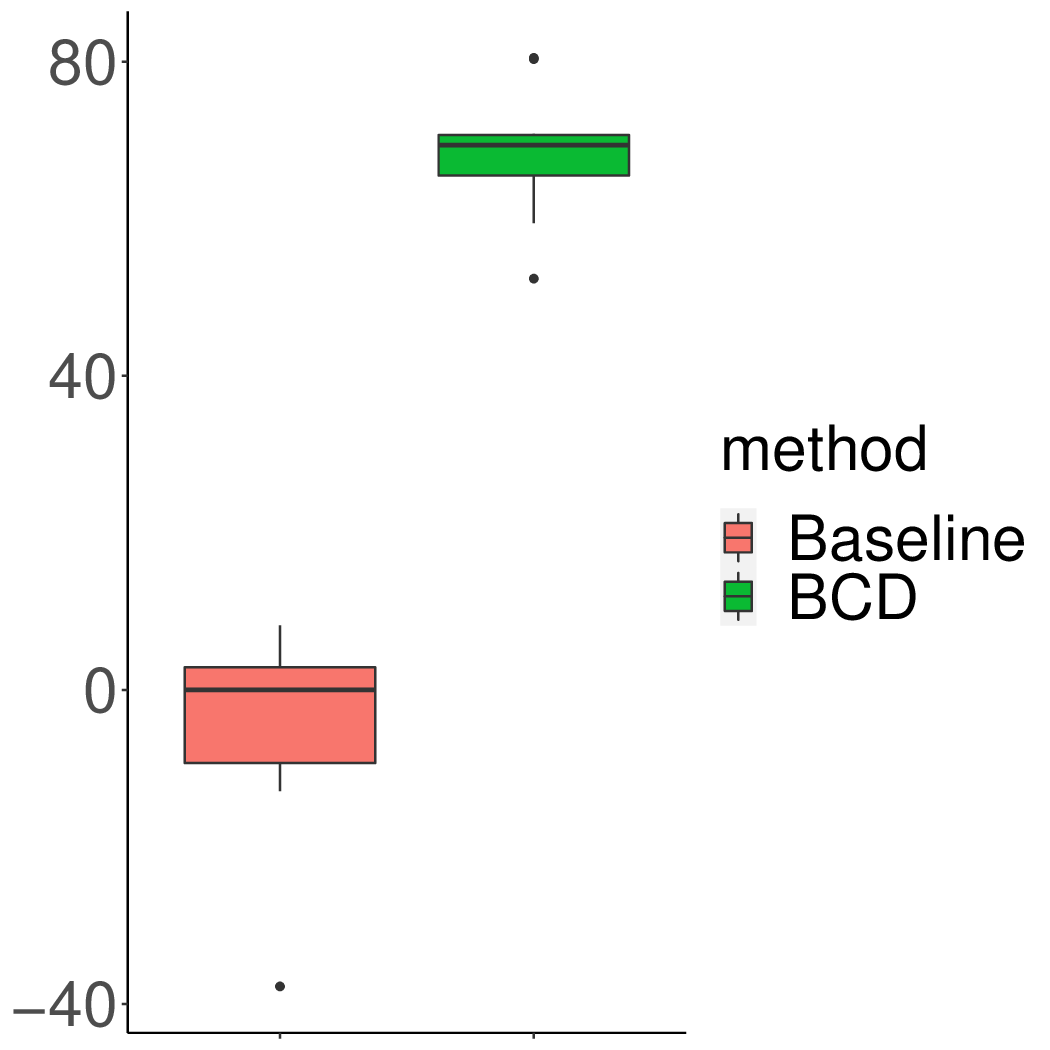}
         \caption*{\texttt{Andes}\\\texttt{Facebook}}
    \end{subfigure}%
    \begin{subfigure}[t]{0.25\textwidth}
        \centering
        \includegraphics[width=\textwidth]{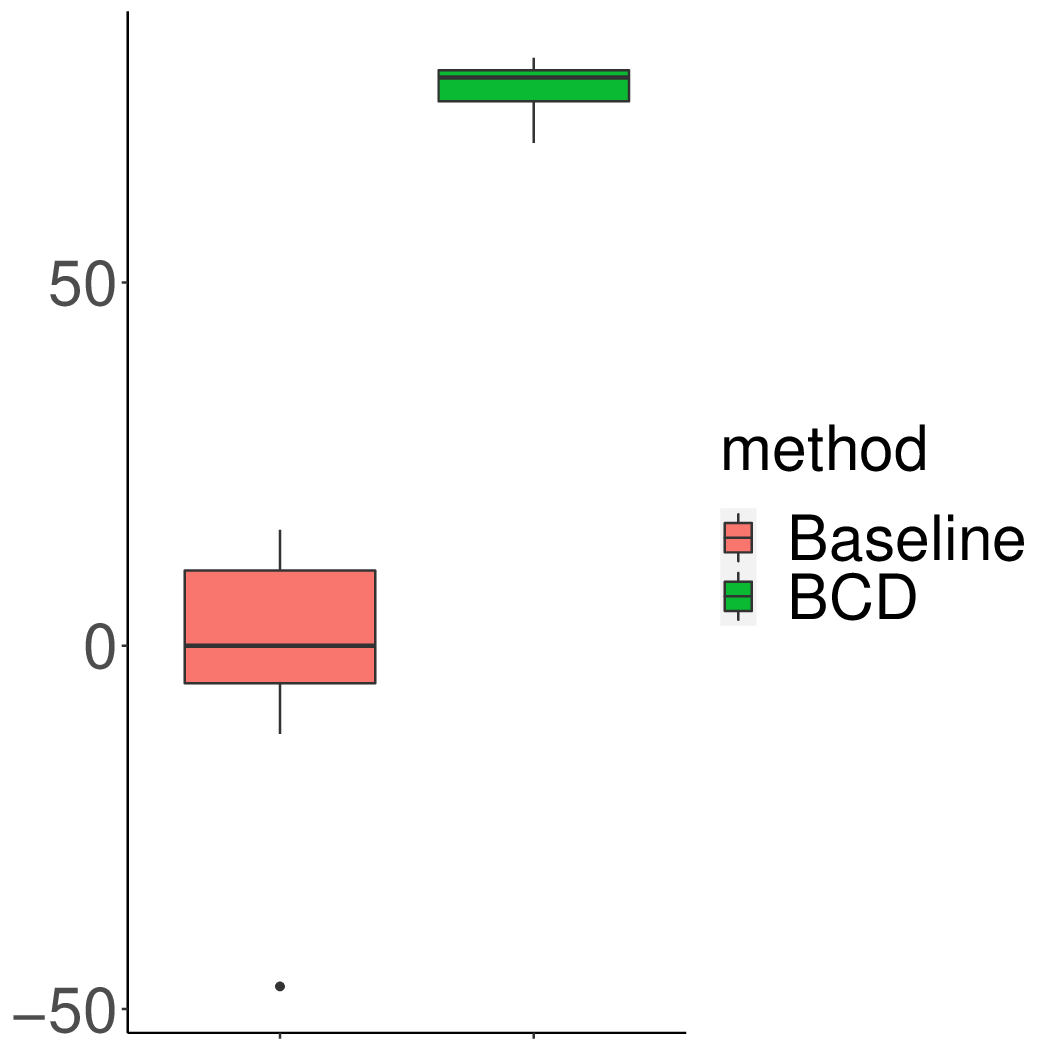}
        \caption*{\quad\texttt{Hailfinder}\\ \texttt{celegans\_n306}}
    \end{subfigure}%
    \begin{subfigure}[t]{0.25\textwidth}
        \centering
        \includegraphics[width=\textwidth]{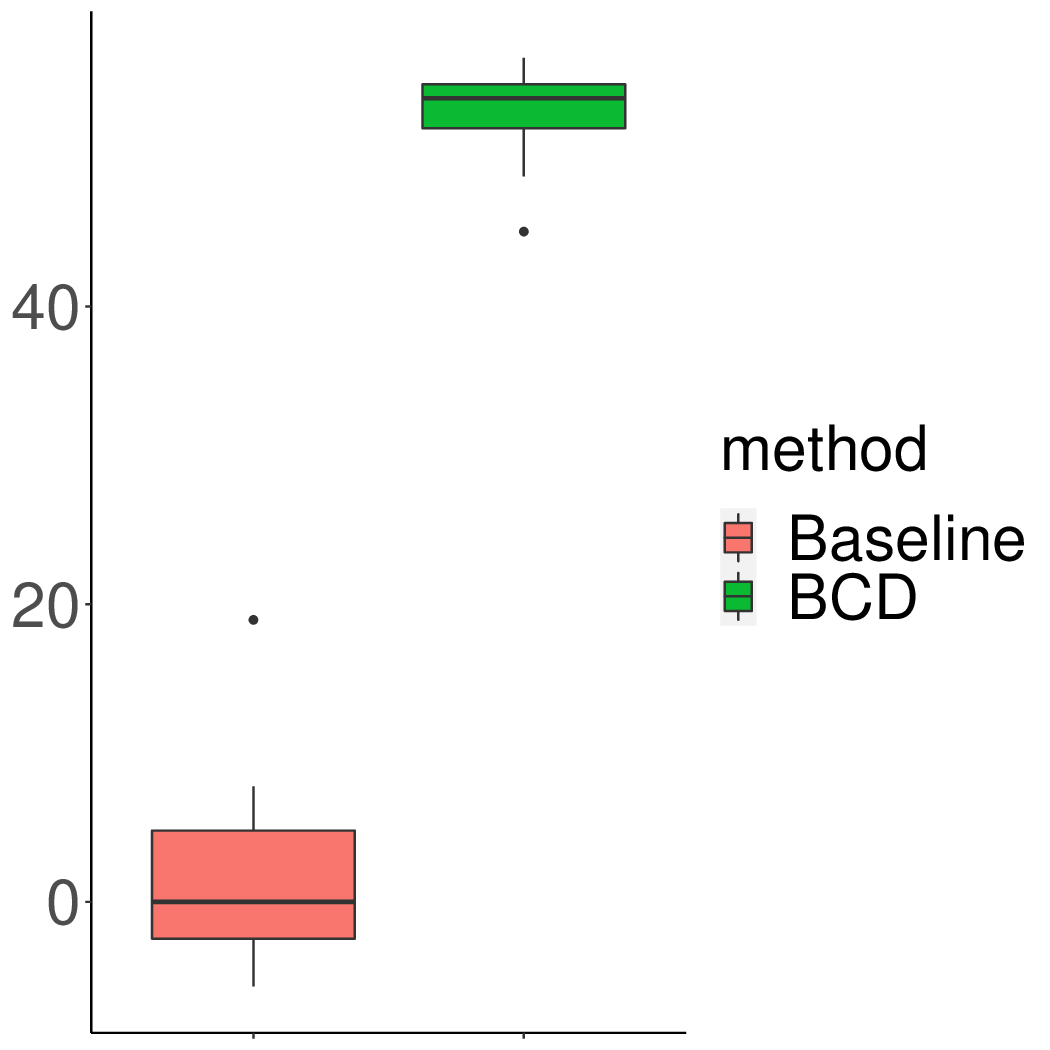}
          \caption*{\texttt{Barley}\\ \texttt{Facebook}}
    \end{subfigure}%
    \begin{subfigure}[t]{0.25\textwidth}
        \centering
        \includegraphics[width=\textwidth]{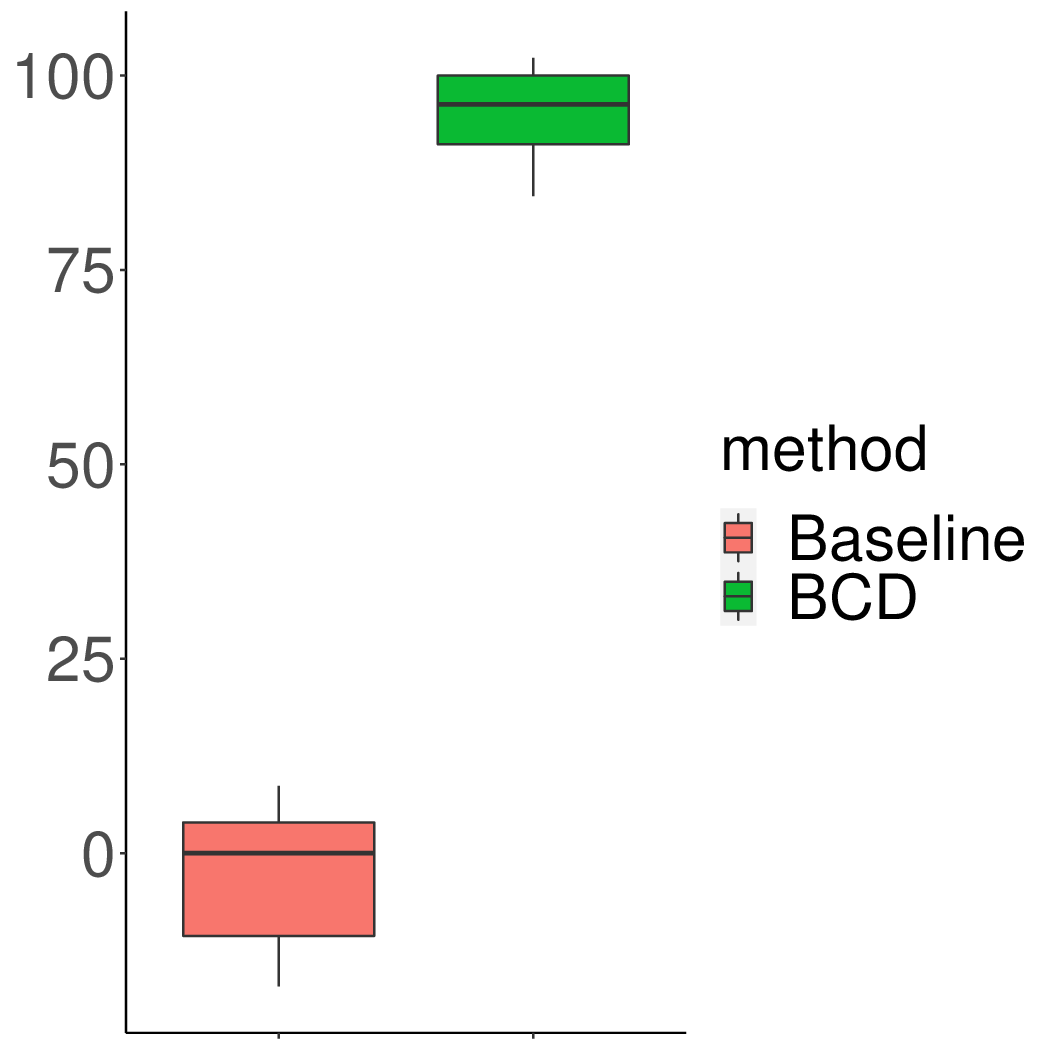}
        \caption*{\texttt{Hepar2}\\\texttt{celegans\_n306}}
    \end{subfigure}%
    
    \begin{subfigure}[t]{0.25\textwidth}
        \centering
        \includegraphics[width=\textwidth]{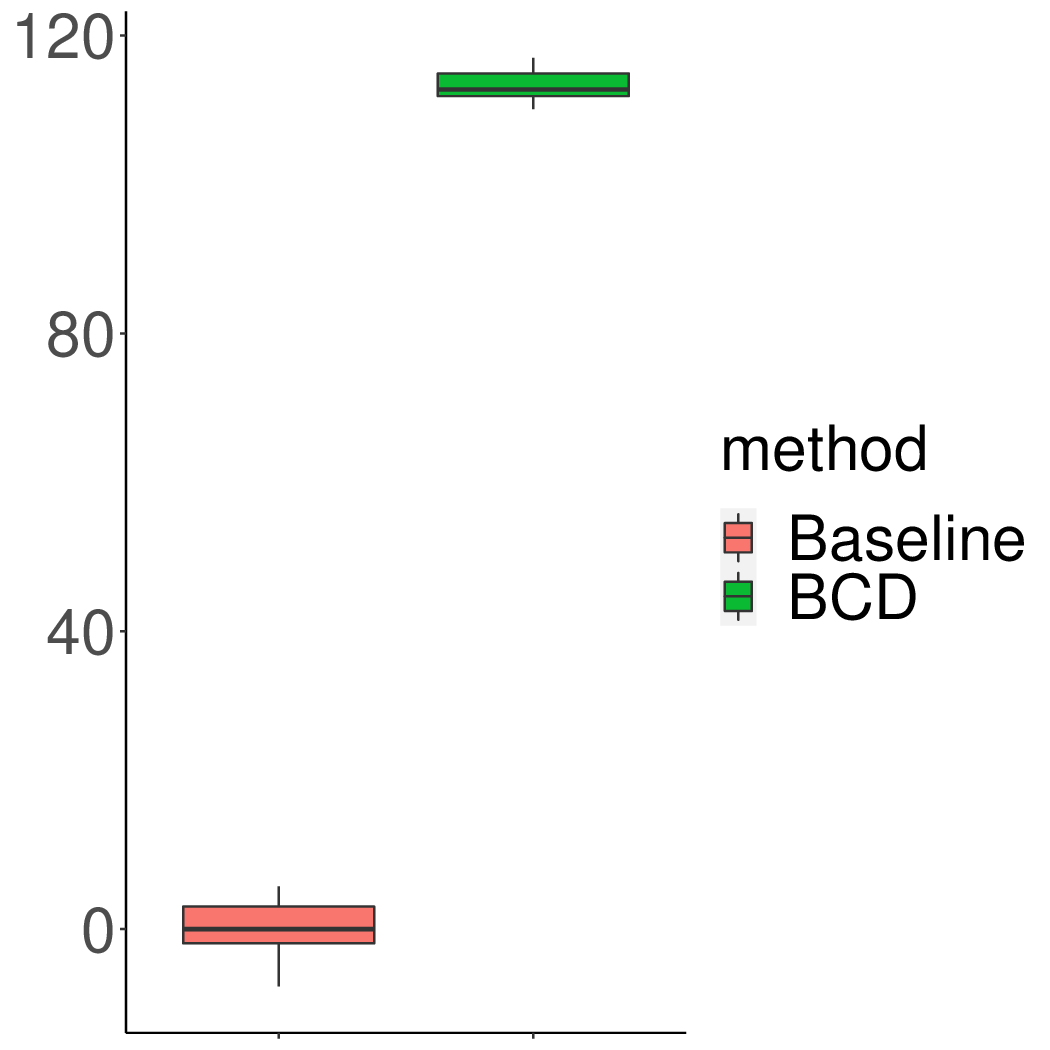}
         \caption*{\texttt{Andes}\\\texttt{Facebook}}
    \end{subfigure}%
    \begin{subfigure}[t]{0.25\textwidth}
        \centering
        \includegraphics[width=\textwidth]{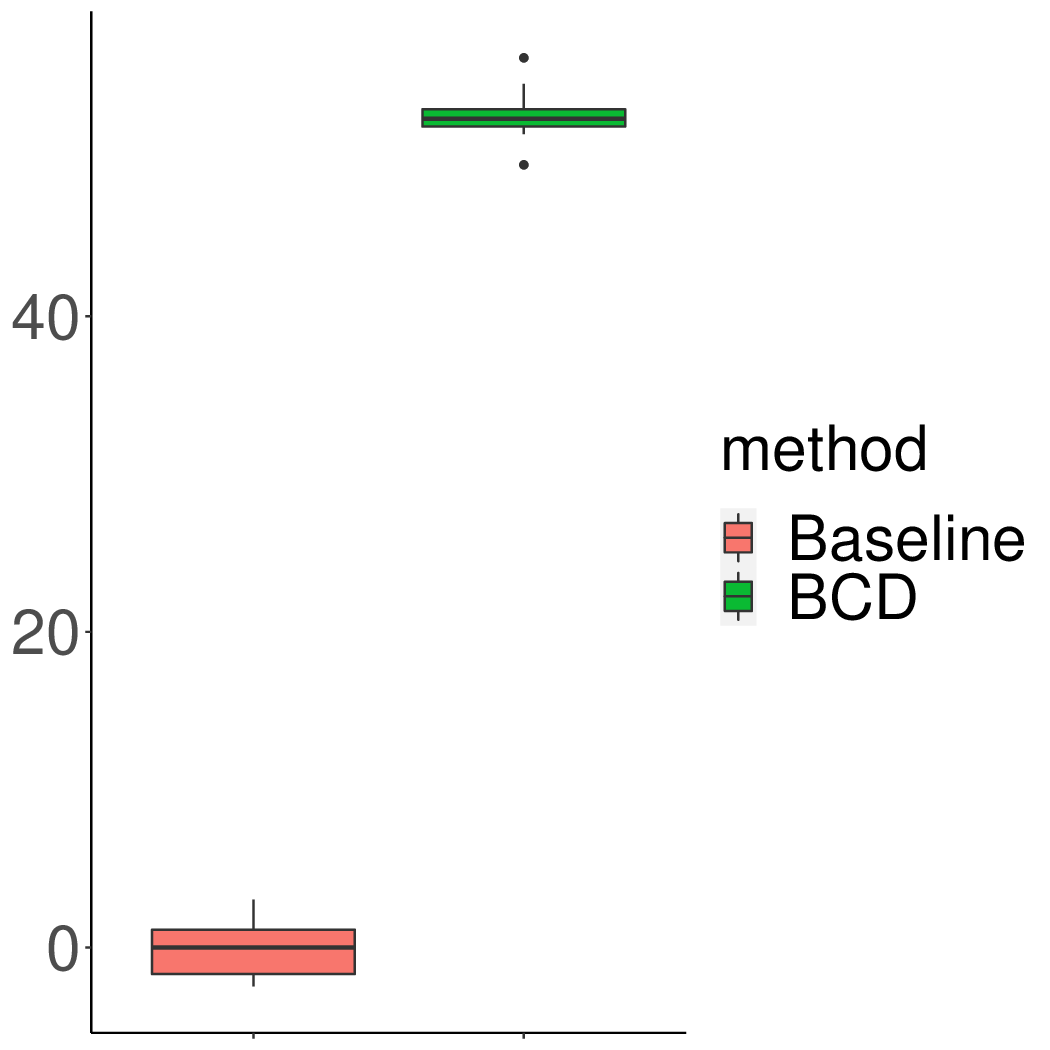}
        \caption*{\quad\texttt{Hailfinder}\\ \texttt{celegans\_n306}}
    \end{subfigure}%
    \begin{subfigure}[t]{0.25\textwidth}
        \centering
        \includegraphics[width=\textwidth]{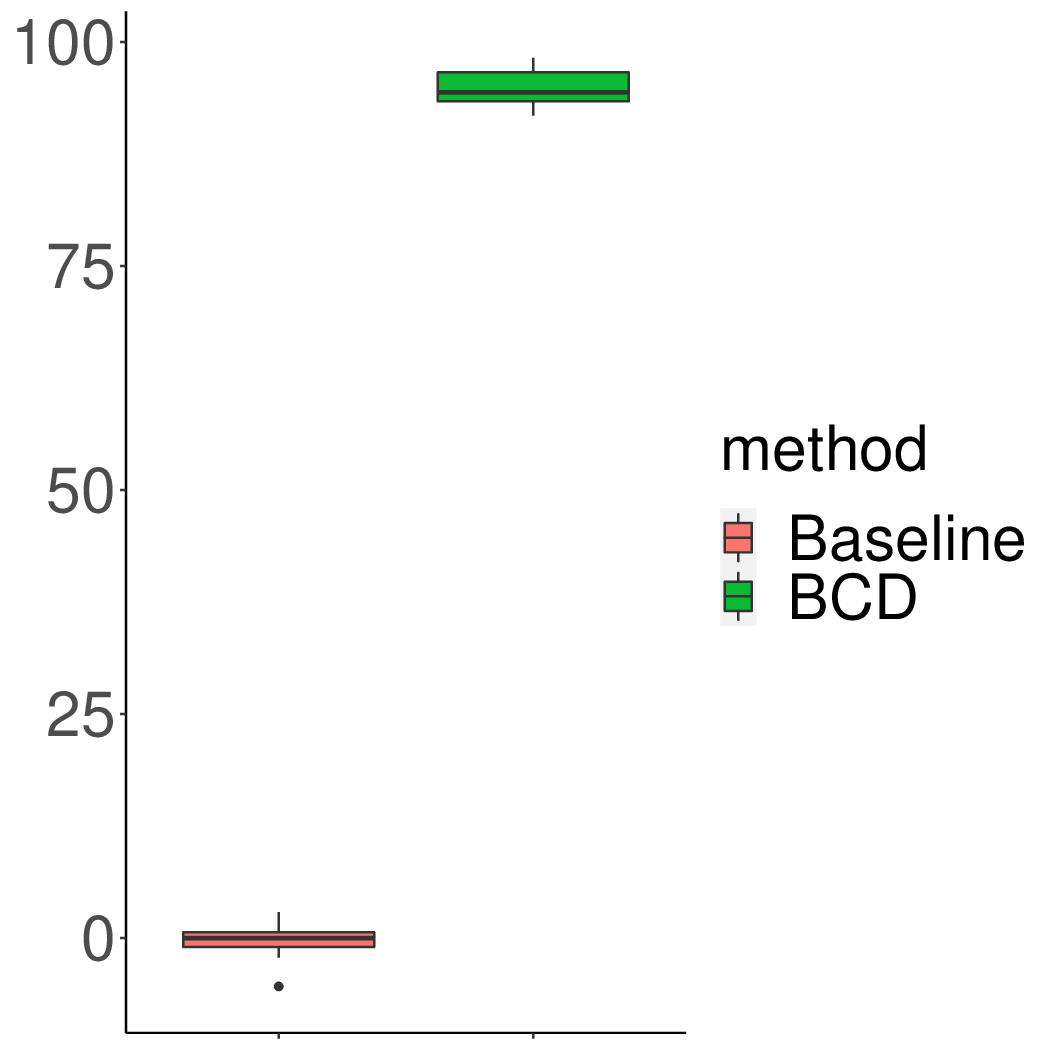}
          \caption*{\texttt{Barley}\\ \texttt{Facebook}}
    \end{subfigure}%
    \begin{subfigure}[t]{0.25\textwidth}
        \centering
        \includegraphics[width=\textwidth]{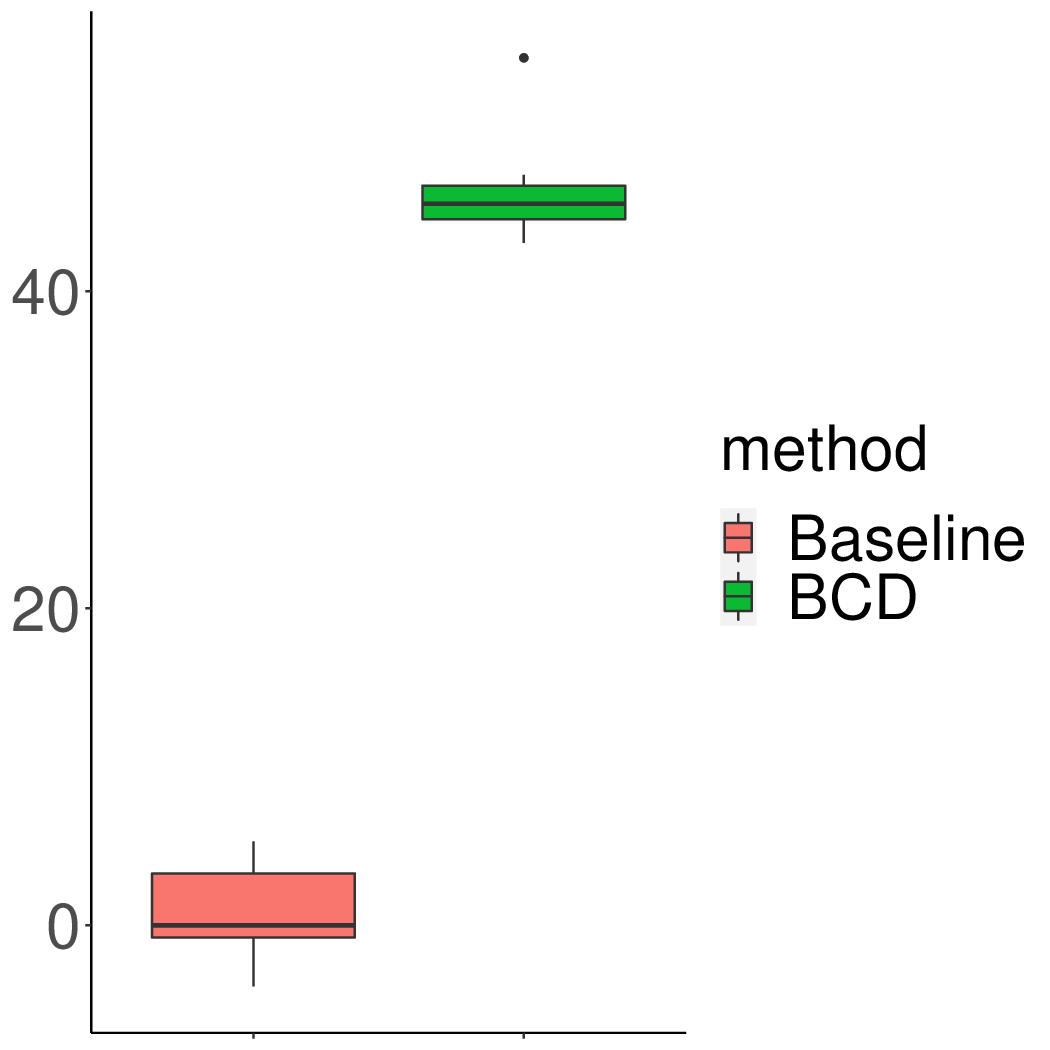}
        \caption*{\texttt{Hepar2}\\\texttt{celegans\_n306}}
    \end{subfigure}%
    \end{adjustbox}
    \caption{Test data log-likelihood normalized by $\sqrt{np}$ on real sorted DAGs. Top row: $n < p$. Bottom row: $n > p$.}
    \label{fig:testllreal1}
\end{figure}

Similar to the previous section, we first looked at the results on ordered DAGs. %In this case $B^*$ is strictly upper triangular and we can identify a DAG from the data. 
We considered four different combinations of network structures as shown in Table \ref{tab:realdag1}, with both $n > p$ and $n < p$. Because KGLasso does not scale well with $n$, we did not include it in our comparisons. BCD continued to outperform the baseline method by modeling the sample correlation. This improvement was more prominent when $n > p$, where BCD significantly reduced the number of false positive edges, achieving higher JI and lower SHD compared to the baseline as well as to itself in the $n < p$ case. Figure \ref{fig:testllreal1} compares the test data log-likelihood of the two methods across 10 simulations, and BCD scored significantly higher test data log-likelihood in all the cases. The ROC curves of the two methods is provided in a figure in the Supplementary Material. Both figures indicate the BCD method indeed gives better DAG estimates than the baseline method.

\begin{table}[!ht]
\centering
\resizebox{1\columnwidth}{!}{
\begin{tabular}{lclc|crrccrc}
  \toprule
  DAG & $\Theta$-Network& Method& ($n,p,s_0$) & E & FN & TP & FDR & JI & SHD & err($\widehat\Theta$) (err($\widehat\Theta^{(1)}$))\\
  \midrule
&&BCD & (100 446, 676)&  440.9 & 314.7 & 361.3 & 0.176 & 0.478 & 394.3 & 0.03458 (0.03564)\\
&&Baseline& (100 446, 676) & 436.6 & 334.2 & 341.8 & 0.211 & 0.444 & 429.0 & ---\\
Andes & facebook &&&&&&&&&\\
\ (2)&&BCD & (500 446, 676)& 500.0 & 197.3 & 478.7 & 0.043 & 0.686 & 218.6 & 0.07816 (0.07758)\\
&&Baseline& (500 446, 676) & 499.1 & 206.4 & 469.6 & 0.058 & 0.666 & 235.9 & ---\\
\midrule
& & BCD & (100, 224, 264) &154.5 & 124.2 & 139.8 & 0.092 & 0.502 & 138.9 & 0.03922 (0.02167)\\
&& Baseline & (100, 224, 264) &154.8 & 126.8 & 137.2 & 0.110 & 0.487 & 144.4 & ---\\
Hailfinder & celegan\_n306 &&&&&&&&&\\
\ (4)&&BCD & (500, 224, 264) &168.4 & 97.1 & 166.9 & 0.009 & 0.629 & 98.6 & 0.02010 (0.02138)\\
&&Baseline & (500, 224, 264) &168.0 & 98.0 & 166.0 & 0.012 & 0.624 & 100.0 & ---\\
\midrule
& & BCD & (100, 192, 336) & 211.0 & 150.5 & 185.5 & 0.119 & 0.513 & 176.0 & 0.01453 (0.00100)\\
&  & Baseline & (100, 192, 336) &211.9 & 156.2 & 179.8 & 0.150 & 0.489 & 188.3 & ---\\
Barley & facebook &&&&&&&&&\\
\ (4)&& BCD & (500, 192, 336) & 260.5 & 91.3 & 244.7 & 0.061 & 0.696 & 107.1 & 0.23144 (0.28172)\\
&& Baseline & (500, 192, 336) &260.2 & 97.6 & 238.4 & 0.083 & 0.666 & 119.4  & ---\\
\midrule
& & BCD & (100, 280, 492) & 366.7 & 234.5 & 257.5 & 0.295 & 0.428 & 343.7 & 0.05101 (0.03104)\\
& & Baseline & (100, 280, 492) & 373.9 & 238.0 & 254.0 & 0.314 & 0.416 & 357.9 & ---\\
Hepar2 & celegan\_n306 &&&&&&&&&\\
\ (4)& & BCD & (500, 280, 492) & 417.5 & 122.2 & 369.8 & 0.114 & 0.685 & 169.9 & 0.00759 (0.00755)\\
& & Baseline & (500, 280, 492) & 417.1 & 126.0 & 366.0 & 0.122 & 0.674 & 177.1 & ---\\
\bottomrule
\end{tabular}}
\caption{Results for ordered DAGs on real network data. Block size is 20 for $n < p$ and 50 for $n > p$. %The numbers of replicated DAG are: Andes (2), Hailfinder (4), Barley (4), Hepar2 (4). 
The number under each DAG reports the number of times it is duplicated to form a large DAG.
All numbers represent the average over 10 simulations.}
\label{tab:realdag1}
\end{table}

\subsubsection{Learning with de-correlation}\label{sec:real_random}
\begin{figure}[ht]
    \centering
    \begin{adjustbox}{minipage=\linewidth,scale=0.9}
    \begin{subfigure}[t]{0.25\textwidth}
        \centering
        \includegraphics[width=\textwidth]{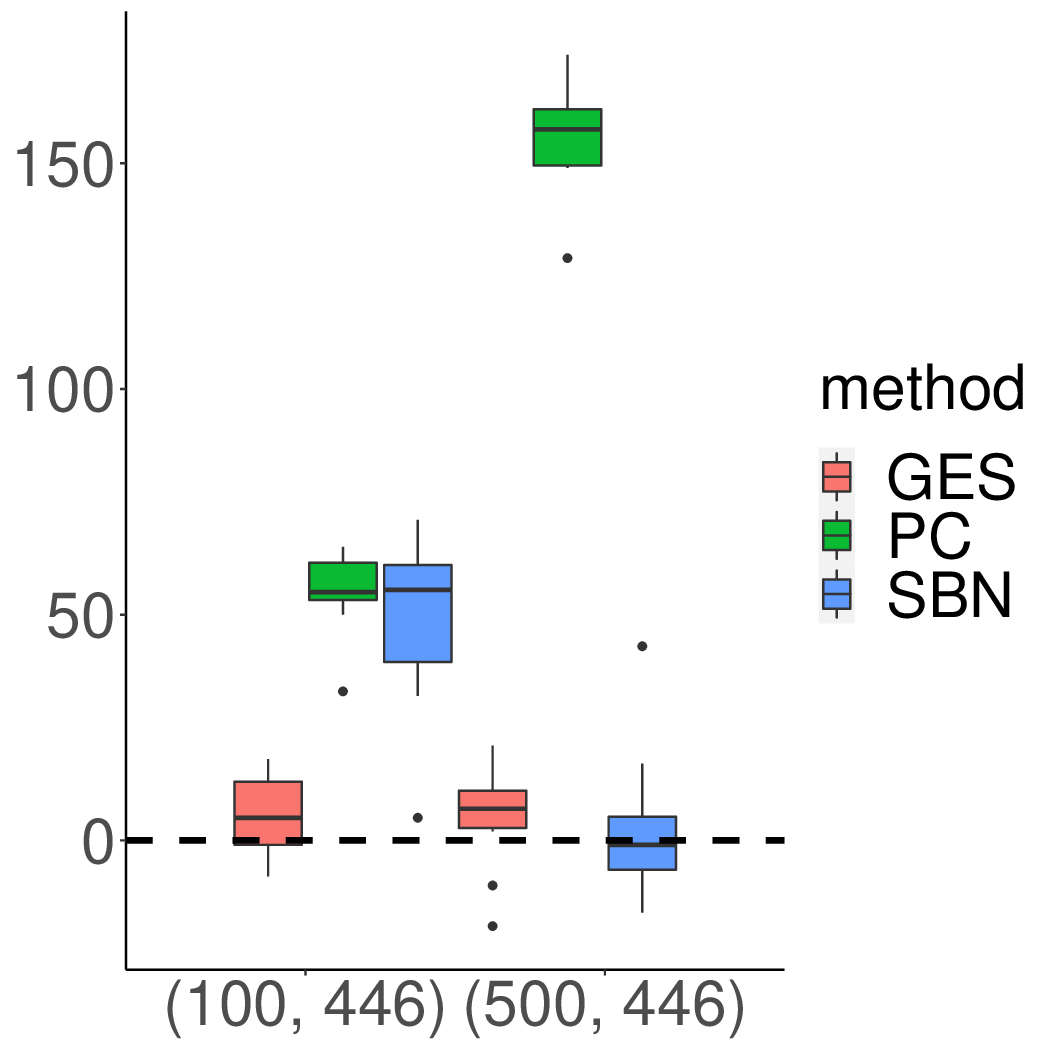}
        \caption*{\texttt{Andes}\\ \texttt{facebook}}
    \end{subfigure}%
    \begin{subfigure}[t]{0.25\textwidth}
        \centering
        \includegraphics[width=\textwidth]{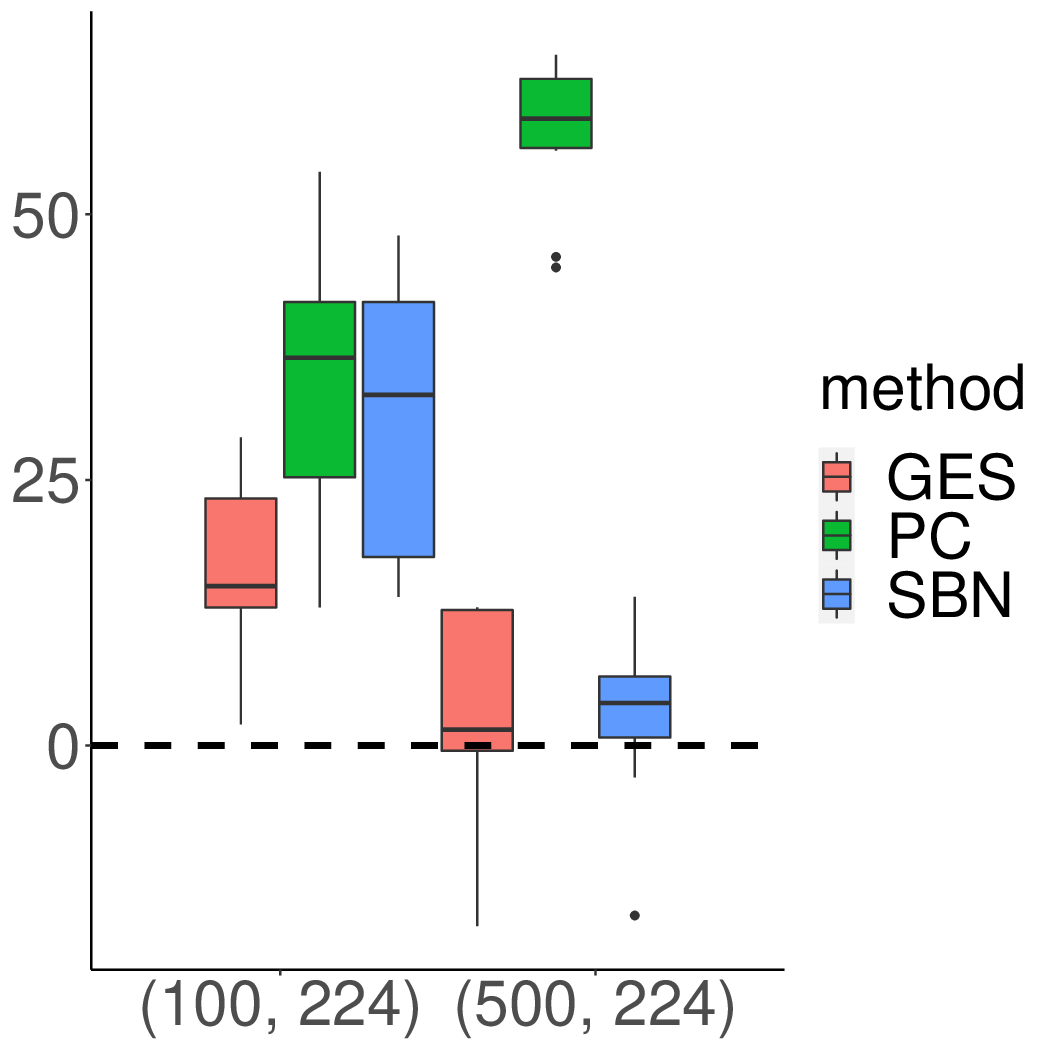}
        \caption*{\quad\texttt{Hailfinder}\\ \texttt{celegans\_n306}}
    \end{subfigure}%
    \begin{subfigure}[t]{0.25\textwidth}
        \centering
        \includegraphics[width=\textwidth]{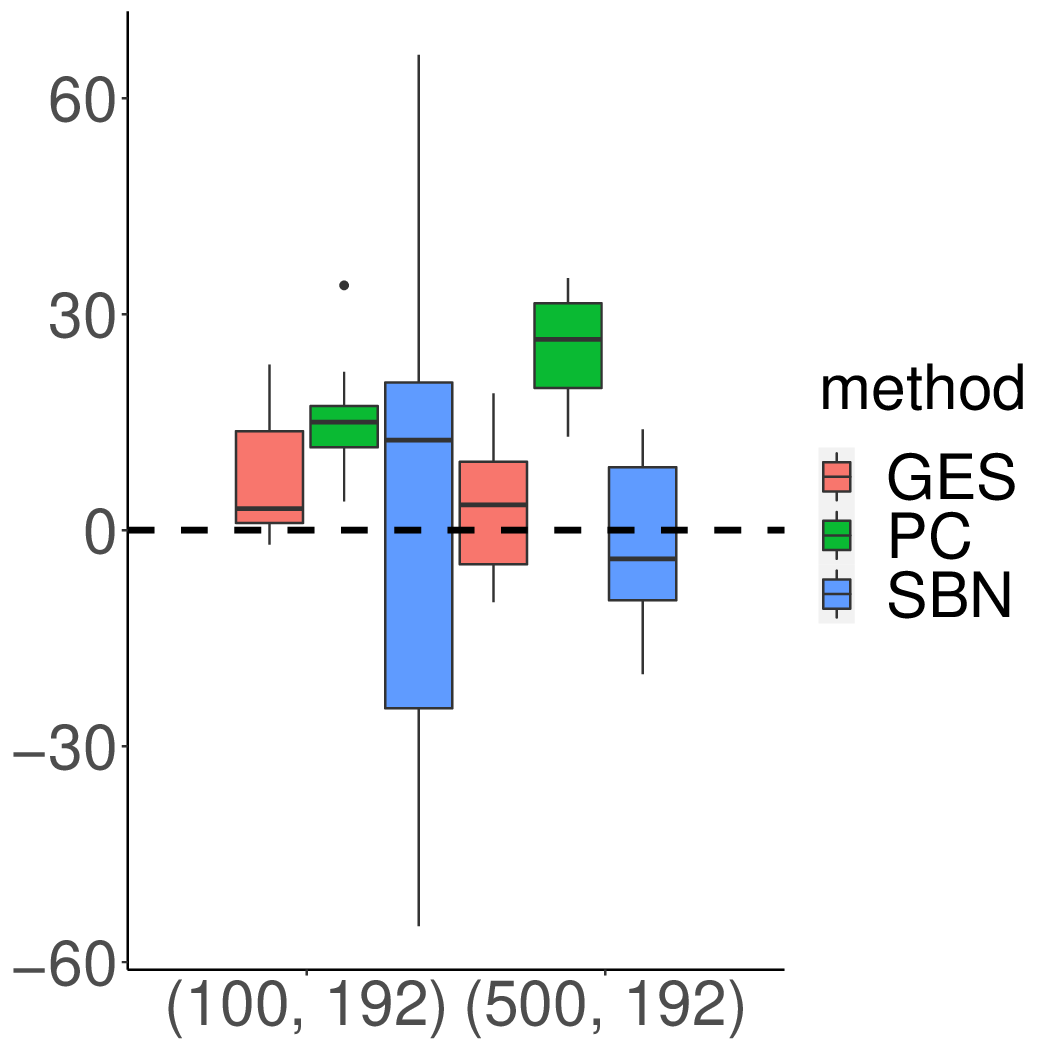}
        \caption*{\texttt{Barley}\\ \texttt{facebook}}
    \end{subfigure}%
    \begin{subfigure}[t]{0.25\textwidth}
        \centering
        \includegraphics[width=\textwidth]{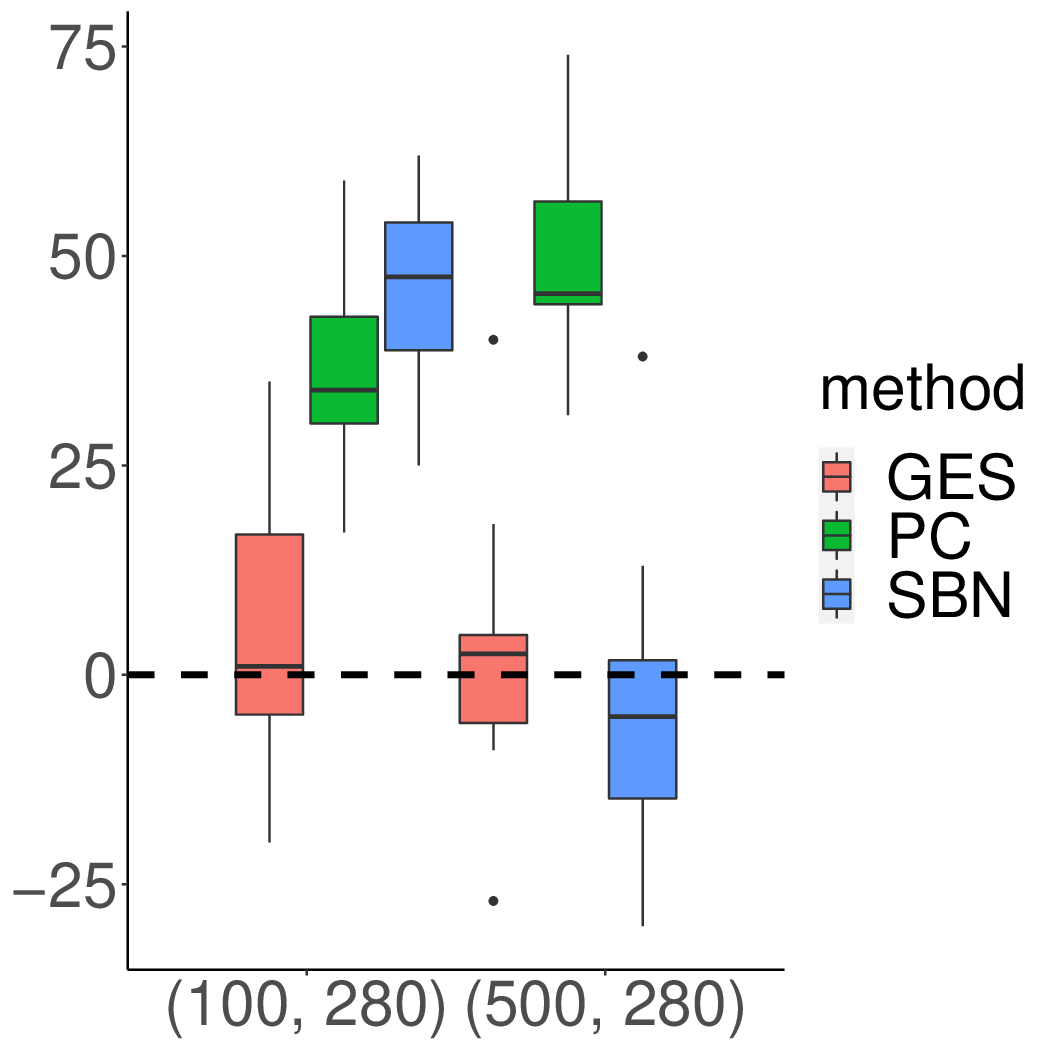}
        \caption*{\texttt{Hepar2}\\\texttt{celegans\_n306}}
    \end{subfigure}%
    
        \begin{subfigure}[t]{0.25\textwidth}
        \centering
        \includegraphics[width=\textwidth]{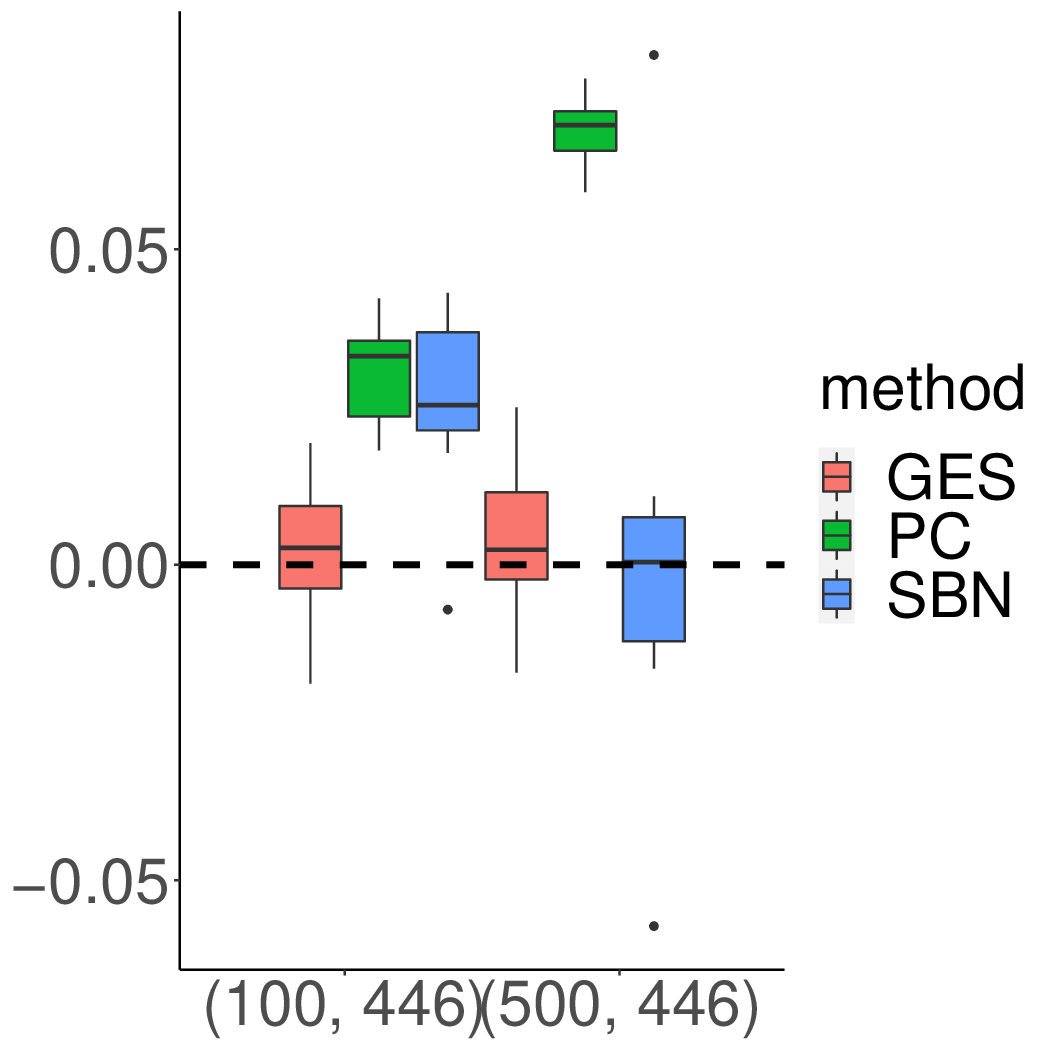}
        \caption*{\texttt{Andes}\\ \texttt{facebook}}
    \end{subfigure}%
    \begin{subfigure}[t]{0.25\textwidth}
        \centering
        \includegraphics[width=\textwidth]{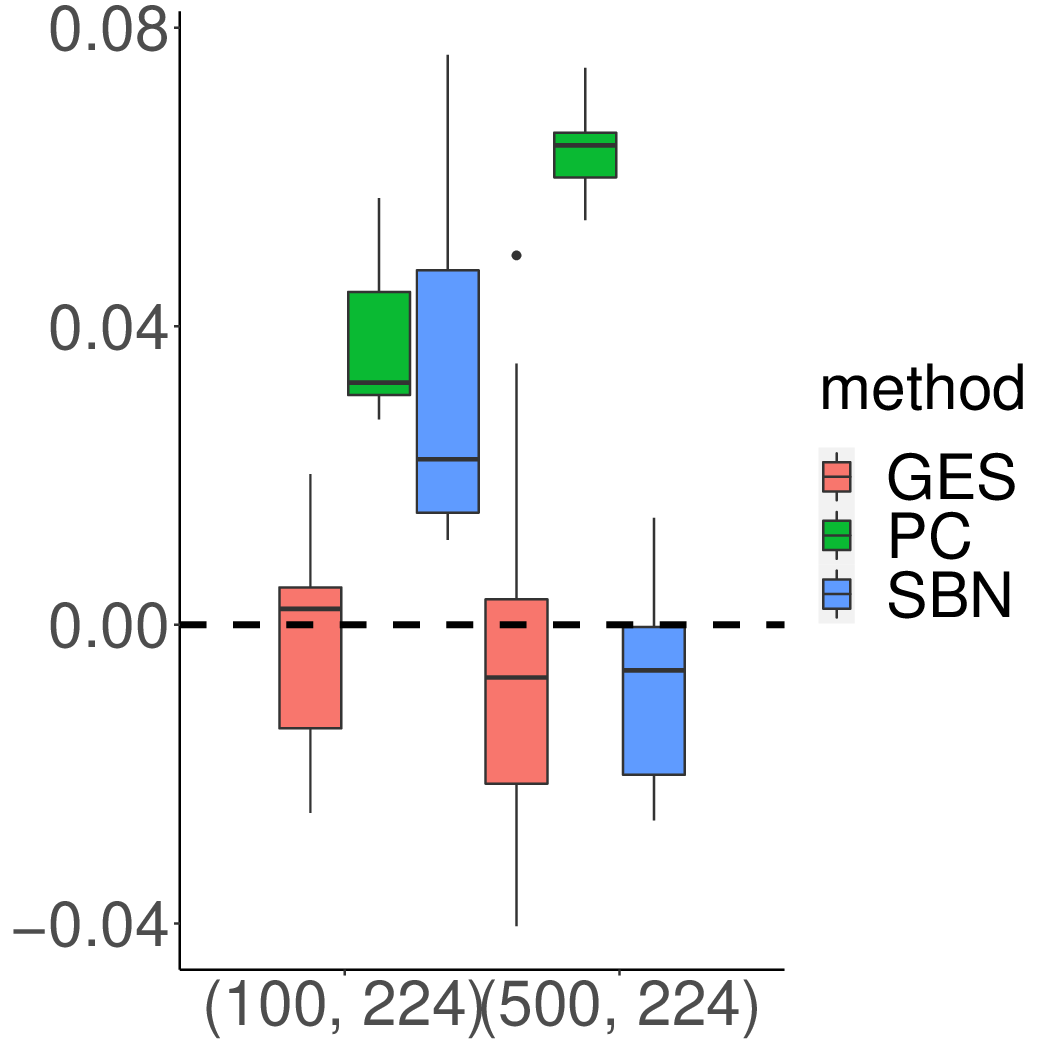}
        \caption*{\quad\texttt{Hailfinder}\\ \texttt{celegans\_n306}}
    \end{subfigure}%
    \begin{subfigure}[t]{0.25\textwidth}
        \centering
        \includegraphics[width=\textwidth]{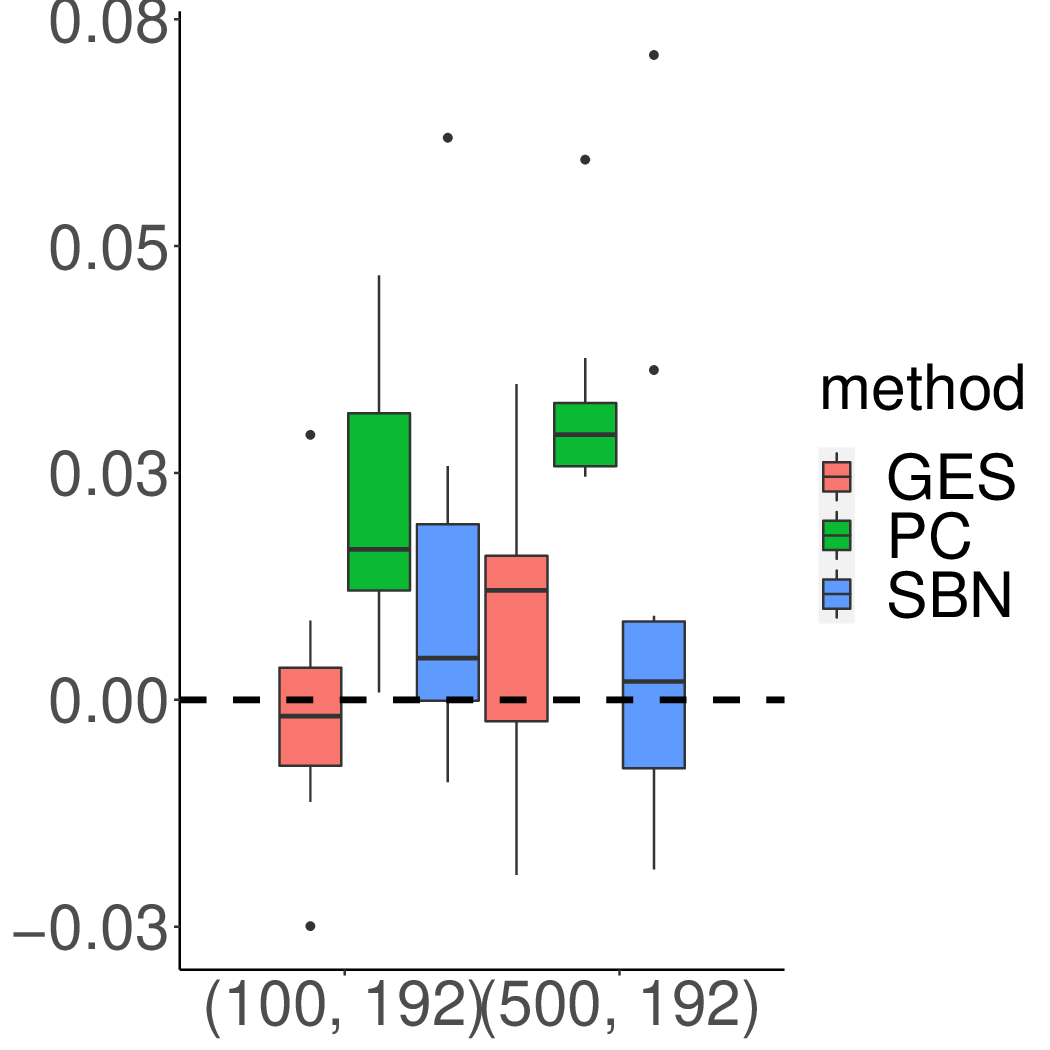}
        \caption*{\texttt{Barley}\\ \texttt{facebook}}
    \end{subfigure}%
    \begin{subfigure}[t]{0.25\textwidth}
        \centering
        \includegraphics[width=\textwidth]{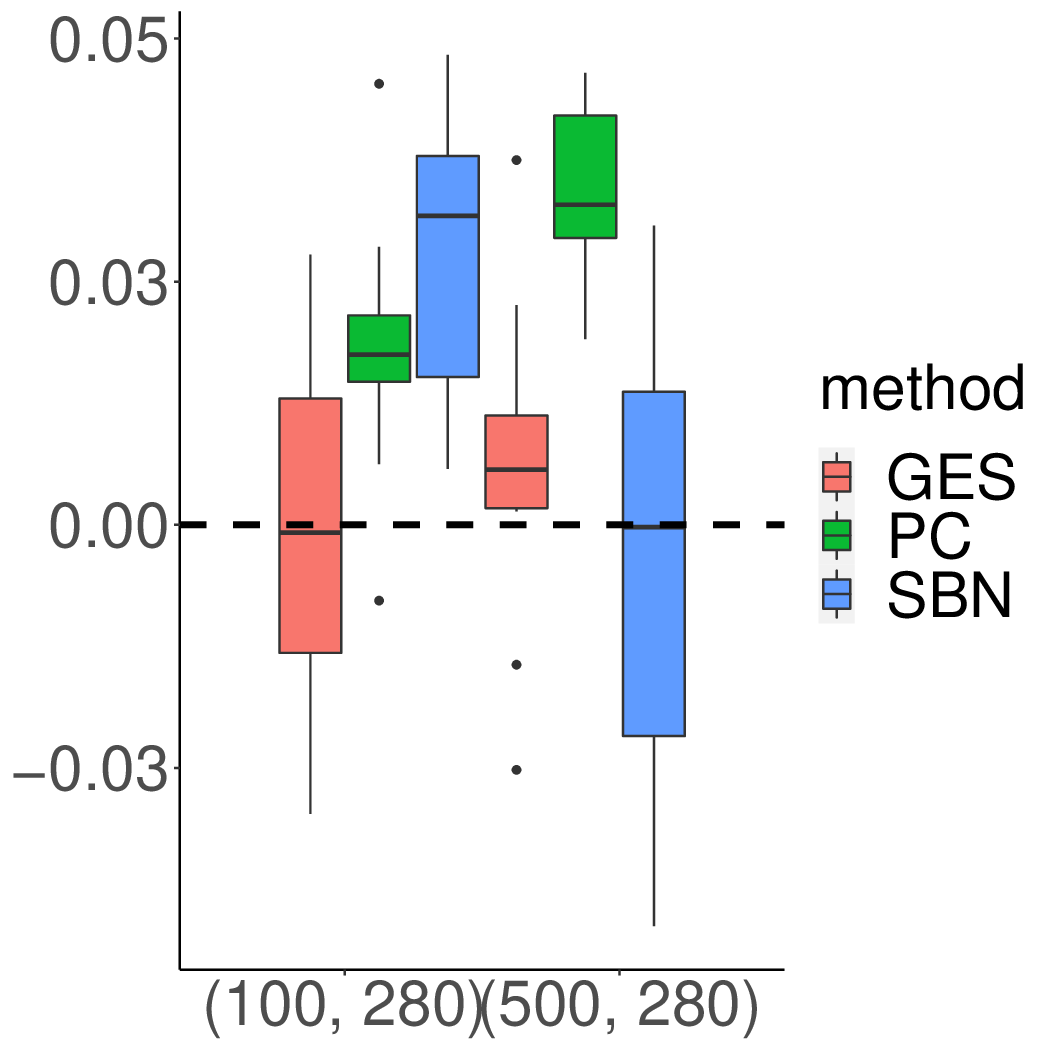}
        \caption*{\texttt{Hepar2}\\\texttt{celegans\_n306}}
    \end{subfigure}%
    \end{adjustbox}
    \caption{Experiments on real unsorted DAGs. Decrease in SHD (top row) and increase in JI (bottom row) via de-correlation for real networks, where the x-axis reports the value of $(n,p)$. %y-axis represents the difference between the SHD (JI) before and after decorrelation. x-axis represents the value of $(n,p)$. 
    In each panel, the three boxplots on the left and the three on the right correspond to cases of $n < p$ and $n > p$, respectively.}% 
    \label{fig:shdreal}
\end{figure}

\begin{figure}
    \centering
    \begin{adjustbox}{minipage=\linewidth,scale=0.8}
    \begin{subfigure}[t]{0.25\textwidth}
        \centering
        \includegraphics[width=\textwidth]{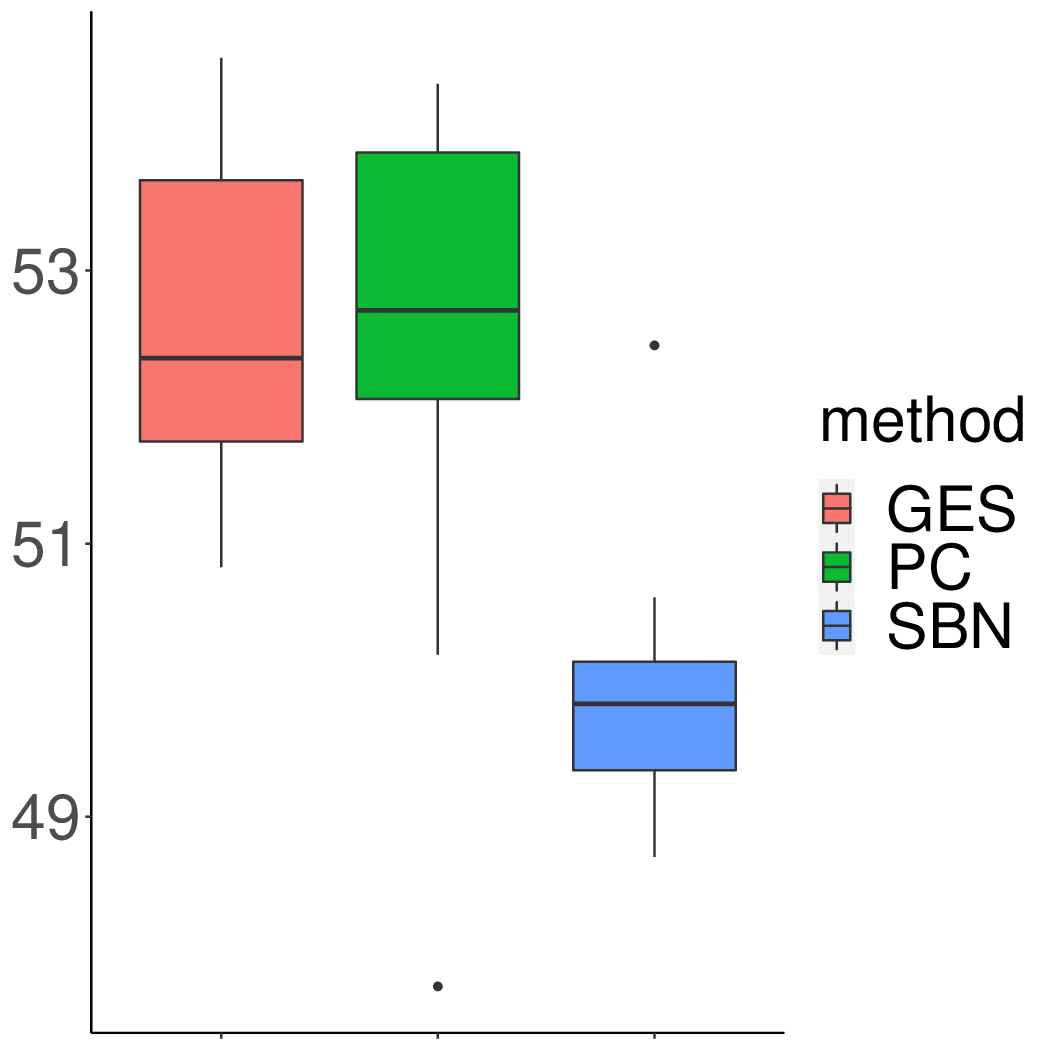}
         \caption*{\texttt{Andes}\\\texttt{Facebook}}
    \end{subfigure}%
    \begin{subfigure}[t]{0.25\textwidth}
        \centering
        \includegraphics[width=\textwidth]{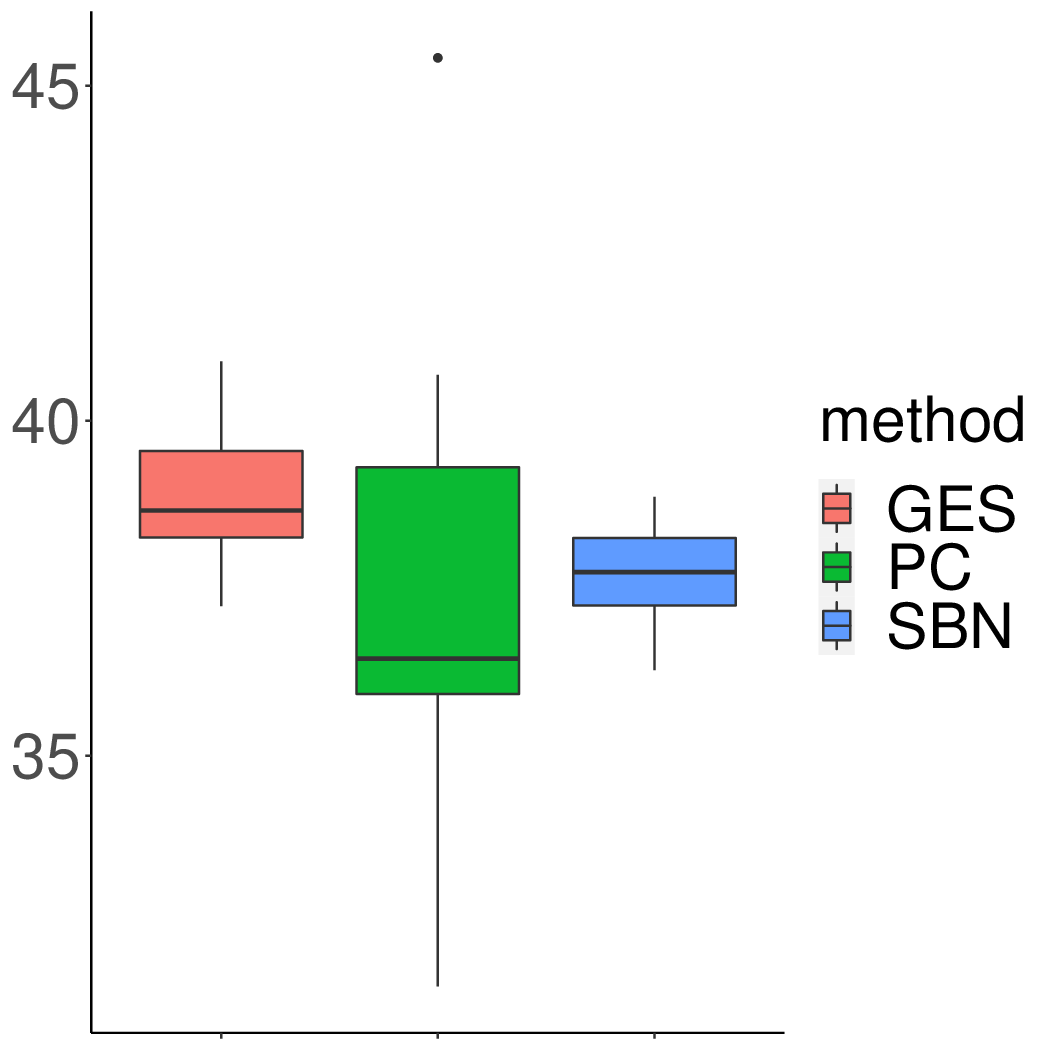}
        \caption*{\quad\texttt{Hailfinder}\\ \texttt{celegans\_n306}}
    \end{subfigure}%
    \begin{subfigure}[t]{0.25\textwidth}
        \centering
        \includegraphics[width=\textwidth]{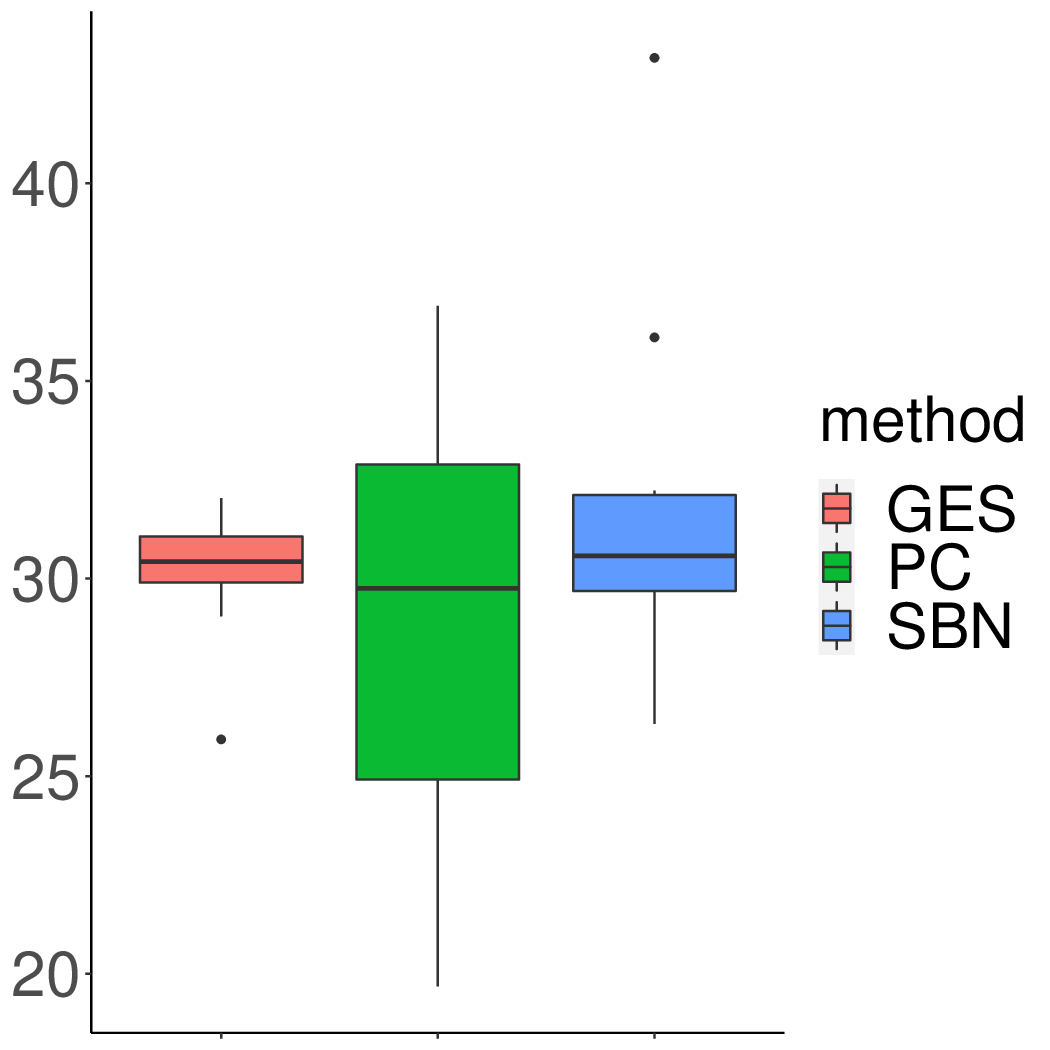}
          \caption*{\texttt{Barley}\\ \texttt{Facebook}}
    \end{subfigure}%
    \begin{subfigure}[t]{0.25\textwidth}
        \centering
        \includegraphics[width=\textwidth]{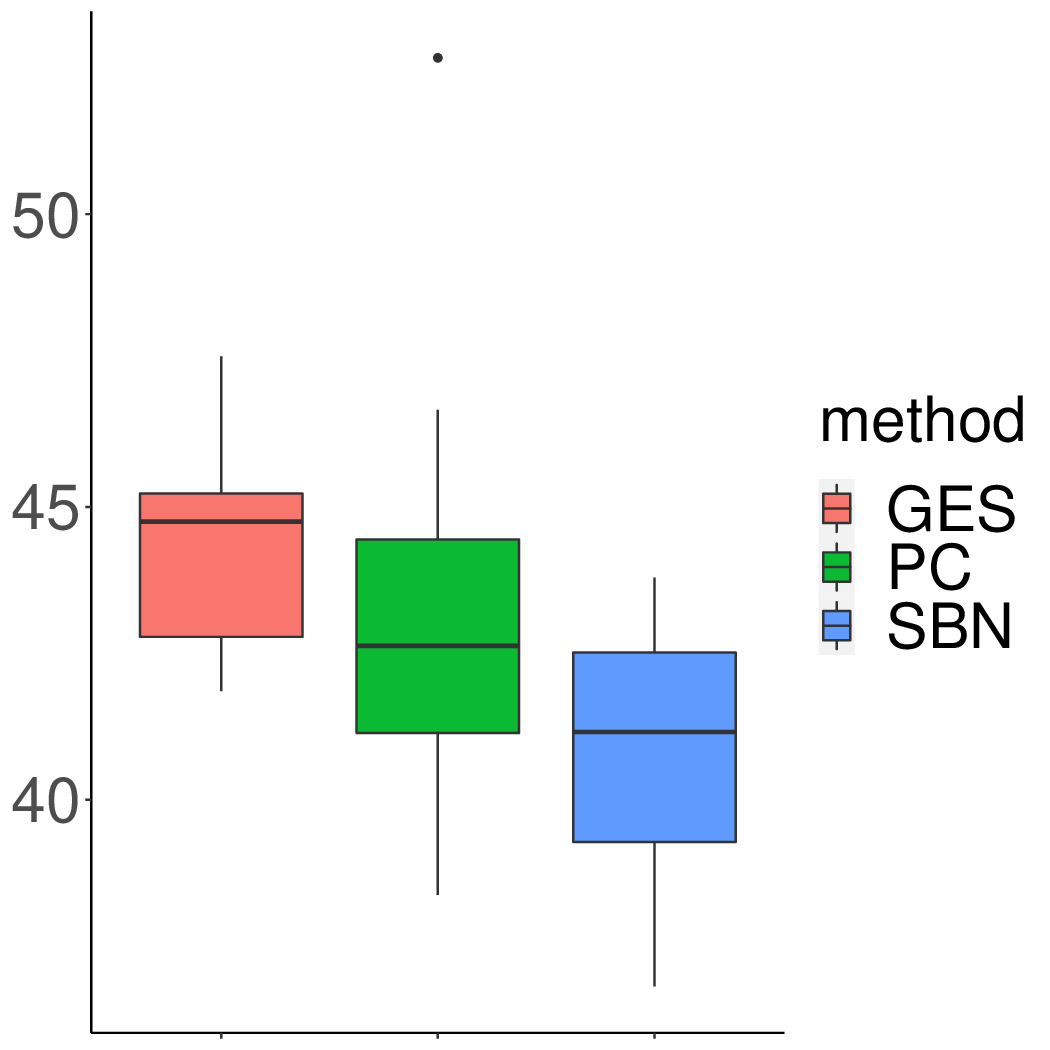}
        \caption*{\texttt{Hepar2}\\\texttt{celegans\_n306}}
    \end{subfigure}%
    
    \begin{subfigure}[t]{0.25\textwidth}
        \centering
        \includegraphics[width=\textwidth]{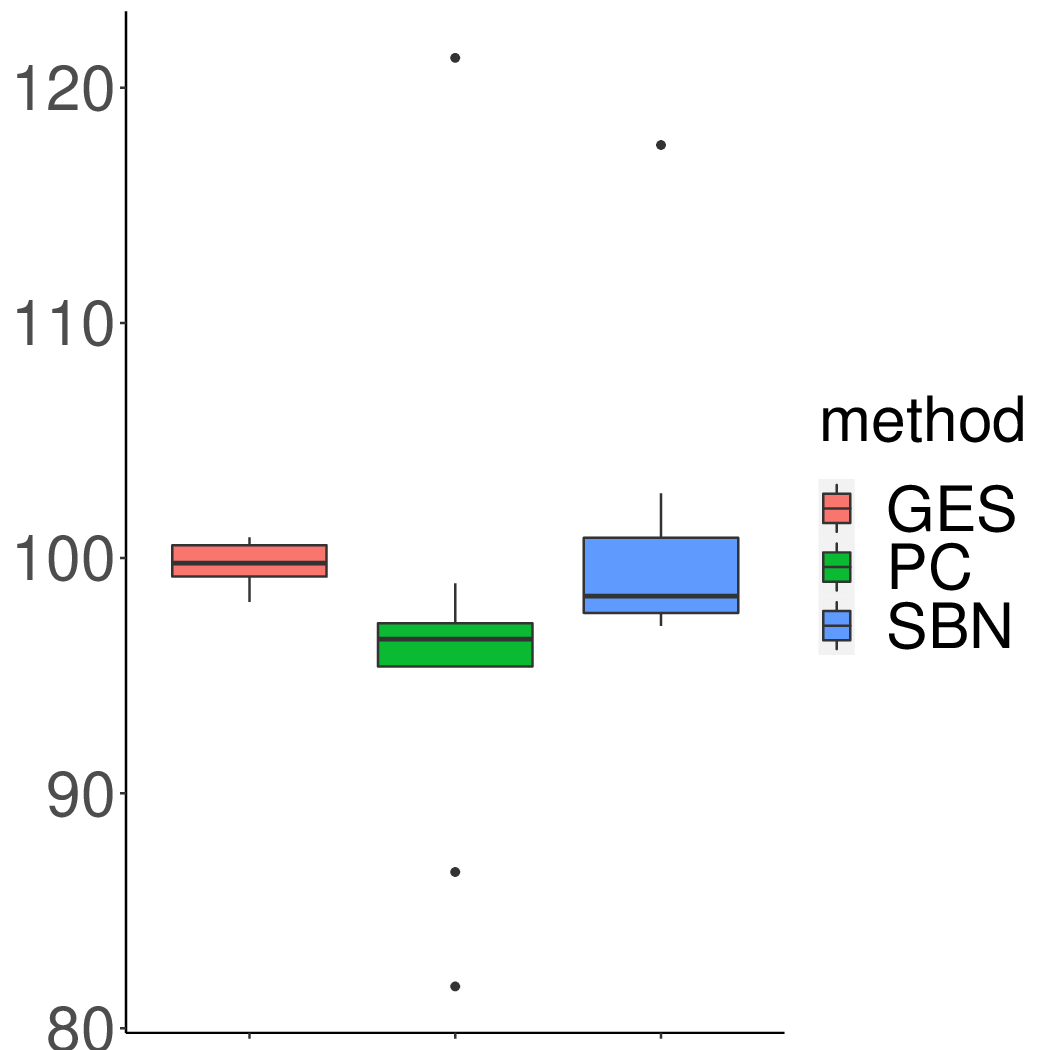}
         \caption*{\texttt{Andes}\\\texttt{Facebook}}
    \end{subfigure}%
    \begin{subfigure}[t]{0.25\textwidth}
        \centering
        \includegraphics[width=\textwidth]{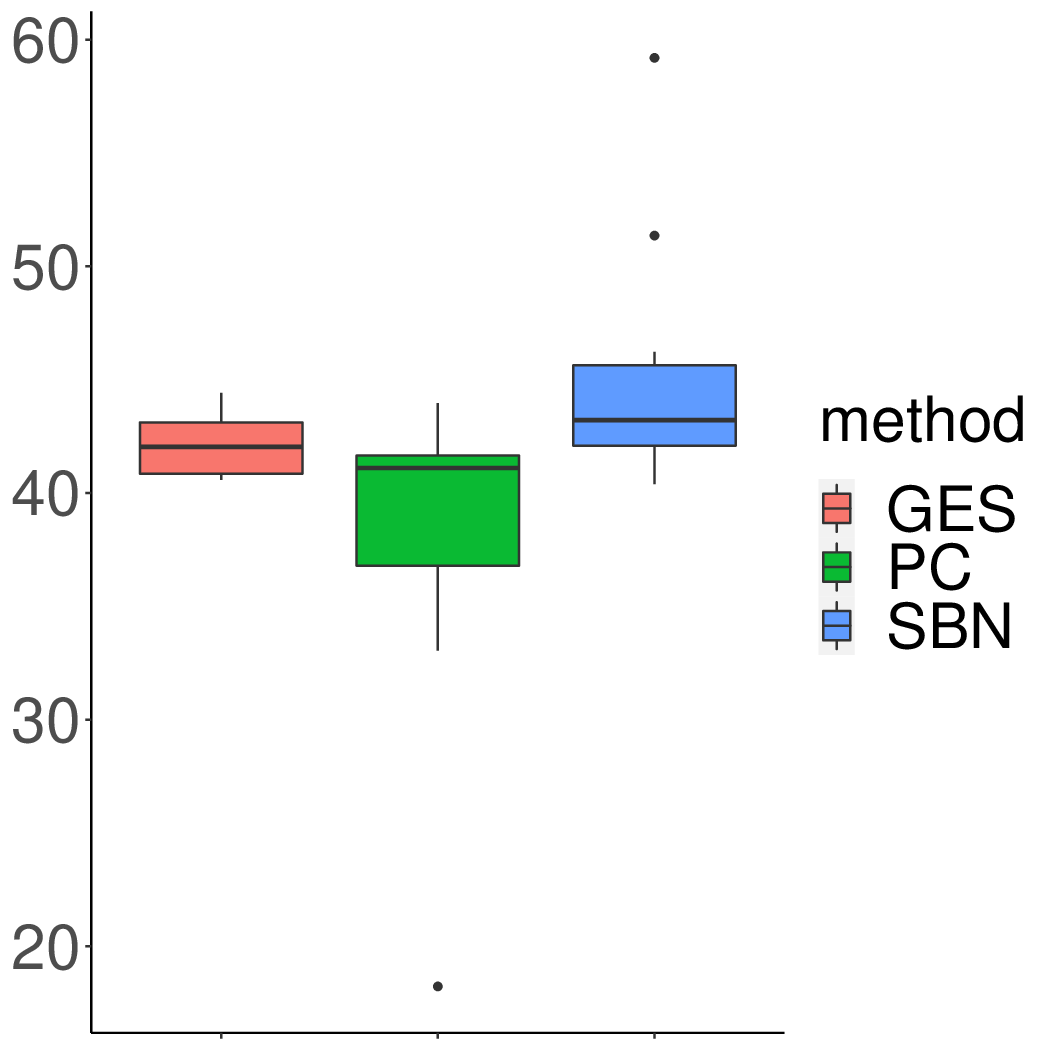}
        \caption*{\quad\texttt{Hailfinder}\\ \texttt{celegans\_n306}}
    \end{subfigure}%
    \begin{subfigure}[t]{0.25\textwidth}
        \centering
        \includegraphics[width=\textwidth]{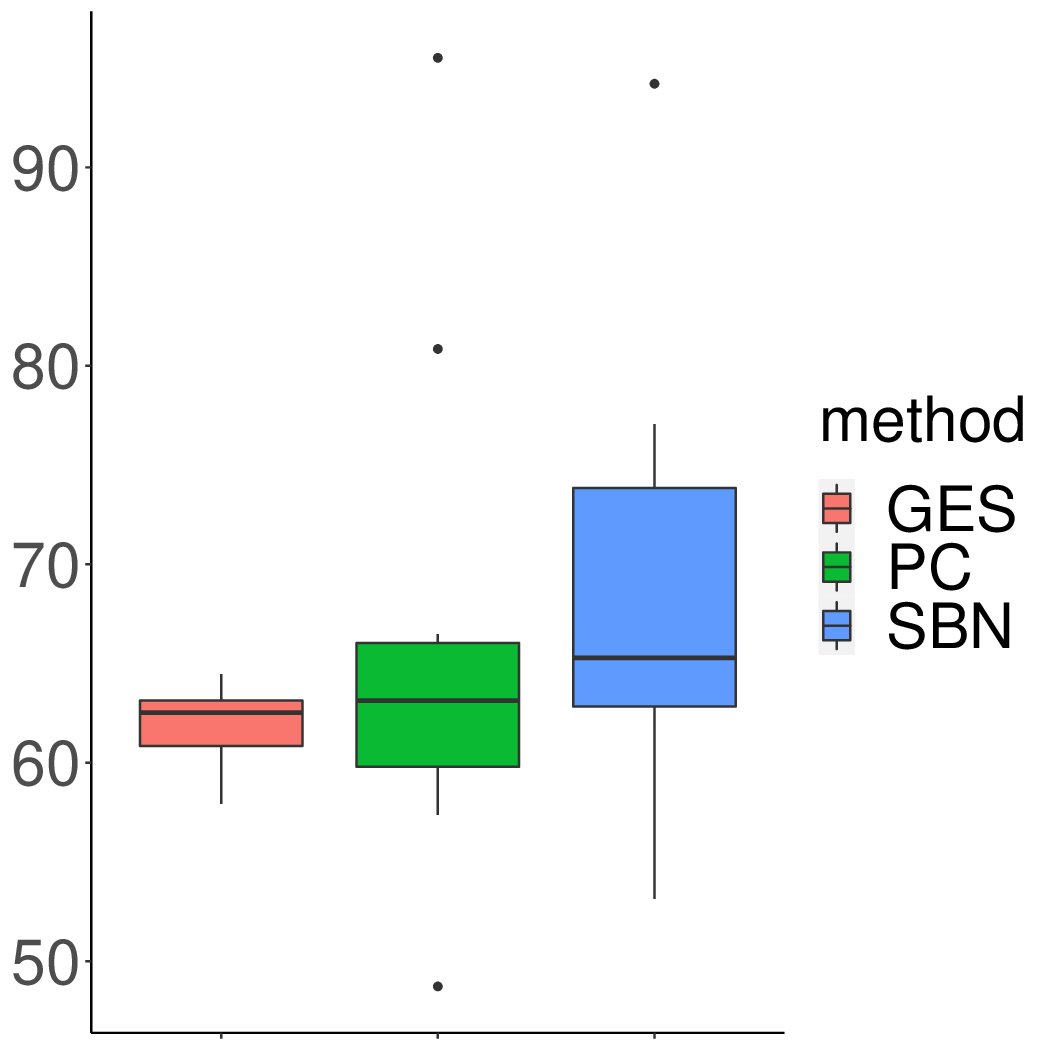}
          \caption*{\texttt{Barley}\\ \texttt{Facebook}}
    \end{subfigure}%
    \begin{subfigure}[t]{0.25\textwidth}
        \centering
        \includegraphics[width=\textwidth]{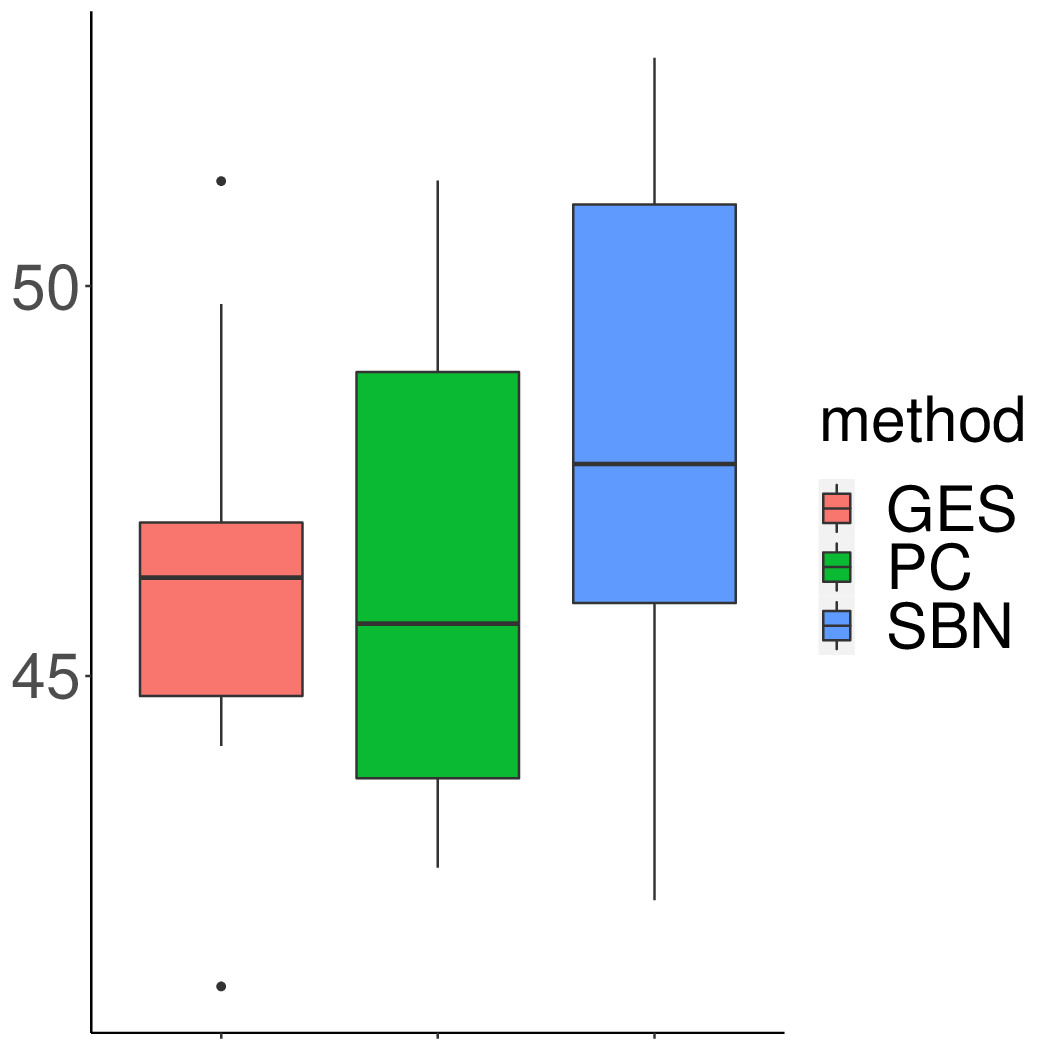}
        \caption*{\texttt{Hepar2}\\\texttt{celegans\_n306}}
    \end{subfigure}%
    \end{adjustbox}
    \caption{Increases in test data log-likelihood on real unsorted DAGs. Top row: $n < p$. Bottom row: $n > p$. Some outliers in the top panels are not shown for a better view of the boxplots.}
    \label{fig:testllreal2}
\end{figure}

When the ordering of the DAG nodes is not given, we compared the effect of decorrelation as in Section \ref{sec:decor}. All network parameters were generated in the same way as before but we randomly shuffled the columns of $X$. The decrease in the structural Hamming distance and increase in Jaccard index from decorrelation over 10 simulations are summarized as boxplots in Figure \ref{fig:shdreal}. PC performed uniformly better after decorrelation compared to before. GES and sparsebn also improved after decorrelation in most cases. The changes in the test data log-likelihood are shown in Figure \ref{fig:testllreal2}, which are positive for almost all date sets, except two outliers (removed from plots) of sparsebn in the second and fourth panels in the top row.

\subsubsection{Learning under general covariance structure}\label{sec:real_general}
In the following experiments, we generated the support of $\Theta^*$ without the block-diagonal constraint by directly sampling the real undirected networks \texttt{facebook} and \texttt{celegans\_n306}.  In other words, the underlying undirected network may have only one connected subgraph where all individuals are dependent. This setup poses major challenge particularly for the estimation of $\Theta$ because its support becomes much larger, %the effect sample size becomes much smaller, 
and as a result, we will need to impose stronger regularization in the graphical Lasso step when $n>p$. For simplicity, we focus on the setting $p > n$ so that we can still fix $\lambda_2 = 0.01$. Proceeding as before, we generated $B^*$ from real DAGs with duplications: \texttt{Andes}, \texttt{Hailfinder}, \texttt{Barley}, and \texttt{Hepar2}.

\begin{table}[ht]
\centering
\resizebox{1\columnwidth}{!}{
\begin{tabular}{lclc|crrccrc}
  \toprule
  DAG & $\Theta$-Network& Method& ($n,p,s_0$) & E & FN & TP & FDR & JI & SHD & err($\widehat\Theta$) (err($\widehat\Theta^{(1)}$))\\
  \midrule
Andes & facebook &BCD & (100 446, 676)& 424.1 & 319.9 & 356.1 & 0.155 & 0.478 & 387.9&   0.01531 ( 0.00288)\\
\ $(2)$&&Baseline& (100 446, 676) & 422.2 & 326.2 & 349.8 & 0.165 & 0.467 & 398.6 & ---\\
\\
Hailfinder &celegans\_n306  & BCD & (100,224,264) &122.0 & 163.0 & 101.0 & 0.172 & 0.354 & 184.0 & 0.08678 (0.08309)\\
\ $(4)$&& Baseline &  (100,224,264) & 122.0 & 161.0 & 103.0 & 0.156 & 0.364 & 180.0 & ---\\
\\
Barley & facebook & BCD & (100,192,336) & 252.6 & 147.3 & 188.7 & 0.248 & 0.471 & 211.2 & 0.08021 (0.08555)\\
\ $(4)$&& Baseline & (100,192,336) & 249.4 & 155.1 & 180.9 & 0.268 & 0.448 & 223.6  & ---\\
\\
Hepar2 & celegans\_n306 & BCD &(100, 420, 738) & 492.3 & 386.7 & 351.3 & 0.285 & 0.399 & 527.7 &0.03725 (0.03614)\\
\ $(6)$&& Baseline & (100, 420, 738) & 490.8 & 397.8 & 340.2 & 0.303 & 0.384 & 548.4  & ---\\
\bottomrule
\end{tabular}}
\caption{Results for ordered DAGs on real network data without block structure.}
\label{tab:generaltheta}
\end{table}

 \begin{figure}[ht]
    \centering
    \begin{adjustbox}{minipage=\linewidth,scale=0.8}
    \begin{subfigure}[t]{0.25\textwidth}
        \centering
        \includegraphics[width=\textwidth]{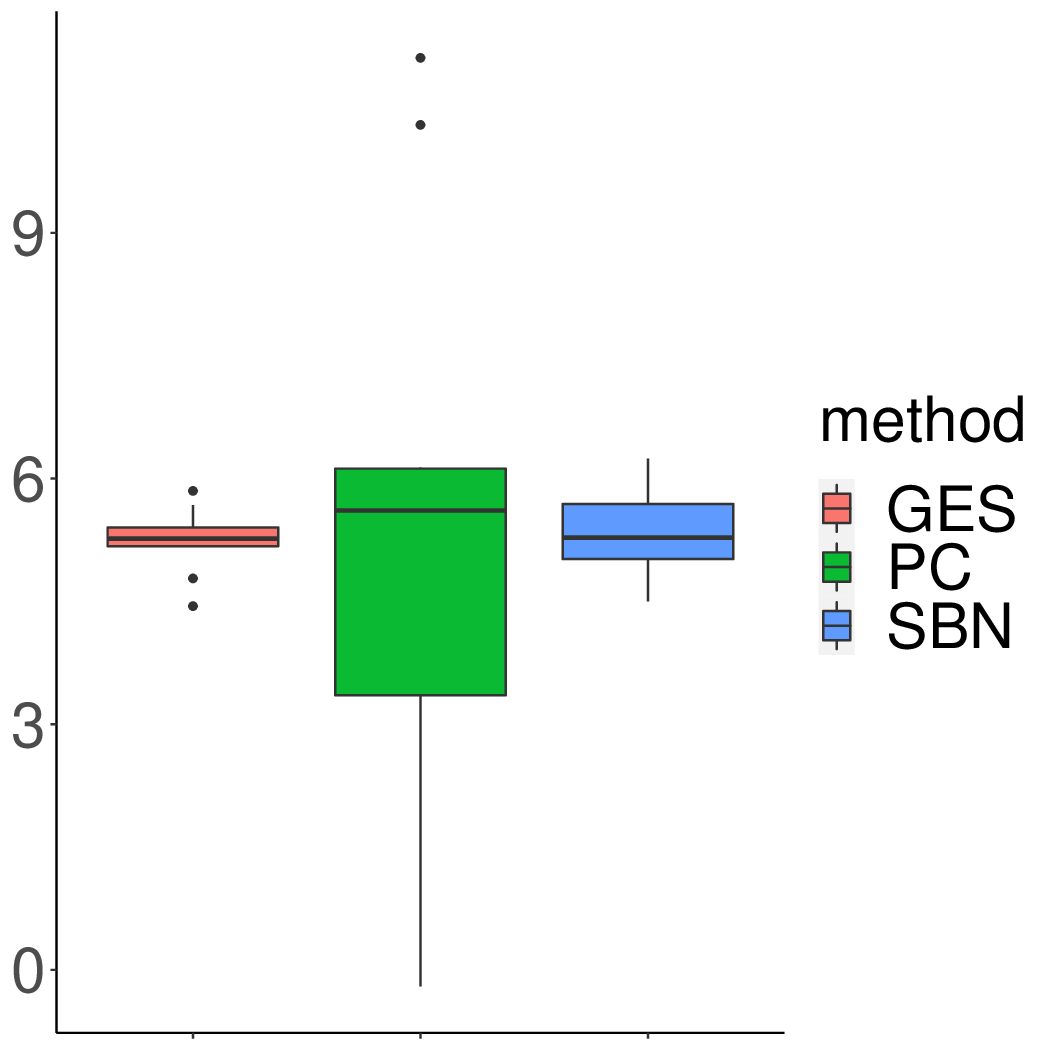}
         \caption*{\texttt{Andes}\\\texttt{Facebook}}
    \end{subfigure}%
    \begin{subfigure}[t]{0.25\textwidth}
        \centering
        \includegraphics[width=\textwidth]{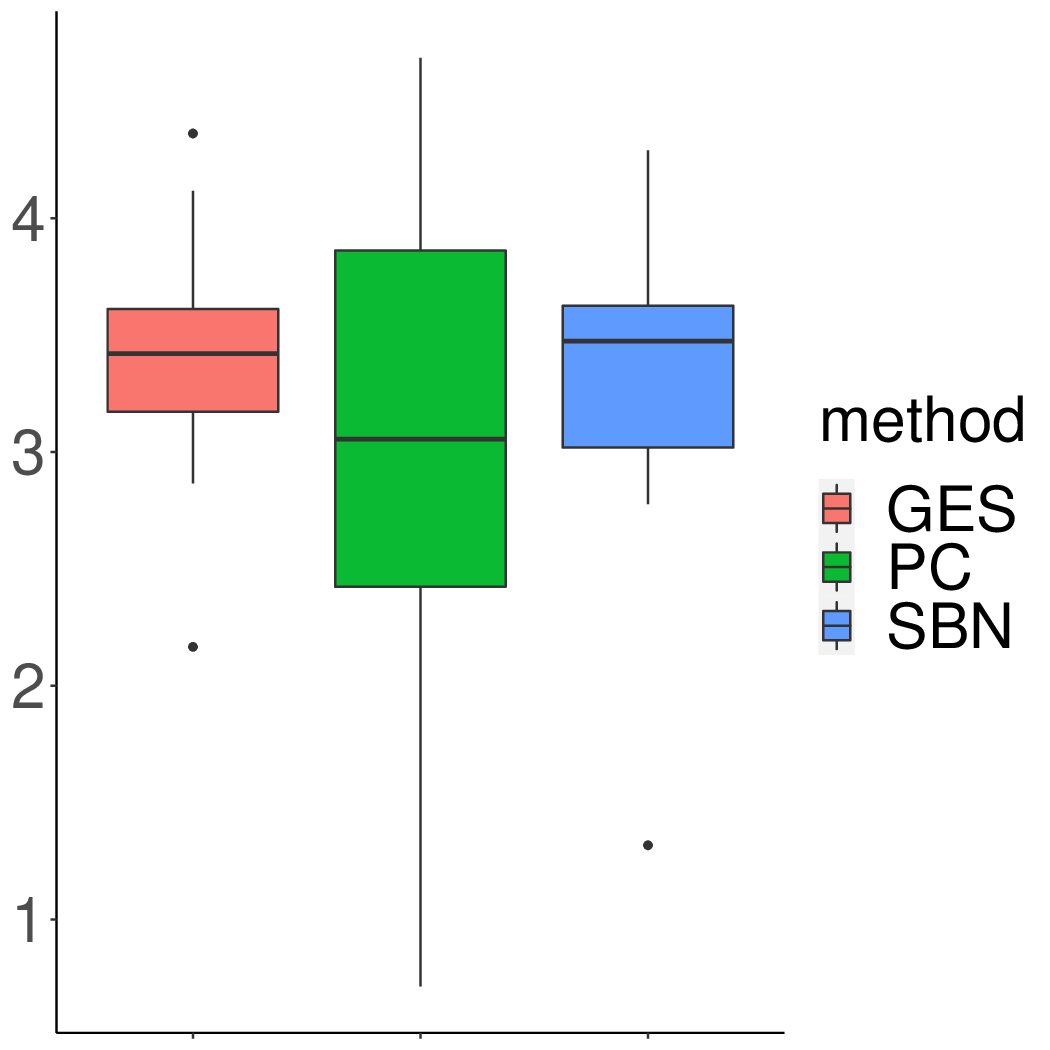}
        \caption*{\texttt{Barley}\\ \texttt{Facebook}}
    \end{subfigure}%
    \begin{subfigure}[t]{0.25\textwidth}
        \centering
        \includegraphics[width=\textwidth]{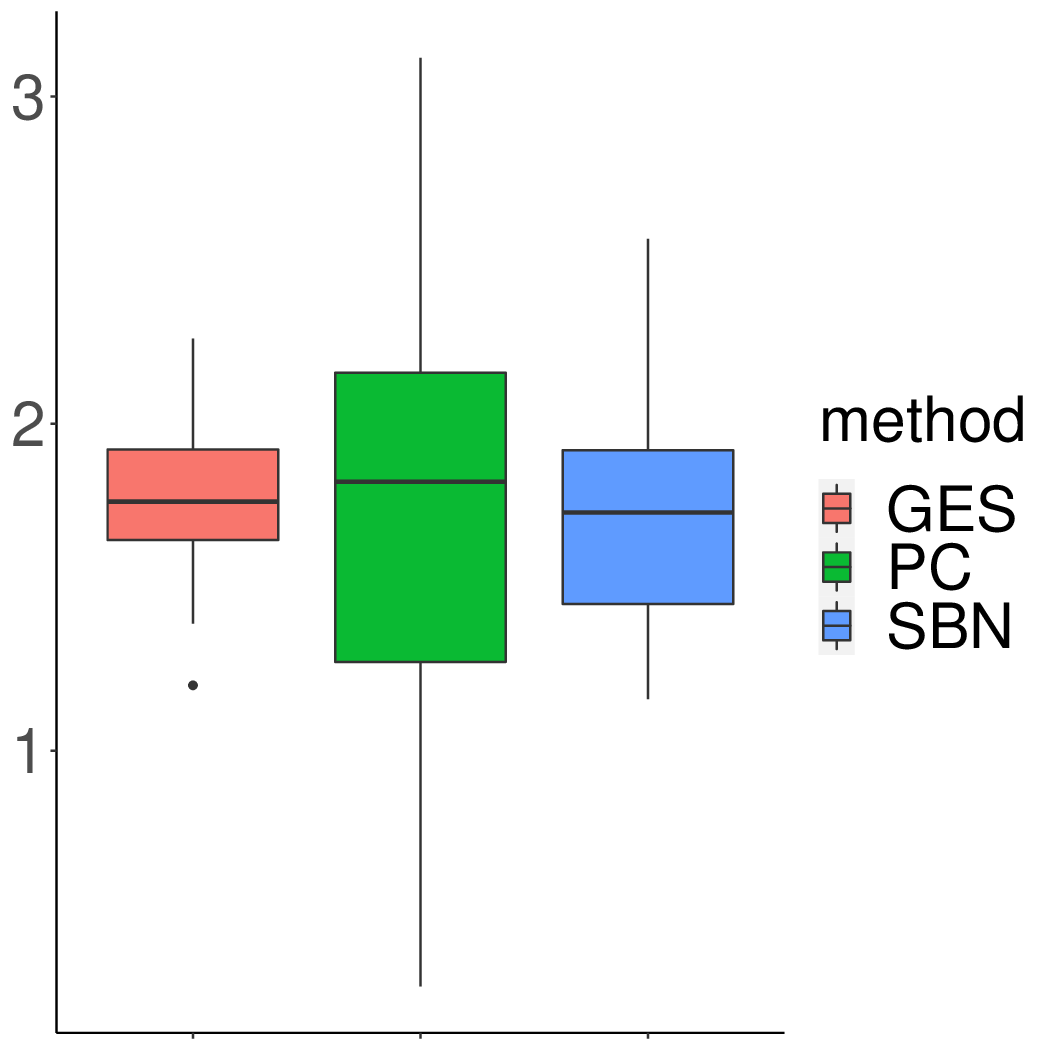}
         \caption*{\qquad\texttt{Hailfinder}\\ \qquad\texttt{celegans\_n306}}
    \end{subfigure}%
    \begin{subfigure}[t]{0.25\textwidth}
        \centering
        \includegraphics[width=\textwidth]{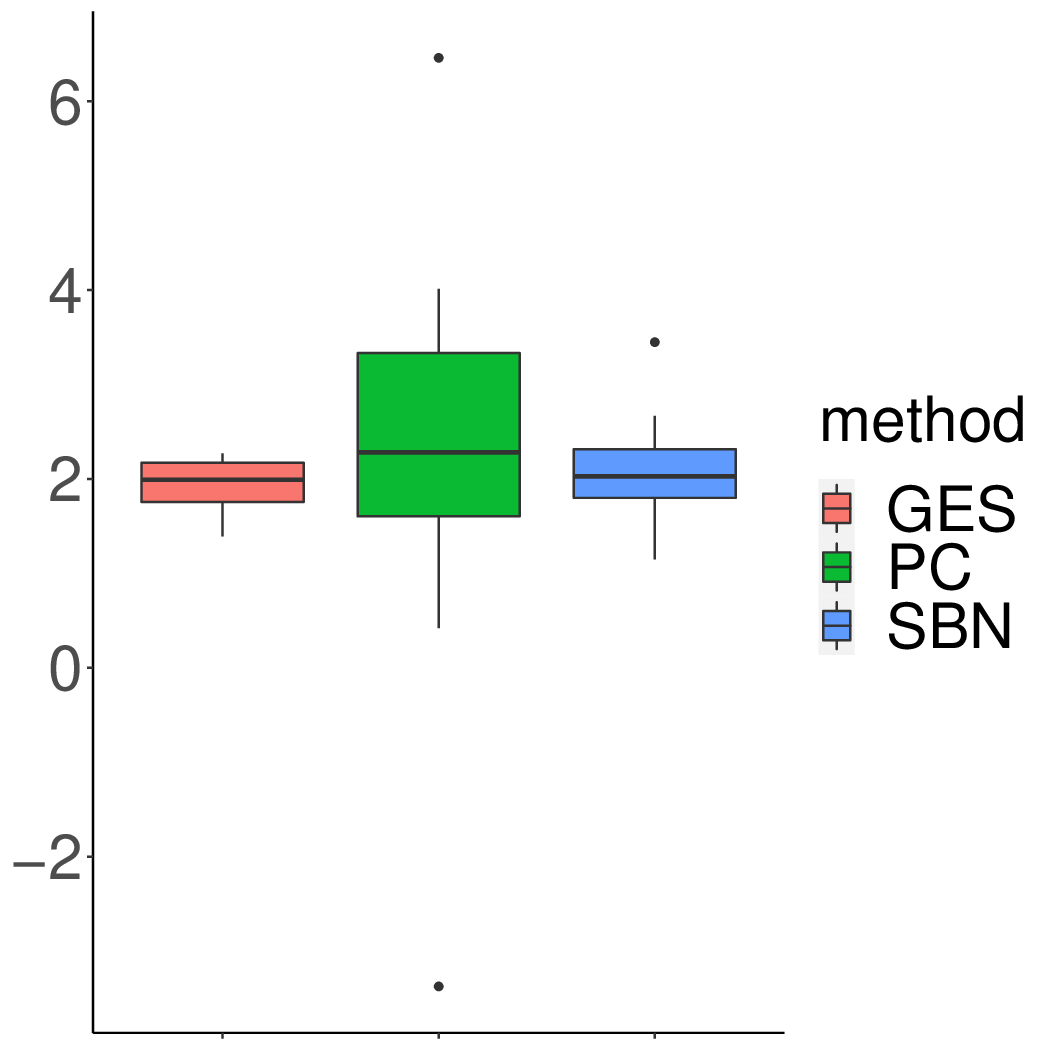}
        \caption*{\qquad\texttt{Hepar2}\\ \qquad\texttt{celegans\_n306}}
    \end{subfigure}%
    
    \begin{subfigure}[t]{0.25\textwidth}
        \centering
        \includegraphics[width=\textwidth]{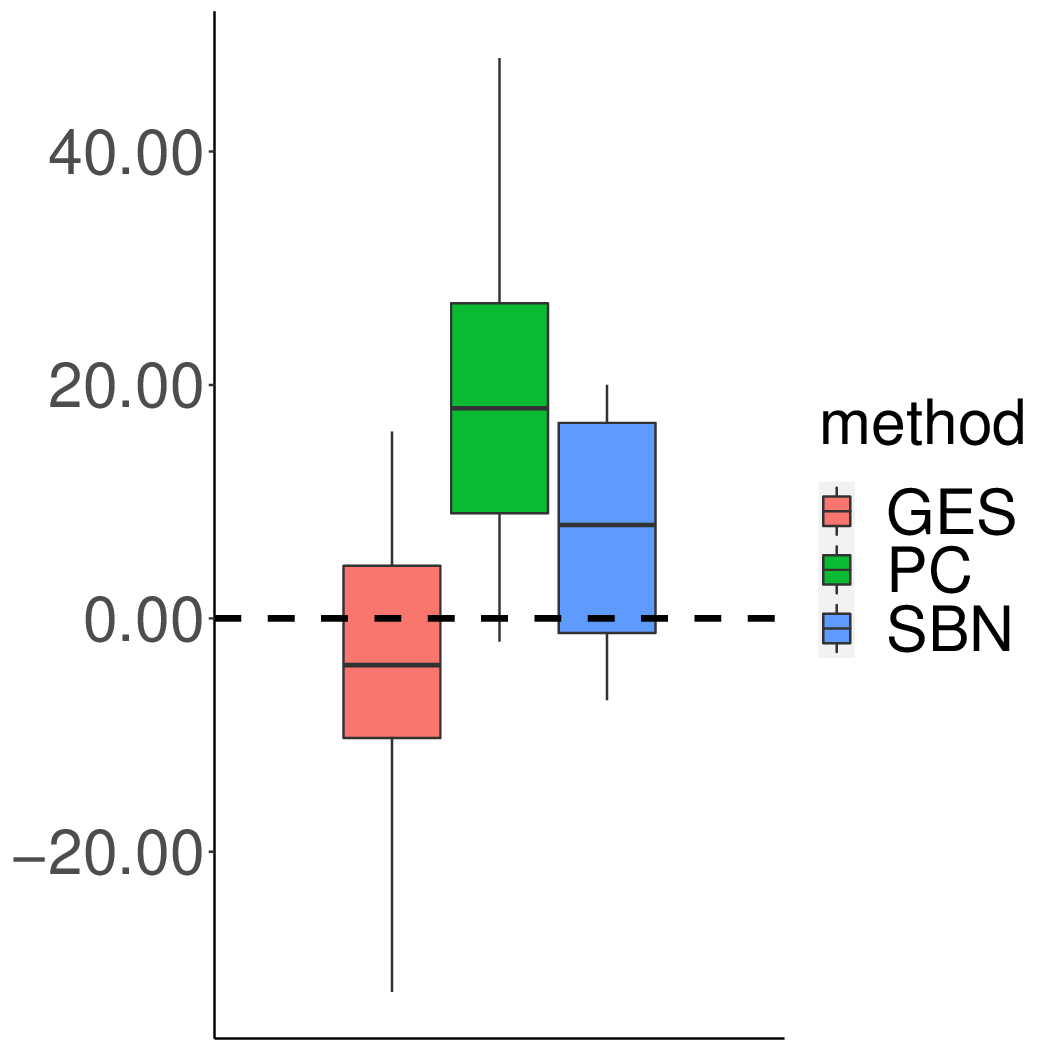}
        \caption*{\texttt{Andes}\\\texttt{Facebook}}
    \end{subfigure}%
    \begin{subfigure}[t]{0.25\textwidth}
        \centering
        \includegraphics[width=\textwidth]{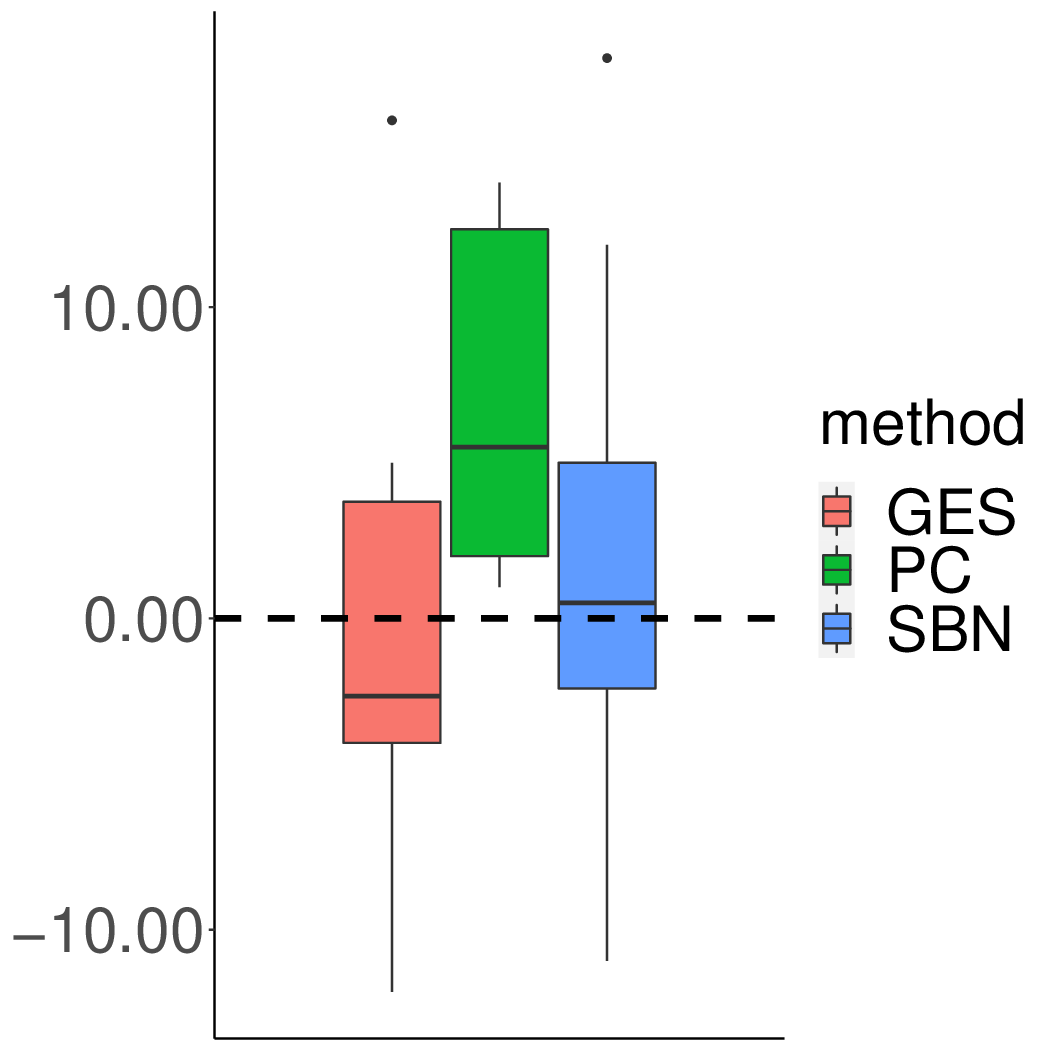}
        \caption*{\texttt{Barley}\\\texttt{Facebook}}
    \end{subfigure}%
    \begin{subfigure}[t]{0.25\textwidth}
        \centering
        \includegraphics[width=\textwidth]{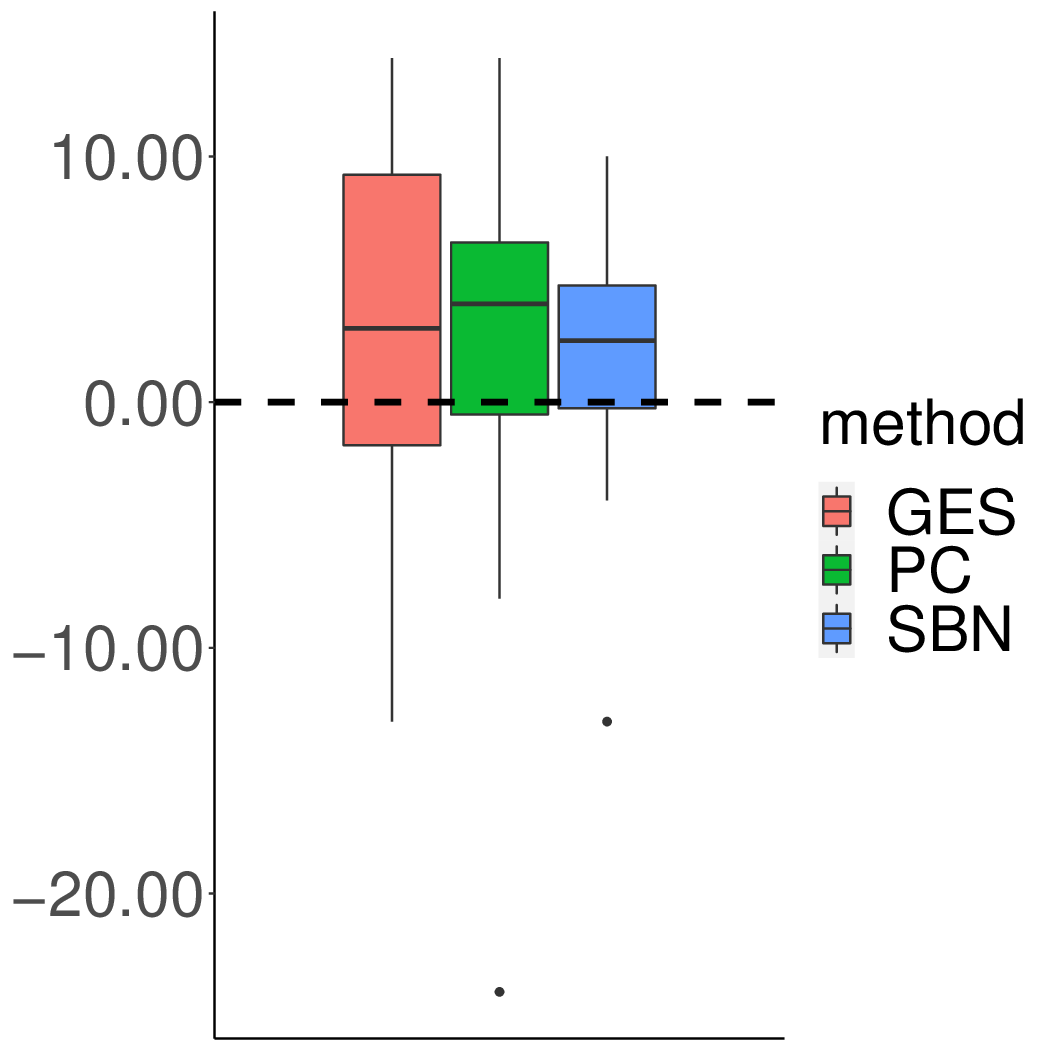}
        \caption*{\qquad\texttt{Hailfinder}\\\texttt{celegans\_n306}}
    \end{subfigure}%
    \begin{subfigure}[t]{0.25\textwidth}
        \centering
        \includegraphics[width=\textwidth]{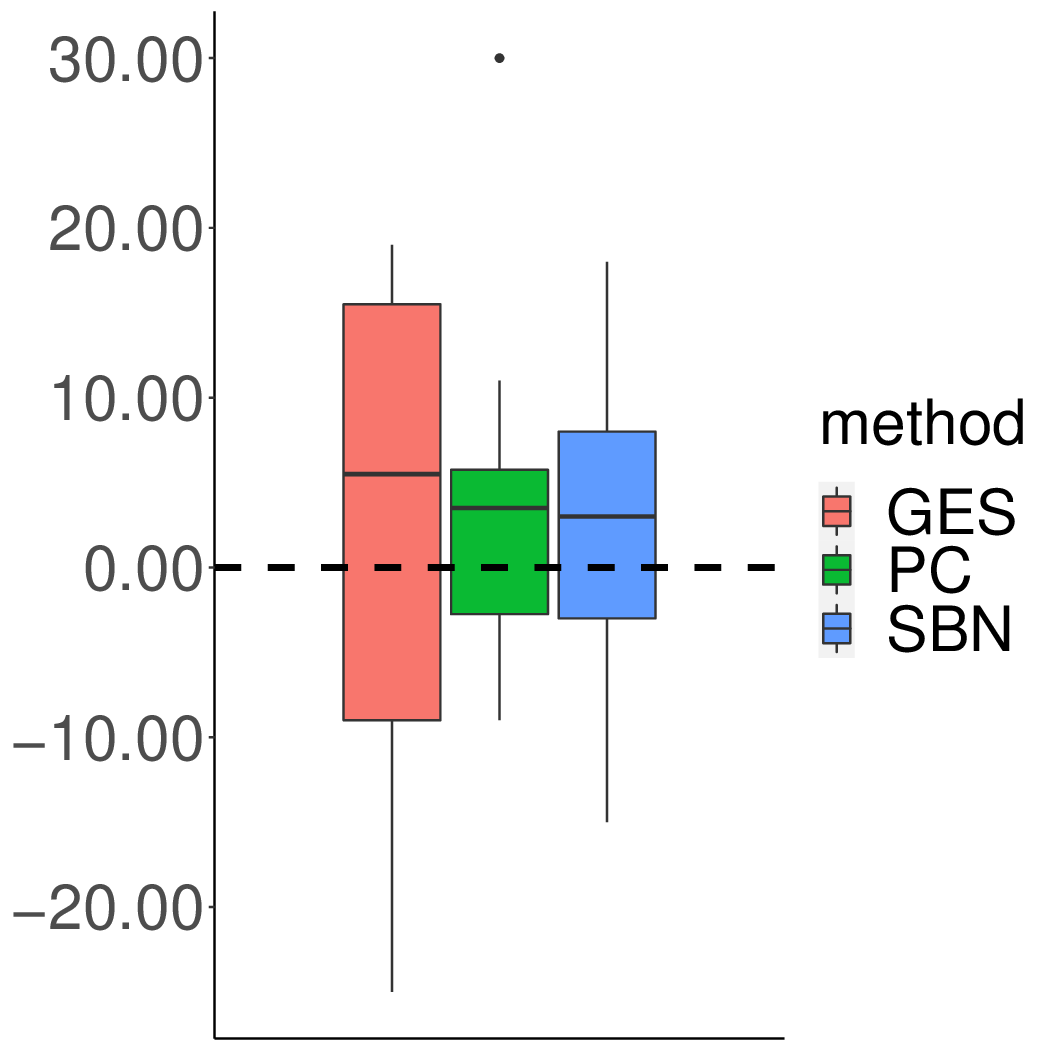}
        \caption*{\qquad\texttt{Hepar2}\\\texttt{celegans\_n306}}
    \end{subfigure}%
    \end{adjustbox}%
    \caption{Results on real unsorted DAGs with general $\Theta$. Top row: increase in normalized test data log-likelihood after de-correlation. Bottom row: Decrease in SHD after de-correlation.}
    \label{fig:generalthetaunordered}
\end{figure}

First we assume that the DAG ordering is known and compare the BCD method against the baseline method. In three out of the four cases we considered, BCD gave better estimates of the DAG structure in terms of Jaccard index and structural Hamming distance as shown in Table \ref{tab:generaltheta}. Next, without assuming a known DAG ordering, we compare the performance of GES, PC, and sparsebn before and after de-correlation. The first row in Figure \ref{fig:generalthetaunordered} shows the increase in the normalized test data log-likelihood after de-correlation, and the increases are positive across all 10 simulations for each of the four scenarios. The second row shows the distribution of the decrease in SHD across 10 simulations after de-correlation. In most cases, all three methods gave much more accurate estimates after de-correlation. We defer the additional tables and figures containing more detailed results to the Supplementary Material. The above results confirm that our methods can indeed improve the accuracy in DAG estimation even $\Theta^*$ is not block-diagonal, as suggested by the theoretical results in Section~\ref{sec:theory}.

% \section{Experiments with DAGs from bnlearn respository}
% \label{sec:bnlearn}
% \input{bnlearn}

\section{Application on RNA-seq Data}
\label{sec:gene}
Gene regulatory networks (GRNs) enable biologists to examine the causal relations in gene expression during different biological processes, and are usually estimated from gene expression data. Recent advances in single-cell RNA sequencing technology has made it possible to trace cellular lineages during differentiation and to identify new cell types by measuring gene expression of thousands of individual cells. A key question arises now is whether we can discover the GRN that controls cellular differentiation and drives transitions from one cell type to another using this type of data. Such GRNs can be interpreted as causal networks among genes, where nodes correspond to different genes and a directed edge encodes a direct causal effect of one gene on another.

The RNA-seq data set used in this section can be found in NCBI's Gene Expression Omnibus and is accessible through GEO series accession number GSE75748. The data set contains gene expression measurements of around 20,000 genes from $n=1018$ cells. Before conducting the experiment, we processed the data according to \cite{scimpute} by imputing missing values and applying $\log$ transformation. %so that the data we use follow matrix normal distribution approximately. 
In this study, we focus on estimating a GRN among $p=51$ target genes selected by \cite{singlecell}, while the rest of the genes, which we call background genes, are used to estimate an undirected network for all 1018 cells. 

\subsection{Pre-estimate the Undirected Network}

\begin{figure}
    \centering
    \begin{adjustbox}{minipage=\linewidth,scale=1}
        \begin{subfigure}{0.5\textwidth}
        \centering
        \includegraphics[width=\textwidth]{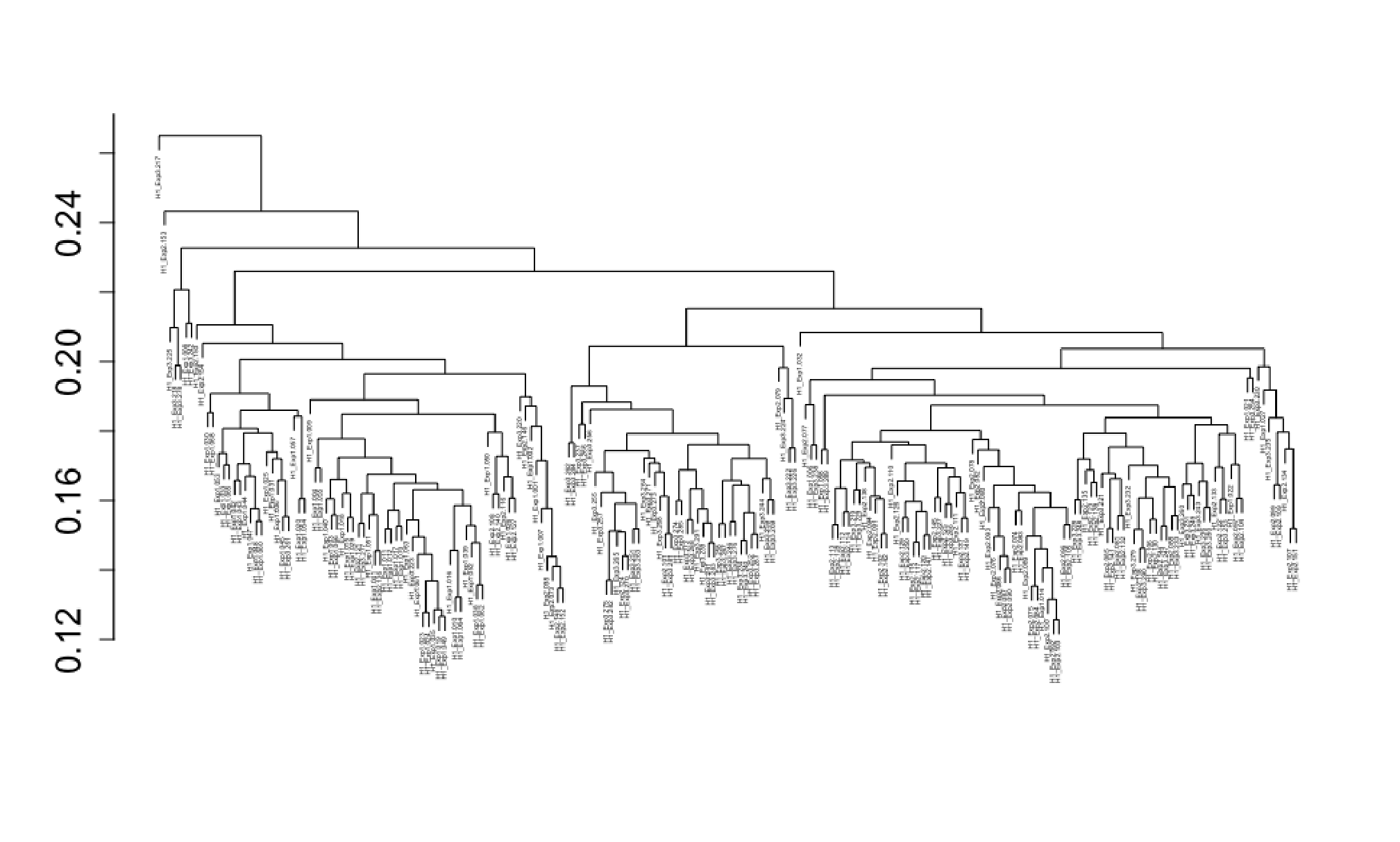}
        \caption{H1}
    \end{subfigure}%
    \begin{subfigure}{0.5\textwidth}
        \centering
        \includegraphics[width=\textwidth]{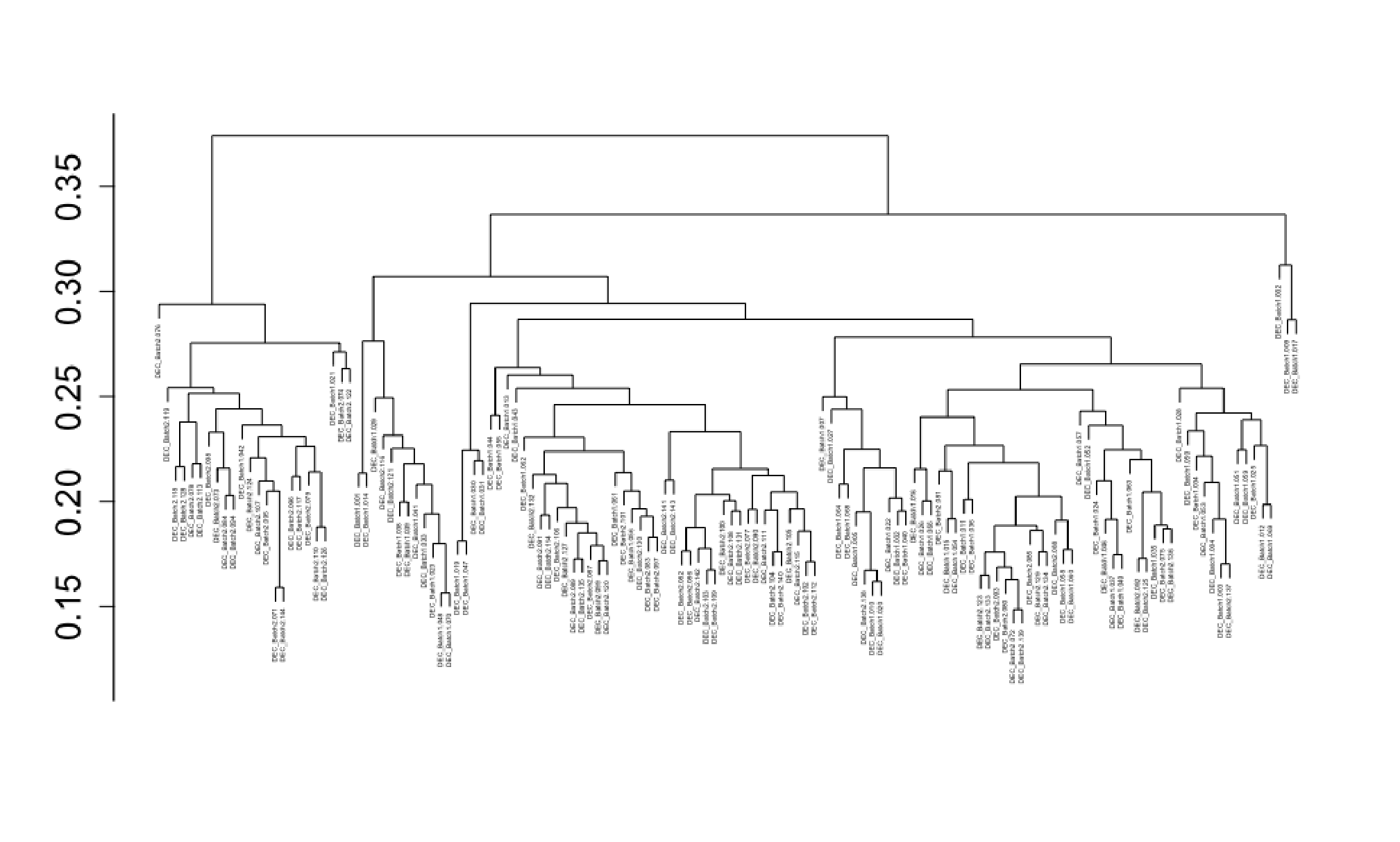}
        \caption{DEC}
    \end{subfigure}
    \end{adjustbox}
    \caption{Cluster dendrograms of H1 and DEC cells from hierarchical clustering. The y-axis represents $1-\rho$ and the leaf nodes are individual cells.}
    \label{fig:dend}
\end{figure}

An essential input to Algorithm \ref{alg1} is $A(G^*)$, %$\supp(\Theta^*)$, 
the adjacency matrix of a known undirected network of observations (cells in this case). The 1018 cells in our data come from 7 distinct cell types (H1, H9, HFF, TB, NPC, DEC, and EC), and it is reasonable to assume that the similarity between cells of different types is minimal. Therefore, we posit that the network of the 1018 cell consists of at least 7 connected components, i.e. $N\geq 7$, where $N$ denotes the number of diagonal blocks in $\Theta^*$ as in Section \ref{sec:theory}. Since it is unlikely that all cells of the same type are strongly associated with one another, we further divided each type of cells into smaller clusters by applying classical clustering algorithm on the background genes. More specifically, we randomly selected 8000 genes from the background genes and applied hierarchical clustering on each type of cells. In this experiment, we used hierarchical clustering with complete-linkage and a distance metric between two cells defined as $1-\rho$, where $\rho$ is the correlation between their observed gene expression levels. We verified our choice of clustering algorithm by applying it on the entire data set, and it clearly grouped the cells into 7 groups, coinciding with the 7 cell types. At the end of the hierarchical clustering step, we needed to pick cutoff levels in order to finish clustering. Because the levels of dependence among cells are quite different across cell types as shown in Figure \ref{fig:dend}, we picked cutoff points separately for each cell type. Generally, we chose the cutoff thresholds such that the largest cluster is smaller than $p=51$, so $\lambda_2$ can be set to 0.01. By shifting the cutoff levels, we also obtained different number $N$ of blocks in $\Theta^*$. In the end, this clustering process returns an adjacency matrix $A$ of the estimated network defined by the $N$ clusters. In our experiments, we compared results from three choices of $N \in \{383, 519, 698\}$. The cluster size varied from 1 (singleton clusters) to 43 across the three cases.

\subsection{Model Evaluation}
\begin{figure}
    \centering
    \begin{adjustbox}{minipage=\linewidth,scale=1}
        \begin{subfigure}{0.5\textwidth}
        \centering
        \includegraphics[width=\linewidth]{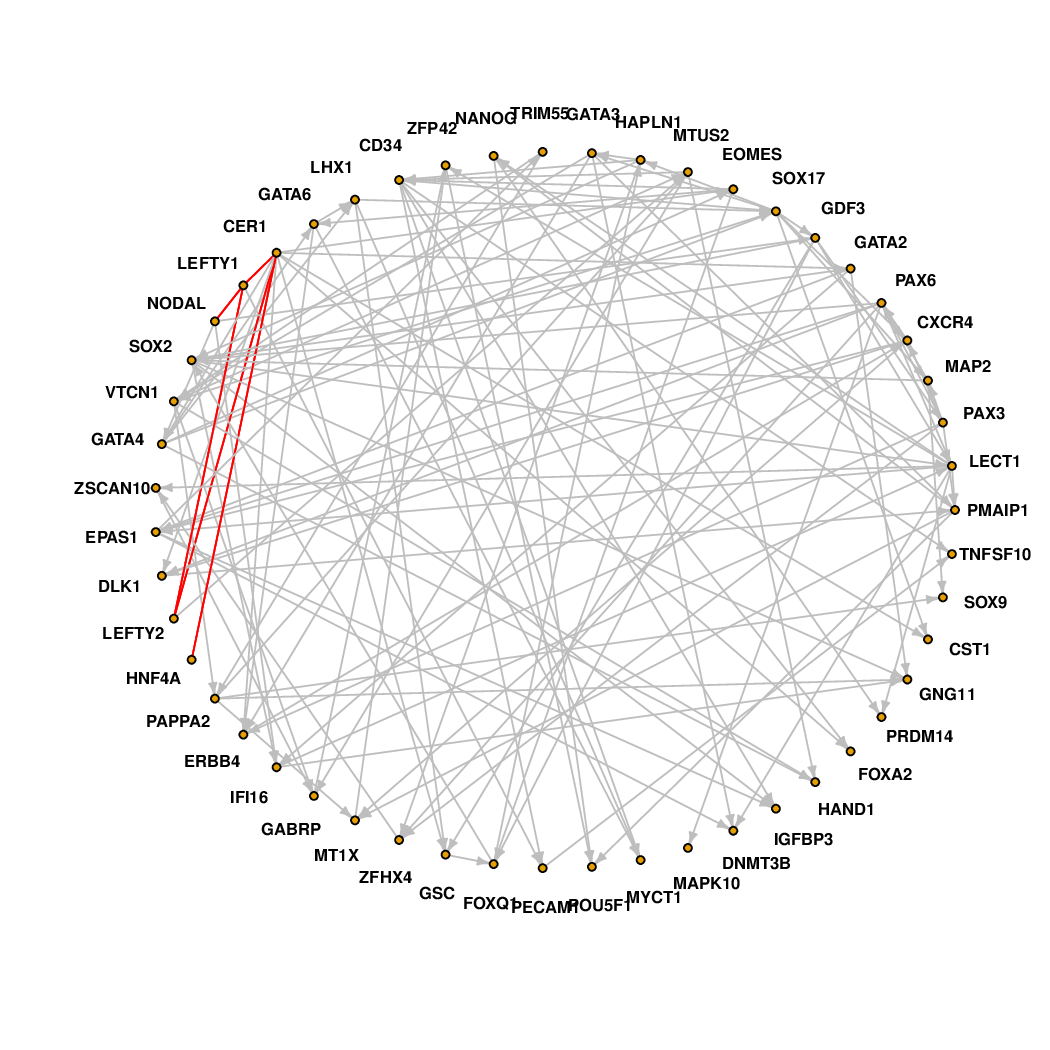}
        \caption{GES ($E=131, U=5$)}
    \end{subfigure}%
    \begin{subfigure}{0.5\textwidth}
        \centering
        \includegraphics[width=\linewidth]{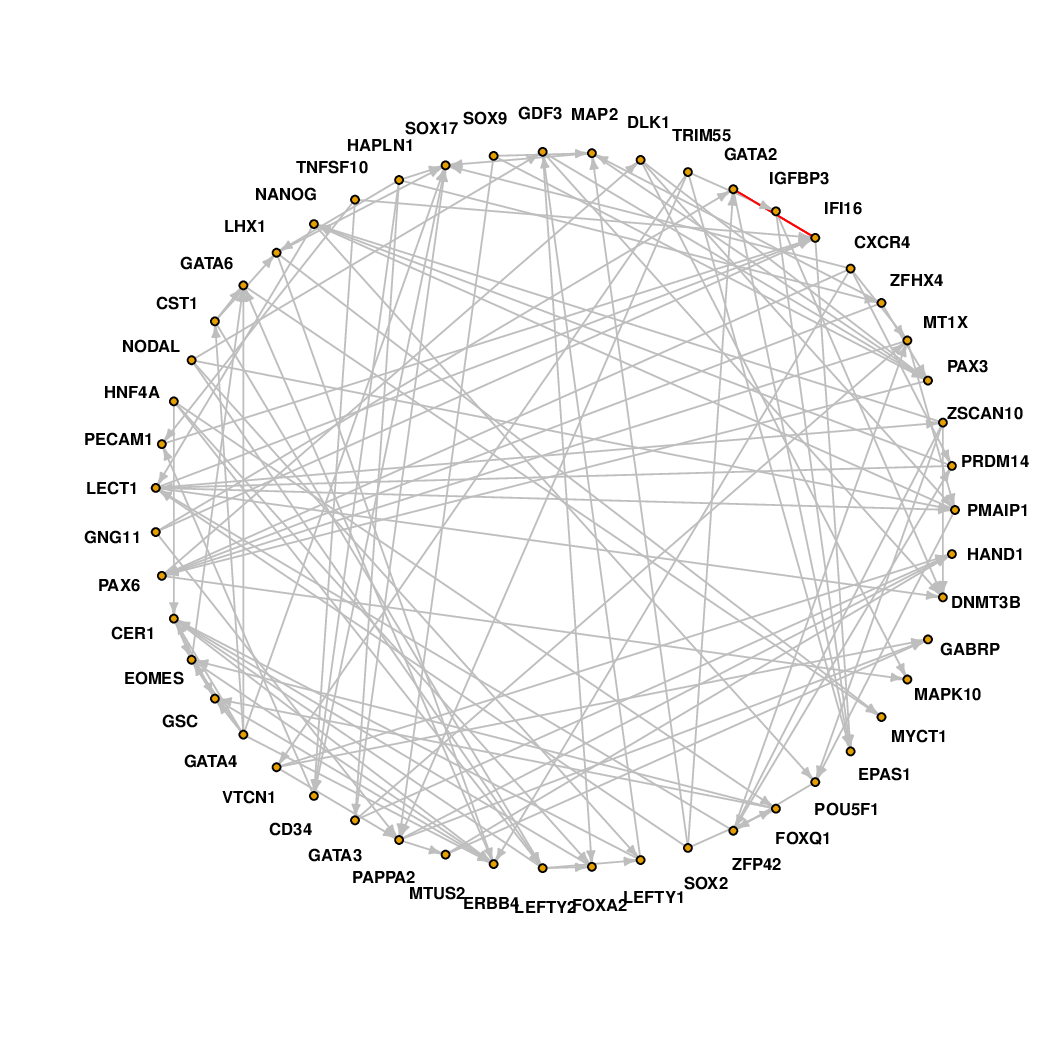}
        \caption{PC ($E = 119, U = 1$)}
    \end{subfigure}
    
     \begin{subfigure}{\textwidth}
        \centering
        \includegraphics[width=0.5\linewidth]{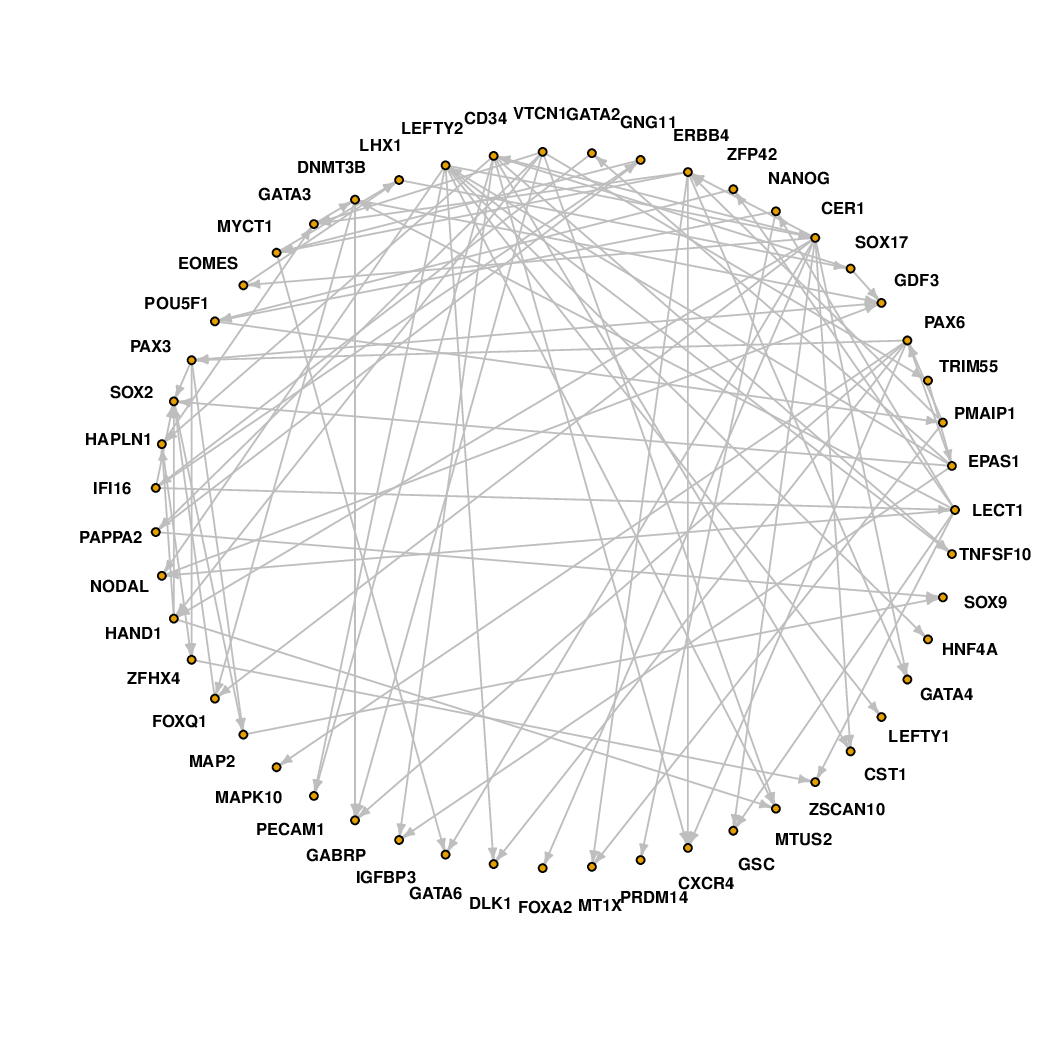}
        \caption{sparsebn ($E = 90, U = 0$)}
    \end{subfigure}
    \end{adjustbox}
    \caption{Estimated gene regulatory networks (CPDAGs) after de-correlation, with $E$ edges and $U$ undirected edges colored in red.}
    \label{fig:gene}
\end{figure}

In this experiment, the input to the BCD algorithm is a data matrix $X_{1018\times 51}$ and $\supp(\Theta^*)\subseteq A$ estimated by the above hierarchical clustering. The matrix of error variances $\widehat\Omega$ was obtained following the method described in Section \ref{sec:natural}. The output is a solution path of $(\widehat B,\widehat\Theta)$ for a range of $\lambda_1$'s. We computed the corresponding MLEs $(\widehat{B}^{MLE},\widehat\Omega^{MLE})$ given the support of each $\widehat{B}$ on the solution path. We picked the $(\widehat{B}^{MLE},\widehat\Omega^{MLE})$ with the smallest BIC from the solution path as in Section \ref{sec:exp} and used the corresponding $\widehat\Theta$ to de-correlate $X$. Table \ref{tab:sc} shows the results of GES, PC, and sparsebn before and after de-correlation. In each case, we computed the BIC of the estimated GRN and we used the Likelihood Ratio (LR) test to determine whether the increase in log-likelihood from de-correlation is significant or not. The LR test statistic is defined as follows: 
$$\text{LR}=\-2(\log p(X \mid \widehat\Theta, \widehat B_{decor}^{MLE}, \widehat\Omega_{decor}^{MLE}) - \log p(X\mid I_n, \widehat B_{baseline}^{MLE}, \widehat\Omega_{baseline}^{MLE})),$$
where $(\widehat B_{baseline}^{MLE}, \widehat\Omega_{baseline}^{MLE})$ and $(\widehat B_{decor}^{MLE}, \widehat\Omega_{decor}^{MLE})$ denote the MLEs given the estimated graph structures from GES, PC, and sparsebn before and after de-correlation, respectively. %and $\widehat\Omega^{MLE}$ is the \revision{MLE of the error covariance matrix given $\widehat B^{(\infty)}$}. 
If the baseline model (before de-correlation) is true, then the LR statistic follows approximately a $\chi^2$ distribution with degrees of freedom
$$df=\frac{|\supp(\widehat\Theta)| - n}{2} + |\supp(\widehat B_{decor}^{MLE})| - |\supp(\widehat B_{baseline}^{MLE})|.$$ 
In most cases, we saw significant improvements, in terms of both the BIC and the $\chi^2$ statistic, in all three DAG estimation methods by de-correlating $X$ using the estimated $\widehat\Theta$ from the BCD algorithm. This confirms the dependence among individual cells and implies that our proposed network model fits this real-world data better. %Lastly, we used BIC to select the estimated CPDAG of the gene expression data, and 
Figure \ref{fig:gene} shows the estimated CPDAGs after de-correlation for the case $N=383$, which corresponds to the minimum BIC for all three methods in our experiments. It is interesting to note that a directed edge NANOG$\to$POU5F1, between the two master regulators in embryonic stem cells, appears in all three estimated CPDAGs, consistent with previously reported gene regulatory networks \citep{chen2008integration, Zhou16438}. 
% \qzcmt{cite Zhou et al (2007) PNAS, and Chen et al (2008), Cell, 133: 1106–1117.}

\begin{table}
    \centering
    \resizebox{\linewidth}{!}{
    \begin{tabular}{lc|ccrrrrc}
    \toprule
    model\_type & $N$ & BIC(baseline) & BIC(decor)& ll(baseline) & ll(decor) & LR $\chi^2$ & df & p-value  \\
    \midrule
    GES&\multirow{3}{*}{383}&-35036.60&-41799.39&17593.30&22667.70& 10148.79&3386 &0\\
    PC&&-14102.81 & -32274.01&7102.91&17901.01&21596.19&3425&0\\
    sparsebn&&-28553.26 &-37894.85&14336.13&20697.42&12722.59&3381&0\\
    \midrule
    GES&\multirow{3}{*}{519}&-35036.60&-33312.95&17593.30&17597.47&8.34&1732&1\\
    PC&&-14102.81&-18738.62&7102.91&10297.31&6388.81&1753&0\\
    sparsebn&&-28553.26&-28797.87&14336.13&15324.43&1976.61&1732&0\\
    \midrule
    GES&\multirow{3}{*}{698}&-35036.60&-34921.63 &17593.30 &17879.82& 573.03 & 688 &1\\
    PC&&-14102.81&-16959.60 &7102.91& 8879.30&3552.79&696 & 0\\
    sparsebn&&-28553.26&-29751.59&14336.13&15273.80&1875.33&677&0\\
    \bottomrule
    \end{tabular}
    }
    \caption{BIC scores and log-likelihood(ll) values from GES, PC, and sparsebn before and after de-correlation. $N$ denotes the number of clusters among cells.}
    \label{tab:sc}
\end{table}

\section{Discussion}
\label{sec:disc}
In this paper  our main goal is to generalize the existing Gaussian DAG model to dependent data. We proposed to model the covariance between observations by assuming a non-diagonal covariance structure of the noise vectors. This generalization is related to the semi-Markovian assumption in causal DAG models. Our main contributions include the development of a consistent structural learning method for the DAG and the sample network under sparsity assumptions and finite-sample guarantees for the estimators. 

Our proposed BCD algorithm is built upon existing Lasso regression and graphical Lasso covariance estimation methods. When a topological ordering of the true DAG is known, it estimates the covariance between the observations $\Sigma$ and the WAM of the DAG $B$ in an iterative way. The method is fast and often converges in a few iterations. Our theoretical analysis shows that the estimates after one iteration are $\ell_2$-consistent under various asymptotic frameworks including both $n\ll p$ and $n\gg p$, assuming a proper initialization of the precision matrix $\widehat\Theta^{(0)}$. The estimate of the DAG WAM $\widehat{B}^{(1)}$ achieves the optimal rate as Lasso estimators. The estimate of the precision matrix $\widehat{\Theta}^{(1)}$ achieves the same optimal rate as the graphical Lasso method when $ n \gg p$ and there are sufficiently many independent subgroups within the data. Otherwise, it has a slightly worse rate due to the bias of the sample covariance matrix. When the DAG ordering is unknown, we showed the covariance $\Sigma$ is invariant under permutations of the DAG nodes. Therefore, if the true DAG is sparse, our BCD algorithm can still give a good estimate of $\Theta$ which can be used to decorrelate the data. In addition to the theoretical analysis, we conducted extensive experiments on both synthetic and real data to compare our method with existing methods. When a true ordering of the DAG was given, the BCD algorithm significantly improved the structural estimation accuracy compared to the baseline method which ignored the sample dependency. When the ordering was unknown, classical DAG learning methods, such as GES, PC, and sparsebn, can all be greatly improved with respect to structural learning of CPDAGs by using our proposed de-correlation method based on the BCD algorithm. In all cases, our estimation methods under the proposed network Gaussian DAG model yielded significantly higher test data log-likelihood compared to other competing methods, indicating better predictive modeling performance. %for DAG structural learning in practice. 

%For synthetic data, we looked at performance under different designs of covariance matrices and sample sizes, and for real data we used different combinations of real network structures. The performance of all the methods were evaluated both in terms of both structural learning accuracy measured by SHD, FDR, JI, etc. as well as model fitting and generalization accuracy such as BIC and test data log-likelihood. 

There are several unexplored directions from our research. First, the current error bounds and consistency results are based upon a known topological ordering of the true DAG. In practice, however, it can be hard to obtain the ordering in advance. It would be interesting to see if we can combine the method of estimating partial orders such as \cite{partialorder} with our method and extend the theoretical results. Second, part of the reason that the current model relies on a known DAG ordering is the lack of experimental data. From purely observational data, it is impossible to orient some of the edges and find a topological ordering of the true DAG. In the next step, we would like to extend our method to handle both observational and experimental data sets. Finally, there are recent methods that use continuous optimization for DAG learning without imposing the acyclicity constraint, such as NOTEARS \citep{zheng2018dags}. It is a promising future direction to incorporate such ideas into DAG learning on network data.

% Acknowledgements should go at the end, before appendices and references

\acks{We would like to acknowledge support for this project
from NSF grant DMS-1952929.}

% Manual newpage inserted to improve layout of sample file - not
% needed in general before appendices/bibliography.
\appendix

\section{Technical Details of Section \ref{sec:theory}}\label{appendixB}

\subsection{Some Auxiliary Results}\label{auxres}
Here we introduce four lemmas that we use to establish the error bounds of $\widehat{\Theta}^{(1)}$ and $\widehat B^{(1)}$. Let us start by deriving an upper bound on the $\ell_2$ deviation of $\widehat{L}^{(0)}$ from $L^*$ under Assumption \ref{assump: theta}.
\begin{lemma}\label{lem:2.1}
Suppose Assumption \ref{assump: theta} holds and let $L^*, \widehat{L}^{(t)}$ be the Cholesky factors of $\Theta^*$ and $\widehat{\Theta}^{(t)}$, respectively. Let $\widehat{\Delta}_{chol}^{(t)} = \widehat{L}^{(t)} - L^*$. Then,
\begin{align*}
    % \|\widehat{\Delta}_{chol}\|_2 &\leq \frac{\|\widehat\Delta_{prec}\|_2}{2\sigma_{\min}(L^*)},  \\
    \|\widehat{\Delta}_{chol}^{(0)}\|_2 &\leq \frac{M}{2\sigma_{\min}(L^*)},
\end{align*}
where $\sigma_{\min}(L^*)$ is the smallest singular value of $L^*$, and $M$ is from Assumption \ref{assump: theta}. 
\end{lemma}
To generalize the basic bound on $\|\hat\beta^{lasso}-\beta_j^*\|_2$ from \cite{bhlmann} to dependent data, we need to control the $\ell_\infty$-norm of an empirical process component $2X^\top\widehat\Theta^{(0)}\epsilon_j/n$. Let us start with the case when the data are independent. Define the following events, where $\widetilde{X}= L^* X$ represents the independent data as explained in Section \ref{sec:theory}:
    \begin{align}\label{eq:eventE}
        \mathscr{E} &:= \bigcap_{k=1}^p\bigg\{\|\widetilde{X}_k\|_2 \leq 6\bar\psi\sqrt{n}\bigg\}.
    \end{align}
Then the following lemma follows from $\widetilde{X}$ being sub-Gaussian.
\begin{lemma}\label{lem:xnorm}
Let $\alpha > 2$ be an integer. If $n > 2\sqrt{2\alpha}\log p$, then the event $\mathscr{E}$ defined in \eqref{eq:eventE} holds with probability at least $1 - 1/p^{\alpha-1}$.
    % \end{align}
\end{lemma}
Next, define the following events that depend on $\lambda_n$ and are standard in the classical Lasso problem,
%on the empirical process for a given $\lambda_n$ as in the classical Lasso problem.
    \begin{align}
        \mathscr{T}_j &:= \left\{ 2\|\widetilde{X}^\top \Tilde{\varepsilon}_j\|_\infty/n\leq \lambda_n \right\},\quad j = 1,\ldots, p \nonumber\\
        \mathscr{T} &:= \bigcap_{j=1}^p  \mathscr{T}_j \label{event:T}.
    \end{align}{}
\begin{lemma}\label{lem:2}
  Let $\widetilde{X}_k$ consist of $n$ i.i.d sub-Gaussian random variables with parameter $\bar\psi^2$ for $k=1,\ldots,p$. If 
    \begin{align*}
      \lambda_n = 12\bar\psi\bar\omega\left(\sqrt{\frac{2\log p}{n}} + \sqrt{\frac{2\log 2 + 4\log p}{n}}\right),
    \end{align*}
    then the probability of $\mathscr{T}$ satisfies
    \begin{align*}
        \mathbb{P}(\mathscr{T}) \geq \left(1 - \frac{1}{p}\right)^2.
    \end{align*}
\end{lemma}
Lemma \ref{lem:2} implies that if $\lambda_n \asymp \sqrt{\frac{\log p}{n}}$, then the error terms will be uniformly under control with high probability, especially when both $n$ and $p$ are large. \\
\begin{lemma}[Maximal inequality]\label{lem:maximal}
Let $x_i = \left(x_{i1},\ldots, x_{ip}\right)$ be a random vector where each element $x_{ij}$ is $\text{sub-Gaussian}$ with parameter $\bar\psi^2$, then 
\begin{align*}
    \mathbb{P}\left(\|x_i\|_\infty \geq 2\bar\psi\sqrt{\log p}\right) \leq 2/p.
\end{align*}
\end{lemma}

From model \eqref{eq:matrixnormal}, it is clear that each row in $\widetilde{X}$ is sub-Gaussian with parameter $\bar\psi^2$. By Lemma \ref{lem:maximal}, we have $\|\tilde{x}_i\|_\infty \lesssim \sqrt{\log p}$ w.h.p.

% \begin{proof}
% By union bound, we obtain for $\delta < 1/v_*$,
% \begin{align}
%     \mathbb{P}\left(\|\widehat{\Sigma} - \Sigma^*\|_\infty \geq \delta \right) \leq n^2/f(p,\delta)
% \end{align}{}
% Letting $\bar\delta_f(p;n^\tau) = \delta$ have we
% $$\mathbb{P}\left(\|\widehat{\Sigma} - \Sigma^*\|_\infty\geq \bar\delta_f(p;n^\tau)\right)\leq n^2/n^\tau = 1/n^{\tau-2}$$
% \end{proof}
% If $p$ is large enough, we will have $\bar\delta_f(p;n^\tau) \leq 1/v_*$ and Lemma \ref{lem:sampleNoise} shows $\mathcal{A}$ in \eqref{eventA} holds with high probability.\\

% Conditional on event $\mathcal{A}$, we have
% \begin{align*}
%     \|\widehat W\|_\infty &:= \|\widehat{S} - \Sigma^*\|_\infty \leq \|\widehat{\Sigma} - \Sigma^*\|_\infty + \|\widehat R\|_\infty\leq \bar\delta_f(p;n^\tau) + \|\widehat R\|_\infty
% \end{align*}
% We need to show that $ \|R\|_\infty$ is dominated by  $ \bar\delta_f(p;n^\tau) $ for sufficiently large $(p, n)$.

\subsection{Error Bound of  \texorpdfstring{$\widehat{B}^{(1)}$}{hatB}} 
The estimation error bound for the classical Lasso problem where samples are i.i.d. was established by choosing a penalty coefficient that dominates the measurement error term. Specifically, as shown in Lemma \ref{lem:2}, this can be achieved with high probability by setting  $\lambda_n\asymp\sqrt{\frac{\log p}{n}}$. In order to prove the consistency of $\widehat B^{(1)}$ from Algorithm \ref{alg1}, we need to control a similar error term which depends on $\widehat\Theta^{(0)}$. Notably, such error can be controlled under the same rate as $\lambda_n$, see Theorem \ref{thm_inf}.

\begin{theorem}[Control the empirical process]\label{thm_inf}
Let $\lambda_n$ be the same as in Lemma \ref{lem:2}. Suppose the initial estimator $\widehat{\Theta}^{(0)}$ satisfies Assumption \ref{assump: theta}. Then 
\begin{align}\label{event:basic}
    \mathbb{P}\left(\sup_{j\in[p]}2\|X^\top \widehat{\Theta}^{(0)}\varepsilon_j\|_\infty / n \leq \lambda_n\right) \geq \left(1-\frac{1}{p}\right)\left(1 - \frac{2}{p}\right).
\end{align}
\end{theorem}
Next, we show that the random matrix $\widehat L^{(0)} X$ satisfies the Restricted Eigenvalue (RE) condition \citep{wainwright_2019} w.h.p. Towards that end, we define the event   $\mathscr{K}$ as in Theorem 7.16 from \cite{wainwright_2019} given as
%Let $\mathscr{K}$ be the event that $\|\widetilde{X}\beta\|_2^2/n$ is bounded as in Theorem 7.16 from \citep{wainwright_2019} for all $\beta \in\mathbb{R}^{p}$.
\begin{align}\label{eventK}
    \mathscr{K} := \left\{\|\widetilde{X}\beta\|_2^2/n \geq  \tilde{c}_1\|\sqrt{\Psi^*}\beta\|_2^2 - \tilde{c}_2\rho^2(\Psi^*)\frac{\log p}{n}\|\beta\|_1^2\right\},
\end{align}
where $\rho^2(\Psi^*)$ is the maximum diagonal entry of $\Psi^*$ and $\widetilde{X} = L^*X$ is the de-correlated data.

\begin{lemma}[Restricted eigenvalue condition]\label{lem:condition2}
Consider a random matrix $X\in\mathbb{R}^{n\times p}$, which is drawn from a $\mathcal{N}_{n\times p}(0, \Sigma^*, \Psi^*)$ distribution. Let $\widehat{\Theta}^{(0)}$ be the initial estimate of $\Theta^* = \Sigma^{*-1}$ satisfying Assumption \ref{assump: theta}, $\widehat{L}^{(0)}$ be the Cholesky factor of $\widehat{\Theta}^{(0)}$, and $\rho^2(\Psi^*)$ be the maximum diagonal entry of $\Psi^*$. Then under event $\mathscr{K}$ defined in \eqref{eventK}, there are universal positive constants $c_1 < 1 < c_2$ such that
\begin{align}\label{eq:condition2}
    \frac{\|\widehat{L}^{(0)}X\beta\|_2^2}{n}\geq  c_1\|\sqrt{\Psi^*}\beta\|_2^2 - c_2\rho^2(\Psi^*)\frac{\log p}{n} \|\beta\|_1^2,
\end{align}
for all $\beta\in\mathbb{R}^{p}$. 
\end{lemma}
The probability of event $\mathscr{K}$ can be found in Theorem 7.16 from \cite{wainwright_2019} . This event is a restriction on the design matrix $\widetilde X$ and it holds with high probability for a variety of matrix ensembles. With Theorem \ref{thm_inf} and Lemma \ref{lem:condition2}, it is possible to prove an \textit{oracle inequality} for the dependent Lasso problem, which yields a family of upper bounds on the estimation error. 
% \subsection{Lasso estimation consistency}
\begin{theorem}[Lasso oracle inequality]\label{thm:oracle}
Consider the Lasso problem in \eqref{eq:1step} for $t=0$. 
Suppose the inequality \eqref{eq:condition2} and the event in \eqref{event:basic} hold. Let $\bar\kappa=\sigma_{\min}(\Psi^*)$. For $j\in[p]$ and any $\beta_j^*\in\mathbb{R}^p$, if 
\begin{align*}
\lambda_n &\geq 12\bar\psi\bar\omega\left(\sqrt{\frac{2\log p}{n}} + \sqrt{\frac{2\log 2 + 4\log p}{n}}\right),
\end{align*}{}
then any optimal solution $\hat\beta_j^{(1)}$ satisfies:
\begin{align}\label{eq:oracle}
    \|\hat\beta_j^{(1)} - \beta_j^*\|_2^2 \leq \frac{768\lambda_n^2}{c_1^2\bar\kappa^2}|S| + \frac{64\lambda_n}{4c_1\bar\kappa}\|\beta^*_{j,S^c}\|_1 + \frac{128c_2}{c_1}\frac{\rho^2(\Psi^*)}{\bar\kappa}\frac{\log p}{n} \|\beta^*_{j,S^c}\|_1^2,
\end{align}{}
for any subset $S$ with cardinality $|S| \leq \frac{c_1}{64c_2}\frac{\bar\kappa}{\rho^2(\Psi^*)}\frac{n}{\log p}$. Let $\widehat{L}^{(0)}$ be the Cholesky factor of $\widehat\Theta^{(0)}$. Then,
\begin{align}\label{pred_bound}
    \|\widehat{L}^{(0)} X\left(\hat\beta_j^{(1)} - \beta_j^*\right)\|_2^2/n &\leq 6\lambda_n\|\beta_j^*\|_1.
\end{align}
\end{theorem}{}
Theorem \ref{thm:oracle} implies $\sup_{j\in[p]} \|\hat\beta_j^{(1)} - \beta_j^*\|_2^2 \leq \frac{768\lambda_n^2}{c_1^2\bar\kappa^2}s \asymp s\frac{\log p}{n}$, where $s$ is the maximum in-degree of the true DAG. 

% \begin{theorem}[Prediction error bound]\label{thm:pred}
% Under the same conditions as in Theorem \ref{thm:oracle}, if $\lambda_n \geq 12\bar\psi\bar\omega\left(\sqrt{\frac{2\log p}{n}} + \sqrt{\frac{2\log 2 + 4\log p}{n}}\right)$,
% then for $j\in[p]$ and any $\beta_j^*\in\mathbb{R}^p$, any optimal solution $\hat\beta_j^{(1)}$ of the Lasso problem in \eqref{eq:1step} when $t=0$ satisfies the bound:
% \end{theorem}
% \qzcmt{You can merge the prediction error into Theorem 16 and remove Theorem 17.}

\subsection{Error Bounds of  \texorpdfstring{$\widehat\Theta^{(1)}$}{hatTheta}}\label{app:theta}
Recall that $s$ denotes the maximum number of nonzero entries in $\beta_j^*$ for $j\in[p]$. In order to control $\|\widehat\Theta^{(1)} - \Theta^*\|_2$, we need to rely on certain type of error bound on $\widehat{S}^{(1)}  - \Sigma^*$, where $\widehat{S}^{(1)} $ is the sample covariance defined in \eqref{glasso} when $t=0$. Therefore, we adopt the definition of tail condition on the sample covariance from \cite{logdet}.
% Lemma \ref{lem:omega} and Theorem \ref{thm:oracle} implies that,
% \begin{align}
% \sup_j\|\hat\beta_j - \beta_j^*\|_2 &\lesssim \lambda_n\sqrt{s}
% \asymp \sqrt{\frac{s\log p }{n}}\\
% \sup_j\abs{\hat\omega_j^2 - \omega_j^{*2}}&\lesssim s\sqrt{\frac{\log p}{N}}
% \end{align}
% In this section, we will reply on these results to prove the consistency of $\widehat{\Theta}^{(1)}$. 
\begin{definition}{(Tail conditions)}\label{def:tail}
We say the $n\times p$ random matrix $X$ from model \eqref{eq:model} satisfies tail condition $\mathcal{T}(f, v_*)$ if there exists a constant $v_*\in(0,\infty]$ and a function $f:\mathbb{N}\times(0,\infty)\to(0,\infty)$ such that for any $(i,j) \in [n]\times [n]$, %$\in V\times V$:
\begin{align*}
    \mathbb{P}\left[|\widehat{S}_{ij}^{(1)} - \Sigma_{ij}^*|\geq \delta\right]\leq 1 / f(p,\delta)\quad \forall \delta\in(0,1/v_*].
\end{align*}
% \qzcmt{$\widehat{S}_{ij}^{(1)}$?}
\end{definition}{}
We require $f(p,\delta)$ to be monotonically increasing in $p$, so for a fixed $\delta > 0$, define the inverse function
\begin{align*}
    \bar{p}_f(\delta; r) := \argmax\left\{p \mid f(p,\delta)\leq r\right\}.
\end{align*}
Similarly, $f$ should be increasing in $\delta$ for each fixed $p$, so we define an inverse function in the second argument:
\begin{align}\label{eq:tail}
    \bar{\delta}_f(p; r) :=\argmax\left\{\delta\mid f(p,\delta)\leq r\right\}.
\end{align}
Under the setting of a Gaussian DAG model, we can derive a sub-Gaussian tail bound.
\begin{lemma}\label{lem:sampleCov}
Let $X$ be a sample from our Gaussian DAG model \eqref{eq:model}. The sample covariance matrix
\begin{align}\label{hatsigma}
    \widehat{\Sigma} &= \frac{1}{p}\sum_{j=1}^p \frac{1}{\omega_j^{2*}}\left(X_j - X \beta_j^*\right)\left(X_j - X \beta_j^*\right)^\top,
\end{align}
satisfies the tail bound
\begin{align*}
    \mathbb{P}\left(\sup_{i,j}|\widehat{\Sigma}_{ij} - \Sigma_{ij}^*|> \delta\right) \leq 4\exp\left\{-\frac{p\delta^2}{3200}\right\},
\end{align*}
for all $\delta \in (0, 40)$.
\end{lemma}
\begin{cor}\label{cor:tail}
If $f(p,\delta) = 4\exp\left\{\frac{p\delta^2}{3200}\right\}$, then the inverse function $\bar\delta_f(p;n^\tau)$ takes the following form,
\begin{align*}
    \bar\delta_f(p;n^\tau) = 40\sqrt{2}\sqrt{\frac{\tau\log n + \log 4}{p}}.
\end{align*}
\end{cor}
Based on the tail bound in Corollary \ref{cor:tail}, we can control the sampling noise $\widehat\Sigma - \Sigma^*$ as in Lemma \ref{lem:sampleNoise}. %\qzcmt{$\widehat\Sigma$ to $\widehat\Sigma^{(1)}$? change all.}
\begin{lemma}[Lemma 8 in \citealt{logdet}]\label{lem:sampleNoise}
Define event 
\begin{align}\label{eventA}
\mathcal{A} = \left\{\|\widehat{\Sigma} -\Sigma^*\|_\infty \leq\bar\delta_f(p;n^\tau) \right\},
\end{align}
where $\bar\delta_f(p;n^\tau) = 40\sqrt{2}\sqrt{\frac{\tau\log n + \log 4}{p}}$.
For any $\tau > 2$ and $(n,p)$ such that $\bar\delta_f(p; n^\tau) \leq 1/40$, we have
\begin{align*}
    \mathbb{P}\left[\mathcal{A}^c\right] \leq \frac{1}{n^{\tau - 2}} \to 0.
\end{align*}{}
\end{lemma}
Recall $r(\widehat\Omega)$ defined in \eqref{def: rnp} and the constant $b$ defined in Lemma~\ref{lem:rho_consistency}. 
\begin{lemma}\label{lem:R}
Suppose $b>0$ and %that $\widehat{B}^{(1)}$ and $\widehat{\Omega}$ satisfy
\begin{align}
\begin{split}\label{R:assump}
    \sup_j\|\hat\beta_j^{(1)} - \beta_j^*\|_2^2 &\leq c\cdot s\frac{\log p}{n},\\
    %r(\widehat\Omega) &:= \sup_k|1/\hat\omega_k^2 - 1/\omega_k^{*2}|,
\end{split}
\end{align}
for a fixed positive constant $c$. Let 
$\widehat{W}^{(1)}:=\widehat{S}^{(1)} - \Sigma^*$. %the difference between the sample covariance matrix $\widehat{S}^{(1)}$ and $\Sigma^*$ be $\widehat{W}^{(1)}$. 
Then, we have
\begin{align*}
    \|\widehat W^{(1)}\|_\infty %&:= \|\widehat{S}^{(1)} - \Sigma^*\|_\infty\\ 
    \leq 40\sqrt{2}\sqrt{\frac{\tau\log n + \log 4}{p}} + \max\left\{6\bar\omega r(\widehat\Omega), \frac{72\bar\omega\bar\psi s}{b}\sqrt{\frac{\log p\log^2 \max\{n,p\}}{n}} \right\}
\end{align*}
with probability at least $1 - 1/n^{\tau-2} - 5n^2 / \max\{n,p\}^4$.
\end{lemma}
% \begin{proposition}[Control of $D_p$, Lemma 6 in \citealt{logdet} ]\label{prop:maxbond}
% Suppose that
% \begin{align*}
%     r := 2\kappa_{\Gamma^*}\left(\|\widehat W^{(1)}\|_\infty + \lambda_p\right) \leq \min\left\{\frac{1}{3\kappa_{\Sigma^*}d}, \frac{1}{3\kappa_{\Sigma^*}^32\kappa_{\Gamma^*}d}\right\}.
% \end{align*}{}
% Then we have the element-wise $\ell_2\infty$ bound
% \begin{align*}
%     \|\widehat D_p^{(1)}\|_\infty := \|\widehat{\Theta}^{(1)} - \Theta^*\|_\infty \leq r.
% \end{align*}{}
% \end{proposition}

\begin{theorem}\label{thm:infnorm}
Assume $\widehat{B}^{(1)}$ %and $\widehat{\Omega}$ 
satisfies \eqref{R:assump} and $b > 0$. Let 
\begin{align*}
\bar{R}(s,p,n) &= \max\left\{6\bar\omega r(\widehat\Omega), \frac{72\bar\omega\bar\psi s}{b}\sqrt{\frac{\log p\log^2 \max\{n,p\}}{n}} \right\},\\
\bar\delta_f(p;n^\tau) &=40\sqrt{2}\sqrt{\frac{\tau\log n + \log 4}{p}}.
\end{align*}
Consider the graphical Lasso estimate $\widehat{\Theta}^{(1)}$ from Algorithm \ref{alg1} with $\lambda_p = \bar\delta_f(p;n^\tau) + \bar{R}$ for $\tau > 2$. Assume
\begin{align}\label{sample_size}
    \bar{p}_f\left(1/\max\left\{160, 24mC\right\}, n^\tau\right) \leq p\quad \text{and}\quad \bar{R} \leq \frac{1}{24C},
\end{align}
where $C=\max\left\{\kappa_{\Sigma^*}\kappa_{\Gamma^*},\kappa_{\Sigma^*}^3\kappa_{\Gamma^*}^2 \right\}$. Then, with probability at least $1 - 1/n^{\tau-2} - 5n^2 / \max\{n,p\}^4$, we have
\begin{align*}
    \|\widehat{\Theta}^{(1)} - \Theta^*\|_\infty &\leq 4\kappa_{\Gamma^*}\left(\bar\delta_f(p;n^\tau) + \bar{R}\right),\\
    \|\widehat\Theta^{(1)} - \Theta^*\|_2 &\leq 4\kappa_{\Gamma^*}(m+1)\left(\bar\delta_f(p;n^\tau) + \bar{R}\right).
\end{align*}
where $m$ is the maximum degree of the undirected network $G^*$. %$G(A)$. 
\end{theorem}

\section{Proofs of Main Results}\label{app:theorem}

We collect the proofs of our main theoretical results here, including the proofs of Theorem~\ref{thm:score_equivalence} in Section~\ref{sec:model}, Proposition~\ref{prop:convergence} in Section~\ref{sec:method}, and Theorem~\ref{thm_main}, Corollary~\ref{cor:p>n} and Corollary~\ref{cor:p<n} in Section~\ref{sec:theory}.

\subsection{Proof of Proposition \ref{prop:convergence}}
\begin{proof}
First, notice that, with probability one, $X$ has full column rank. Also, because the Cholesky factor $\widehat{L}^{(t)}$ is always positive definite for each iteration, $\widehat{L}^{(t)}X$ is in \textit{general position} a.s. Note that the first three terms of (\ref{obj}) are differentiable (regular) and the whole function is continuous. Furthermore, solving (\ref{obj}) with respect to each variable gives a unique coordinate-wise minimum. Therefore, by Theorem 4.1 (c) in \cite{Tseng2001}, the block coordinate descent converges to a stationary point.
\end{proof}

\subsection{Proof of Theorem \ref{thm:score_equivalence}}

\begin{proof}
By Proposition \ref{prop:convergence} from \cite{Chickering:2003:OSI:944919.944933}, there exists a finite sequence of covered edge reversals in $\mathcal{G}_1$ such that at each step $\mathcal{G}_1\backsimeq \mathcal{G}_2$ and eventually $\mathcal{G}_1 = \mathcal{G}_2$. Hence it suffices to show the result for those $\mathcal{G}_1$ and $\mathcal{G}_2$ that differ only by one edge reversal. 

Before we show the main result, let us first prove that given the same initial $\Theta_0$, the BCD algorithm will generate the same limiting estimate $(\widehat\Theta, \widehat B,\widehat \Omega)$. This limit point is a stationary and partial optimal point for \eqref{eq:lik} by proposition \ref{prop:convergence}. To do this, it suffices to show that the sample covariance matrices calculated from \eqref{eq:S} (appear in the trace term $\Tr(S\Theta)$) for the two networks are equal: $S_{\mathcal{G}_1} = S_{\mathcal{G}_2}$. 

Suppose $X_i$ and $X_j$ are two nodes in the DAGs and the edges between them have opposite directions in $\mathcal{G}_1$ and $\mathcal{G}_2$. We assume the nodes follow a topological order, and let $Z$ denote the common parents of $X_i$ and $X_j$. The sample covariance matrix $S(\widehat B, \widehat\Omega)$ is defined for a DAG $\mathcal{G}$ as the following:
$$S_\mathcal{G} = \sum_{k = 1}^p \frac{1}{(\widehat\omega_k^\mathcal{G}){^2}} \varepsilon_k^\mathcal{G}\varepsilon_k^\mathcal{G}{^\top},$$
where $\varepsilon_k$ is the residual after projecting $X_k$ onto its parents (given by a DAG). Also, $\widehat\omega_k^2 = \|\varepsilon_k^\mathcal{G}\|_2^2/n$. In our case, we simply need to show that
\begin{align}\label{eq:lem4}
\frac{1}{(\widehat\omega_i^{\mathcal{G}_1})^2} \varepsilon_i^{\mathcal{G}_1}\varepsilon_i^{\mathcal{G}_1}{^\top} + \frac{1}{(\widehat\omega_j^{\mathcal{G}_1})^2} \varepsilon_j^{\mathcal{G}_1}\varepsilon_j^{\mathcal{G}_1}{^\top} = \frac{1}{(\widehat\omega_i^{\mathcal{G}_2})^2} \varepsilon_i^{\mathcal{G}_2}\varepsilon_i^{\mathcal{G}_2}{^\top} + \frac{1}{(\widehat\omega_j^{\mathcal{G}_2})^2} \varepsilon_j^{\mathcal{G}_2}\varepsilon_j^{\mathcal{G}_2}{^\top},
\end{align}
because all other terms in the summation are equal for both DAGs. \\
Let $X_i^\perp, X_j^\perp$ be the respective residuals in the projection of $X_i$ and $X_j$ to the $\text{span}(Z)$, and $\tilde X_i^\perp, \tilde X_j^\perp$ be the normalized $X_i^\perp$ and $X_j^\perp$. Then if we let $X_i \xrightarrow{} X_j$ in $\mathcal{G}_1$ and $X_j \xrightarrow{} X_i$ in $\mathcal{G}_2$, we have 
\begin{align}
\text{LHS of (\ref{eq:lem4}}) &= \tilde X_i^\perp\tilde X_i^\perp{^\top} + \left(\frac{X_j^\perp  - \langle X_j^\perp,  \tilde X_i^\perp \rangle\tilde X_i^\perp}{\|X_j^\perp  - \langle X_j^\perp,  \tilde X_i^\perp \rangle\tilde X_i^\perp\|}\right) \cdot  \left(\frac{X_j^\perp  - \langle X_j^\perp,  \tilde X_i^\perp \rangle\tilde X_i^\perp}{\|X_j^\perp  - \langle X_j^\perp,  \tilde X_i^\perp \rangle\tilde X_i^\perp\|}\right)^\top \label{eq:lhs}.\\ 
% \text{RHS of (\ref{eq:lem4}}) &=  \tilde X_j^\perp\tilde X_j^\perp{^\top} + \left(\frac{X_i^\perp  - \langle X_i^\perp,  \tilde X_j^\perp \rangle\tilde X_j^\perp}{\|X_i^\perp  - \langle X_i^\perp,  \tilde X_j^\perp \rangle\tilde X_j^\perp\|}\right) \cdot  \left(\frac{X_i^\perp  - \langle X_i^\perp,  \tilde X_j^\perp \rangle\tilde X_j^\perp}{\|X_i^\perp  - \langle X_i^\perp,  \tilde X_j^\perp \rangle\tilde X_j^\perp\|}\right)^\top \label{eq:rhs}
\intertext{Denote $\langle \tilde X_i^\perp, \tilde X_j^\perp \rangle = \cos\theta$, and notice that}
(\ref{eq:lhs}) &=  \tilde X_i^\perp\tilde X_i^\perp{^\top} + \frac{\left(X_j^\perp  - \langle X_j^\perp,  \tilde X_i^\perp \rangle\tilde X_i^\perp\right) \cdot \left(X_j^\perp  - \langle X_j^\perp,  \tilde X_i^\perp \rangle\tilde X_i^\perp\right)^\top}{\|X_j^\perp\|^2\sin^2\theta}\nonumber\\
&= \tilde X_i^\perp\tilde X_i^\perp{^\top} + \frac{\tilde X_j^\perp\tilde X_j^\perp{^\top}}{\sin^2\theta} + \frac{\tilde X_i^\perp\tilde X_i^\perp{^\top}\cos^2\theta}{\sin^2\theta} + \text{A shared term}\nonumber\\
&= \frac{\tilde X_j^\perp\tilde X_j^\perp{^\top} +\tilde X_i^\perp\tilde X_i^\perp{^\top} }{\sin^2\theta} + \text{A shared term}.\label{eq:sym}
\end{align} 
Since (\ref{eq:sym}) is symmetric in $i$ and $j$, we have that $\text{LHS of (\ref{eq:lem4}}) = \text{RHS of (\ref{eq:lem4}}).$
In other words, given the same initial $\widehat\Theta_0$, the iterative algorithm will generate the same sequence of $(\widehat B,\widehat\Omega, \widehat\Theta)$, thus the same limiting point. 

Note that the MLE estimates for $\mathcal{G}_1$ and $\mathcal{G}_2$ are also limiting points given some initial $\Theta$. Let us suppose the MLE exists and $(\widehat\Theta_1, \widehat B_1), (\widehat\Theta_2, \widehat B_2)$ are the MLEs for $\mathcal{G}_1, \mathcal{G}_2$, respectively, then according to the results above we have
$$L_{\mathcal{G}_1} (\widehat\Theta_1, \widehat B_1(\mathcal{G}_1)) = L_{\mathcal{G}_2}(\widehat\Theta_1, \widehat B_1(\mathcal{G}_2)) \leq L_{\mathcal{G}_2}(\widehat\Theta_2, \widehat B_2(\mathcal{G}_2)) = L_{\mathcal{G}_1}(\widehat\Theta_2, \widehat B_2(\mathcal{G}_1)).$$
Therefore, $$L_{\mathcal{G}_1} (\widehat\Theta_1, \widehat B_1(\mathcal{G}_1)) = L_{\mathcal{G}_2}(\widehat\Theta_2, \widehat B_2(\mathcal{G}_2)).$$
Thus, all MLEs for $\mathcal{G}_1$ yield the same likelihood value which is equal to the likelihood value of any MLE for $\mathcal{G}_2$.
\end{proof}

\subsection{Proof of Theorem \ref{thm_main}}
\begin{proof}
We first prove the consistency in $\widehat{B}^{(1)}$. Under Assumption \ref{assump: theta}, Theorem \ref{thm_inf} shows that for the given $\lambda_n$, the empirical process term of the noises can be uniformly bounded with high probability. Therefore, in order to obtain the conclusion in Theorem \ref{thm:oracle}, we only need the inequality \eqref{eq:condition2} in Lemma \ref{lem:condition2} to hold. Since the event $\mathscr{K}$ in \eqref{eventK} holds with high probability by Theorem 7.16 in \cite{wainwright_2019}, \eqref{eq:condition2} holds by Lemma \ref{lem:condition2}. Next, we show $\widehat{\Theta}^{(1)}$ is consistent by invoking Theorem \ref{thm:infnorm}. For the chosen $\lambda_p$ and under the constraint on $(n, p)$ specified in \eqref{main:sample_size}, the sample size requirement in \eqref{sample_size} is satisfied. Therefore, the results follow from Theorem \ref{thm:infnorm}. Combining Theorem \ref{thm:oracle} and \ref{thm:infnorm}, we get the desired results. 
\end{proof}

\subsection{Proof of Corollary \ref{cor:p>n}}
\begin{proof}
The rate of $\sup_j\|\hat\beta_j^{(1)} - \beta_j^*\|_2^2$ follows directly from the choice of $\lambda_n \asymp \sqrt{\frac{\log p}{n}}$. Since $r(\widehat\Omega) = O_p\left(s\sqrt{\frac{\log p}{N}}\right)$ and $p \gg n$,
\begin{align*}
    \|\widehat\Theta^{(1)} - \Theta^*\|_2 &= O_p\left(m\left(\sqrt\frac{\log n}{p} + s\max\left\{ \sqrt{\frac{\log p}{N}}, \sqrt{\frac{\log^3 p}{n}}\right\}\right)\right) \\
    &= O_p\left(ms \max\left\{ \sqrt{\frac{\log p}{N}}, \sqrt{\frac{\log^3 p}{n}}\right\}\right) \quad \text{($n \gtrsim N$)}\\
    &= O_p\left(ms\sqrt{\frac{\log^3 p}{n}}\right) \quad \text{($N\log^2p \gtrsim n$).}
\end{align*}
\end{proof}

\subsection{Proof of Corollary \ref{cor:p<n}}
\begin{proof}
The rate of $\sup_j\|\hat\beta_j^{(1)} - \beta_j^*\|_2^2$ can be derived in the same way as in the proof of Corollary \ref{cor:p>n}.  Since $r(\widehat\Omega) = O_p\left(s\sqrt{\frac{\log p}{N}}\right)$ and $n \gg p$,
\begin{align*}
    \|\widehat\Theta^{(1)} - \Theta^*\|_2 &= O_p\left(m\left(\sqrt\frac{\log n}{p} + s\max\left\{ \sqrt{\frac{\log p}{N}}, \sqrt{\frac{\log p\log^2 n}{n}}\right\}\right)\right) \\
    & = O_p\left(m\left(\sqrt\frac{\log n}{p} + s\max\left\{ \sqrt{\frac{\log p}{N}}, \sqrt{\frac{\log p\log^2 n}{n}}\right\}\right)\right) \\
\end{align*}
$N\gtrsim s^2p \implies \sqrt{\frac{\log n}{p}} \gtrsim \sqrt{\frac{s^2\log p}{N}}$ and $n \gg s^2p\log p\log n \implies \sqrt{\frac{\log n}{p}}  \gtrsim \sqrt{\frac{s^2\log p\log^2 n}{n}}$. Thus,
$$\|\widehat\Theta^{(1)} - \Theta^*\|_2 = O_p\left(m\sqrt{\frac{\log n}{p}}\right).$$
\end{proof}

\section{Proofs of Intermediate Results for Theorem \ref{thm_main}}
We include the proofs for all the intermediates results that lead to Theorem \ref{thm_main} in this section.
\subsection{Proof of Theorem \ref{thm_inf}}
\begin{proof}
For any $j = 1,\ldots, p,$
    \begin{align*}
        \frac{2}{n}\|X^\top\widehat{\Theta}^{(0)}\varepsilon_j\|_\infty &= \frac{2}{n} \|X^\top(\Theta^* + \widehat{\Delta}_{prec}^{(0)})\varepsilon_j \|_\infty
        \leq \frac{2}{n}\|\widetilde{X}^\top\Tilde{\varepsilon}_j\|_\infty + \frac{2}{n}\|X^\top \widehat{\Delta}_{prec}^{(0)}\varepsilon_j\|_\infty \\ 
        &\leq \frac{2}{n}\|\widetilde{X}^\top\Tilde{\varepsilon}_j\|_\infty + \frac{2}{n}\|\widetilde{X}^\top L^{*-\top}\widehat{\Delta}_{prec}^{(0)}L^{*-1}\Tilde{\varepsilon}_j\|_\infty.
   \end{align*}
%  We only need to bound the second term and apply Lemma \ref{lem:2}.
 Let $\widehat{K}^{(0)} = L^{*-\top} \widehat{\Delta}_{prec}^{(0)} L^{*-1}$. Then following Assumption \ref{assump: theta},
\begin{align*}
   \|\widehat{K}^{(0)}\|_2 \leq \|\widehat{\Delta}_{prec}^{(0)}\|_2/\sigma_{\min}^2(L^*) \leq M/\sigma_{\min}^2(L^*). 
\end{align*}
Under event $\mathscr{E}$ defined in \eqref{eq:eventE}, for $j \in [p]$,
\begin{align*}
   \|\widehat{K}^{(0)}\widetilde X_j\|_2 \leq 6\bar\psi M\sqrt{n}/\sigma_{\min}^2(L^*)\leq 6\bar\psi \sqrt{n}.
\end{align*}
 For $j\in[p]$, define the event $\overline{\mathscr{T}}_j$ as the following
 \begin{align*}
     \overline{\mathscr{T}}_j := \left\{2\|\widetilde{X}^\top \widehat{K}^{(0)}\Tilde{\varepsilon}_j\|_\infty/n < \lambda_n/2\right\}.
 \end{align*}
Similar to the proof of Lemma \ref{lem:2}, we can show
$$\mathbb{P}\left(\bigcup_{j=1}^p\overline{\mathscr{T}}_j^c\mid\mathscr{E}\right) \leq \frac{1}{p},$$
 and
 \begin{align*}
     \mathbb{P}\left(\overline{\mathscr{T}}\bigcap\mathscr{T}\right) &\geq \mathbb{P}\left(\overline{\mathscr{T}}\bigcap\mathscr{T}\mid\mathscr{E}\right) \mathbb{P}\left(\mathscr{E}\right) \geq (1 - \frac{1}{p})(1 - \frac{2}{p})\geq (1 - \frac{2}{p})^2,
 \end{align*}
 where $\mathscr{T}$ is defined in Lemma \ref{lem:2}.
\end{proof}

\subsection{Proof of Lemma \ref{lem:condition2}}
\begin{proof}
We observe that
    \begin{align*}
        \|\widehat{L}^{(0)}X\theta\|_2/\sqrt{n} &\geq \|L^*X\theta\|_2\sqrt{n} - \|\widehat{\Delta}^{(0)}_{chol} X\theta\|_2/\sqrt{n}\\
        &=\|L^*X\theta\|_2 /\sqrt{n} - \|\widehat{\Delta}_{chol}^{(0)}L^{*-1}L^{*}X\theta\|_2/\sqrt{n}\\
        &\geq \left(1 - \|\widehat{\Delta}_{chol}^{(0)}\|_2/\sigma_{\min}\left(L^*\right)\right)\|L^*X\theta\|_2/\sqrt{n}\\
        &\geq \left(1 - \frac{M}{2\sigma_{\min}^2\left(L^*\right)}\right)\|\widetilde{X}\theta\|_2/\sqrt{n}\quad\ \text{(By Assumption \ref{assump: theta} and Lemma \ref{lem:2.1})}\\
        &\geq \frac{1}{2\sqrt{n}}\|\widetilde{X}\theta\|_2,
    \end{align*}
    when $n$ is sufficiently large. Since event $\mathscr{K}$ defined in \eqref{eventK} holds, by Theorem 7.16 in \cite{wainwright_2019}, we have
    \begin{align*}
        \frac{\|\widehat{L}^{(0)}X\theta\|_2^2}{n}&\geq \frac{1}{4}\left(\Tilde c_1\|\sqrt{\Psi^*}\theta\|_2^2 - \Tilde c_2\rho^2(\Psi^*)\frac{\log p}{n}\|\theta\|_1^2\right)\\
        &\geq c_1\|\sqrt{\Psi^*}\theta\|_2^2 - c_2\rho^2(\Psi^*)\frac{\log p}{n}\|\theta\|_1^2.
     \end{align*}{}
\end{proof}

\subsection{Proof of Theorem \ref{thm:oracle}}

\begin{proof}
Consider the penalized negative likelihood function from \eqref{eq:1step}:
$$\mathcal{L}(\beta_j,\lambda_n) = \frac{1}{2n}\|\widehat L^{(0)}X_j - \widehat L^{(0)}X\beta_j\|_2^2 + \lambda_n\|\beta_j\|_1.$$
% For simplicity, we drop the subscript $j$ in $\beta_j$ and $\varepsilon_j$ and the superscript $(t)$ in $\hat\beta^{(1)}$ and $\widehat L^{(0)}$. The generic penalized negative likelihood becomes:
% $$\mathcal{L}(\beta,\lambda_n) = \frac{1}{2n}\|\widehat L X_j - \widehat LX\beta\|_2^2 + \lambda_n\|\beta\|_1.$$
For simplicity, we drop the superscript $(t)$ in $\hat\beta^{(1)}$ and $\widehat L^{(0)}$.
Let $\rho$ stand for $\rho(\Psi^*)$, $\beta_j^*\in\mathbb{R}^p$ , and $\widehat\Delta_j = \hat\beta_j - \beta_j^*$. We start from the \textit{basic inequality} \citep{wainwright_2019}:
\begin{align*}
    \mathcal{L}(\hat\beta_j,\lambda_n) &\leq \mathcal{L}(\beta_j^*,\lambda_n) = \frac{1}{2n}\|\widehat{L}\varepsilon_j\|_2^2 + \lambda_n\|\beta_j^*\|_1.
\end{align*}
After rearranging some terms,
\begin{align}
    0 \leq \frac{1}{2n} \|\widehat{L}X\widehat{\Delta}_j\|_2^2 &\leq \frac{\varepsilon_j^\top\widehat{\Theta}X\widehat{\Delta}_j}{n} + \lambda_n \left(\|\beta_j^*\|_1 - \|\hat\beta_j\|_1\right).\label{eq:35}
\end{align}
Next, for any subset $S\subseteq [p]$, we have
\begin{align}
    \|\beta_j^*\|_1 - \|\hat\beta_j\|_1 &= \|\beta^*_{j,S}\|_1 + \|\beta^*_{j,S^c}\|_1 - \|\beta_{j,S}^* + \widehat{\Delta}_{j,S}\|_1 - \|\widehat{\Delta}_{j,S^c} + \beta_{j,S^c}^*\|_1. \label{eq:36}
\end{align}
Combined \eqref{eq:35} with \eqref{eq:36}, and apply triangle and Hölder's inequalities,
\begin{align*}
    0\leq\frac{1}{2n}\|\widehat{L}X\widehat{\Delta}_j\|_2^2 &\leq \frac{1}{n}\varepsilon_j^\top\widehat{\Theta}X\widehat{\Delta}_j + \lambda_n\left(\|\widehat{\Delta}_{j,S}\|_1 - \|\widehat{\Delta}_{j,S^c}\|_1 + 2\|\beta_{j,S^c}^*\|_1\right)\\
    &\leq \|X^\top\widehat{\Theta}\varepsilon_j\|_\infty/n\|\widehat{\Delta}_j\|_1 + \lambda_n\left(\|\widehat{\Delta}_{j,S}\|_1 - \|\widehat{\Delta}_{j,S^c}\|_1 + 2\|\beta_{j,S^c}^*\|_1\right)\\
    &\leq \frac{\lambda_n}{2}\left(\|\widehat{\Delta}_j\|_1 + 2\|\widehat{\Delta}_{j,S}\|_1 - 2\|\widehat{\Delta}_{j,S^c}\|_1 + 4\|\beta^*_{j,S^c}\|_1\right) \\
    &\leq\frac{\lambda_n}{2}\left[ 3\|\widehat{\Delta}_{j,S}\|_1 - \|\widehat{\Delta}_{j,S^c}\|_1 + 4\|\beta^*_{j,S^c}\|_1\right],\numberthis\label{eq:cone}\\
    \|\widehat{\Delta}_j\|_1 &\leq 4\left(\|\widehat{\Delta}_{j,S}\|_1 + \|\beta^*_{j,S^c}\|_1\right).
\end{align*}{}
This inequality implies (apply Cauchy-Schwarz inequality) 
\begin{align}\label{eq:oracle_temp}
    \|\widehat{\Delta}_j\|_1^2 \leq \left( 4\|\widehat{\Delta}_{j,S}\|_1 +  4\|\beta^*_{j,S^c}\|_1\right)^2 \leq 
     32\left(|S|\ \|\widehat{\Delta}_j\|_2^2 + \|\beta^*_{j,S^c}\|_1^2\right).
\end{align}{}
Next, from \eqref{eq:condition2} and \eqref{eq:oracle_temp}, we know,
\begin{align*}
    \frac{\|\widehat{L}X\widehat{\Delta}_j\|_2^2}{n} 
    &\geq \left(c_1\bar{\kappa} - 32c_2\rho^2|S|\frac{\log p}{n} \right)\|\widehat\Delta_j\|_2^2 -32c_2\rho^2\frac{\log p}{n} \|\beta^*_{j,S^c}\|_1^2 \\
    &\geq \frac{c_1\bar\kappa}{2}\|\widehat\Delta_j\|_2^2 - 32c_2\rho^2\frac{\log p}{n} \|\beta^*_{j,S^c}\|_1^2,\numberthis \label{eq:2.39}
\end{align*}{}
where the last inequality comes from the condition $|S| \leq \frac{c_1}{64c_2}\frac{\bar\kappa}{\rho^2(\Psi^*)}\frac{n}{\log p}$. Now let's analyze the following two cases regarding \eqref{eq:2.39}:
\begin{case}
Suppose that $\frac{c_1\bar\kappa}{4}\|\widehat\Delta_j\|_2^2  \geq 32c_2\rho^2\frac{\log p}{n} \|\beta^*_{j,S^c}\|_1^2$, then from \eqref{eq:cone} we can get
\begin{align*}
   \frac{c_1\bar\kappa}{4}\|\widehat\Delta_j\|_2^2 \leq \lambda_n\left(3\sqrt{|S|}\|\widehat{\Delta}_j\|_2 + 4\|\beta^*_{j,S^c}\|_1\right).
\end{align*}{}
Solving for the zeros of this quadratic form in $\|\widehat{\Delta}_j\|_2$ yields
\begin{align*}
    \|\widehat{\Delta}_j\|_2^2 \leq \frac{48\lambda_n^2}{c_1^2\bar\kappa^2}|S| + \frac{16\lambda_n\|\beta^*_{j,S^c}\|_1}{c_1\bar\kappa}.
\end{align*}
\end{case}
\begin{case}
Otherwise, we have $\frac{c_1\bar\kappa}{4}\|\widehat\Delta_j\|_2^2  \leq 32c_2\rho^2\frac{\log p}{n} \|\beta^*_{j,S^c}\|_1^2$. 
\end{case}{}
After combining the two cases, we obtain the claimed bound in \eqref{eq:oracle}. To get the prediction bound in \eqref{pred_bound}, we first show $\|\widehat{\Delta}_j\|_1 \leq 4\|\beta_j^*\|_1$. From basic inequality, we have
\begin{align*}
    0 &\leq \frac{1}{2n}\|\widehat{L}X\widehat{\Delta}_j\|_2^2 \leq \frac{\varepsilon_j^\top\widehat\Theta X\widehat{\Delta}_j}{n} + \lambda_n\left(\|\beta_j^*\|_1 - \|\hat\beta_j\|_1\right).
\end{align*}
By H\"older's inequality and Theorem \ref{thm_inf},
\begin{align*}
    \left|\frac{\varepsilon_j^\top\widehat{\Theta}X\widehat{\Delta}_j}{n}\right| &\leq \left\lVert\frac{X^\top\widehat{\Theta}\varepsilon_j}{n}\right\rVert_\infty\|\widehat\Delta_j\|_1 \leq \frac{\lambda_n}{2}\left(\|\beta_j^*\|_1 + \|\hat\beta_j\|_1\right).
\end{align*}
Combine the two inequalities above, we get
\begin{align*}
    0&\leq \frac{3\lambda_n}{2}\|\beta_j^*\|_1 - \frac{\lambda_n}{2}\|\hat\beta_j\|_1,
\end{align*}
which implies $\|\hat\beta_j\|_1 \leq 3\|\beta_j^*\|_1$. Consequently, we have
$$\|\widehat{\Delta}_j\|_1 \leq \|\hat\beta_j\|_1 + \|\beta_j^*\|_1 \leq 
4\|\beta_j^*\|_1.$$
Return to the basic inequality, we have
\begin{align*}
    \frac{\|\widehat{L}X\widehat{\Delta}_j \|_2^2}{2n}&\leq\frac{\lambda_n}{2}\|\widehat{\Delta}_j\|_1 + \lambda_n\left(\|\beta_j^*\|_1 - \|\beta_j^* + \widehat\Delta_j\|_1\right)\\ &\leq \frac{3}{2}\lambda_n\|\widehat{\Delta}_j\|_1 \leq 6\lambda_n\|\beta_j^*\|_1.
\end{align*}
\end{proof}

% \subsection{Proof of Theorem \ref{thm:pred}}
% \begin{proof}
% \end{proof}

\subsection{Proof of Lemma \ref{lem:sampleCov}}
\begin{proof}
The proof follows a similar approach as the proof for Lemma 1 in \citet{logdet}.
\end{proof}

\subsection{Proof of Corollary \ref{cor:tail}}
\begin{proof}
A little calculation using Lemma \ref{lem:sampleCov} and Definition \ref{def:tail} shows that the corresponding inverse functions for data from the Gaussian DAG model \eqref{eq:model} are:
\begin{align*}
    \bar\delta_f(p;r) = 40\sqrt{\frac{2\log(4r)}{p}}, \quad\text{and}\quad \bar{p}_f(\delta;r) = \frac{3200\log(4r)}{\delta^2}.
\end{align*}
Setting $r = n^\tau$ yields the desired result.
\end{proof}

\subsection{Proof of Lemma \ref{lem:R}}
\begin{proof}
Let $\widehat{S}_{ij}^{(t)}$ denote the $(i,j)$ entry of the sample variance matrix $\widehat S^{(t)}$ defined in \eqref{glasso}. Let $X_{i\cdot}$ and $X_{\cdot j}$ denote the $i$th row and $j$th column of $X$, respectively. Let $\varepsilon_{ik}^{*} := X_{ik} - X_{i\cdot}\beta_{k}^*$ where $\beta_{k}^*$ is the $k$th column of $B^*$, $\rho_k^* = 1/\omega_k^{*2}, \hat\rho_k = 1/\hat\omega_k^{2}$, $\widehat\Delta_k^{(t)} := \hat\beta_k^{(t)} - \beta_k^*$, and
$\hat\delta_k = \hat\rho_k - \rho_k^*$. 
Then,
\begin{align*}
\widehat{S}_{ij}^{(1)} &= \frac{1}{p}\sum_{k=1}^p \hat\rho_{k}\left(X_{ik} - X_{i\cdot}\hat\beta_k^{(1)}\right)\left(X_{jk} - X_{j\cdot}\hat \beta_k^{(1)}\right)\\
&=\frac{1}{p}\sum_{k=1}^p \hat\rho_k\left(\varepsilon_{ik}^* - X_{i\cdot}\widehat\Delta_k^{(1)}\right)\left(\varepsilon_{jk}^* - X_{j\cdot}\widehat\Delta_k^{(1)}\right)\\
&= \widehat\Sigma_{ij} + \frac{1}{p}\sum_{k=1}^p\rho_k^{*}\left(-\varepsilon_{ik}^{*}X_{j\cdot}\widehat\Delta_{k}^{(1)} - \varepsilon_{jk}^{*}X_{i\cdot}\widehat\Delta_k^{(1)} + X_{i\cdot}\widehat\Delta_k^{(1)}X_{j\cdot}\widehat\Delta_k^{(1)}\right) \\ &\quad + \frac{1}{p}\sum_{k=1}^p\hat\delta_{k}\left(\varepsilon_{ik}^* - X_{i\cdot}\widehat\Delta_k^{(1)}\right)\left(\varepsilon_{jk}^* - X_{j\cdot}\widehat\Delta_k^{(1)}\right) \\
&=\widehat\Sigma_{ij} + \frac{1}{p}\sum_{k=1}^p\hat\rho_k\left(-\varepsilon_{ik}^{*}X_{j\cdot}\widehat\Delta_{k}^{(1)} - \varepsilon_{jk}^{*}X_{i\cdot}\widehat\Delta_k^{(1)} + X_{i\cdot}\widehat\Delta_k^{(1)}X_{j\cdot}\widehat\Delta_k^{(1)}\right) + \frac{1}{p}\sum_{k=1}^p \hat\delta_{k}\varepsilon_{ik}^*\varepsilon_{jk}^*.
\end{align*}
If we let $R_{ij}= \frac{1}{p}\sum_{k=1}^p\hat\rho_k\left(-\varepsilon_{ik}^{*}X_{j\cdot}\widehat\Delta_{k}^{(1)} - \varepsilon_{jk}^{*}X_{i\cdot}\widehat\Delta_k^{(1)} + X_{i\cdot}\widehat\Delta_k^{(1)}X_{j\cdot}\widehat\Delta_k^{(1)}\right) + \frac{1}{p}\sum_{k=1}^p \hat\delta_{k}\varepsilon_{ik}^*\varepsilon_{jk}^*$, we can upper bound $|R_{ij}|$ by dividing it into three terms and controlling each term separately.\\\\
Part 1.\\
We observe that
\begin{align*}
\abs{\frac{1}{p}\sum_{k=1}^p\hat\rho_k\varepsilon_{ik}^{*}X_{j\cdot}\widehat\Delta_{k}^{(1)}} &= \abs{\frac{1}{p}\sum_{k=1}^p\left(\rho_k^{*} +\hat\delta_k \right)\varepsilon_{ik}^{*}X_{j\cdot}\widehat\Delta_{k}^{(1)}}\\
&\leq \abs{\frac{1}{p}\sum_{k=1}^p\rho_k^{*}\varepsilon_{ik}^{*}X_{j\cdot}\widehat\Delta_{k}^{(1)}} + \abs{\frac{1}{p}\sum_{k=1}^p \hat\delta_k\varepsilon_{ik}^{*}X_{j\cdot}\widehat\Delta_{k}^{(1)}}.
\end{align*}
If $\widehat\Delta_k^{(1)}$ and $\hat\delta_k$ satisfy \eqref{R:assump}, both $\rho_k^{*}$ and $\hat\delta_k$ can be bounded by positive constants ($r(\widehat\Omega) \ll 1$). Define the following events:
\begin{align*}
    \mathcal{B}_1 &= \bigcup_{i=1}^n\left\{\|\varepsilon_{i\cdot}\|_\infty \geq 6\bar\omega\sqrt{\log \max\{n, p\}}\right\},\\
    \mathcal{B}_2 &= \bigcup_{k=1}^n\bigg\{\|\varepsilon_{k\cdot}^*\|_2 \geq 6\bar\omega\sqrt{p}\bigg\},\\
    \mathcal{B}_3 &= \bigcup_{k=1}^n\bigg\{\|X_{k\cdot}\|_\infty \geq 6\bar\psi\sqrt{\log \max\{n, p\}}\bigg\}.
\end{align*}
Under event $\mathcal{B}_1$, $\mathcal{B}_2$, and $\mathcal{B}_3$,
\begin{align*}
    \abs{\frac{1}{p}\sum_{k=1}^p\hat\rho_k\varepsilon_{ik}^{*}X_{j\cdot}\widehat\Delta_{k}^{(1)}} &\leq 
     \frac{2}{b}\sup_k |\varepsilon_{ik}^{*}X_{j\cdot}\widehat\Delta_k^{(1)}| \\
    &\leq \frac{12\bar\omega}{b}\sqrt{\log \max\{n,p\}} \sup_k\|X_{j\cdot}\|_\infty\|\widehat\Delta_k^{(1)}\|_1 \quad\text{(By H\"older's Inequality and $\mathcal{B}_1$)}\\
    &\leq \frac{12\bar\omega s\sqrt{\log p\log\max\{n,p\}}}{b\sqrt{n}}\|X_{j\cdot}\|_\infty \quad\text{(From $\|\widehat\Delta_k^{(1)}\|_1 \leq 4\sqrt{2s}\|\widehat\Delta_k^{(1)}\|_2$)}\\
    &\leq \frac{72\bar\omega\bar\psi s}{b}\sqrt{\frac{\log p\log^2 \max\{n,p\}}{n}} \quad\text{(By event $\mathcal{B}_3$).}
\end{align*}
The second last inequality comes from $\|\widehat\Delta^{(1)}\|_1 \leq 4\sqrt{2s}\|\widehat\Delta^{(1)}\|_2$ in the proof of Theorem \ref{thm:oracle}.\\\\
Part 2\\
Notice that
\begin{align*}
    \abs{\frac{1}{p}\sum_{k=1}^p\hat\rho_kX_{i\cdot}\widehat\Delta_k^{(1)}X_{j\cdot}\widehat\Delta_k^{(1)}} &\leq \frac{2}{b}\sup_k|X_{j\cdot}\widehat\Delta_{k}^{(1)}||X_{i\cdot}\widehat\Delta_{k}^{(1)}|\\
    &\leq \frac{2s^2\log p}{bn}\|X_{j\cdot}\|_\infty\|X_{i\cdot}\|_\infty\\
    &\leq \frac{12\bar\psi s^2}{b}\sqrt{\frac{\log^2 p\log^2\max\{n,p\}}{n^2}}.
\end{align*}
\\
Part 3\\
By Lemma \ref{lem:rho_consistency}, $\|\hat\delta\|_\infty = \sup_k|\hat\rho_k - \rho_k^{*}| = \sup_k|1/\hat\omega_k^2 - 1/\omega_k^{*2}| = r(\widehat\Omega)$. Combining with $\mathcal{B}_2$, we have
% \frac{2\bar\omega}{b^4}s\sqrt{\frac{\log p}{N}}
\begin{align*}
    \abs{\frac{1}{p}\sum_{k=1}^p \hat\delta_k\varepsilon_{ik}^{*}\varepsilon_{jk}^{*}} &\leq  \frac{1}{p} \|\hat\delta\|_\infty \sum_{k=1}^p|\varepsilon_{ik}^{*}\varepsilon_{jk}^{*}| \leq\frac{1}{p} \|\delta\|_\infty \|\varepsilon_{i\cdot}^{*}\|_2\|\varepsilon_{j\cdot}^{*}\|_2 \leq 6\bar{\omega} \|\hat\delta\|_\infty = 6\bar{\omega}r(\widehat\Omega).
\end{align*}
Combine all three parts, we have 
\begin{align*}
     |R_{ij}| \leq \max\left\{6\bar\omega r(\widehat\Omega), \frac{72\bar\omega\bar\psi s}{b}\sqrt{\frac{\log p\log^2 \max\{n,p\}}{n}} \right\}.
\end{align*}
Using Lemma \ref{lem:maximal} and Lemma \ref{lem:xnorm}, we can derive the upper bound for the probabilities of $\mathcal{B}_1, \mathcal{B}_2, \mathcal{B}_3$:
\begin{align*}
    \mathbb{P}\left(\mathcal{B}_1\right) &\leq 2/\max\{n,p\}^4 ,\\
    \mathbb{P}\left(\mathcal{B}_2\right) &\leq 1/\max\{n,p\}^4\quad\text{if $n > 2\sqrt{10}\log\max\{n,p\}$},\\
    \mathbb{P}\left(\mathcal{B}_2\right) &\leq 2/\max\{n,p\}^4, \\
    \mathbb{P} \left(\bigcup_{l=1}^3\mathcal{B}_i\right) &\leq 5 / \max\{n,p\}^4.
\end{align*}
Applying union bound,
\begin{align*}
    \|R\|_\infty \leq \max\left\{6\bar\omega r(\widehat\Omega), \frac{72\bar\omega\bar\psi s}{b}\sqrt{\frac{\log p\log^2 \max\{n,p\}}{n}} \right\},
\end{align*}
with probability at least $1 - \frac{5n^2}{\max\{n,p\}^4}$. Take event $\mathcal{A}$ from Lemma \ref{lem:sampleNoise} into account and apply union bound one more time, we arrive at the desired conclusion.
\end{proof}

\subsection{Proof of Theorem \ref{thm:infnorm}}
\begin{proof}
Let $\bar{R}(s, p, n)$ and $\bar{\delta}_f(p;n^\tau)$ be defined as stated, then the monotonicity of the inverse tail function \eqref{eq:tail} and condition \eqref{sample_size} on $(n, p)$ implies that $\bar{\delta}_f(p;n^\tau) \leq 1/40$. Lemma \ref{lem:sampleNoise} and Lemma \ref{lem:R} imply that the event $\mathcal{A}$ defined in \eqref{eventA} and the events $\mathcal{B}_1$,$\mathcal{B}_2$,$\mathcal{B}_3$ defined in the proof of Lemma \ref{lem:R} hold with high probability. Conditioning on these events, we have
$$\|\widehat W^{(1)}\|_\infty \leq \bar{\delta}_f(p;n^\tau) + \bar{R}(s, p, n).$$ 
Choose $\lambda_p = \bar\delta_f(p;n^\tau) + \bar{R}$. By Lemma \ref{lem:R} and condition \eqref{sample_size} we have that
\begin{align*}
2\kappa_{\Gamma^*}\left(\|\widehat W^{(1)}\|_\infty + \lambda_p\right) &\leq 4\kappa_{\Gamma^*}\left(\bar\delta_f(p;n^\tau) + \bar{R}\right)\leq \min\left\{\frac{1}{3\kappa_{\Sigma^*}m}, \frac{1}{3\kappa_{\Sigma^*}^32\kappa_{\Gamma^*}m}\right\}.
\end{align*}
Applying Lemma 6 in \cite{logdet} we obtain
\begin{align*}
    \|\widehat\Theta^{(1)} - \Theta^*\|_\infty &\leq 4\kappa_{\Gamma^*}\left(\bar{\delta}_f(p;n^\tau) + \bar{R}\right),\\
     \|\widehat\Theta^{(1)} - \Theta^*\|_2 &\leq \|A\|_2 \|\widehat\Theta^{(1)} - \Theta^*\|_\infty \leq (m+1) \|\widehat\Theta^{(1)} - \Theta^*\|_\infty.
\end{align*}
\end{proof}

\section{Proofs of Other Auxiliary Results}
This section includes the proofs for Lemma \ref{lem:omega} and \ref{lem:rho_consistency} as well as the four lemmas introduced in Section \ref{auxres}.
\subsection{Proof of Lemma \ref{lem:omega}}
\begin{proof}
From Lemma 1 in \cite{jacob} we know that if $\lambda_N \geq N^{-1} \|X^{(B)\top}\varepsilon_j^{(B)}\|_\infty$, then 
$$\abs{\hat\omega_j^2 - N^{-1}\|\varepsilon_j^{(B)}\|_2^2} \leq 2\lambda_N\|\beta_j^*\|_1 \leq \lambda_Ns\bar\beta.$$
Pick $\lambda_N = 12\bar\psi\bar\omega\left(\sqrt{\frac{2\log p}{N}} + \sqrt{\frac{2\log 2 + 6\log p}{N}}\right)$, then applying the Chernoff bound for sub-Gaussian random variables (e.g., see proof of Lemma \ref{lem:2}) we can show that 
$$\lambda_N \geq \sup_jN^{-1} \|X^{(B)\top}\varepsilon_j^{(B)}\|_\infty.$$
holds with probability at least $(1 - \frac{1}{p})^2$. This proves the first inequality.
To prove the second inequality, notice that
\begin{align*}
    \sup_j\abs{\hat\omega_j^2 - \omega_j^{*2}} &= \sup_j\abs{\hat\omega_j^2 - \frac{\|\varepsilon_j^{(B)}\|_2^2}{N}+ \frac{\|\varepsilon_j^{(B)}\|_2^2}{N} - \omega_j^{*2}}\\
    &\leq \sup_j\abs{\hat\omega_j^2 - \frac{\|\varepsilon_j^{(B)}\|_2^2}{N}} + \sup_j\abs{\frac{\|\varepsilon_j^{(B)}\|_2^2}{N} - \omega_j^{*2}}.
\end{align*}
From $\chi^2$ concentration inequality, (e.g. \citealt{wainwright_2019} Example 2.11)
\begin{align*}
    \abs{\omega_j^{*2} - \frac{1}{N}\|\varepsilon_j^{(B)}\|_2^2} &\geq 2\sqrt{2}\bar\omega\sqrt{\frac{\log2 + 3\log p}{N}} \quad\text{with probability at most $ 1/p^3$,}\\
    \sup_j\abs{\omega_j^{*2} - \frac{1}{N}\|\varepsilon_j^{(B)}\|_2^2} &\geq 2\sqrt{2}\bar\omega\sqrt{\frac{\log2 + 3\log p}{N}} \quad\text{with probability at most $ 1/p^2$.}
\end{align*}
Combining all the inequalities, we can show that the second inequality holds with probability at least $(1 - 1/p)^2 - 1/p$.
\end{proof}
\subsection{Proof of Lemma \ref{lem:rho_consistency}}
\begin{proof}
Simply notice that
\begin{align*}
 \sup_{1\leq j\leq p}\abs{\frac{1}{\hat\omega_j^2} - \frac{1}{\omega_j^{*2}}} = \sup_{1\leq j\leq p}\abs{\frac{1}{\hat\omega_j^2\omega_j^{*2}}}\sup_{1\leq j\leq p}\abs{\omega_j^{*2} - \hat\omega_j^2} \leq \frac{1}{b^4}\sup_{1\leq j\leq p}|\omega_j^{*2} - \hat\omega_j^2|. 
\end{align*}
\end{proof}

\subsection{Proof of Lemma \ref{lem:2.1}}
\begin{proof}
The claim follows since
\begin{align*}
\|\widehat \Delta_{prec}\|_2 &= \|\widehat\Theta - \Theta^*\|_2 \\
&=  \|\left(L^* + \widehat{\Delta}_{chol}\right)^\top \left(L^* + \widehat{\Delta}_{chol}\right) - L^{*\top}L^*\|_2 \\
&= \|L^{*\top}\widehat{\Delta}_{chol} + \widehat{\Delta}_{chol}^\top L^* + \widehat{\Delta}_{chol}^\top\widehat{\Delta}_{chol}\|_2 \\
&\geq \max_{x\in\mathbb{S}^{n-1}} x^\top \left(L^{*\top}\widehat{\Delta}_{chol} + \widehat{\Delta}_{chol}^\top L^* + \widehat{\Delta}_{chol}^\top\widehat{\Delta}_{chol}\right)x\\
&\geq \max_{x\in\mathbb{S}^{n-1}} x^\top \left(L^{*\top}\widehat{\Delta}_{chol} + \widehat{\Delta}_{chol}^\top L^*\right)x\quad\text{(as $\widehat{\Delta}_{chol}^\top\widehat{\Delta}_{chol} \succcurlyeq 0$.)}\\
&= \|L^{*\top}\widehat{\Delta}_{chol} + \widehat{\Delta}_{chol}^\top L^*\|_2\\
&\geq 2\sigma_{\min}\left(L^*\right) \|\widehat{\Delta}_{chol}\|_2.
\end{align*}
\end{proof}

\subsection{Proof of Lemma \ref{lem:xnorm}}
\begin{proof}
Notice that
$\widetilde{X}_k \in\mathbb{R}^n$ is a sub-Gaussian random vector with variance smaller than $\bar\psi$. By Theorem 1.19 in \cite{mit}, we have that 
\begin{align*}
    \mathbb{P}\left(\|\widetilde{X}_k\|_2 > 4\bar\psi \sqrt{n} + 2\bar\psi\sqrt{2\log(1/\delta)}\right) \leq \delta.
\end{align*}
Setting $\delta = 1/p^\alpha$ and using union bound we obtain the desired conclusion.
\end{proof}

\subsection{Proof of Lemma \ref{lem:2}}
\begin{proof}
    In other words, with probability at least $1 - 1/p$, 
\begin{align}\label{eq:normbound}
    \|\widetilde{X}_k\|_2 \leq 4\bar\psi\sqrt{n} + 2\bar\psi\sqrt{2\log p} \leq 6\bar\psi\sqrt{n}, 
\end{align}
for all $k$. Under event the $\mathscr{E}$ defined in \eqref{eq:eventE}, $\|\widetilde{X}_{[j-1]}^\top \Tilde{\varepsilon}_j\|_\infty/n$ corresponds to the absolute maximum of $j-1$ zero-mean Gaussian variables, each with variance at most $36\bar\psi^2\bar\omega^2/n$. Next, we calculate the probability of the event $\mathscr{T}\cap\mathscr{E}$, where $\delta = 1/p^2$. We also let
\begin{align*}
    t &= \sqrt{\frac{2\log 2 + 4\log p}{n}}, \\
    \lambda_n &= 12\bar\psi\bar\omega\left(\sqrt{\frac{2\log p}{n}} + t\right).
\end{align*}
Because both $\widetilde{X}$ and $\Tilde{\varepsilon}$ are random, we use the equivalence: $p(y) = \mathbb{E}_{p(x)} \left[p(y\mid x)\right]$ to apply the properties of fixed-design Lasso: Let $X_{[j-1]}$ denote the first $j-1$ columns in $X$,
    \begin{align*}
        1 - \mathbb{P}(\mathscr{T}_j\mid\mathscr{E}) &= \mathbb{E}_{X_{[j-1]}}\mathbb{P}\left\{2\|\widetilde{X}_{[j-1]}^\top \Tilde{\varepsilon}_j\|_\infty/n > \lambda_n\mid X_{[j-1]},\mathscr{E}\right\} \\
        &= \mathbb{E}_{X_{[j-1]}}\mathbb{P}\left\{2\|\widetilde{X}_{[j-1]}^\top \Tilde{\varepsilon}_j\|_\infty/n > 6\bar\psi\bar\omega\left(\sqrt{\frac{2\log p}{n}} + t\right)\mid X_{[j-1]},\mathscr{E}\right\} \\
        &\leq 2\exp\left\{-\frac{nt^2}{2}\right\} = 1/p^2. 
    \end{align*}
where in the last inequality we apply the Chernoff standard Gaussian tail bound. Hence,
\begin{align}\label{eq:conditional}
        1 - \mathbb{P}\left(\mathscr{T}\mid\mathscr{E}\right) = 1 - \mathbb{P}\left(\bigcap_{j=1}^p\mathscr{T}_j\mid\mathscr{E}\right) 
        = \mathbb{P}\left(\bigcup_{j=1}^p\mathscr{T}_j^c\mid\mathscr{E}\right) \leq \frac{1}{p}.  
\end{align}
Finally, by Lemma \ref{lem:xnorm} with $\alpha = 2$ we get
    \begin{align}\label{eq:eventEE}
        \mathbb{P}(\mathscr{T}) \geq \mathbb{P}(\mathscr{E}) \mathbb{P}\left(\mathscr{T} \mid \mathscr{E}\right) \geq \left(1 - \frac{1}{p}\right)^2.
\end{align}
\end{proof}

\subsection{Proof of Lemma \ref{lem:maximal}}
\begin{proof}
By the sub-Gaussian maximal inequality (e.g., Theorem 1.14 in \citealt{mit}), we know that  if $X_1, \dots X_N$ are random variables such that $X_i\sim$ sub-Gaussian with parameter $\sigma^2$, then for any $t > 0$,
$$\mathbb{P}\left(\max_{1\leq i\leq N}|X_i| \geq t\right) \leq 2N\exp\left(-\frac{t^2}{2\sigma^2}\right).$$
Letting $t = \sqrt{4\bar\psi^2\log p}$ and taking $\sigma^2 = \bar\psi^2$, we arrive at  the desired result. 
\end{proof}

% \subsection{Proof of Corollary \ref{cor:l2norm}}
% \begin{proof}
% Let $A$ be the adjacency matrix of the undirected network $G(A)$. Then
% \begin{align*}
%     \|\widehat\Theta - \Theta^*\|_2 &\leq \|A\|_2\cdot \kappa_{\Gamma^*}\left(\bar\delta_f(p;n^\tau) + \bar{R}\right)\\ 
%     &\leq \kappa_{\Gamma^*}m\left(\bar\delta_f(p;n^\tau) + \bar{R}\right),
% \end{align*}
% from the property of adjacency matrices. 
% \end{proof}

% Note: in this sample, the section number is hard-coded in. Following
% proper LaTeX conventions, it should properly be coded as a reference:

%In this appendix we prove the following theorem from
%Section~\ref{sec:textree-generalization}:

\vskip 0.2in
% \newpage

\bibliography{main.bib}

\end{document}